\documentclass{article}

\usepackage{arxivtemplate}
\usepackage{graphicx}
\usepackage{algorithm}
\usepackage{algorithmic}
\usepackage{xcolor}

\usepackage[utf8]{inputenc} 
\usepackage[T1]{fontenc}    
\usepackage{etoc}
\usepackage{hyperref}      
\usepackage{url}        
\usepackage{tcolorbox}
\usepackage{booktabs}      
\usepackage{amsfonts}     
\usepackage{nicefrac}     
\usepackage{microtype}    
\usepackage{xcolor}        
\usepackage{amsmath}
\usepackage{amsthm}
\usepackage{float}
\usepackage{wrapfig}
\usepackage{caption}
\usepackage{array}
\usepackage{natbib}
\usepackage[export]{adjustbox} 
\usepackage{subfig} 
\usepackage{multirow}
\usepackage{tabularx}
\usepackage{fontawesome5}
\usepackage{enumitem}
\usepackage{caption}
\newcolumntype{Y}{>{\centering\arraybackslash}p{2.4cm}} 
\newcolumntype{Z}{>{\centering\arraybackslash}p{1.5cm}} 
\newcolumntype{M}{>{\centering\arraybackslash}p{1.8cm}} 

\newcommand{\blfootnote}[1]{%
  \begingroup
  \renewcommand\thefootnote{}\footnote{#1}%
  \addtocounter{footnote}{-1}%
  \endgroup
}

\newcommand{\starfootnote}[1]{%
  \begingroup
  \renewcommand\thefootnote{\fnsymbol{footnote}}%
  \setcounter{footnote}{1}%
  \footnotetext[1]{#1}%
  \endgroup
}

\title{Superclass-Guided Representation Disentanglement for Spurious Correlation Mitigation}

\author{\fontsize{11}{13}\selectfont
Chenruo Liu$^{1}$\textsuperscript{*}\qquad
Hongjun Liu$^{1}$\textsuperscript{*}\qquad
Zeyu Lai$^{3}$\qquad
Yiqiu Shen$^{1,2}$\qquad
Chen Zhao$^{1}$\qquad
Qi Lei$^{1}$\\[1.5ex]
\fontsize{10}{13}\selectfont
$^{1}$New York University\quad
$^{2}$NYU Grossman School of Medicine\quad
$^{3}$Zhejiang University
}
\date{}

\begin{document}

\maketitle
\setcounter{footnote}{0}

\blfootnote{Accepted at the Third Conference on Parsimony and Learning (CPAL 2026).}
\starfootnote{Equal contribution.}

\begin{abstract}
    To enhance group robustness to spurious correlations, prior work often relies on auxiliary group annotations and assumes identical sets of groups across training and test domains. To overcome these limitations, we propose to leverage superclasses---categories that lie higher in the semantic hierarchy than the task’s actual labels---as a more intrinsic signal than group labels for discerning spurious correlations. Our model incorporates superclass guidance from a pretrained vision-language model via gradient-based attention alignment, and then integrates feature disentanglement with a theoretically supported minimax-optimal feature-usage strategy. As a result, our approach attains robustness to more complex group structures and spurious correlations, without the need to annotate any training samples. Experiments across diverse domain generalization tasks show that our method significantly outperforms strong baselines and goes well beyond the vision-language model's guidance, with clear improvements in both quantitative metrics and qualitative visualizations.
\end{abstract}

\section{Introduction}
\label{sec:introduction}
When the underlying group composition of the training and test distributions differs, certain input features may exhibit strong correlations with the target label during training, yet these correlations often fail to remain stable when evaluated on test data. When training machine learning models, such spurious correlations often lead to significantly degraded domain generalization performance~\citep{ye2024spurious, liu2025bridging}.

To improve model robustness across different groups under spurious correlations, many existing methods leverage group information to capture core features for prediction, including upweighting minority groups \citep{sagawa2019distributionally}, downsampling majority groups \citep{deng2023robust}, group distributionally robust optimization \citep{sagawa2019distributionally}, and progressive data expansion \citep{deng2023robust}. When group annotations are unknown, another line of works aims to infer latent groups or identify biased samples during training~\citep{nam2020learning, liu2021just, zhang2022correct, han2024improving}.

However, these methods for mitigating spurious correlation typically fail or become substantially less effective when (1) group labels are unavailable because obtaining them is costly or even infeasible, or (2) certain test-time groups are absent from the training data, situations in which spurious correlations can become unidentifiable and significantly more severe. Moreover, the sensitivity of these methods to the specification of group information and to changes in the set of groups is further exacerbated by the unreliability of group labels in practice: although commonly used spurious correlation benchmarks deliberately define groups as combinations of $(\text{label}, \text{spurious feature})$ tuples~\citep{sagawa2019distributionally, liu2015deep, arjovsky2019invariant}, in real-world settings such clean partitions are rare, and the available grouping often fails to faithfully capture the underlying sources of spurious correlation.
Consequently, this paper seeks to address the following question: 

\begin{center}
\begin{tcolorbox}[left=2pt, right=2pt, top=2pt, bottom=2pt, width=0.9\textwidth]
    \centering
    \emph{What serves as a more intrinsic and reliable signal than group information to provide a fundamental criterion for discerning spurious correlations?}
\end{tcolorbox}
\end{center}

We propose that the answer lies in the superclass label, i.e., a label higher in the semantic hierarchy than the task’s original class labels~\citep{ni2021superclass}.

\begin{wrapfigure}{r}{0.35\textwidth}  
\vspace{-10pt}
\centering
\small  

\begin{tabular}{@{}c@{\hspace{3mm}}c@{\hspace{3mm}}c@{}}
    \includegraphics[width=0.2\linewidth]{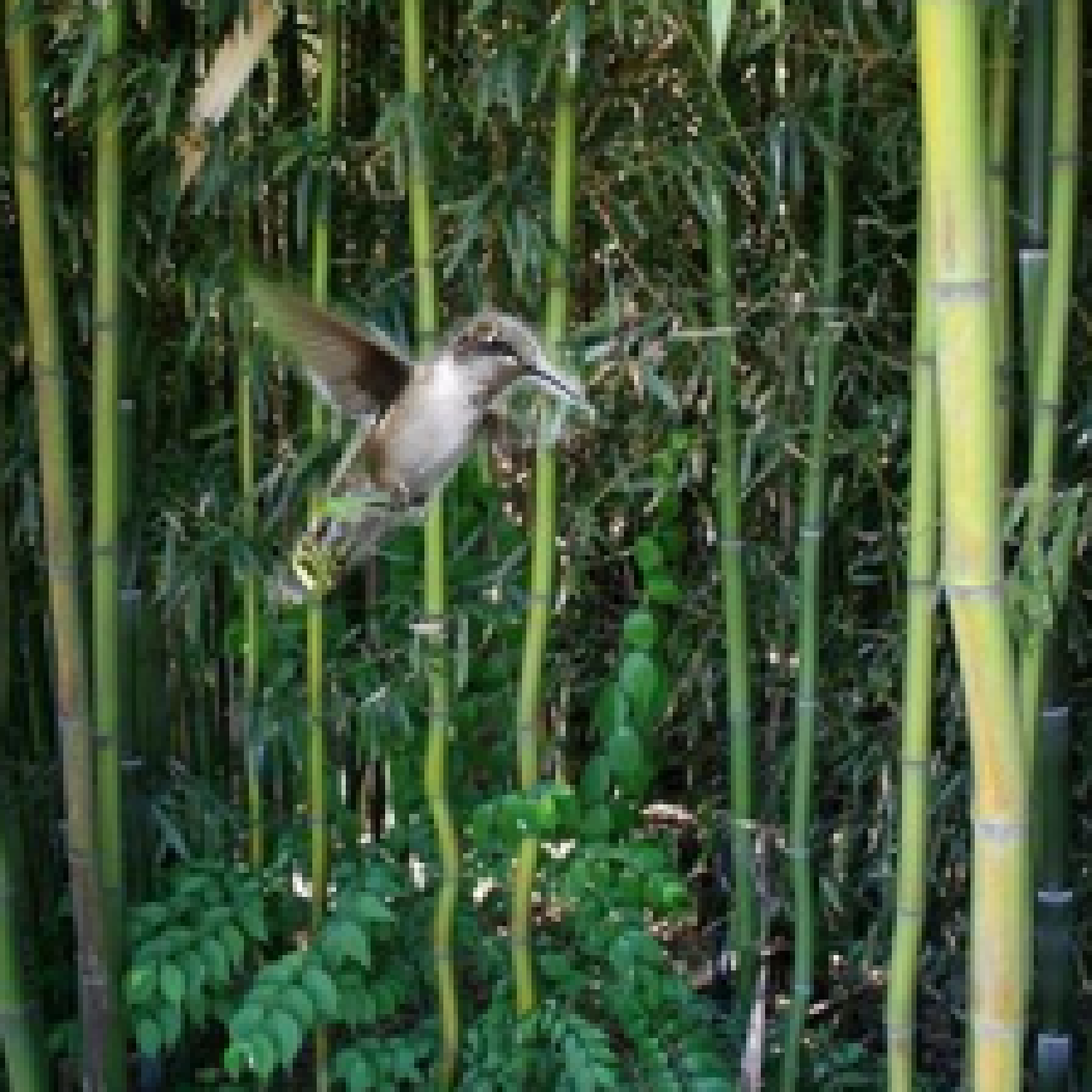} &
    \includegraphics[width=0.2\linewidth]{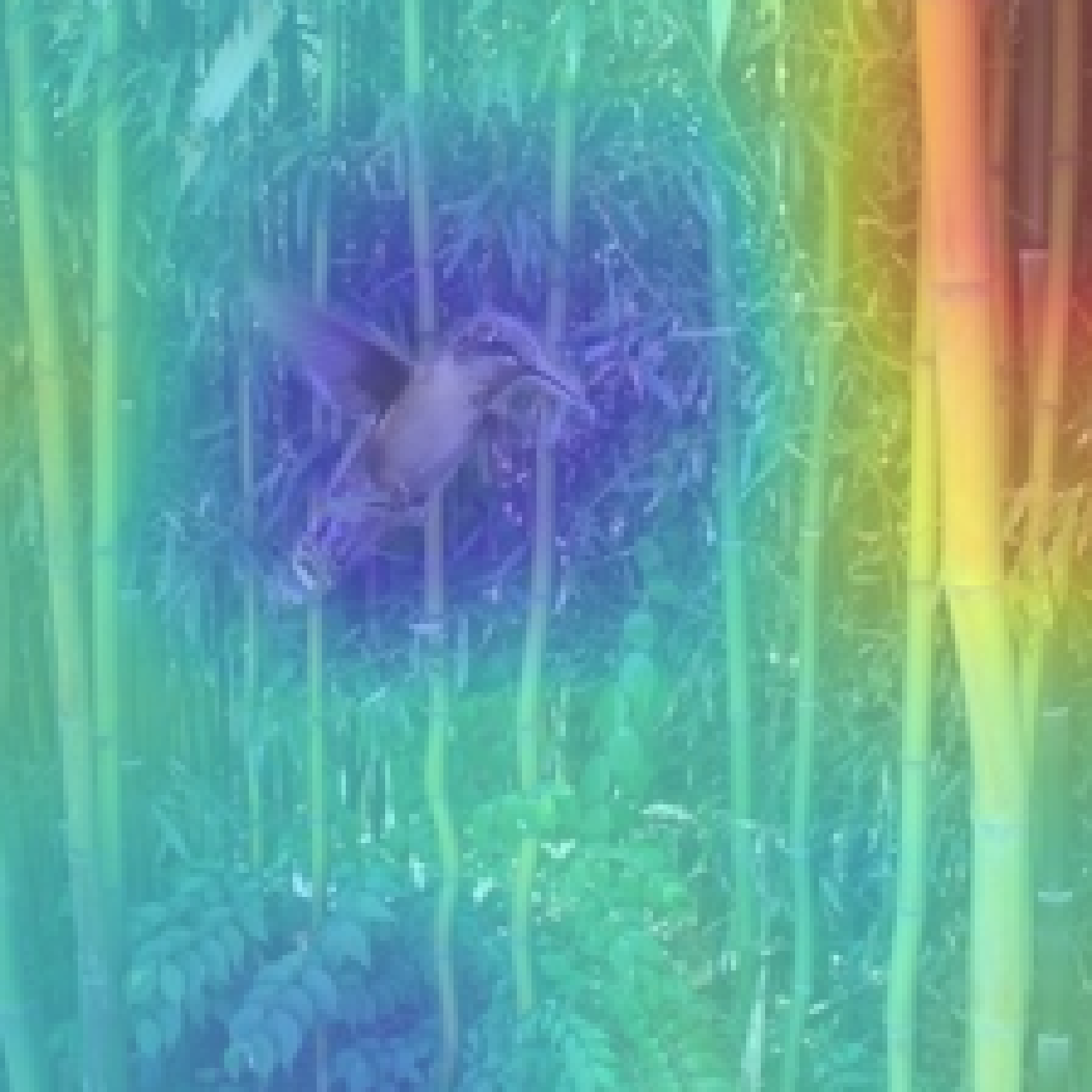} &
    \includegraphics[width=0.2\linewidth]{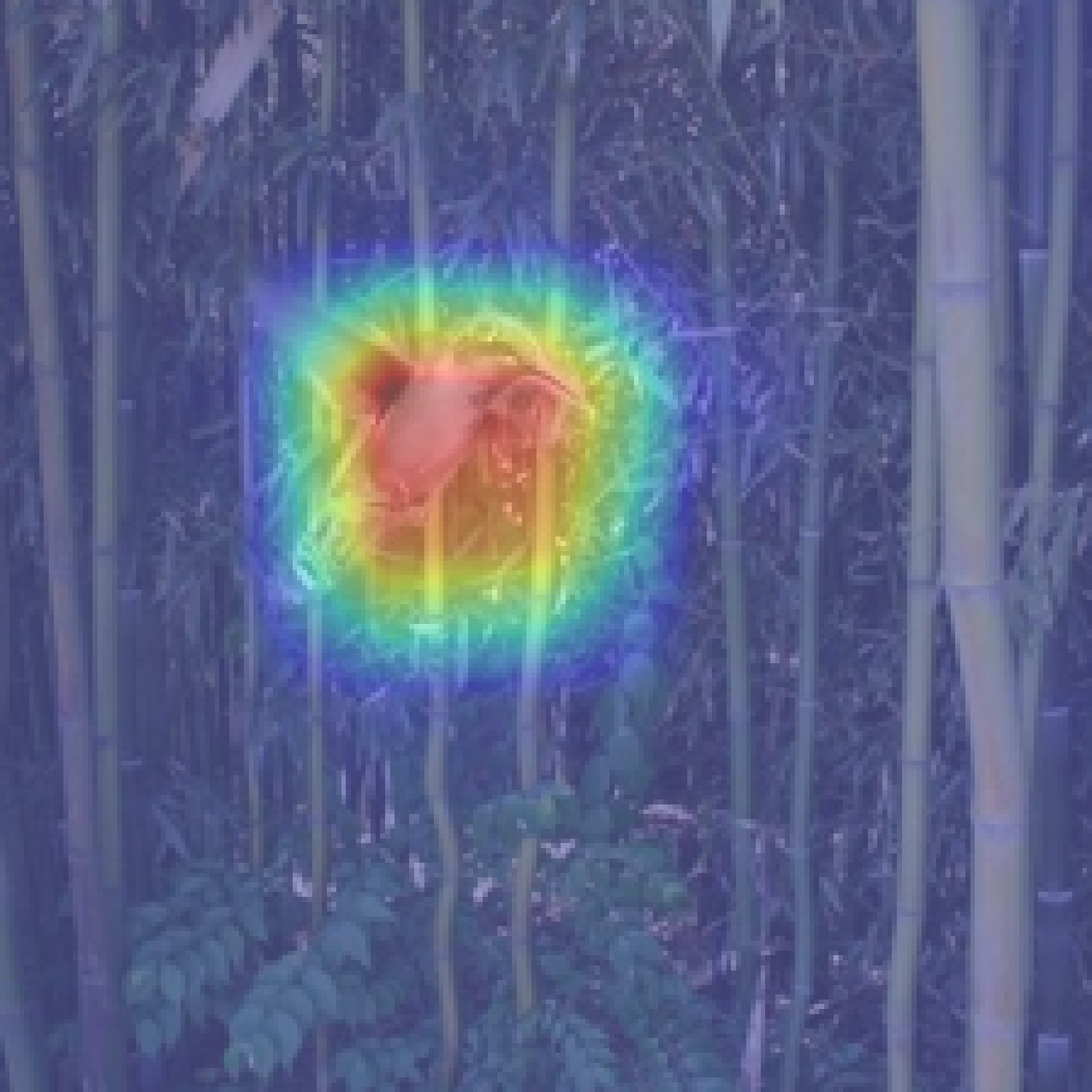} \\[0.4em]
    
    \includegraphics[width=0.2\linewidth]{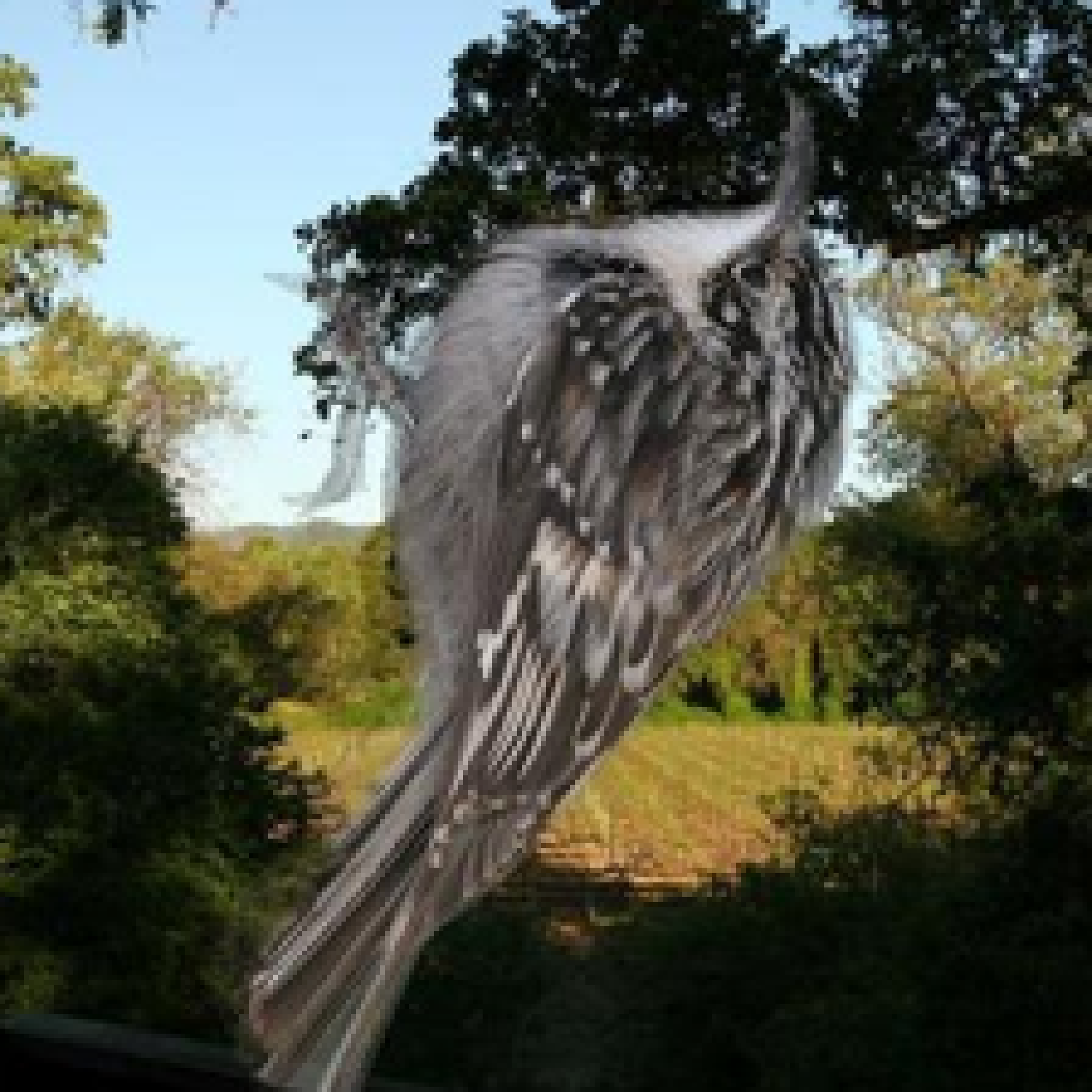} &
    \includegraphics[width=0.2\linewidth]{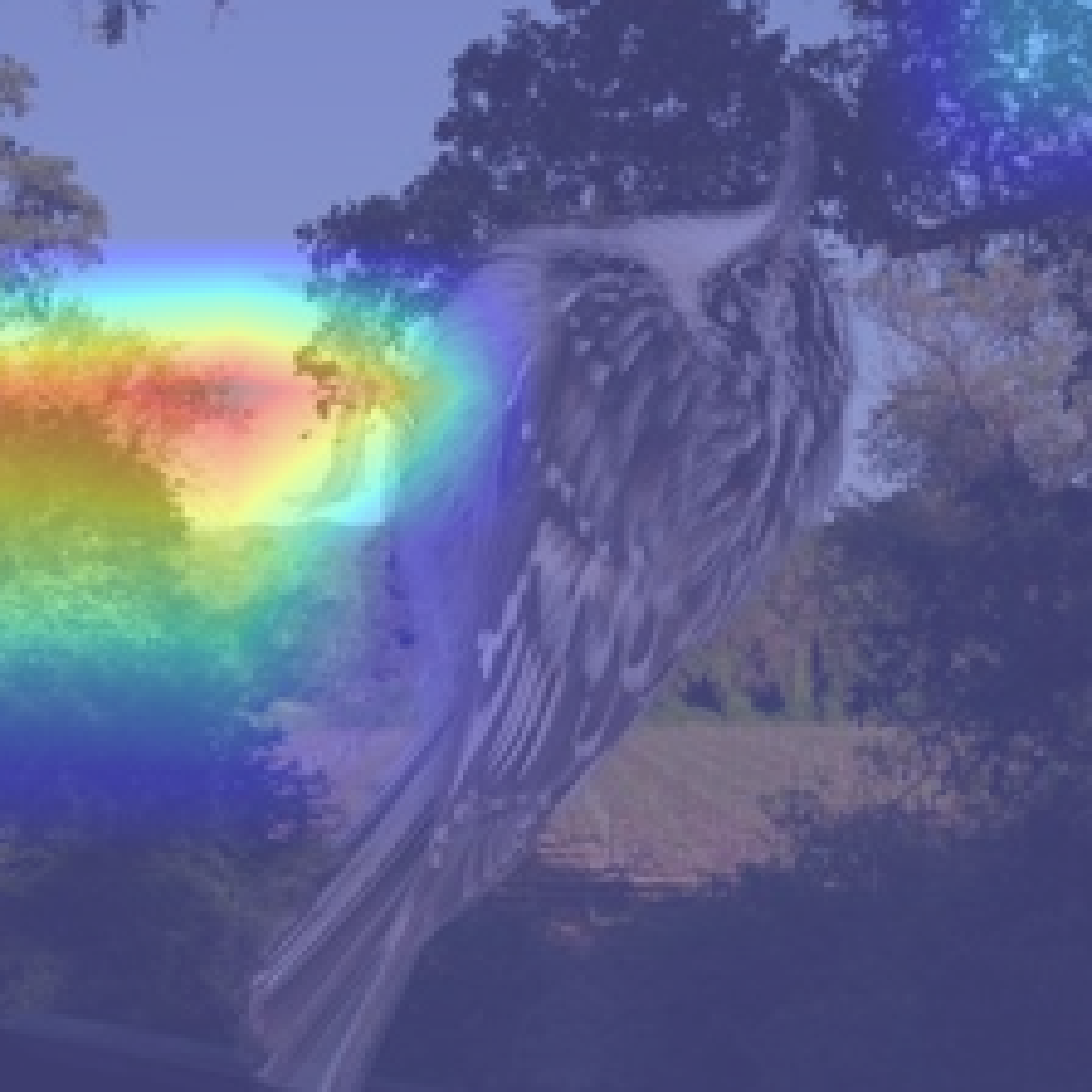} &
    \includegraphics[width=0.2\linewidth]{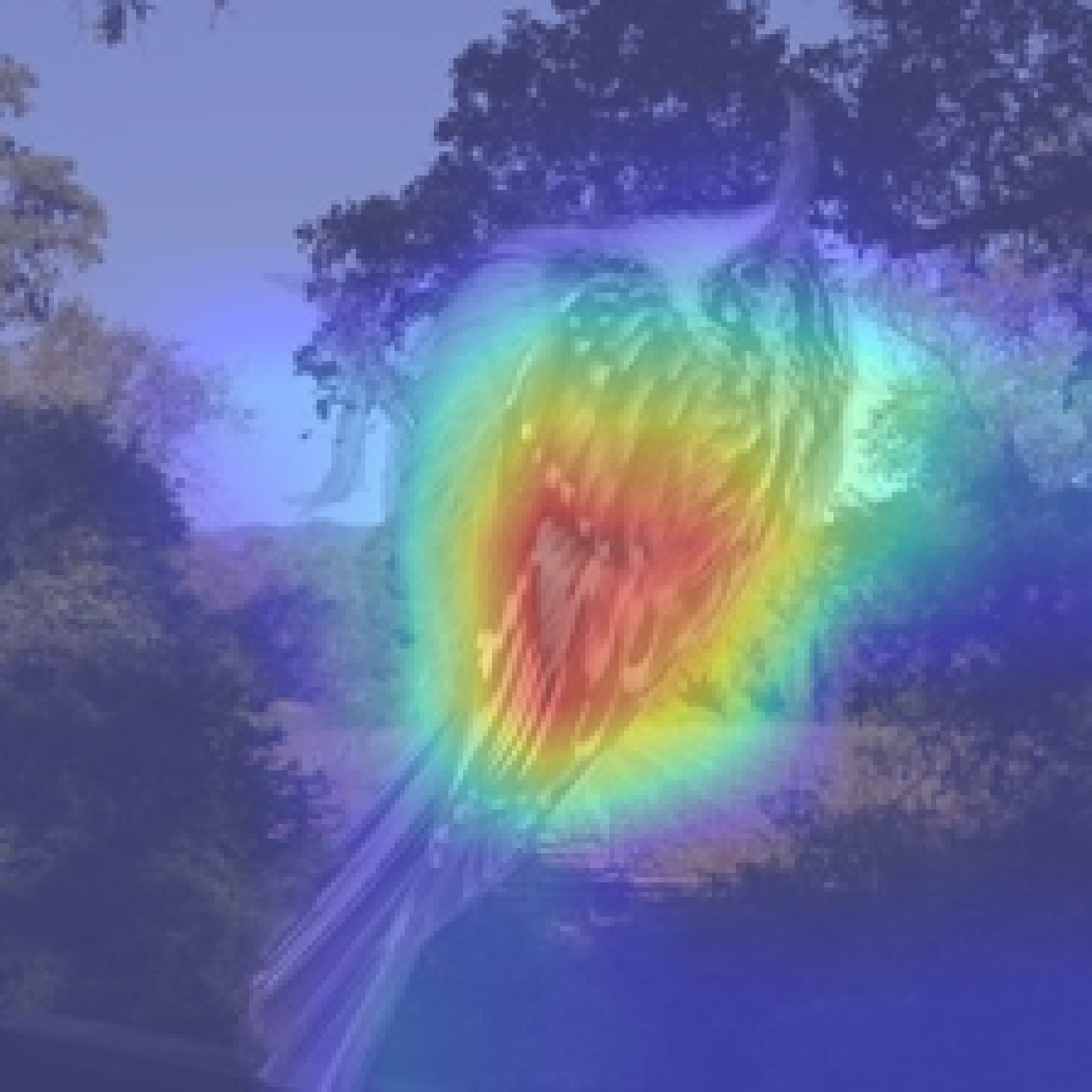} \\[0.4em]
    
    (a) Original & (b) ERM & (c) SupER \\
\end{tabular}

\caption{
GradCAM maps of ERM baseline and our SupER approach on Waterbirds dataset.
(a) original bird images, (b) ERM's GradCAM maps, and (c) SupER's GradCAM maps.
Our approach focuses on core features for classification, while ERM tends to rely on spurious features.
}
\label{fig:intro_fig}
\vspace{-5pt}
\end{wrapfigure}

Consider a thought experiment in waterbirds (\(Y=0\)) and landbirds (\(Y=1\)) image classification, where during training we only observe two groups of bird–background combinations, represented as the tuples \((\text{waterbird}, \text{water background})\) and \((\text{landbird}, \text{land background})\). As shown in experiments with the Waterbirds dataset \citep{sagawa2019distributionally, liu2021just}, a model trained using Empirical Risk Minimization (ERM) will tend to make predictions based on backgrounds, which leads to spurious correlations (see Figure~\ref{fig:intro_fig}(b)). However, if we redefine the task as a background classification (\(Y=0\) for water background, \(Y=1\) for land background), all numeric labels remain the same and the group set is also unchanged, yet the background features we focus on are now non-spurious. 
This indicates that the meaning of the label \(Y\) goes far beyond the numeric values \(\{0,1\}\). In particular, unlike group annotations, the superclass label (``bird'' versus ``background'' for these two tasks) captures the essential semantic content of \(Y\) by specifying what is being classified, and therefore serve as a key factor that can be exploited by appropriate representation and classifier design to mitigate spurious correlations.

\textbf{Our Contributions.} We propose a novel approach, \textbf{Sup}erclass-guided \textbf{E}mbedding \textbf{R}epresentation (SupER), which uses superclass semantic as a more intrinsic and reliable signal than group information for mitigating spurious correlations. Our contributions are summarized as follows:

\begin{itemize}[leftmargin=*]
    \item SupER combines feature disentanglement with superclass guidance from a pretrained vision–language model using gradient-based attention alignment, together with principled usage of different superclass-relevant and superclass-irrelevant features. To our knowledge, this is the first group-label-free framework to provide a formal extension and systematic evaluation for addressing group robustness against spurious correlations in settings where certain test-time groups are absent from the training data.

    \item We provide a theoretical analysis under a reasonable simplified setting, showing that our feature-usage strategy is minimax-optimal under superclass supervision.

    \item Extensive experiments indicate that SupER significantly outperforms baselines across multiple domain generalization benchmarks, especially when new groups appear at test time or spurious correlations become more severe.

    \item We further conduct visualizations and quantitative measurements to examine the behavior of SupER, showing that it effectively mitigates biases in the guiding model and achieves substantial improvements over it.
\end{itemize}

\section{Related Work}
Various methods have been proposed to mitigate ML models’ reliance on spurious features for better domain generalization, including invariant learning \citep{peters2016causal,arjovsky2019invariant,creager2021environment}, distributionally robust optimization \citep{sagawa2019distributionally,krueger2021out}, causal relationship studies \citep{makar2022causally,sun2021recovering}, fine-tuning methods \citep{kirichenko2022last,labonte2024towards}, contrastive learning \citep{zhang2022correct}, and the use of vision-language models \citep{petryk2022guiding,zhang2023diagnosing,yenamandra2023facts}. Among these, two lines of research are particularly relevant to our approach.

\textbf{Group robustness to spurious correlation.} To mitigate spurious correlations caused by group imbalance, when group labels are accessible, methods employ strategies such as upweighting minority groups \citep{sagawa2019distributionally}, downsampling majority groups \citep{deng2023robust}, group distributionally robust optimization \citep{sagawa2019distributionally}, and progressive data expansion \citep{deng2023robust}, with the shared goal of balancing performance across groups. When group information is not available, another line of work attempts to infer group labels or identify biased samples \citep{nam2020learning,liu2021just,zhang2022correct,yenamandra2023facts,han2024improving}, or leverage auxiliary information such as knowledge of spurious attributes \citep{puli2021out,puli2022nuisances, makar2022causally}. However, these methods become ineffective when the sets of groups across training and test domains differ, as spurious correlations can not be reliably identified.

\textbf{Feature learning through disentangled representation.} Disentangled representation learning aim to separate independent generative factors of data variation \citep{bengio2013representation}. Building on this principle, various approaches have sought to disentangle representations of $X$ into core and spurious features, and then use only core features for prediction \citep{lee2021learning,zhang2022towards,yang2022chroma}. Additionally, sparsity-based methods \citep{lachapelle2023synergies,fumero2023leveraging} and diverse classifier training \citep{teney2022evading,pagliardini2022agree} have shown effectiveness in feature disentanglement and enhancing generalization. However, these approaches still rely on group or environment annotations, or become less effective when the test domain contains groups that do not appear during training.

\section{Method}
\label{sec:method}
\subsection{Problem Setup}
\label{subsec:problem setup}

We study a prediction task with inputs \(X\in\mathcal{X}\) and labels \(Y\in\mathcal{Y}\). 
The training dataset \(\mathcal{D}_s\) (drawn from \(P_s\)) and test dataset \(\mathcal{D}_t\) (drawn from \(P_t\)) consist of groups collected in the sets \(\mathcal{G}_s\) and \(\mathcal{G}_t\), with each group specified by a label \(y\in\mathcal{Y}\) and an attribute \(z\in\mathcal{Z}\). 
When the mixture weights of these groups differ, \(P_s\neq P_t\), and \(z\) may correlate spuriously with \(y\). 
We assume that the label set \(\mathcal{Y}\) is shared between \(\mathcal{D}_s\) and \(\mathcal{D}_t\). 
Our goal is to leverage the superclass label \(y^{\text{super}}\), defined as a label higher in the semantic hierarchy than the task’s original class labels \(\mathcal{Y}\)~\citep{ni2021superclass}, to learn a predictor on \(\mathcal{D}_s\) that maximizes worst group accuracy on \(\mathcal{D}_t\).
Unlike prior work that often assumes \(\mathcal{G}_s=\mathcal{G}_t\), we consider a more general setting: (1) \(\mathcal{G}_s\) and \(\mathcal{G}_t\) may differ, allowing unseen groups at test time (or equivalently, missing groups during training). In this case, the spurious correlation between \(z\) and \(y\) in \(\mathcal{D}_s\) can become more severe, or \(z\) may fail to faithfully capture the underlying sources of spurious correlation.
(2) no group information is available during training.

\subsection{Feature Disentanglement with Superclass Guidance} \label{sec:superclass-disentangle}

\textbf{Superclass-guided feature disentanglement.}  
For each $(\mathbf{x}, y) \in \mathcal{D}_{s}$, we use a $\beta$-VAE \citep{higgins2017beta} to facilitate the disentanglement of the latent feature representation $\mathbf{z}$ of $\mathbf{x}$ by maximizing
\begin{align}
\label{eq:betaVAE}
\mathcal{L}^{\text{Beta}}_{\theta, \phi}(\mathbf{x})
  = \mathbb{E}_{\mathbf{z}\sim q_{\phi}(\mathbf{z}|\mathbf{x})}[\log p_{\theta}(\mathbf{x}|\mathbf{z})] - \beta D_{KL}(q_{\phi}(\mathbf{z}|\mathbf{x}) \,\|\, p(\mathbf{z})),
\end{align}
where $p_{\theta}(\mathbf{x}|\mathbf{z})$ is modeled by a decoder, $q_{\phi}(\mathbf{z}|\mathbf{x})$ approximates the posterior distribution as Gaussian $\mathcal{N}(\mathbf{z}|\boldsymbol{\mu}_{\phi}(\mathbf{x}), \boldsymbol{\Sigma}_{\phi}(\mathbf{x}))$, and the prior $p(\mathbf{z})$ is the standard normal distribution $\mathcal{N}(\mathbf{0}, \mathbf{I})$. 
This objective promotes feature disentanglement by encouraging $\mathbf{z}$ to capture independent generative factors of $\mathbf{x}$.

Gradient-based visual attention have been shown to provide visual explanations by highlighting regions that the model attends to during inference \citep{selvaraju2017grad, Chattopadhay_2018}. Meanwhile, CLIP \citep{radford2021learning} possesses strong capability in mapping semantic information from superclass text descriptions into a shared latent space with images.
Therefore, to further incorporate superclass guidance from \(y^{\text{super}}\), our insight is to leverage CLIP’s attention mechanism to guide the partition 
\(\mathbf{z} = [\mathbf{z}_{\mathrm{rel}};\, \mathbf{z}_{\mathrm{irr}}]\),  
such that  
\(\mathbf{z}_{\mathrm{rel}}\) captures superclass-relevant information  
and  
\(\mathbf{z}_{\mathrm{irr}}\) captures superclass-irrelevant information.

Formally, for any $(\mathbf{x}, y) \in \mathcal{D}_{s}$ and text prompt $\mathbf{T}$, we compute a normalized gradient-based attribution map  
\(L_{\text{CLIP}}^{\mathbf{T}}(\mathbf{x})\)  
that reveals the regions CLIP attends to when classifying $\mathbf{x}$ as $\mathbf{T}$ (details in Appendix~\ref{subsec:A.1}).  
Since $\mathbf{z}_{\mathrm{rel}}$ and $\mathbf{z}_{\mathrm{irr}}$ are intended to extract features relevant and irrelevant to the superclass, respectively, the text prompts $\mathbf{T}$ must correspond to these semantic aspects.  
Specifically, we use $m$ text prompts  
\(\mathbf{T}^1,\dots,\mathbf{T}^{m}\)  
that are semantically aligned with the superclass label \(y^{\text{super}}\) (e.g., “\(\mathtt{a/an\ [superclass]}\)”; see Table~\ref{tab:prompt_variants}), and average their attribution maps to obtain
\( L_{\text{CLIP}}^{\mathbf{T}_{\mathrm{rel}}}(\mathbf{x}) = \frac{1}{m}\sum_{i=1}^{m} L_{\text{CLIP}}^{\mathbf{T}^i}(\mathbf{x}), \)
which guides the extraction of superclass-relevant features.  
For attention guidance of superclass-irrelevant features, given that spurious features are unknown in our setting and previous studies \citep{nie2024out} have shown CLIP's limitations in understanding negative prompts (e.g., the meaning of the word ``not''), we instead define \( L_{\text{CLIP}}^{\mathbf{T}_{\mathrm{irr}}}(\mathbf{x}) = \mathbf{J}- L_{\text{CLIP}}^{\mathbf{T}_{\mathrm{rel}}}(\mathbf{x}) \) as the attribution map, where $\mathbf{J}$ represents an all-ones matrix.

To align the CLIP attribution maps with the attribution maps derived from $\mathbf{z}_{\mathrm{rel}}$ and $\mathbf{z}_{\mathrm{irr}}$, we train two different classifiers $\omega_{\mathrm{rel}}$ and $\omega_{\mathrm{irr}}$ on $\boldsymbol{\mu}_{\mathrm{rel}}$ and $\boldsymbol{\mu}_{\mathrm{irr}}$ (i.e., the means of $\mathbf{z}_{\mathrm{rel}}$ and $\mathbf{z}_{\mathrm{irr}}$), respectively, by minimizing the cross-entropy losses $\mathcal{L}^{\text{CE}}_{\phi, \omega_{\mathrm{rel}}}(\mathbf{x}, y)$ and $\mathcal{L}^{\text{CE}}_{\phi, \omega_{\mathrm{irr}}}(\mathbf{x}, y)$. 
For each sample $(\mathbf{x}, y)$, we then compute gradient-based attribution maps  
\(L_{\phi,\omega_{\mathrm{rel}}}(\mathbf{x}, y)\)  
and  
\(L_{\phi,\omega_{\mathrm{irr}}}(\mathbf{x}, y)\) with respect to the true label $y$ 
(details in Appendix~\ref{subsec:A.2}).  
Finally, superclass-guided feature disentanglement is fulfilled by minimizing the alignment regularization loss:
\begin{align}
\mathcal{L}^{\text{ATT}}_{\phi, \omega_{\mathrm{rel}},\omega_{\mathrm{irr}}}(\mathbf{x},y) 
= \big\| L_{\text{CLIP}}^{\mathbf{T}_{\mathrm{rel}}}(\mathbf{x}) -
        L_{\phi,\omega_{\mathrm{rel}}}(\mathbf{x}, y) \big\|_F^2  + \big\| L_{\text{CLIP}}^{\mathbf{T}_{\mathrm{irr}}}(\mathbf{x}) -
        L_{\phi,\omega_{\mathrm{irr}}}(\mathbf{x}, y) \big\|_F^2 .
\label{eq:7}
\end{align}

The combination of \(\mathcal{L}^{\text{Beta}}_{\theta, \phi}(\mathbf{x})\) and \(\mathcal{L}^{\text{ATT}}_{\phi, \omega_{\mathrm{rel}},\omega_{\mathrm{irr}}}(\mathbf{x},y)\) successfully achieves the feature disentanglement of \(\mathbf{x}\), the separation between \(\mathbf{z}_{\mathrm{rel}}\) and \(\mathbf{z}_{\mathrm{irr}}\), as well as the gradient-based attention supervision based on superclass semantic information.

\begin{figure*}[t]
\vskip 0.2in
\begin{center}
% \centerline{ \phantom{\rule{8cm}{6cm}}}
\includegraphics[width = 0.75\linewidth]{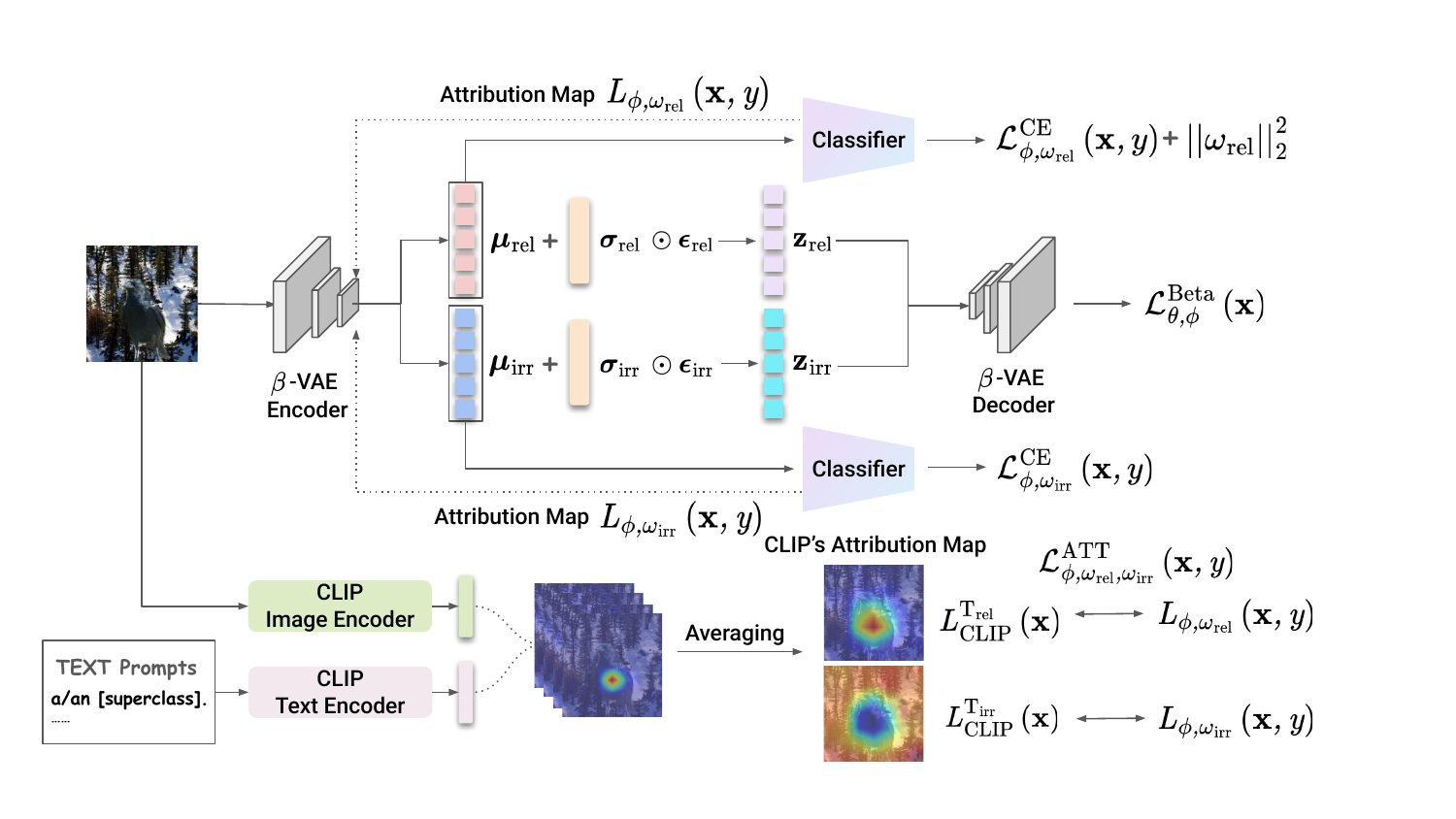}
\caption{Overview of SupER architecture. The model processes each input image $(\mathbf{x}, y)$ through four key components:  
(1) A $\beta$-VAE architecture that optimizes $\mathcal{L}^{\text{Beta}}_{\theta, \phi}(\mathbf{x})$ to disentangle the input into latent features $\mathbf{z} = [\mathbf{z}_{\mathrm{rel}}; \mathbf{z}_{\mathrm{irr}}]$.  
(2) Two classifiers, $\omega_{\mathrm{rel}}$ and $\omega_{\mathrm{irr}}$, are trained separately to predict the label $y$ from $\boldsymbol{\mu}_{\mathrm{rel}}$ (mean of $\mathbf{z}_{\mathrm{rel}}$) and $\boldsymbol{\mu}_{\mathrm{irr}}$ (mean of $\mathbf{z}_{\mathrm{irr}}$), by optimizing $\mathcal{L}^{\text{CE}}_{\phi, \omega_{\mathrm{rel}}}(\mathbf{x}, y)$ and $\mathcal{L}^{\text{CE}}_{\phi, \omega_{\mathrm{irr}}}(\mathbf{x}, y)$, respectively.
(3) A CLIP-guided mechanism that generates attribution maps through text-image alignment, which guides $\mathbf{z}_{\mathrm{rel}}$ and $\mathbf{z}_{\mathrm{irr}}$ to capture superclass-relevant and irrelevant features via $\mathcal{L}^{\text{ATT}}_{\phi,\omega_{\mathrm{rel}},\omega_{\mathrm{irr}}}(\mathbf{x})$.  
(4) An $L_2$ regularization term $\|\omega_{\mathrm{rel}}\|^2_2$ that encourages the utilization of all superclass features during classification.}
\label{fig:placeholder}
\end{center}
\vskip -0.2in
\end{figure*}

\textbf{Robustness to inherent biases in the guiding model.}
While CLIP may exhibit inherent biases toward spurious correlations~\citep{agarwal2021evaluating,yang2023mitigating} (see Table~\ref{tab:clip_vs_super}), both superclass-level guidance and feature disentanglement play crucial roles in mitigating such biases. 
First, since the superclass label \(y^{\text{super}}\) is shared across all samples \((\mathbf{x},y)\in\mathcal{D}_{s}\), it does not provide any discriminative features that contribute to distinguishing specific task labels.
This superclass-level guidance therefore avoids CLIP’s severe spurious correlations that arise when conditioning on the fine-grained label \(y\in\mathcal{Y}\).
Second, because the $\beta$-VAE encourages independent latent factors with semantic structure~\citep{wang2024disentangled}, once a latent dimension of \(\mathbf{z}\) predominantly represents a semantic component under CLIP’s guidance, occasional attribution errors from CLIP are effectively overridden by the dominant semantic signal (see Figures~\ref{fig:gradcam_vis}, \ref{fig:gradcam_vis_1}, and \ref{fig:gradcam_vis_clipbias}). Overall, this weak form of guidance from CLIP significantly reduces the impact of CLIP’s own biases while granting SupER sufficient autonomy to learn robust features. Therefore, combined with the theoretically supported minimax-optimal feature-usage strategy presented in Section~\ref{sec:theory}, as shown in Table~\ref{tab:clip_vs_super}, SupER’s performance goes far beyond the limitations of the guiding model itself.

\subsection{Principled feature-usage with Theoretical Support}
\label{sec:theory}

\begin{algorithm}[t]
{\fontsize{9.5pt}{10pt}\selectfont
   \caption{SupER Model Training}
   \label{alg:training_dafd}
\begin{algorithmic}
   \STATE {\bfseries Input:} $\mathcal{D}_{\text{s}}$, initial model parameters $\phi, \theta, \omega_{\mathrm{rel}}, \omega_{\mathrm{irr}}$, learning rate $\eta$, epochs $T$, batch size $B$, $\lambda_1, \lambda_2, \lambda_3$
   \FOR{epoch $t = 1$ {\bfseries to} $T$}
       \STATE Shuffle $\mathcal{D}_{\text{s}}$ into mini-batches $\{\mathcal{B}\}$ with batch size $B$
       \FOR{each mini-batch $\mathcal{B}$}
           \FOR{each sample $(\mathbf{x}, y) \in \mathcal{B}$}
               \STATE Compute $\mathcal{L}^{\text{Beta}}_{\theta, \phi}(\mathbf{x})$ according to Equation~(\ref{eq:betaVAE})
               \STATE Compute $\mathcal{L}^{\text{ATT}}_{\phi, \omega_{\mathrm{rel}}, \omega_{\mathrm{irr}}}(\mathbf{x}, y)$ according to Equation~(\ref{eq:7})
               \STATE Compute cross-entropy losses $\mathcal{L}^{\text{CE}}_{\phi, \omega_{\mathrm{rel}}}(\mathbf{x}, y)$ and $\mathcal{L}^{\text{CE}}_{\phi, \omega_{\mathrm{irr}}}(\mathbf{x}, y)$
           \ENDFOR
           \STATE Compute
           \[
            \mathcal{L}(\mathcal{B})=\sum_{(\mathbf{x}, y) \in \mathcal{B}} \left( 
           \mathcal{L}^{\text{CE}}_{\phi, \omega_{\mathrm{rel}}}(\mathbf{x}, y) + 
           \mathcal{L}^{\text{CE}}_{\phi, \omega_{\mathrm{irr}}}(\mathbf{x}, y) - 
           \lambda_1 \mathcal{L}^{\text{Beta}}_{\theta, \phi}(\mathbf{x}) + 
           \lambda_2 \mathcal{L}^{\text{ATT}}_{\phi, \omega_{\mathrm{rel}}, \omega_{\mathrm{irr}}}(\mathbf{x}, y) + 
           \lambda_3 \|\omega_{\mathrm{rel}}\|_2^2
           \right)
           \]
           \STATE Update parameters:
           \(
           \phi, \theta, \omega_{\mathrm{rel}}, \omega_{\mathrm{irr}} \leftarrow \phi, \theta, \omega_{\mathrm{rel}}, \omega_{\mathrm{irr}} - \eta \nabla_{\phi, \theta, \omega_{\mathrm{rel}}, \omega_{\mathrm{irr}}} \mathcal{L}(\mathcal{B})
           \)
       \ENDFOR
   \ENDFOR
\end{algorithmic}
}
\end{algorithm}

Unlike existing approaches that rely on group labels to balance risks across groups~\citep{sagawa2019distributionally}, or on environment labels to enforce invariant feature learning~\citep{arjovsky2019invariant}, SupER provides a new feature-usage strategy that leverages the more intrinsic superclass label. 
Building on the superclass-guided feature disentanglement in Section~\ref{sec:superclass-disentangle}, classifier heads $\omega_{\mathrm{rel}}$ and $\omega_{\mathrm{irr}}$ respectively make predictions from \(\mathbf{z}_{\mathrm{rel}}\) and \(\mathbf{z}_{\mathrm{irr}}\) by exploiting those components that are strongly correlated with the label \(y\) on \(\mathcal{D}_{s}\). Our objective, therefore, is to determine an appropriate strategy for using different superclass-relevant and superclass-irrelevant features, so that the resulting predictor achieves optimal performance in the worst case in our setting.

For simplicity, we denote by \(Z_1, Z_2 \in \mathbb{R}^p\) two distinct latent features of \(X\) extracted from \(q_{\phi}\), and let \(Y \in \mathbb{R}\) be the label. 
Under Assumption~\ref{ass:lin} on the linear generative model, we analyze two specific scenarios to provide insight into how \(\omega_{\mathrm{rel}}\) and \(\omega_{\mathrm{irr}}\) should utilize different latent features for prediction.

\begin{itemize}[leftmargin=*]
    \item (1) The generating index is known to be $c=1$. In this case, $Z_1$ is treated as a superclass-relevant feature that generates $Y$, while $Z_2$ is treated as a superclass-irrelevant feature. 
This interpretation follows from the fact that a feature unrelated to the superclass \(y^{\text{super}}\) cannot serve as a core feature. 

    \item (2) The generating index is uncertain with $c\in\{1,2\}$. In this case, both $Z_1$ and $Z_2$ are viewed as distinct superclass-relevant features. 
This reflects the general setting in Section~\ref{subsec:problem setup}: when both features aligned with the superclass \(y^{\text{super}}\), it is unknown in this more general case whether their relationship to the label remains consistent across \(P_s\) and \(P_t\).
\end{itemize}

Further, let $Z_1^s,Z_2^s \in \mathbb{R}^{n\times p}$ denote the fixed matrices obtained by stacking the $n$ training samples of $Z_1$ and $Z_2$ row-wise, respectively. Define the empirical second moments and cross-moment under the training distribution \(P_s\) by \(\hat{\Sigma}_1^s=(Z_1^s)^\top Z_1^s/n\), \(\hat{\Sigma}_2^s=(Z_2^s)^\top Z_2^s/n\), and \(\hat{\Sigma}_{1,2}^s=(Z_1^s)^\top Z_2^s/n=(\hat{\Sigma}_{2,1}^s)^\top\). For the test distribution \(P_t\), let \(\Sigma_1^t=\mathbb{E}_{Z_1\sim P_t}[Z_1Z_1^\top]\), \(\Sigma_2^t=\mathbb{E}_{Z_2\sim P_t}[Z_2Z_2^\top]\), and \(\Sigma_{1,2}^t=\mathbb{E}_{Z_1, Z_2\sim P_t}[Z_1Z_2^\top]=(\Sigma_{2,1}^t)^\top\). Then, Assumptions~\ref{ass:spur}-\ref{ass:commute} formalize the notion of spurious correlation through the strong correlation between features during training (and consequently also between the features and the label), whereas this correlation vanishes under the test distribution.

\begin{assumption}\label{ass:lin}
There exists an index \(c\in\{1,2\}\) and $\boldsymbol{\beta}_c\in\mathbb{R}^p$ such that
$Y=Z_c^\top\boldsymbol{\beta}_c+\varepsilon$, with $\varepsilon\sim\mathcal{N}(0,\sigma^2)$ independent of $(Z_1,Z_2)$.
\end{assumption}

\begin{assumption}\label{ass:spur}
Under the training distribution $P_s$, 
$\hat{\Sigma}_{1,2}^s=\rho\,(\hat{\Sigma}_1^s)^{1/2}(\hat{\Sigma}_2^s)^{1/2}$
with $|\rho|\in(0,1)$ close to 1, while under the test distribution $P_t$ we have $\Sigma_{1,2}^t=0$.
\end{assumption}

\begin{assumption}\label{ass:commute}
$\hat{\Sigma}_1^s,\hat{\Sigma}_2^s,\Sigma_1^t,\Sigma_2^t$ commute with each other, with positive eigenvalues
$\{d_{1,i}^s\}_{i=1}^p$, $\{d_{2,i}^s\}_{i=1}^p$, $\{d_{1,i}^t\}_{i=1}^p$, and $\{d_{2,i}^t\}_{i=1}^p$ (the same $i$ refers to the same common eigen-direction).
\end{assumption}

Consider three prediction strategies: 
$S_1$ predicts with $Z_1$, $S_2$ with $Z_2$, and $S_{1,2}$ with $Z=(Z_1,Z_2)$. 
Then, Theorem~\ref{thm:informal} provides an intuitive characterization of the optimal
feature-usage strategy.  
Specifically, among the features that exhibit strong correlations with \(Y\) during training, SupER should  
(i) discard all superclass-irrelevant features (part~(1)), and  
(ii) exploit as diverse a set of superclass-relevant features as possible (part~(2)).  This selection rationale is intuitive: the superclass label acts as a semantic prior regarding feature validity, and SupER embodies a principle of elimination under certainty and diversification under uncertainty.
Consequently, SupER follows this strategy by excluding the classifier $\omega_{\mathrm{irr}}$ during evaluation, and by adding an $L_2$ penalty $\|\omega_{\mathrm{rel}}\|_2^2$ on the classifier head $\omega_{\mathrm{rel}}$ during training, which encourages smoother and more evenly distributed weights across diverse informative superclass-relevant features~\citep{hastie2009elements}.

To summarize, SupER effectively leverages the semantic information in the superclass label \(y^{\text{super}}\) through CLIP-based guidance, and combines superclass-guided feature disentanglement with a theoretically supported form of feature usage.  
As a result, under the setting of Section~\ref{subsec:problem setup}, SupER attains robustness to spurious correlations without relying on group signals. The detailed training algorithm is presented in Algorithm~\ref{alg:training_dafd}, and the complete pipeline is illustrated in Figure~\ref{fig:placeholder}.

\begin{theorem}[Informal result (formally in Theorems~\ref{thm:formal-restatement}-\ref{thm:random-formal})]
\label{thm:informal}
Let $\mathcal{E}(S;c,\boldsymbol{\beta}_c)$ denote the excess risk under $P_t$ of the ordinary least-squares predictor trained on $P_s$, given strategy $S$ and generating index $c$. Under Assumptions~\ref{ass:lin}-\ref{ass:commute}:
\begin{enumerate}[label=\arabic*., leftmargin=*, nosep]
  \item (Discard superclass-irrelevant features despite strong correlations with \(Y\) on \(P_s\).) If $c=1$ is known, then
\[
\min_{S\in\{S_1,S_{1,2}\}}\mathcal{E}(S;1,\boldsymbol{\beta}_1)
= \frac{\sigma^2}{n}\sum_{i=1}^p \frac{d_{1,i}^t}{d_{1,i}^s},
\]
achieved by \(S=S_1\).
\item (Retain all superclass-relevant features with strong correlations with \(Y\) on \(P_s\).) If $c\in\{1,2\}$ is uncertain, let
$\mathcal{B}_r^{(c)}=\{\boldsymbol{\beta}\in\mathbb{R}^p:\boldsymbol{\beta}^\top\Sigma_c^t\boldsymbol{\beta}=r\}$, then
there exists a constant $C$ such that whenever $nr>C$,
\[ \begin{aligned}
\min_{S\in\{S_1,S_2,S_{1,2}\}}\ \max_{c\in\{1,2\}}\ \max_{\boldsymbol{\beta}_c\in\mathcal{B}_r^{(c)}}\ \mathcal{E}(S;c,\boldsymbol{\beta}_c) 
=\frac{\sigma^2}{n(1-\rho^2)}\sum_{i=1}^p\!\left(\frac{d_{1,i}^t}{d_{1,i}^s}+\frac{d_{2,i}^t}{d_{2,i}^s}\right)
\end{aligned} \]
achieved by \(S=S_{1,2}\).
\end{enumerate}
\end{theorem}

\begin{remark}
Unlike prior works on spurious correlations \citep{arjovsky2019invariant,zhou2021examining, wang2022causal, wang2024effect} that study how the presence of spurious and core features of \(X\) affects generalization, our analysis makes a finer partition using superclass information. 
The part~(1) of Theorem~\ref{thm:informal} shares a similar spirit with prior findings that spurious features can harm generalization, but specifically restricts the scope to superclass-relevant features.
More importantly, to our knowledge, the minimax analysis in part~(2) of Theorem~\ref{thm:informal} regarding the use of different superclass-relevant features has not appeared in earlier spurious correlation theory, and this constitutes the main message conveyed by our theorem.
\end{remark}

\section{Experiments}
\label{sec:exp}
In this section, we empirically evaluate SupER on a diverse set of datasets characterized by distinct types of spurious features and underlying group structures. We compare the performance of SupER with different baseline models that make use of group information to different extents. The code for SupER is available at \url{https://github.com/crliuuuuu/SupER}.

\subsection{Datasets and Baselines}
\textbf{Datasets.} We evaluate SupER on datasets that cover diverse spurious correlation structures caused by group imbalance, including \textbf{Waterbirds-95\%}~\citep{sagawa2019distributionally}, \textbf{Waterbirds-100\%}~\citep{petryk2022guiding}, \textbf{MetaShift}~\citep{liang2022metashift, phan2024controllable}, \textbf{Spawrious}~\citep{lynch2023spawrious}, and \textbf{SpuCo Dogs}~\citep{joshi2023towards}.
Waterbirds-95\% and SpuCo Dogs exhibit a strong correlation ($\sim$95\%) between background and label during training. 
Waterbirds-100\% represents an extreme case where two groups are entirely absent during training. 
MetaShift consists of four subsets, each introducing different degrees of spurious correlation and testing on groups unseen in training. 
Spawrious contains six subsets and is used to assess performance under two correlation regimes: 
one-to-one, where each class correlates with a unique attribute, 
and many-to-many, where multiple classes correlate with multiple attributes.
Detailed dataset splits are provided in Appendix~\ref{subapp:ds}.
We defer additional evaluation on more datasets to Appendix~\ref{app:internal_spurious}.

\begin{table*}[b]
  \centering
  {\fontsize{8pt}{8pt}\selectfont
  \caption{Mean $\pm$ std of worst and average group accuracy (\%) for Waterbirds datasets. 
  As a “ceiling” reference with spurious features fully removed, we include results from ERM trained and evaluated on a bird-only region with backgrounds removed. It achieves 85.6$_{\pm0.4}$\% worst group accuracy on Waterbirds-95\% and 83.0$_{\pm0.5}$\% on Waterbirds-100\%. Several reported baselines exceed this reference on Waterbirds-95\%, suggesting potential benchmark overfitting.
  \textbf{Bold} indicates the best across all baselines; \underline{Underlined} indicates the best among methods without group information.
}
  \label{tab:wb_results}
  \begin{tabular*}{\textwidth}{@{\extracolsep{\fill}} l  c  c  c  c  c  c}
    \toprule
    \multirow{2}{*}{Method}
      & \multirow{2}{*}{\shortstack{Group\\Info}}
      & \multirow{2}{*}{\shortstack{Train\\Twice}}
      & \multicolumn{2}{c}{Waterbirds-95\%}
      & \multicolumn{2}{c}{Waterbirds-100\%} \\
    \cmidrule(lr){4-5} \cmidrule(lr){6-7}
      &  & 
      & Worst & Avg 
      & Worst & Avg \\
    \midrule
    ERM          & $\times$     & $\times$     
                 & 64.9$_{\pm1.5}$  & 90.7$_{\pm1.0}$
                 & 46.4$_{\pm6.9}$  & 74.8$_{\pm3.0}$ \\
    CVaR\,DRO    & $\times$     & $\times$     
                 & 73.1$_{\pm7.1}$ & 90.7$_{\pm0.7}$
                 & 58.0$_{\pm2.2}$ & 79.0$_{\pm1.2}$ \\
    LfF          & $\times$     & $\times$     
                 & 79.1$_{\pm2.5}$  & \underline{91.9}$_{\pm0.7}$ 
                 & 61.5$_{\pm2.8}$ & 80.6$_{\pm1.2}$ \\
    GALS         & $\times$     & $\times$     
                 & 75.4$_{\pm2.2}$  & 89.0$_{\pm0.5}$
                 & 55.0$_{\pm5.5}$ & 79.7$_{\pm0.4}$ \\
    JTT          & $\times$     & $\checkmark$ 
                 & 86.4$_{\pm1.0}$ & 89.5$_{\pm0.5}$ 
                 & 61.3$_{\pm 5.5}$ & 79.7$_{\pm 3.0}$ \\
    CnC          & $\times$     & $\checkmark$ 
                 & \underline{86.5}$_{\pm 5.9}$  & 91.0$_{\pm 0.5}$ 
                 & 62.1$_{\pm 0.9}$ & 81.9$_{\pm 1.5}$ \\
    SupER (Ours)        & $\times$     & $\times$     
                 & 84.4$_{\pm2.3}$  & 87.3$_{\pm0.6}$ 
                 & \textbf{\underline{79.7}}$_{\pm1.7}$ & \textbf{\underline{85.0}}$_{\pm1.4}$ \\
    \midrule
    UW           & $\checkmark$ & $\times$     
                 & 89.3$_{\pm1.5}$  & \textbf{94.5}$_{\pm0.9}$ 
                 & 56.4$_{\pm2.3}$ & 78.6$_{\pm0.8}$ \\
    IRM          & $\checkmark$ & $\times$     
                 & 76.2$_{\pm6.3}$  & 89.4$_{\pm0.9}$ 
                 & 57.0$_{\pm5.4}$ & 80.5$_{\pm5.0}$ \\
    GroupDRO     & $\checkmark$ & $\times$     
                 & 87.2$_{\pm1.3}$ & 93.2$_{\pm0.4}$
                 & 56.5$_{\pm 1.4}$ & 79.4$_{\pm 0.3}$ \\
    DFR          & $\checkmark$ & $\checkmark$ 
                 & \textbf{89.7}$_{\pm2.4}$ & 93.6$_{\pm0.6}$ 
                 & 48.2$_{\pm 0.4}$ & 76.4$_{\pm 0.2}$ \\
    \bottomrule
  \end{tabular*}
  }
\end{table*}

\textbf{Baselines and training.}
Given that SupER does not use any group information during training, our primary comparisons are against baseline methods that do not require group annotations. These include \textbf{ERM}, \textbf{CVaR DRO}~\citep{levy2020large}, \textbf{LfF}~\citep{nam2020learning}, \textbf{JTT}~\citep{liu2021just}, \textbf{CnC}~\citep{zhang2022correct}, and \textbf{GALS}~\citep{petryk2022guiding}, all of which have been widely adopted in prior work.
For completeness, we additionally compare SupER with methods that explicitly leverage group labels to mitigate distribution shifts, such as \textbf{GroupDRO}~\citep{sagawa2019distributionally}, \textbf{UW}~\citep{sagawa2019distributionally}, and \textbf{DFR}~\citep{kirichenko2022last}. We also consider multi-source environment methods such as \textbf{IRM}~\citep{arjovsky2019invariant}.
For SupER, we instantiate both the CLIP and $ \beta$-VAE components with a ResNet-50 backbone architecture~\citep{he2016deep}. For each dataset, we train SupER following Algorithm~\ref{alg:training_dafd}. Additional training details, as well as the choices of hyperparameters and superclass labels $y^{\text{super}}$, are provided in Appendix~\ref{subapp:hs}.

\begin{table*}[t]
  \centering
  {\fontsize{8pt}{8pt}\selectfont
  \caption{Mean $\pm$ std of worst group accuracy (\%) on Spawrious.  
We compare methods that use group information, as they typically outperform methods that do not use group labels.
The final column reports the mean $\pm$ std of worst group accuracy across all six subsets.    
\textbf{Bold} indicates the best among these methods.}
  \label{tab:sc_results}
  \begin{tabular*}{\textwidth}{@{\extracolsep{\fill}} l  c  c  c  c  c  c  c  c}
    \toprule
    \multirow{2}{*}{Method}
      & \multirow{2}{*}{\shortstack{Group\\ Info}}
      & \multicolumn{3}{c}{One--To--One}
      & \multicolumn{3}{c}{Many--To--Many}
      & \multirow{2}{*}{Average} \\
    \cmidrule(lr){3-5} \cmidrule(lr){6-8}
      &  
      & Easy & Medium & Hard
      & Easy & Medium & Hard
      &  \\
    \midrule
    ERM       & $\times$     
              & 78.4$_{\pm 1.8}$ & 63.4$_{\pm 2.3}$ & 71.1$_{\pm 3.7}$
              & 72.9$_{\pm 1.3}$ & 52.7$_{\pm 2.9}$ & 50.7$_{\pm 1.0}$ 
              & 64.9$_{\pm 11.3}$ \\
    SupER (Ours)     & $\times$     
              & 82.7$_{\pm 2.0}$ & \textbf{80.3}$_{\pm 4.6}$ & \textbf{83.8}$_{\pm 3.4}$
              & \textbf{87.4}$_{\pm 1.3}$ & \textbf{83.4}$_{\pm 2.3}$ & \textbf{79.9}$_{\pm 4.7}$ 
              & \textbf{82.9}$_{\pm 2.7}$ \\
    \midrule
    UW        & $\checkmark$  
              & \textbf{87.4}$_{\pm 1.1}$ & 67.9$_{\pm 2.1}$ & 75.9$_{\pm 2.9}$
              & 72.9$_{\pm 1.3}$ & 52.7$_{\pm 2.9}$ & 50.7$_{\pm 1.0}$ 
              & 67.9$_{\pm 14.1}$ \\
    IRM       & $\checkmark$  
              & 78.4$_{\pm 1.0}$ & 64.5$_{\pm 3.2}$ & 64.9$_{\pm 2.2}$
              & 77.9$_{\pm 3.7}$ & 57.1$_{\pm 2.9}$ & 50.7$_{\pm 1.1}$ 
              & 65.6$_{\pm 11.1}$ \\
    GroupDRO  & $\checkmark$  
              & 86.7$_{\pm 1.2}$ & 67.2$_{\pm 0.7}$ & 76.4$_{\pm 2.2}$
              & 74.3$_{\pm 0.9}$ & 55.7$_{\pm 1.4}$ & 49.9$_{\pm 0.8}$ 
              & 68.3$_{\pm 13.7}$ \\
    DFR       & $\checkmark$      
              & 79.1$_{\pm 5.2}$ & 64.3$_{\pm 1.9}$ & 70.0$_{\pm 1.9}$
              & 76.4$_{\pm 1.9}$ & 58.7$_{\pm 2.2}$ & 54.1$_{\pm 2.2}$ 
              & 67.1$_{\pm 9.9}$ \\
    \bottomrule
  \end{tabular*}
  }
  \vspace{-0.2cm}
\end{table*}

\begin{table*}[t]
  \centering
  {\fontsize{8pt}{8pt}\selectfont
 \caption{Mean $\pm$ std of worst group accuracy (\%) for the MetaShift dataset using baselines that do not require group information. A larger value of $d$ indicates a greater distribution shift. The final column reports the mean $\pm$ std of worst group accuracy across the four datasets.
    \textbf{Bold} indicates the best among these methods.}
  \label{tab:ms_results}
  \begin{tabular*}{\textwidth}{@{\extracolsep{\fill}} l  c  c  c  c  c  c  c}
    \toprule
    \multirow{2}{*}{Method}
      & \multirow{2}{*}{\shortstack{Group\\Info}}
      & \multirow{2}{*}{\shortstack{Train\\Twice}}
      & \multicolumn{4}{c}{MetaShift Subsets}
      & \multirow{2}{*}{Average} \\
    \cmidrule(lr){4-7}
      &  & 
      & (a) $d=0.44$ & (b) $d=0.71$ & (c) $d=1.12$ & (d) $d=1.43$ 
      &  \\
    \midrule
    ERM       & $\times$     & $\times$ 
              & 78.8$_{\pm 1.0}$ & 75.8$_{\pm 0.8}$ & 61.9$_{\pm 5.9}$ & 52.6$_{\pm 2.6}$ 
              & 67.3$_{\pm 12.2}$ \\
    CVaR DRO  & $\times$     & $\times$ 
              & 77.8$_{\pm 2.5}$ & 72.5$_{\pm 2.8}$ & 65.1$_{\pm 0.2}$ & 54.7$_{\pm 3.2}$ 
              & 67.5$_{\pm 10.0}$ \\
    LfF       & $\times$     & $\times$ 
              & 77.2$_{\pm 1.7}$ & 73.9$_{\pm 0.6}$ & 69.5$_{\pm 1.0}$ & 59.5$_{\pm 3.1}$ 
              & 70.0$_{\pm 7.7}$ \\
    GALS      & $\times$     & $\times$ 
              & 74.8$_{\pm 3.9}$ & 68.8$_{\pm 2.0}$ & 70.6$_{\pm 2.2}$ & 50.0$_{\pm 0.9}$ 
              & 66.0$_{\pm 11.0}$ \\
    JTT       & $\times$     & $\checkmark$      
              & 76.7$_{\pm 2.3}$ & 73.2$_{\pm 0.8}$ & 67.1$_{\pm 4.6}$ & 53.0$_{\pm 1.6}$ 
              & 67.5$_{\pm 10.4}$ \\
    CnC       & $\times$ & $\checkmark$      
              & \textbf{81.1}$_{\pm 1.4}$ & 71.4$_{\pm 2.4}$ & 65.4$_{\pm 6.8}$
              & 49.6$_{\pm 1.6}$ & 66.9$_{\pm 13.2}$ \\
    SupER (Ours)     & $\times$     & $\times$ 
              & 79.8$_{\pm 3.6}$ & \textbf{78.4}$_{\pm 1.9}$ & \textbf{77.6}$_{\pm 2.1}$ & \textbf{71.4}$_{\pm 2.1}$ 
              & \textbf{76.8}$_{\pm 3.7}$ \\
    \bottomrule
  \end{tabular*}
  }
  \vspace{-0.4cm}
\end{table*}

\subsection{Main Results}
\label{sec:main results}

\subsubsection{Comparison of Accuracy Across Groups}

We report the worst group accuracy, average accuracy, and variance of accuracy across groups for SupER and baseline methods.
Tables~\ref{tab:wb_results}-\ref{tab:variance_wb_sc} present selected results. Comprehensive results are available in Appendix~\ref{app:WGAAAV}.

\textbf{SupER achieves strong performance on worst group accuracy.}
As shown in Tables~\ref{tab:wb_results}-\ref{tab:ms_results}, for most datasets—including Waterbirds-100\%, the last five subsets of the Spawrious dataset, and the last three subsets of the MetaShift dataset—SupER's worst group accuracy exceeds that of all selected baseline methods, regardless of whether they require group information. For the remaining datasets, SupER still outperforms the majority of the baselines that do not rely on group labels. 

\textbf{SupER exhibits superior robustness to complex spurious correlations, especially in the presence of unseen groups at test time.} 
Spawrious and Metashift provide chances to investigate model performance under various levels of spurious correlations. Results in Tables~\ref{tab:wb_results}-\ref{tab:ms_results} show that, despite increasing complexity of spurious correlations, the standard deviation of worst group accuracy across the six Spawrious datasets is 2.7\%, and across the four MetaShift datasets is 3.7\%, significantly lower than other baseline methods. Moreover, Waterbirds-100\%, Spawrious, and MetaShift all contain certain test groups that are entirely absent during training, and they induce different types of spurious correlations. SupER performs particularly well on these datasets. For instance, compared to all selected baselines, SupER’s worst group accuracy exceeds the best competing baseline by 17.6\% on Waterbirds-100\%, 11.9\% on MetaShift (d), 25.8\% on Spawrious M2M-hard, and 7.4\% on Spawrious O2O-hard.

\textbf{SupER shows smaller generalization gaps among groups.}
Table~\ref{tab:variance_wb_sc} shows the variance of accuracy across groups for SupER and other baseline methods on selected datasets. SupER exhibits more consistent test accuracy across different groups within the same dataset. This indicates that, under the guidance of superclass information, the model consistently focuses on features with semantic meaning and becomes less influenced by spurious features. 

\begin{figure*}[t]
\vskip 0.2in
    \centering
    \begin{tabular}{@{}c@{\hspace{1.5mm}}c@{\hspace{3.5mm}}c@{\hspace{4.5mm}}c@{\hspace{4.5mm}}c@{\hspace{8mm}}
                    c@{\hspace{1.5mm}}c@{\hspace{3.5mm}}c@{\hspace{4.5mm}}c@{\hspace{4.5mm}}c@{}}

        % Row 1: Waterbirds-95%
        \includegraphics[width=0.065\textwidth]{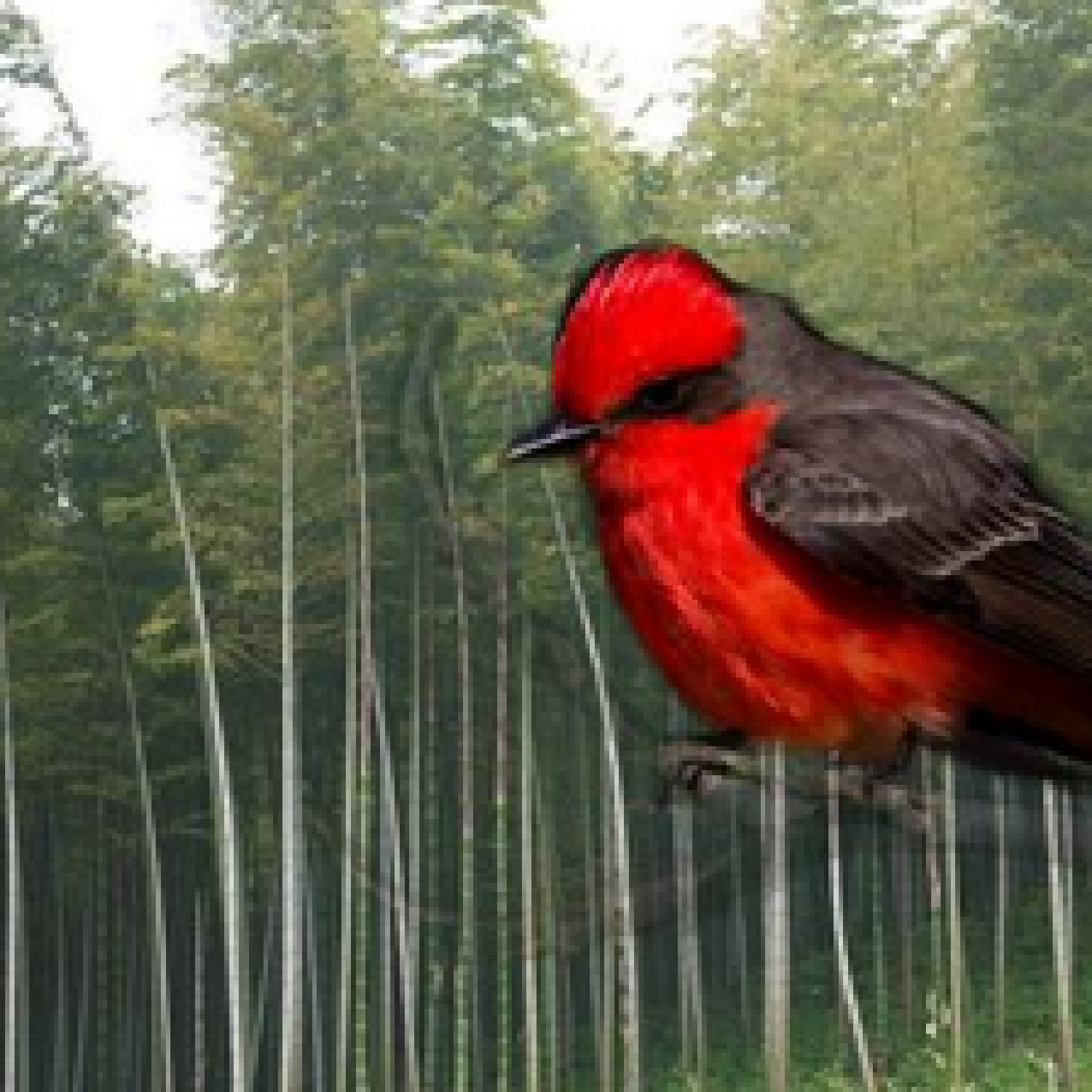} &
        \includegraphics[width=0.065\textwidth]{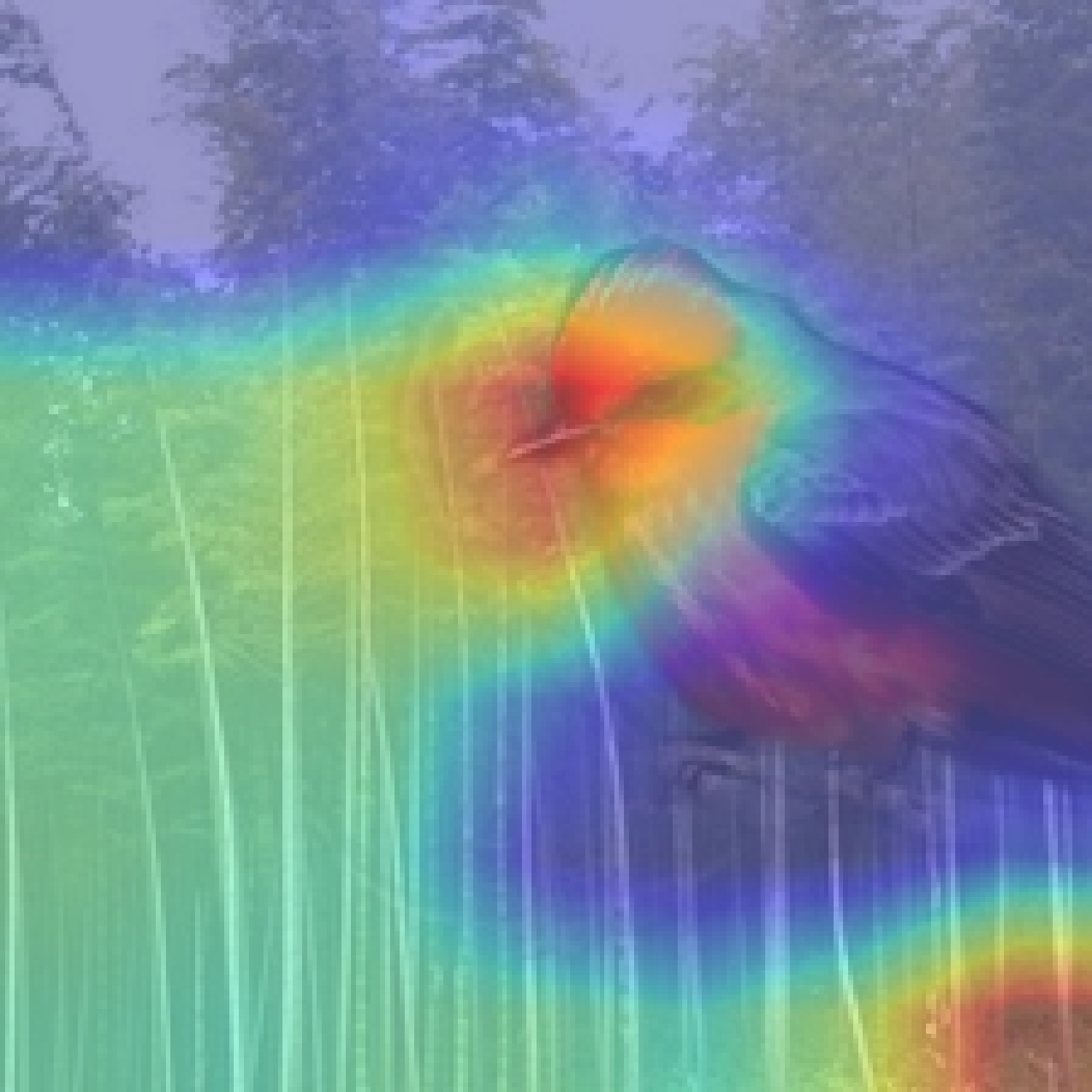} &
        \includegraphics[width=0.065\textwidth]{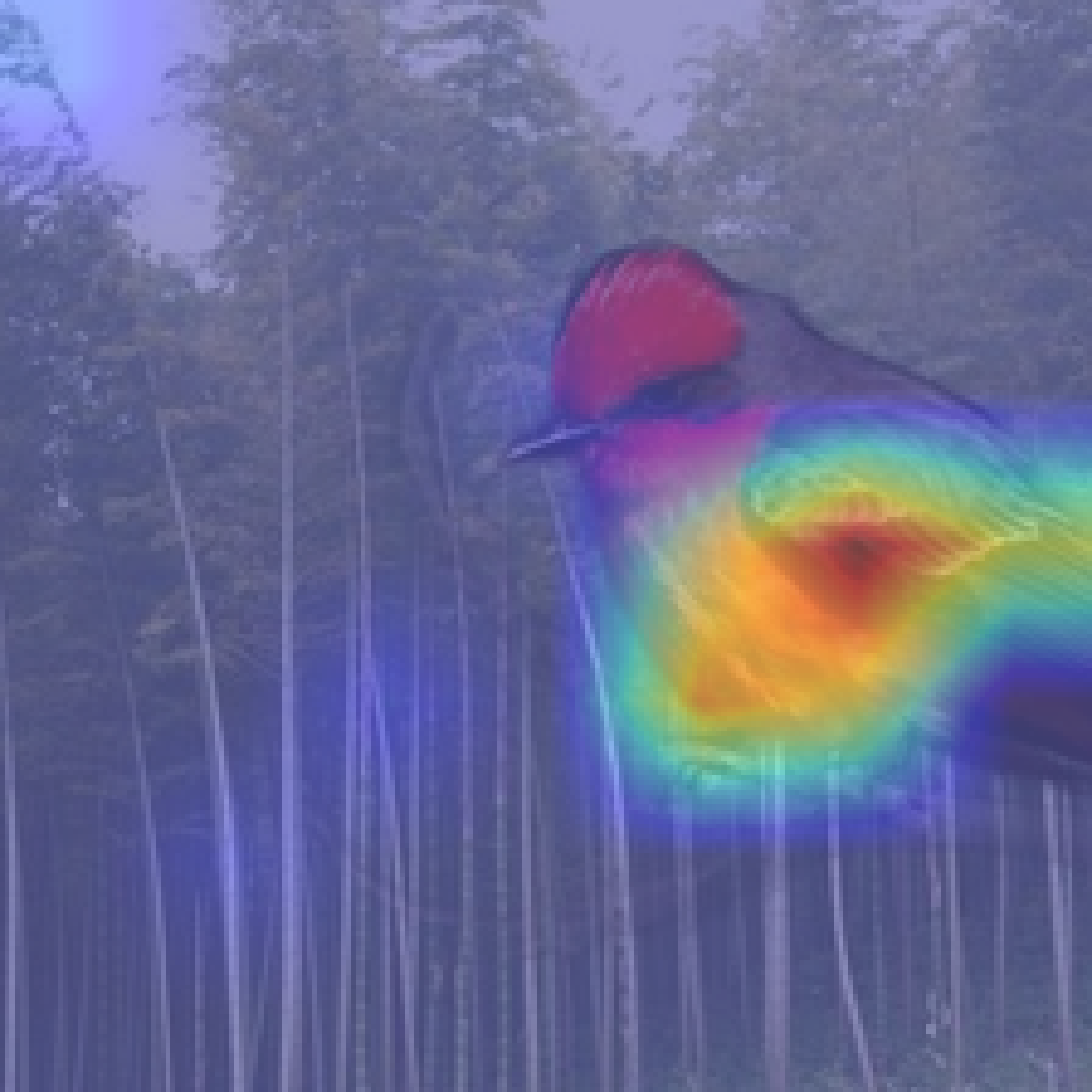} &
        \includegraphics[width=0.065\textwidth]{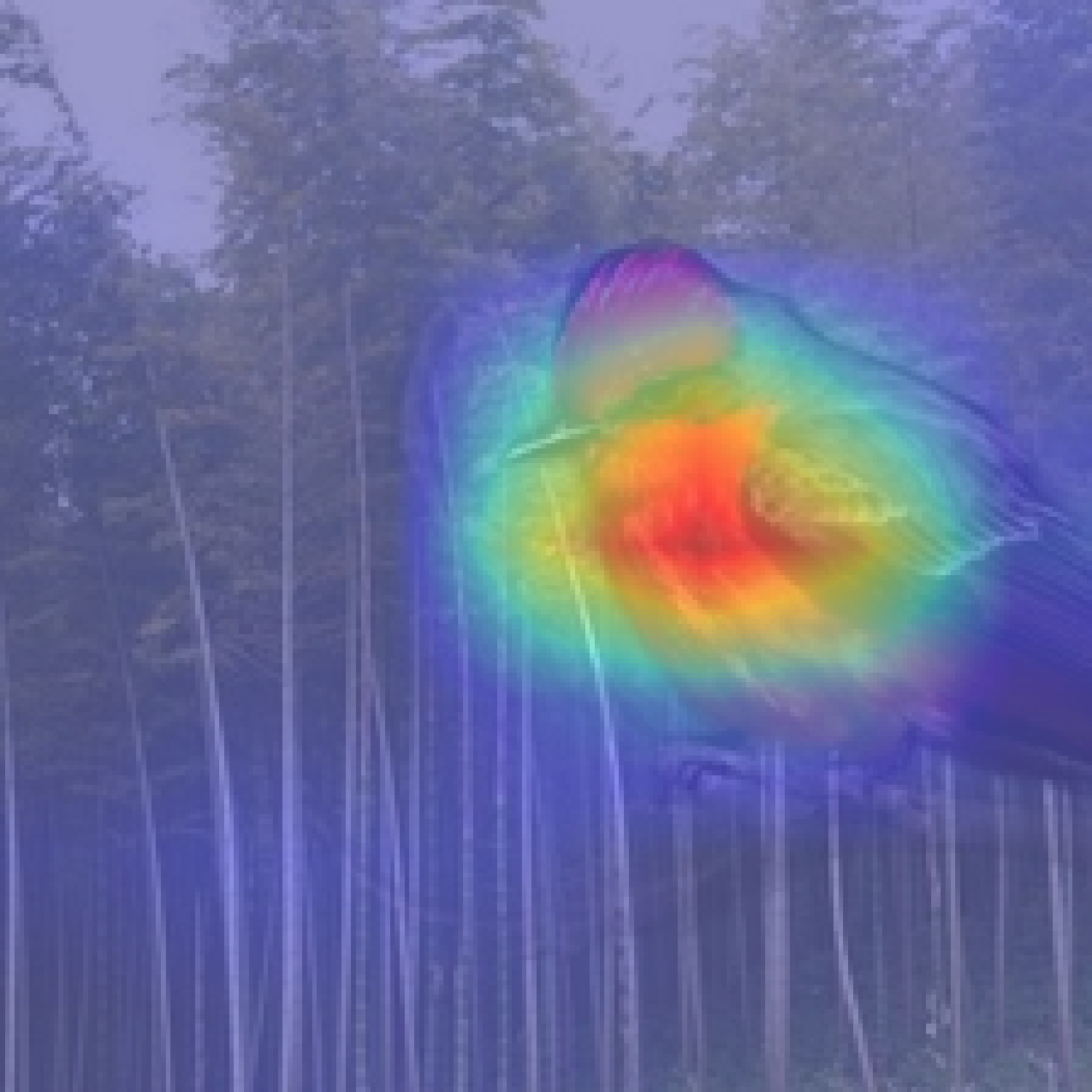} &
        \includegraphics[width=0.065\textwidth]{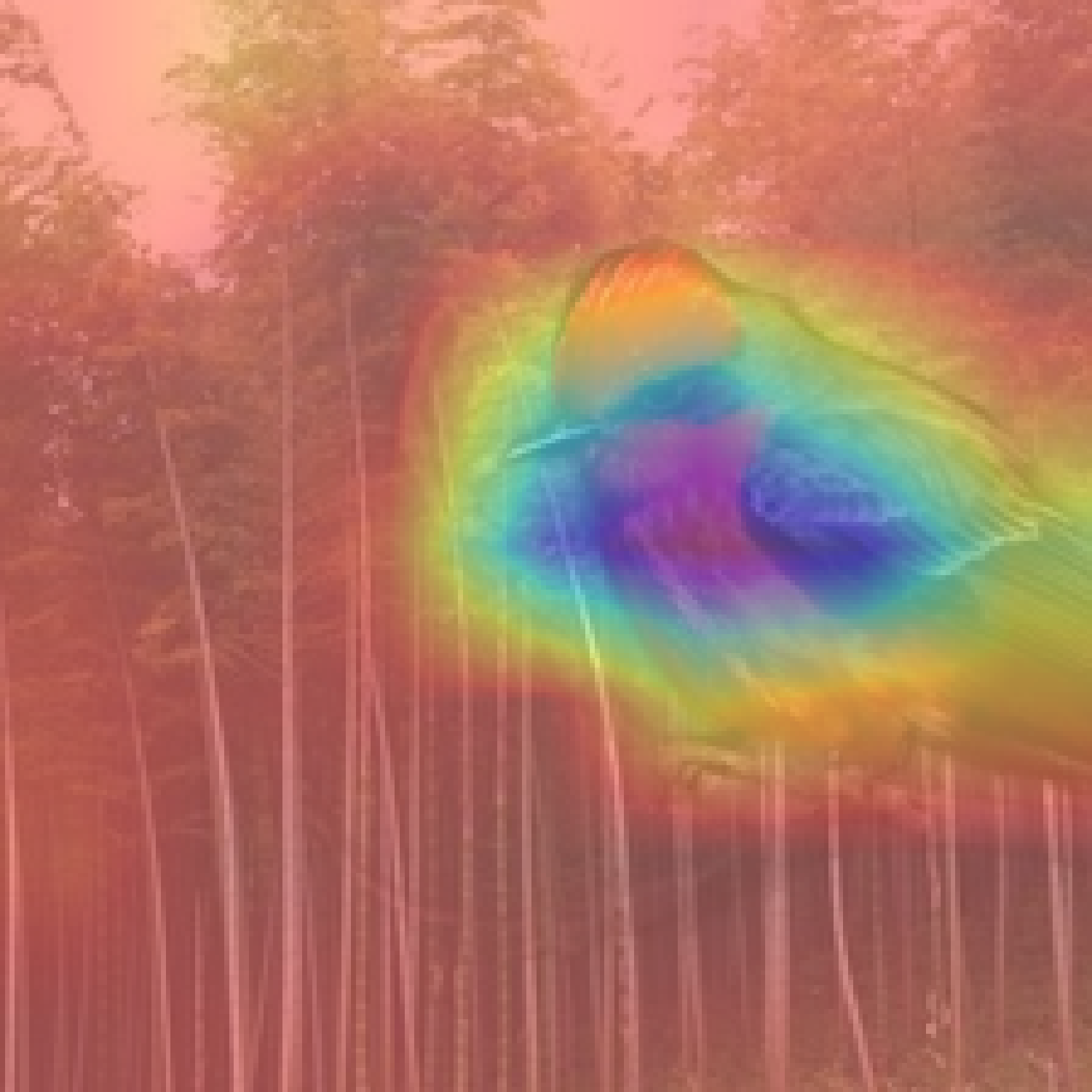} &
        \includegraphics[width=0.065\textwidth]{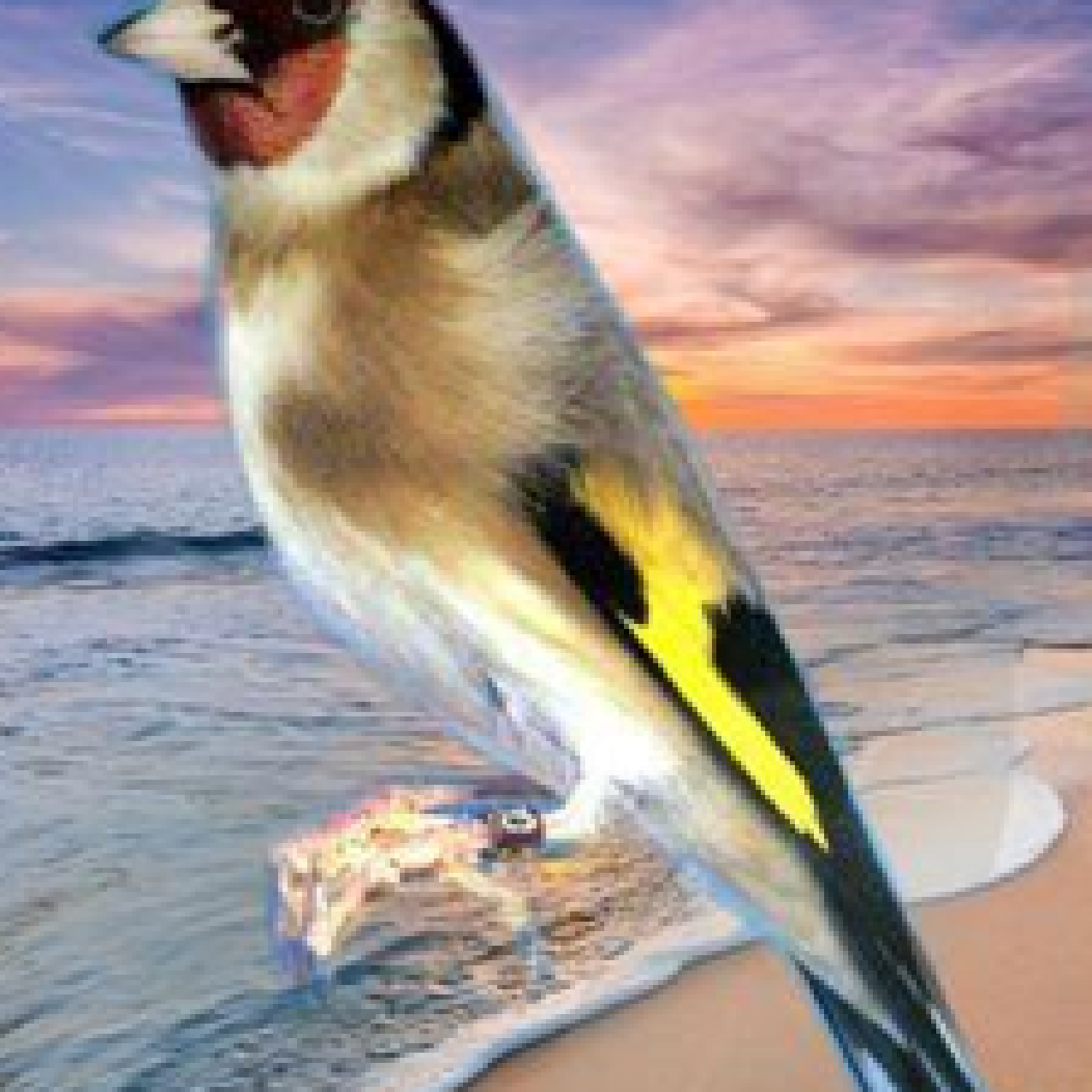} &
        \includegraphics[width=0.065\textwidth]{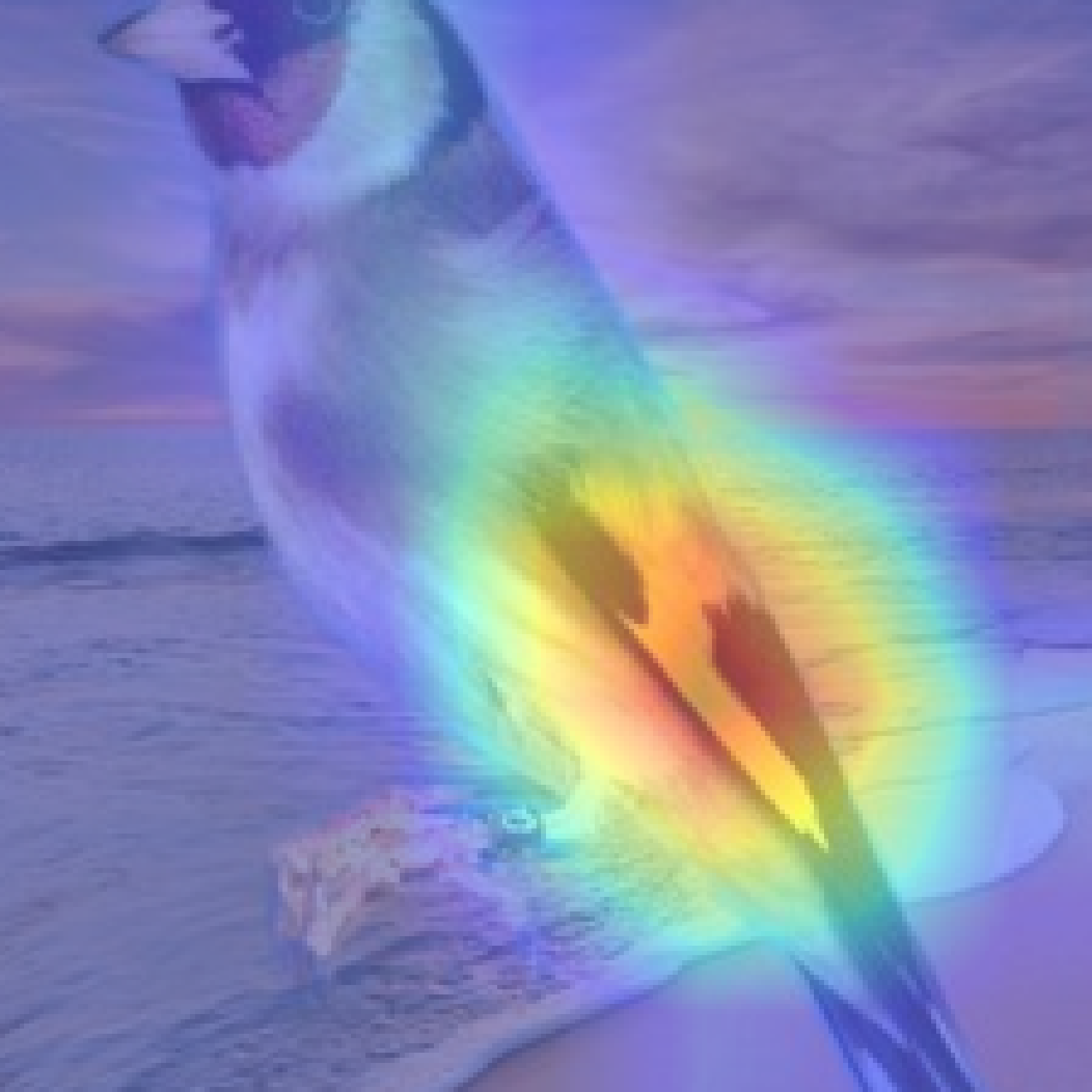} &
        \includegraphics[width=0.065\textwidth]{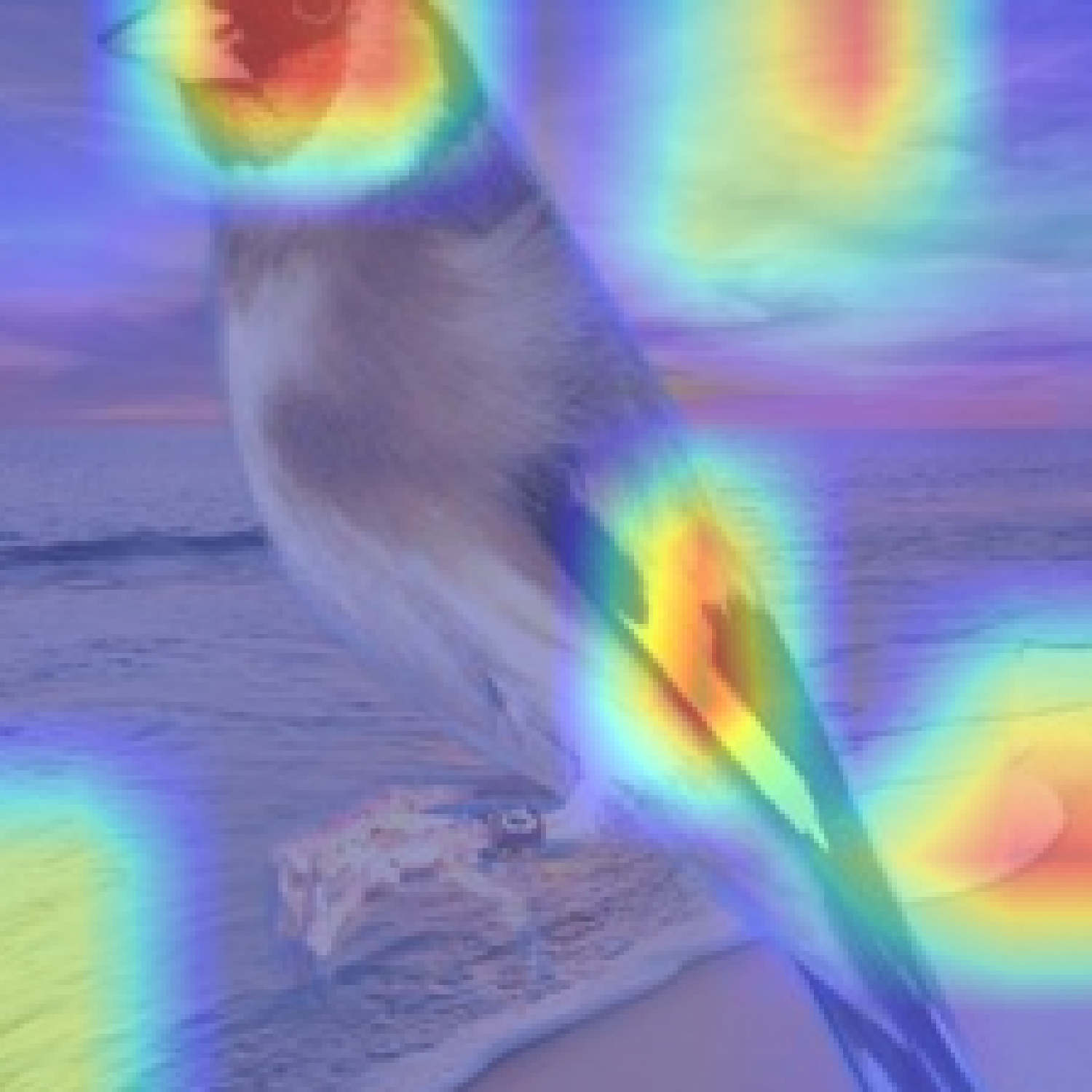} &
        \includegraphics[width=0.065\textwidth]{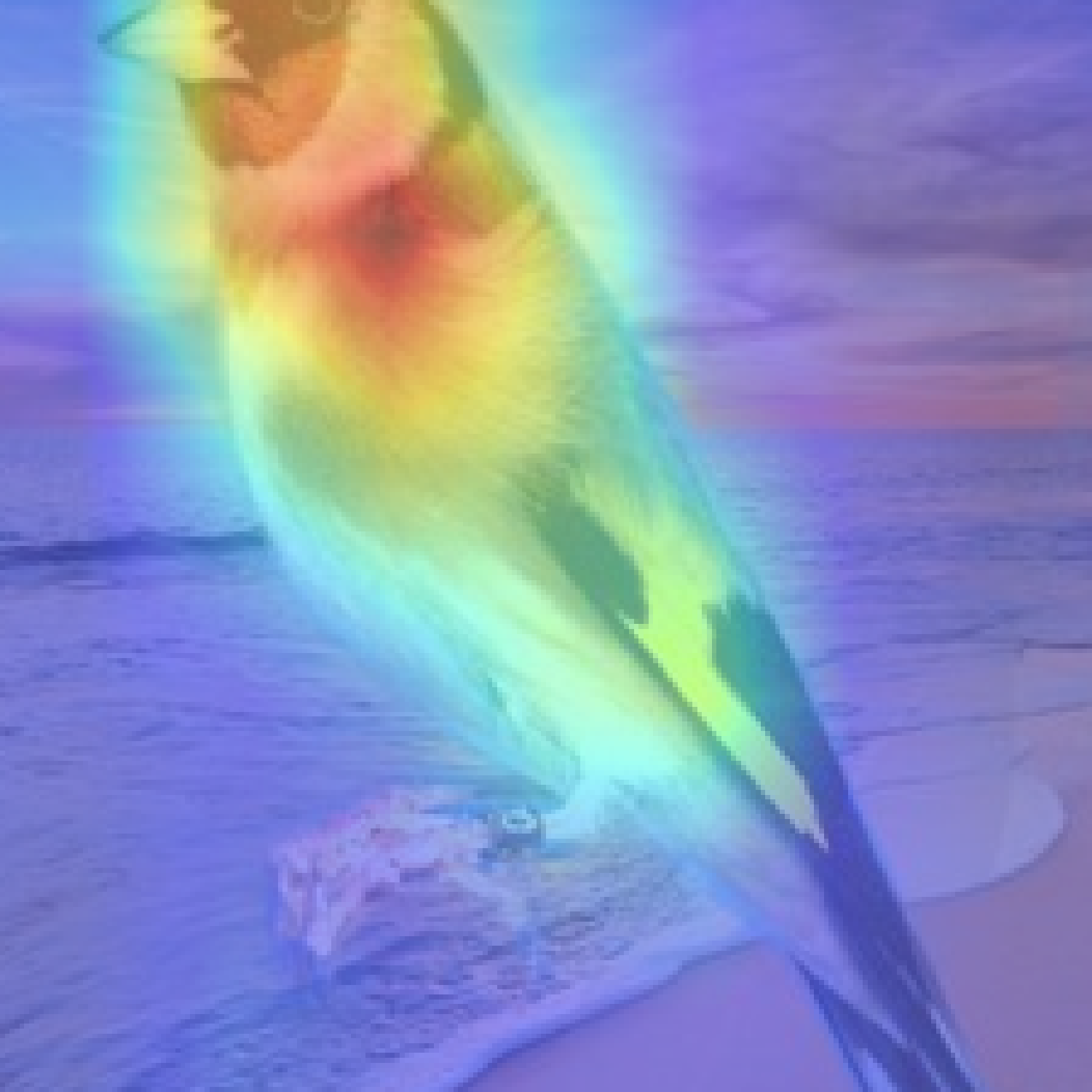} &
        \includegraphics[width=0.065\textwidth]{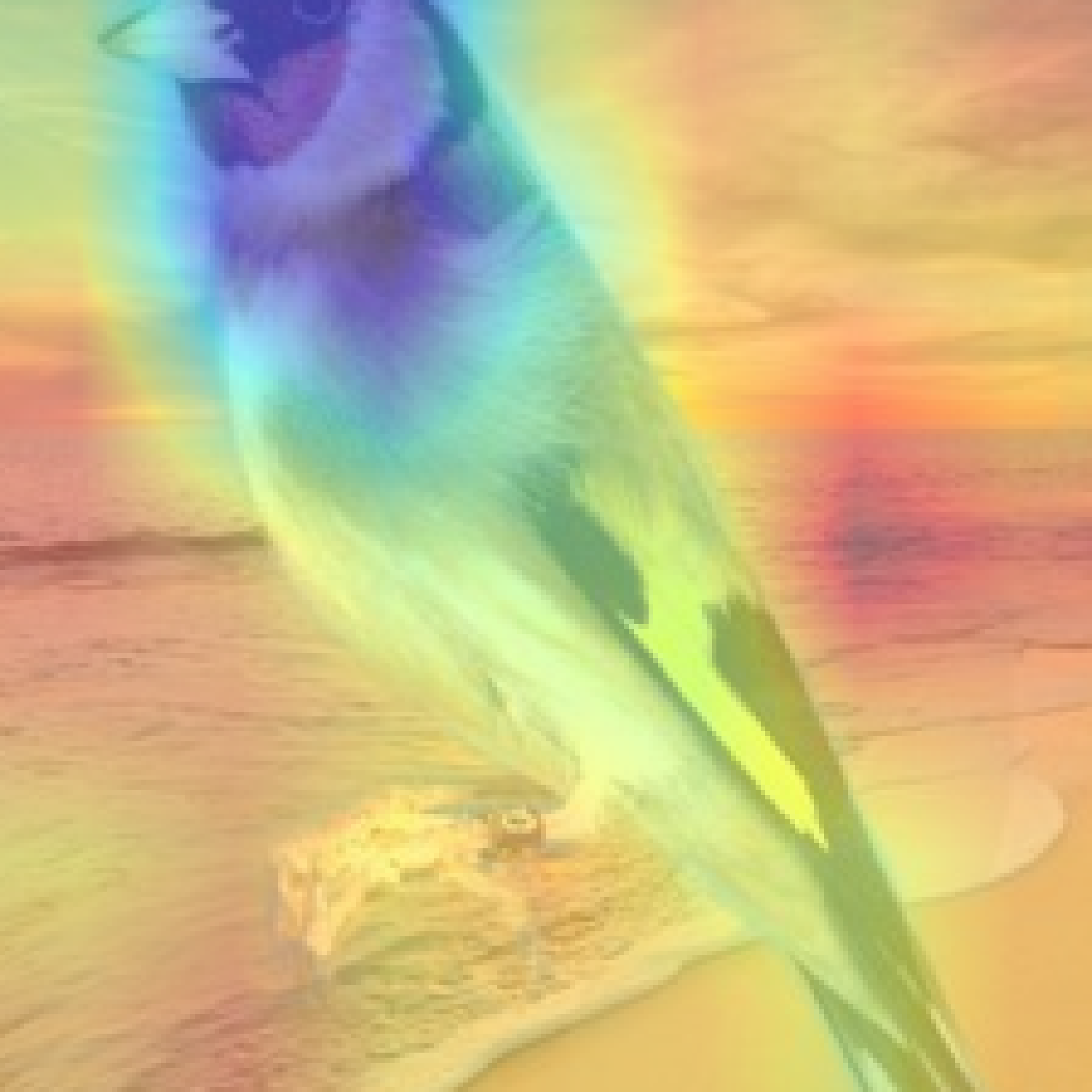} \\[0.3em]

        % Row 2: Waterbirds-100%
        \includegraphics[width=0.065\textwidth]{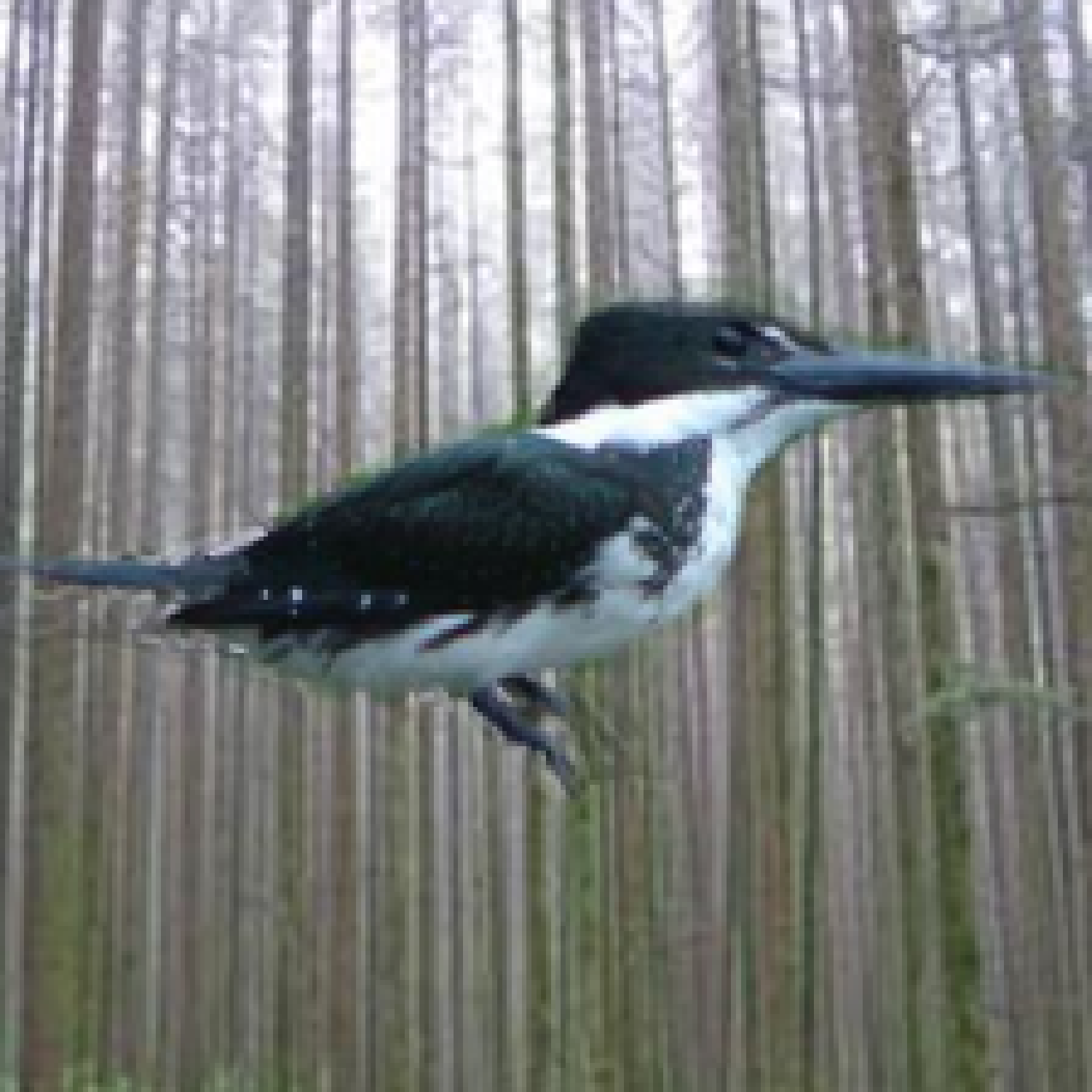} &
        \includegraphics[width=0.065\textwidth]{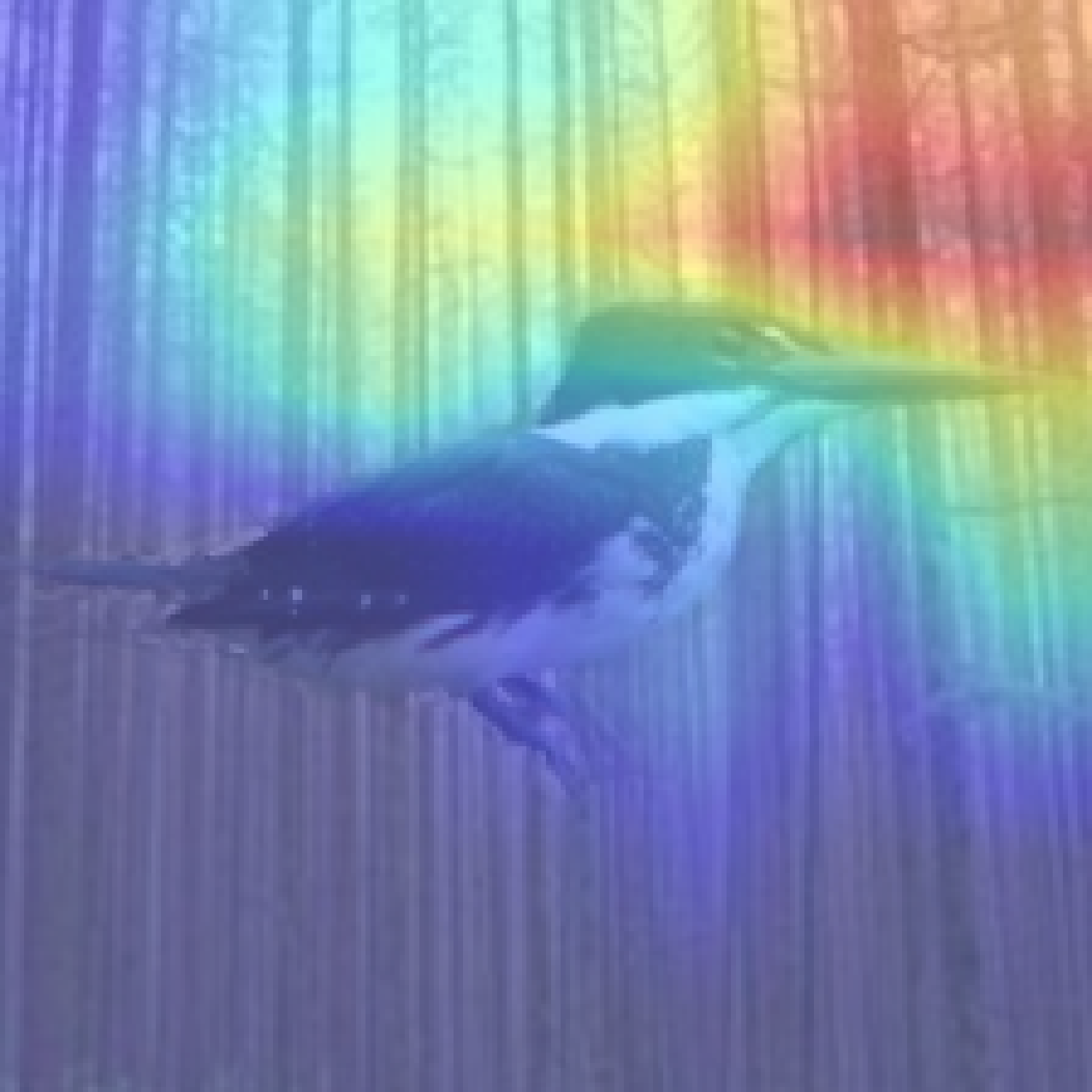} &
        \includegraphics[width=0.065\textwidth]{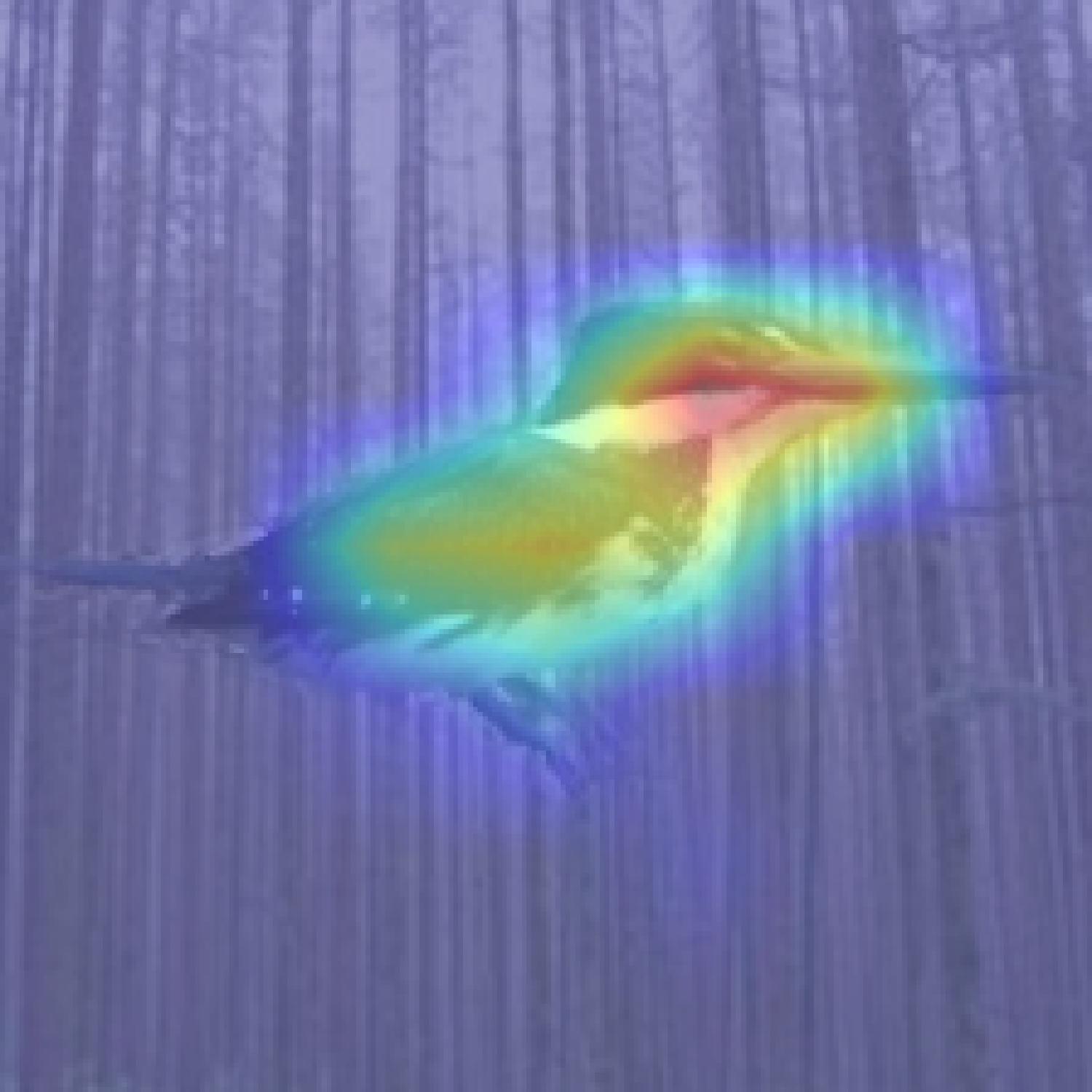} &
        \includegraphics[width=0.065\textwidth]{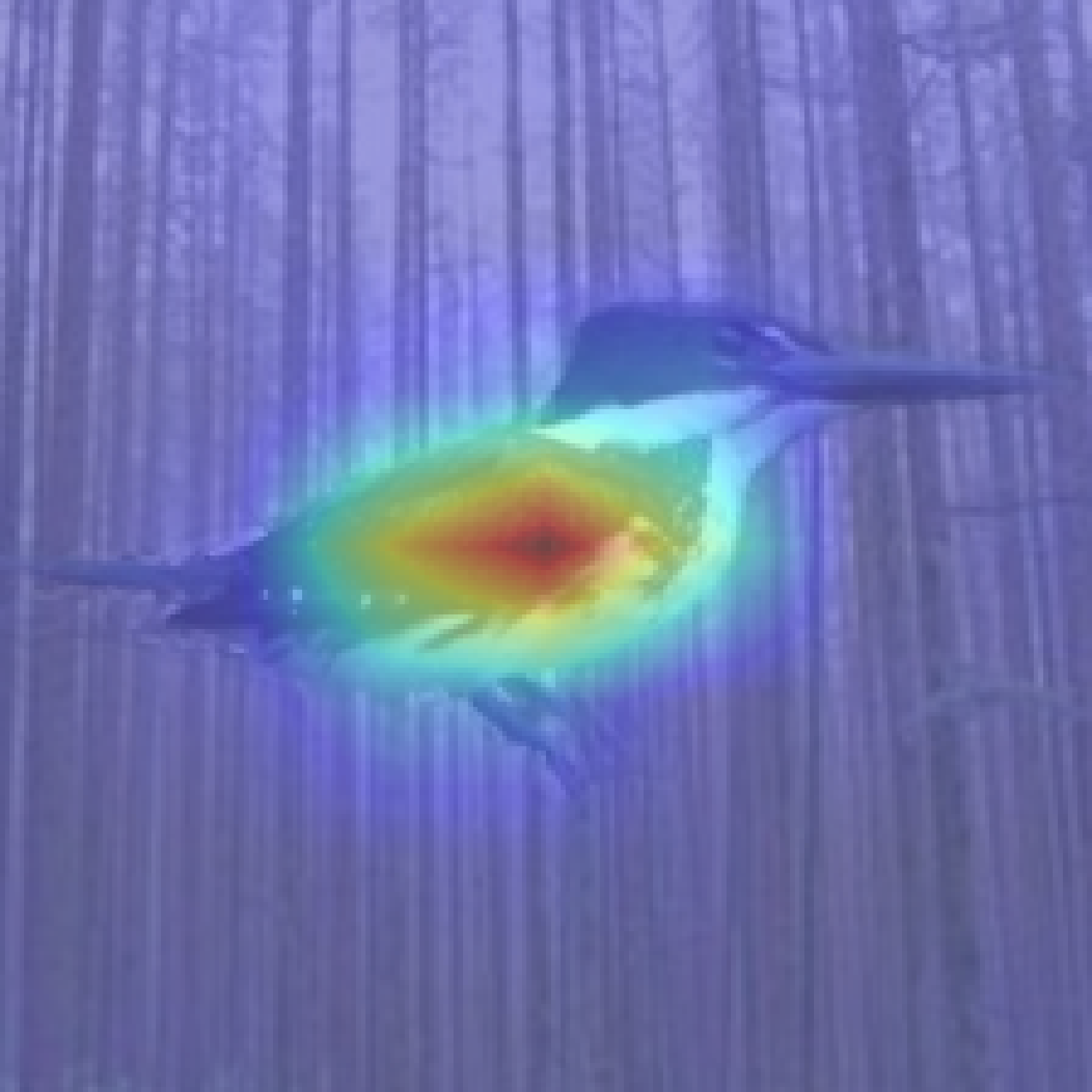} &
        \includegraphics[width=0.065\textwidth]{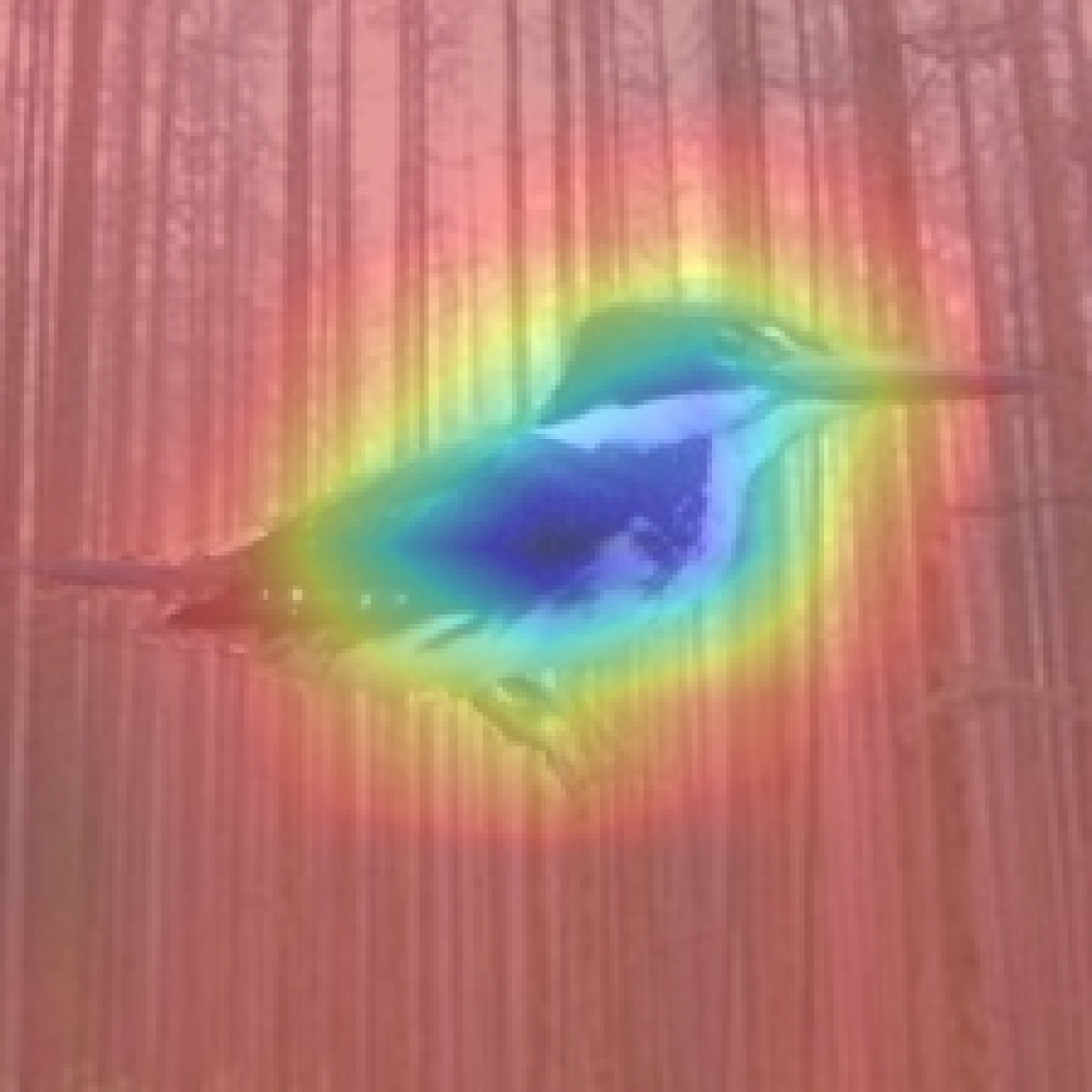} &
        \includegraphics[width=0.065\textwidth]{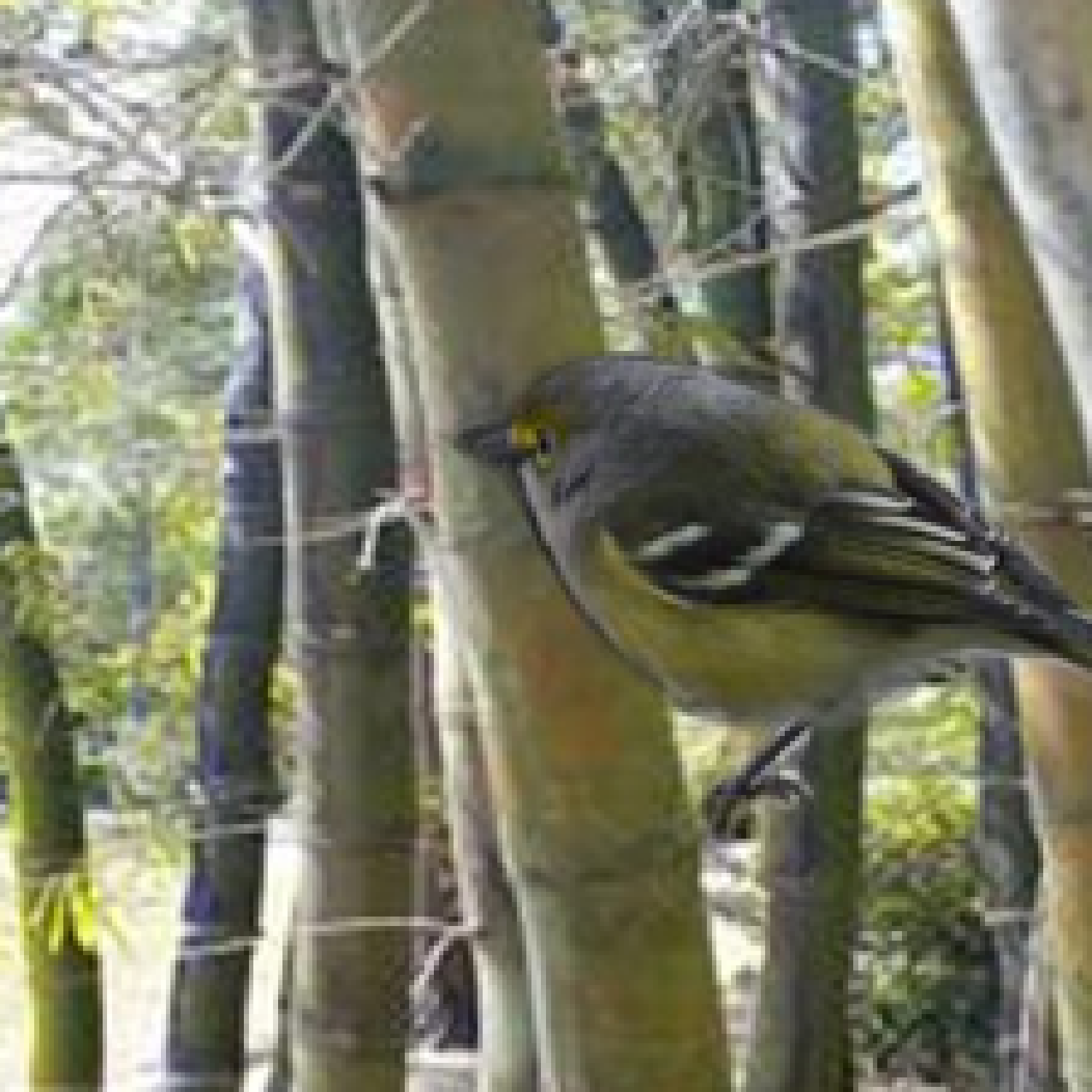} &
        \includegraphics[width=0.065\textwidth]{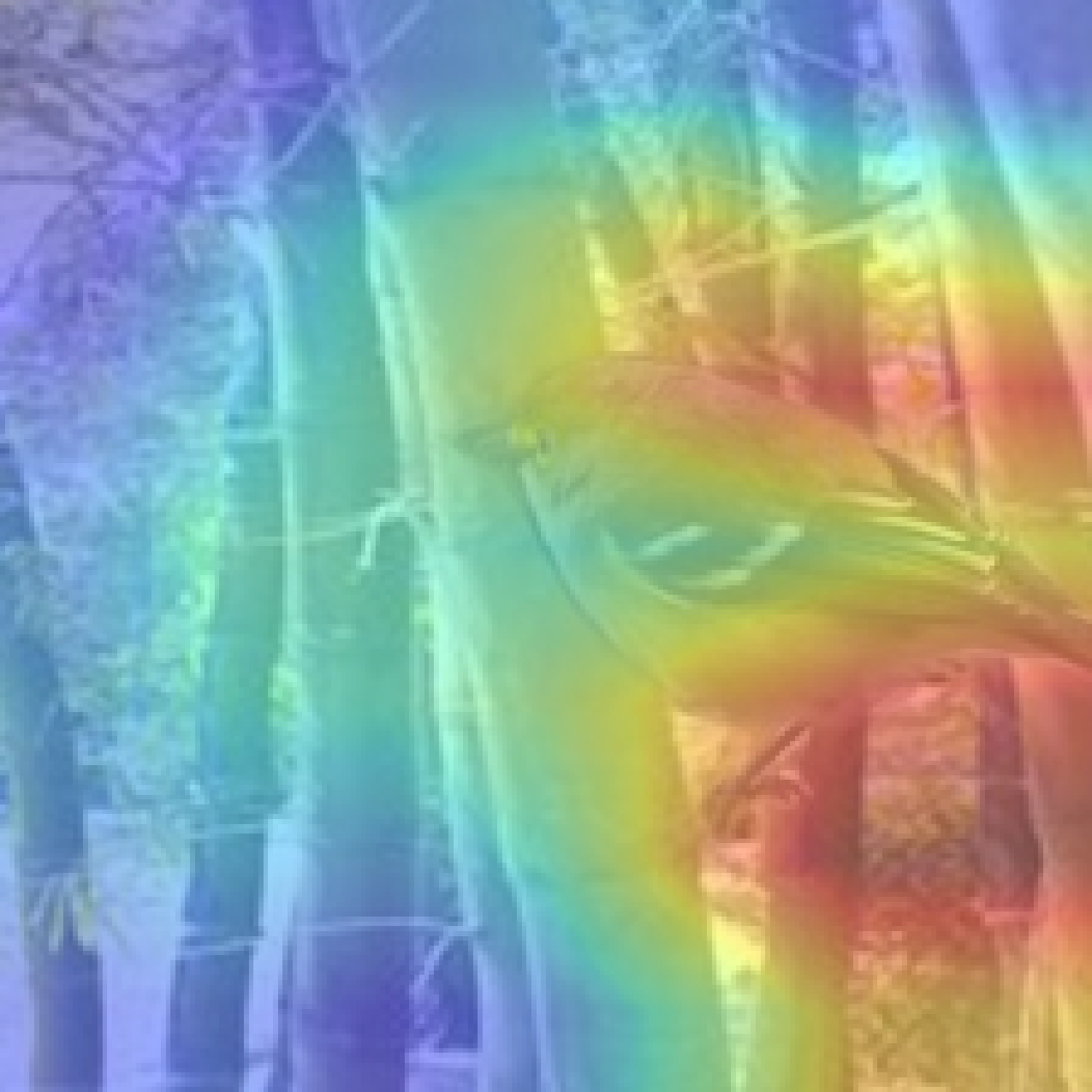} &
        \includegraphics[width=0.065\textwidth]{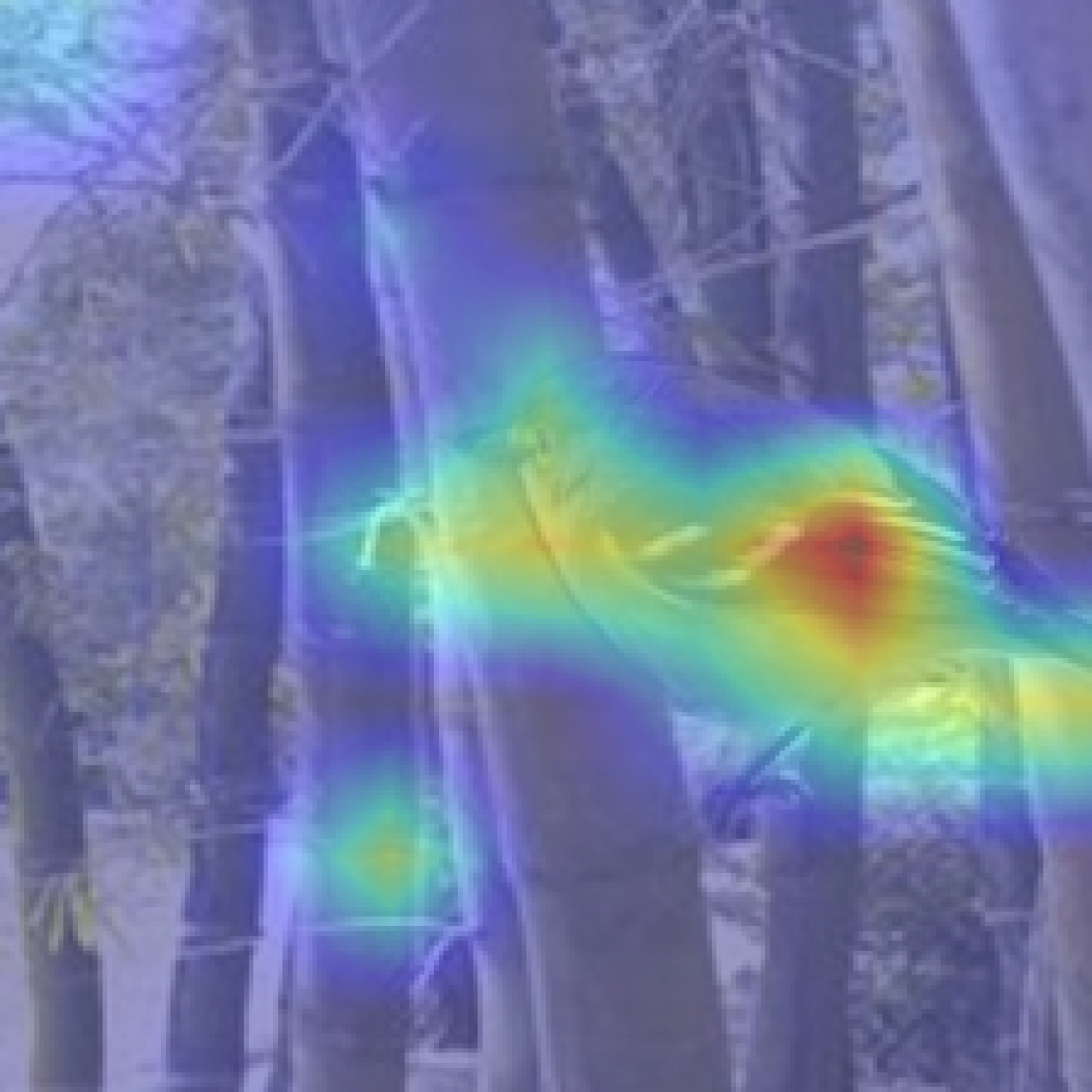} &
        \includegraphics[width=0.065\textwidth]{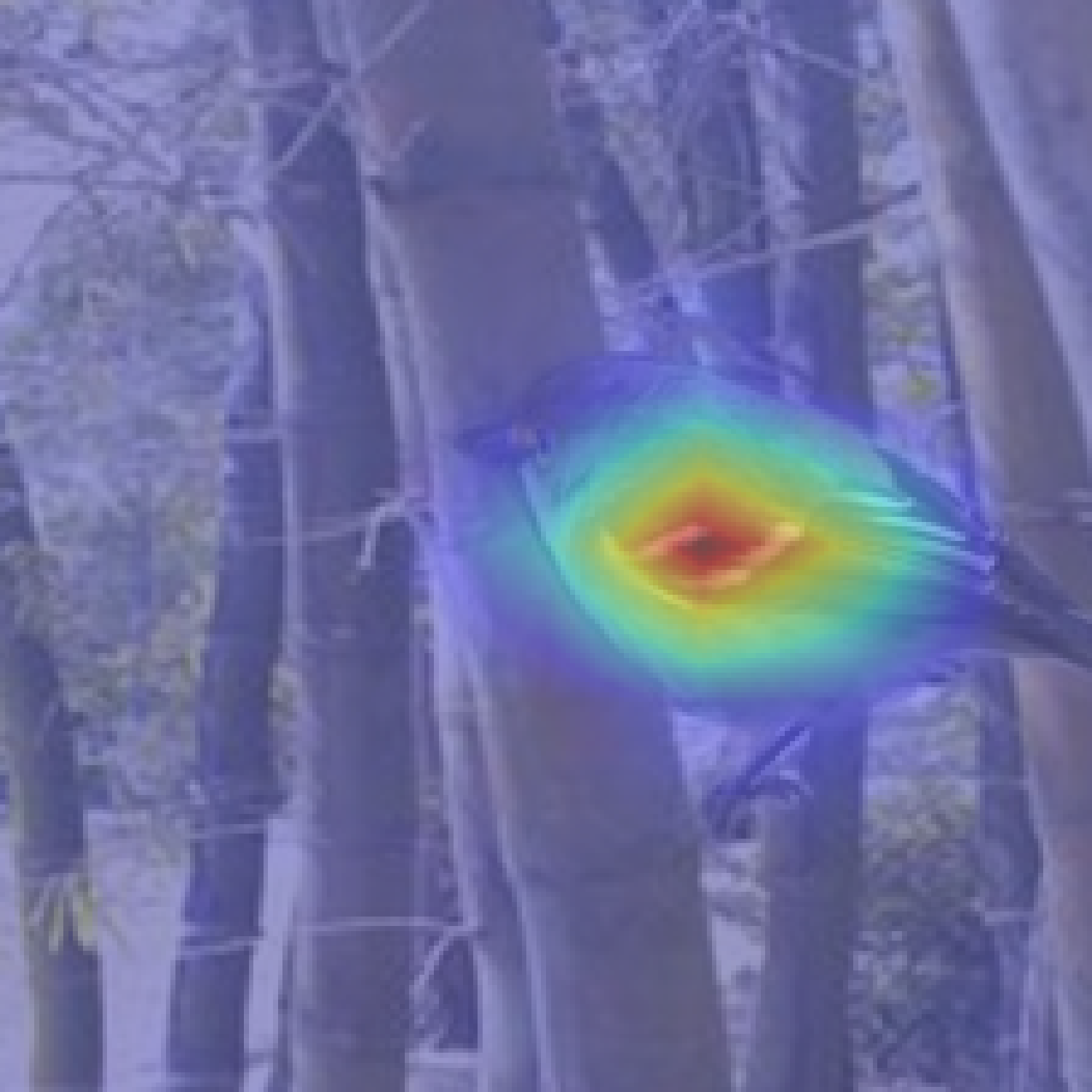} &
        \includegraphics[width=0.065\textwidth]{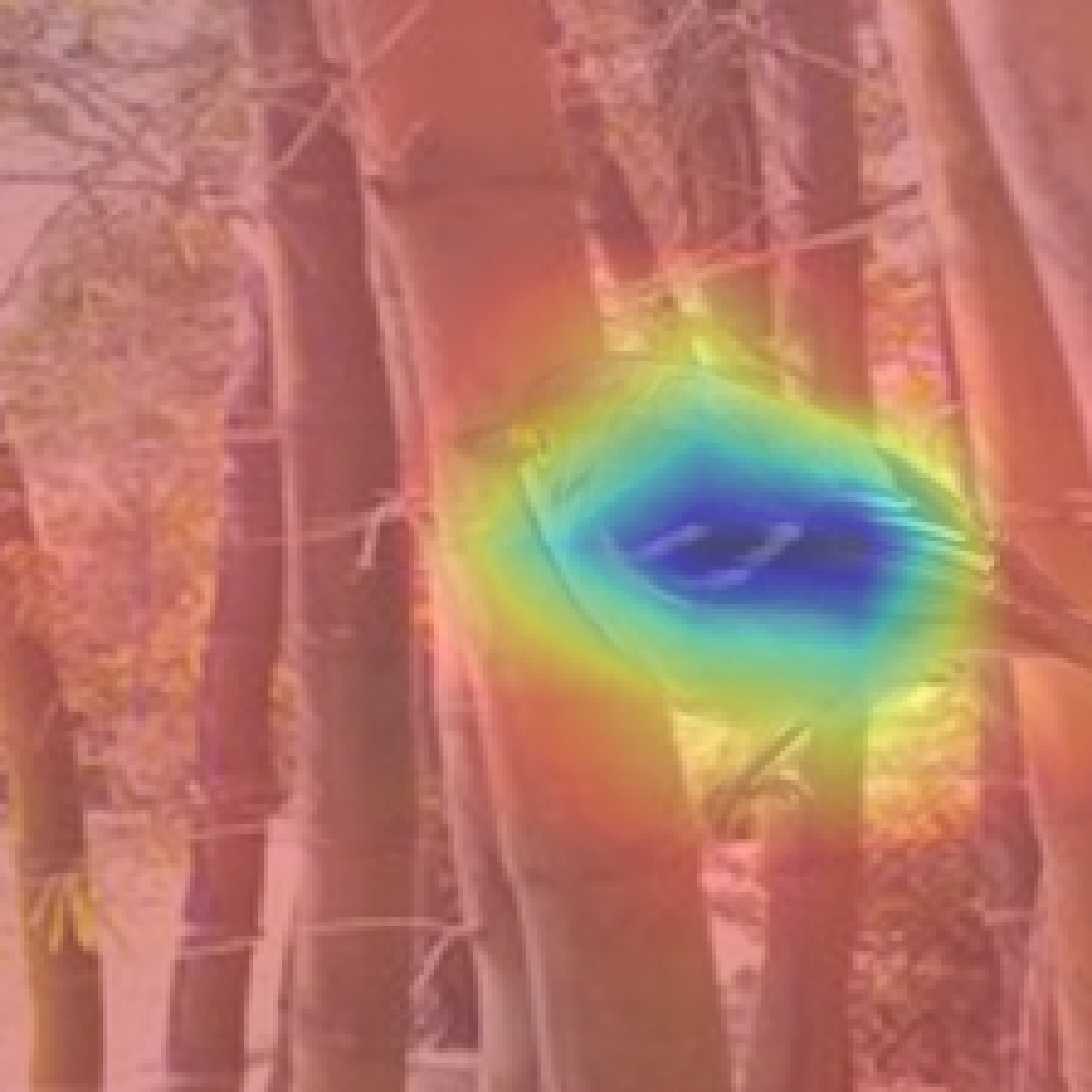} \\[0.3em]

        % Row 3: MetaShift (placeholder)
        \includegraphics[width=0.065\textwidth]{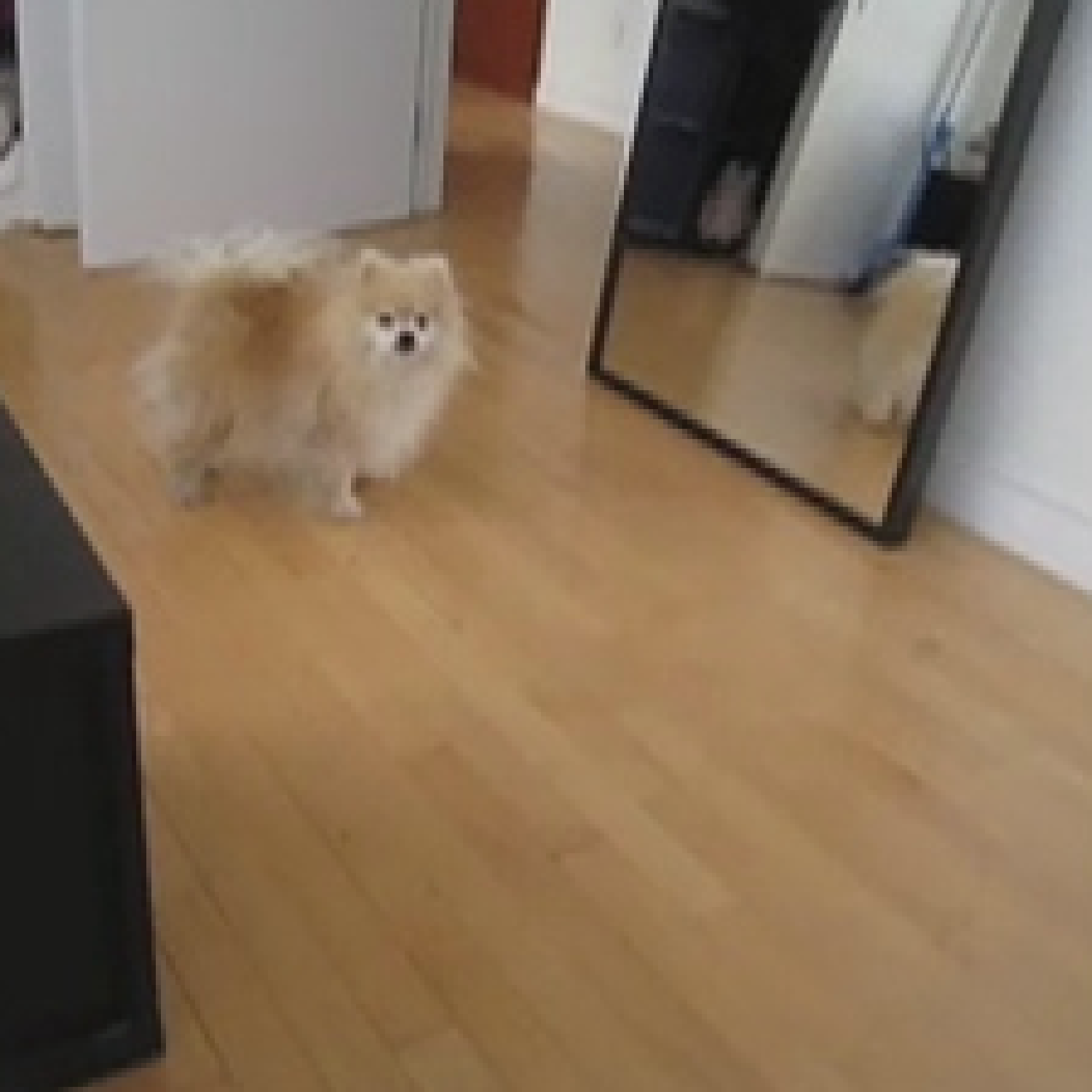} &
        \includegraphics[width=0.065\textwidth]{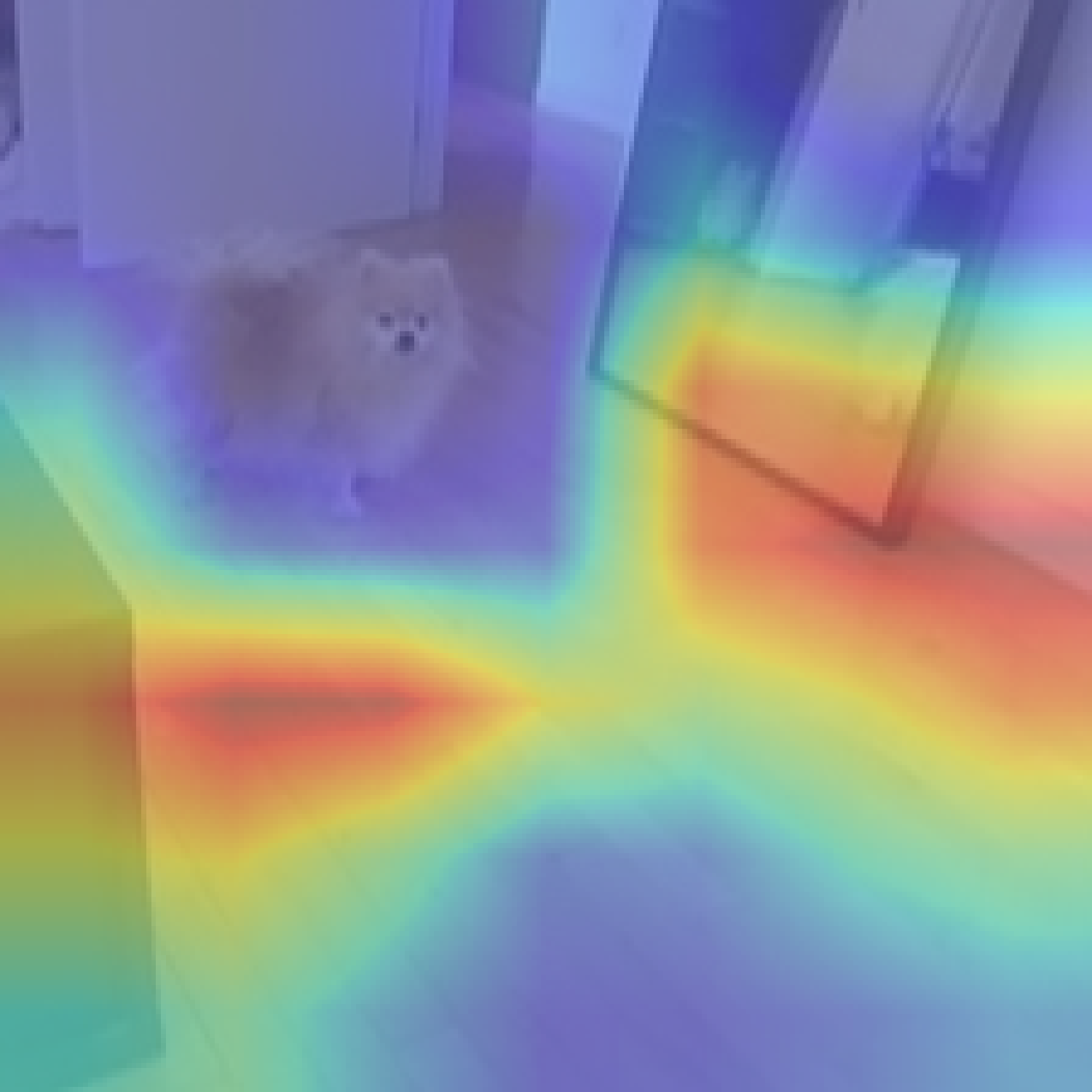} &
        \includegraphics[width=0.065\textwidth]{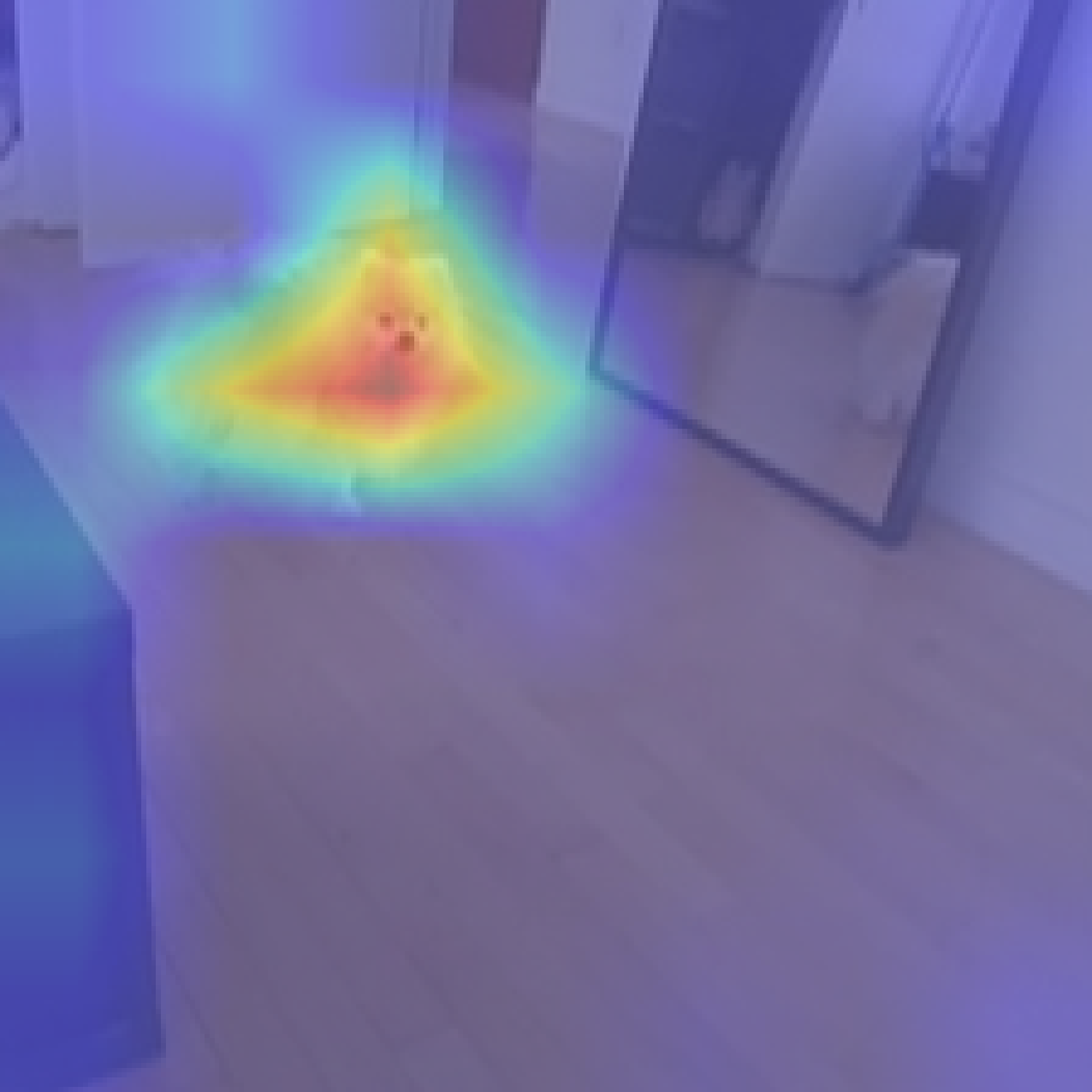} &
        \includegraphics[width=0.065\textwidth]{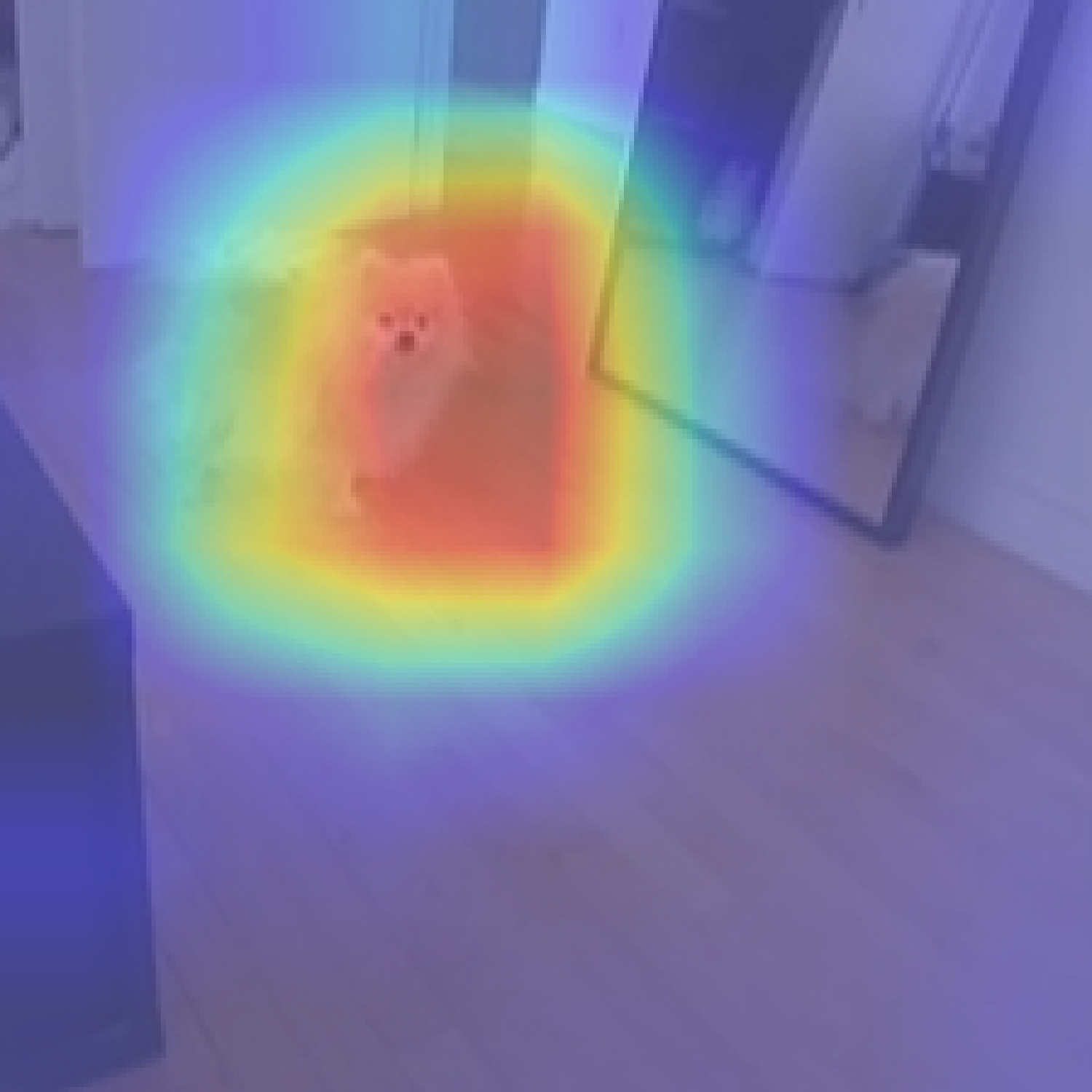} &
        \includegraphics[width=0.065\textwidth]{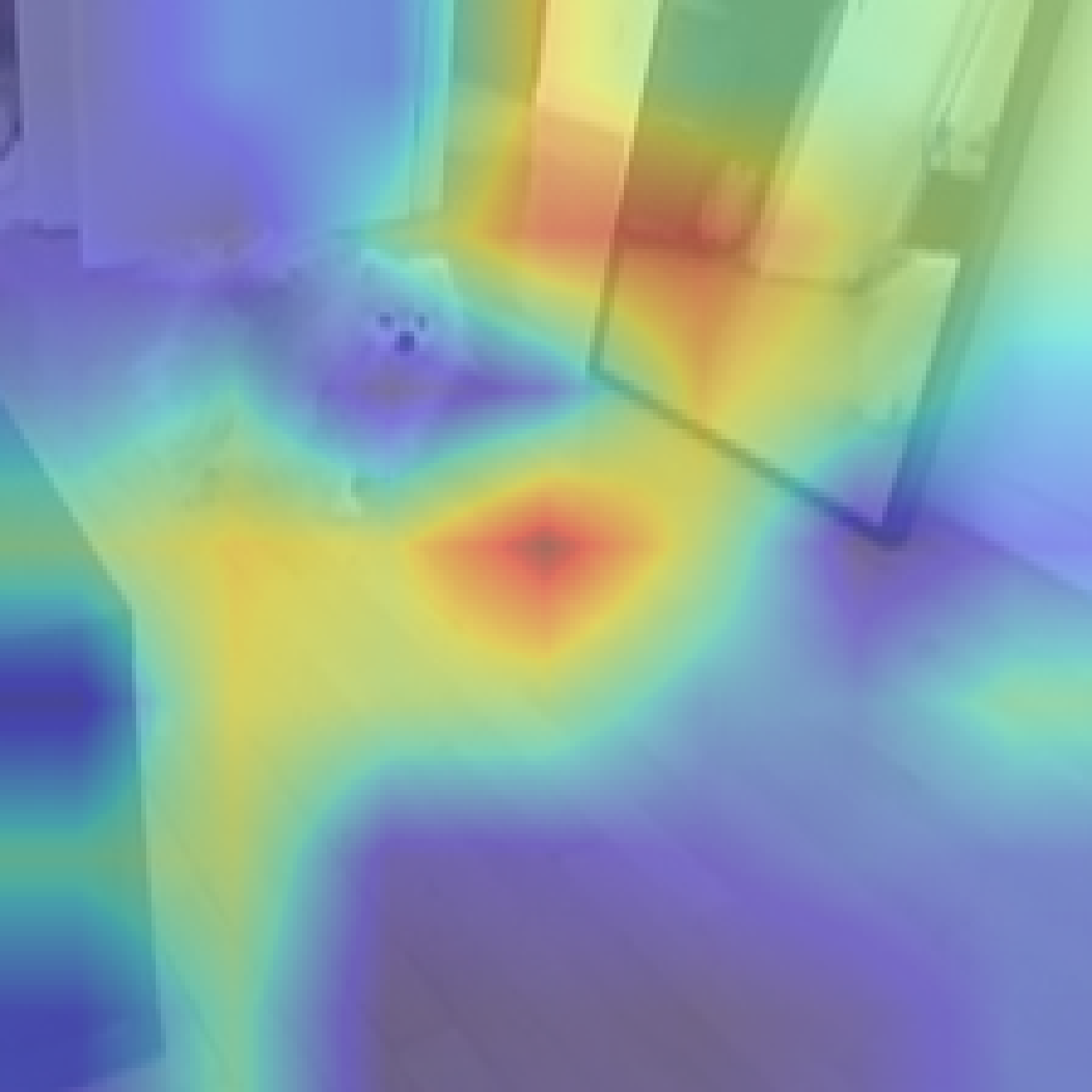} &
        \includegraphics[width=0.065\textwidth]{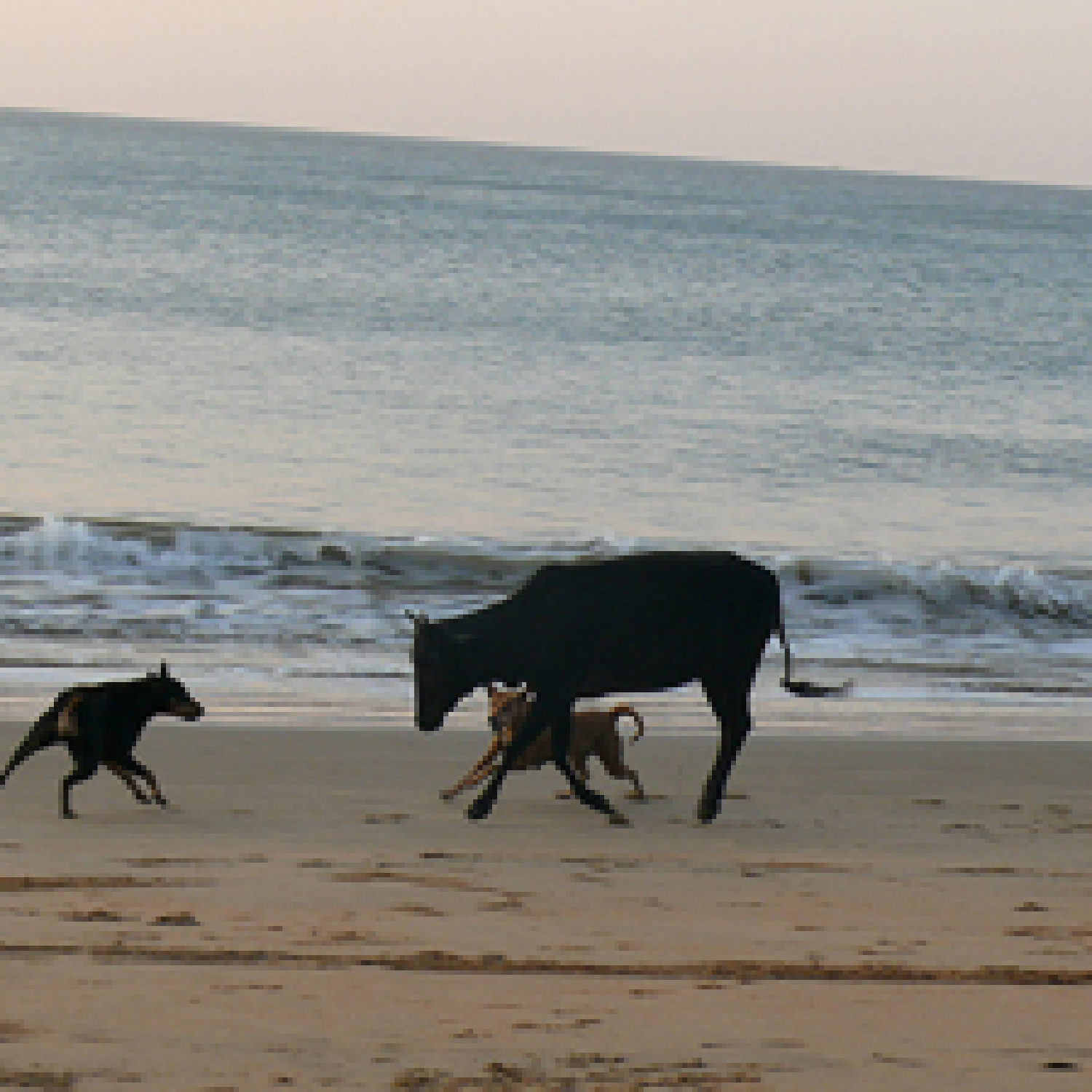} &
        \includegraphics[width=0.065\textwidth]{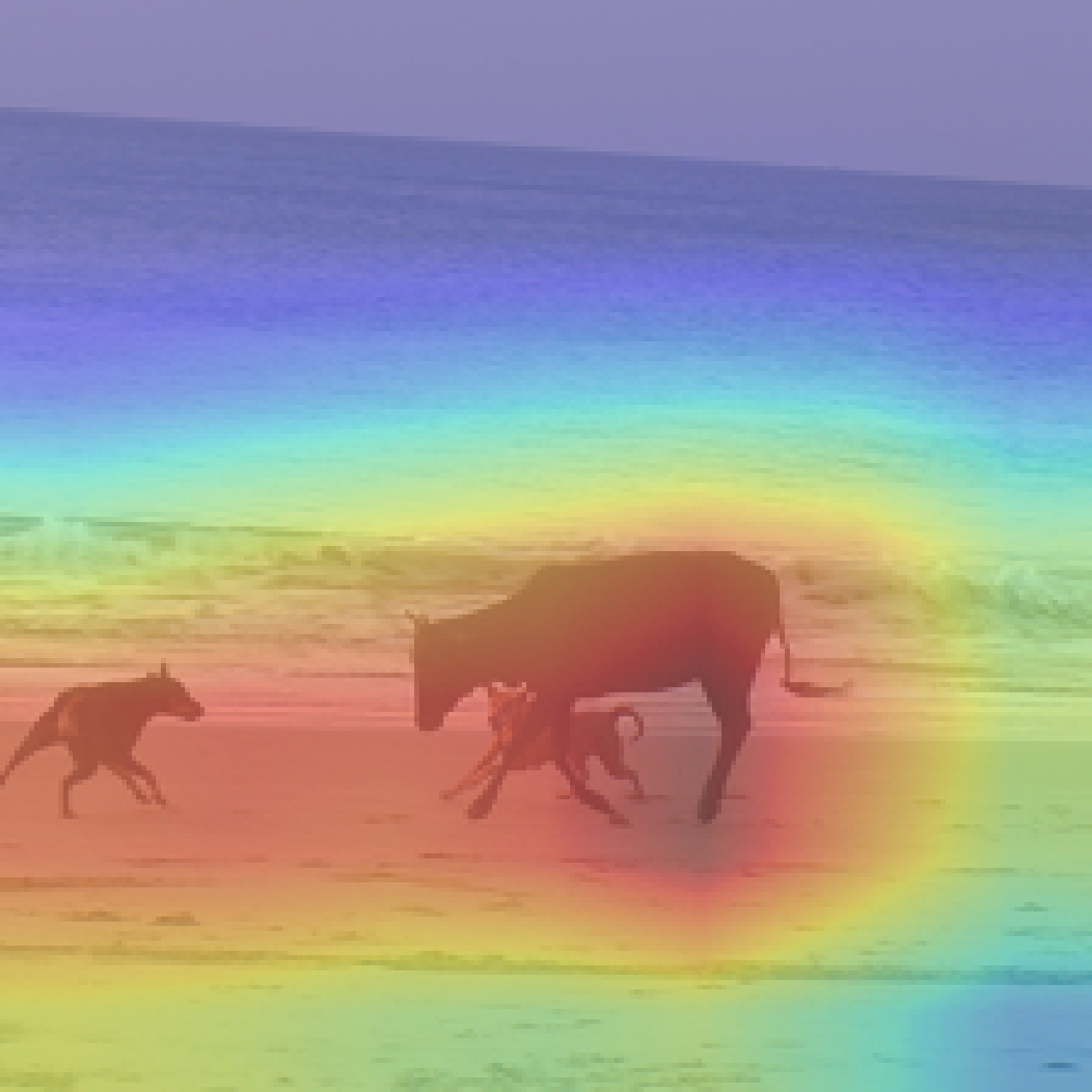} &
        \includegraphics[width=0.065\textwidth]{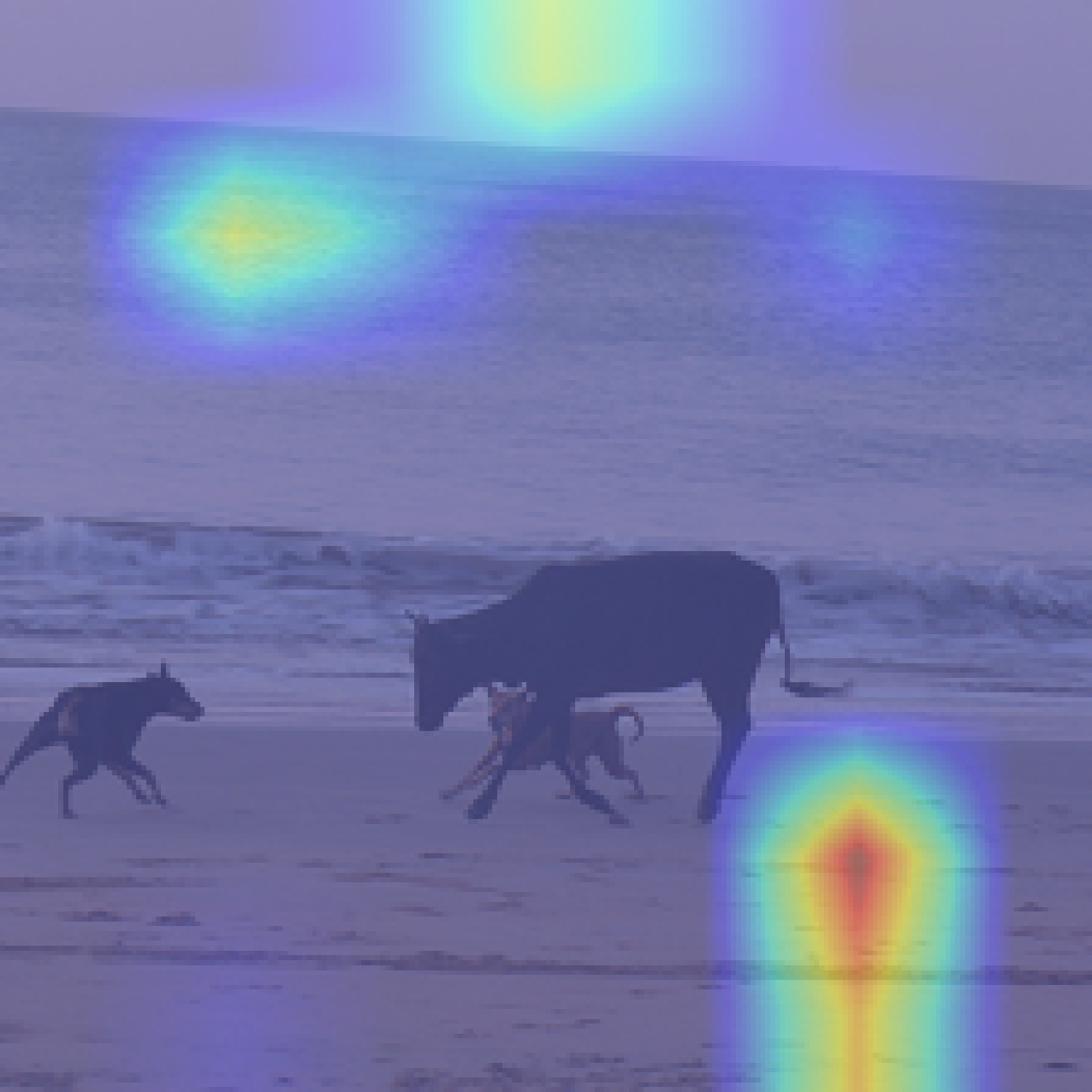} &
        \includegraphics[width=0.065\textwidth]{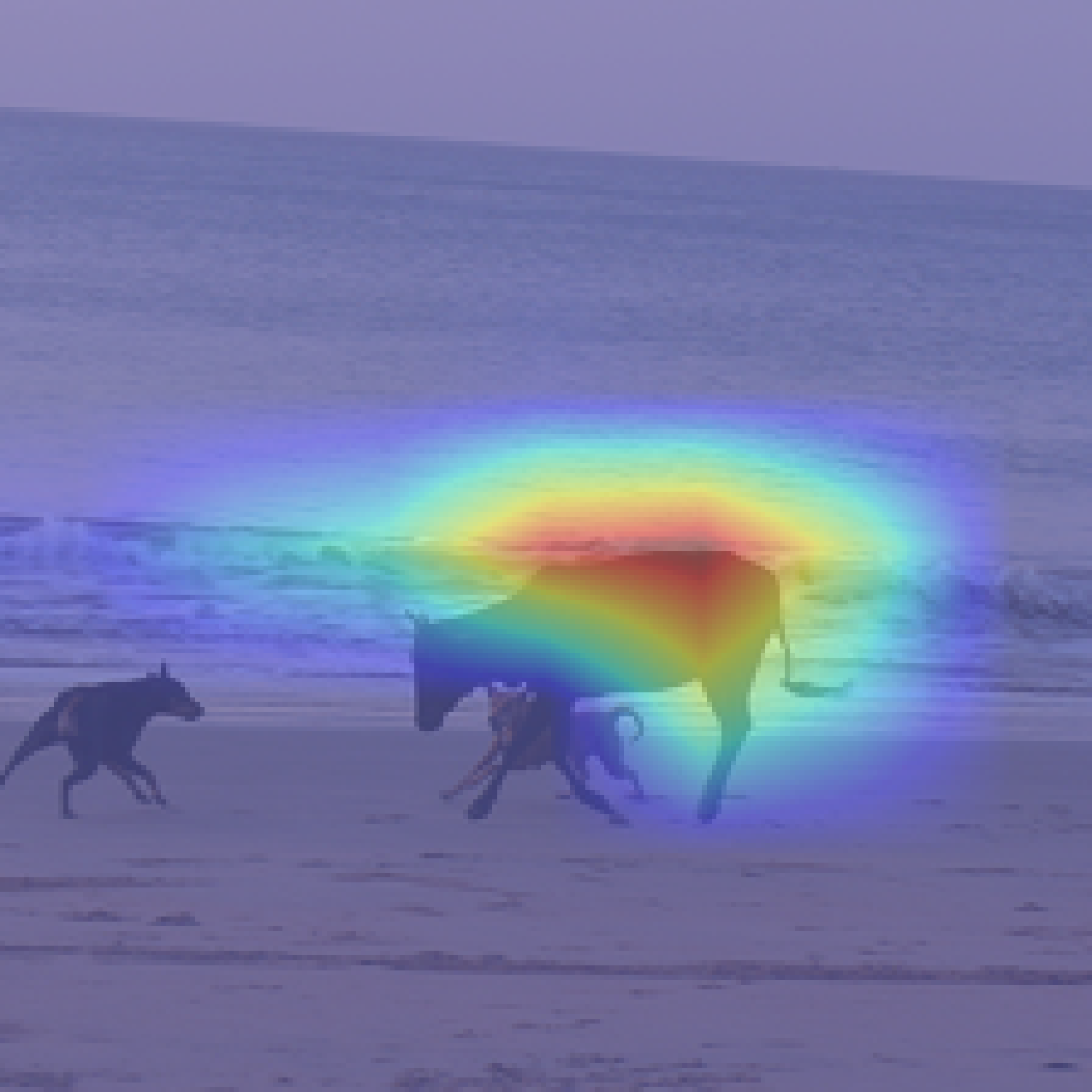} &
        \includegraphics[width=0.065\textwidth]{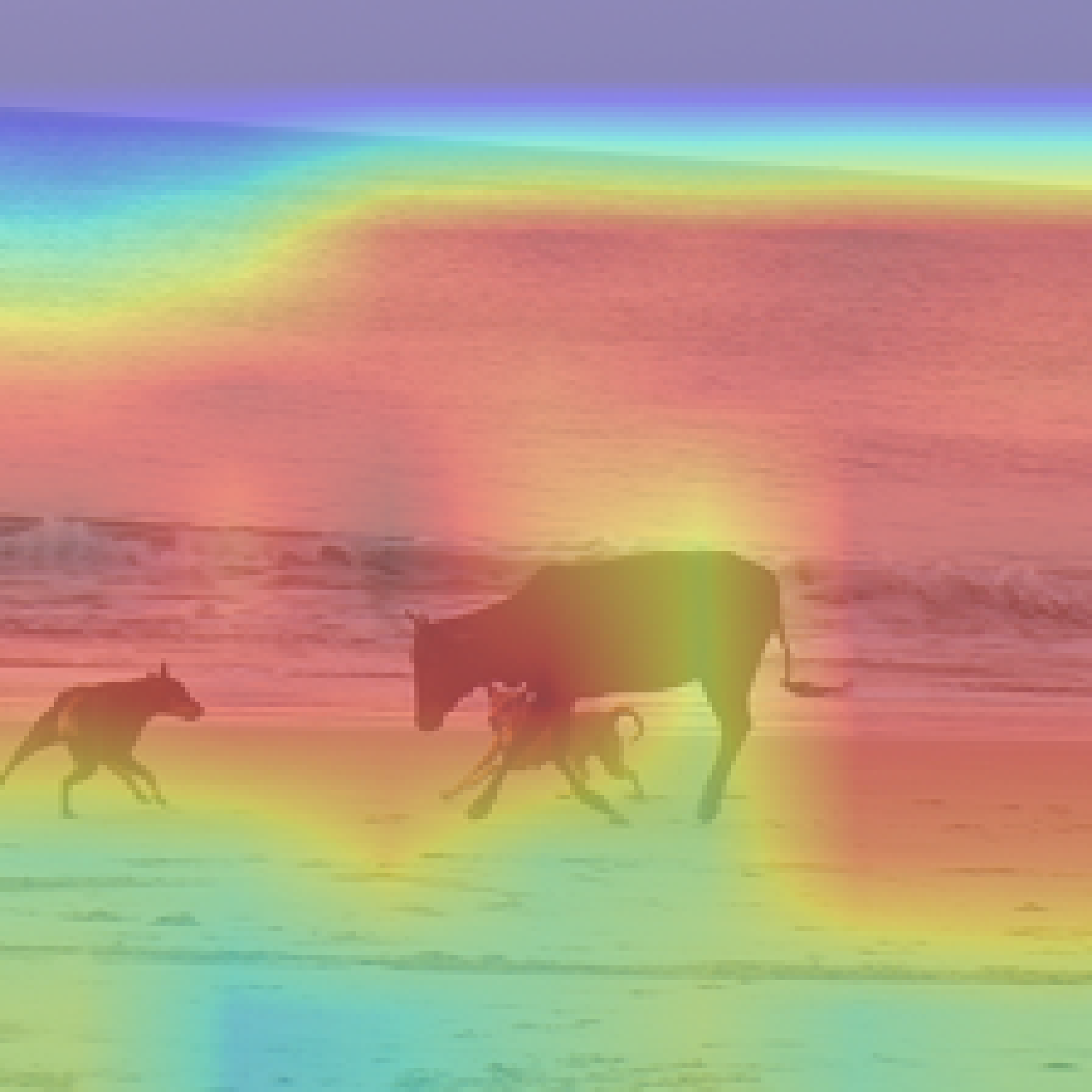} \\[0.3em]

        % Row 4: Spawrious (placeholder)
        \includegraphics[width=0.065\textwidth]{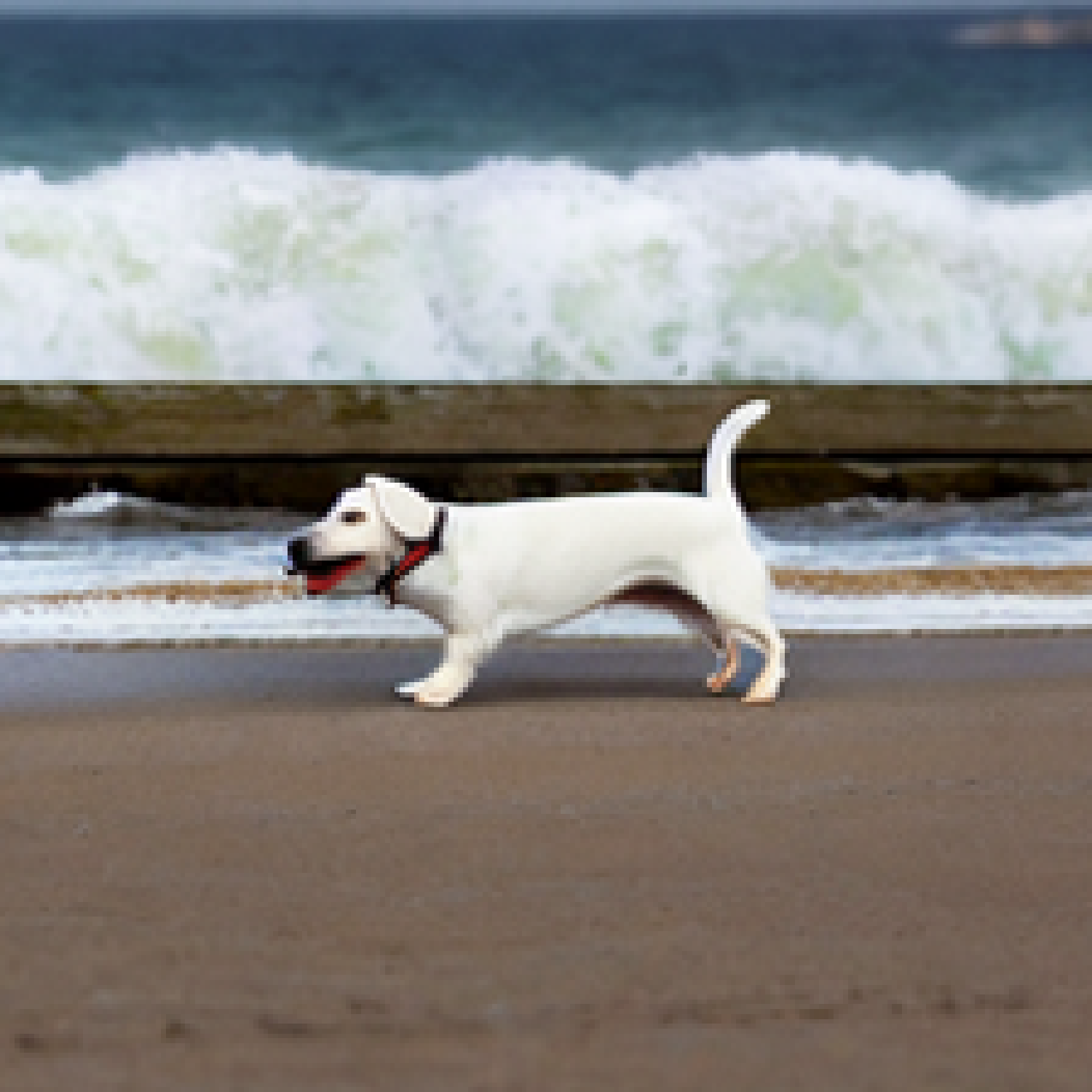} &
        \includegraphics[width=0.065\textwidth]{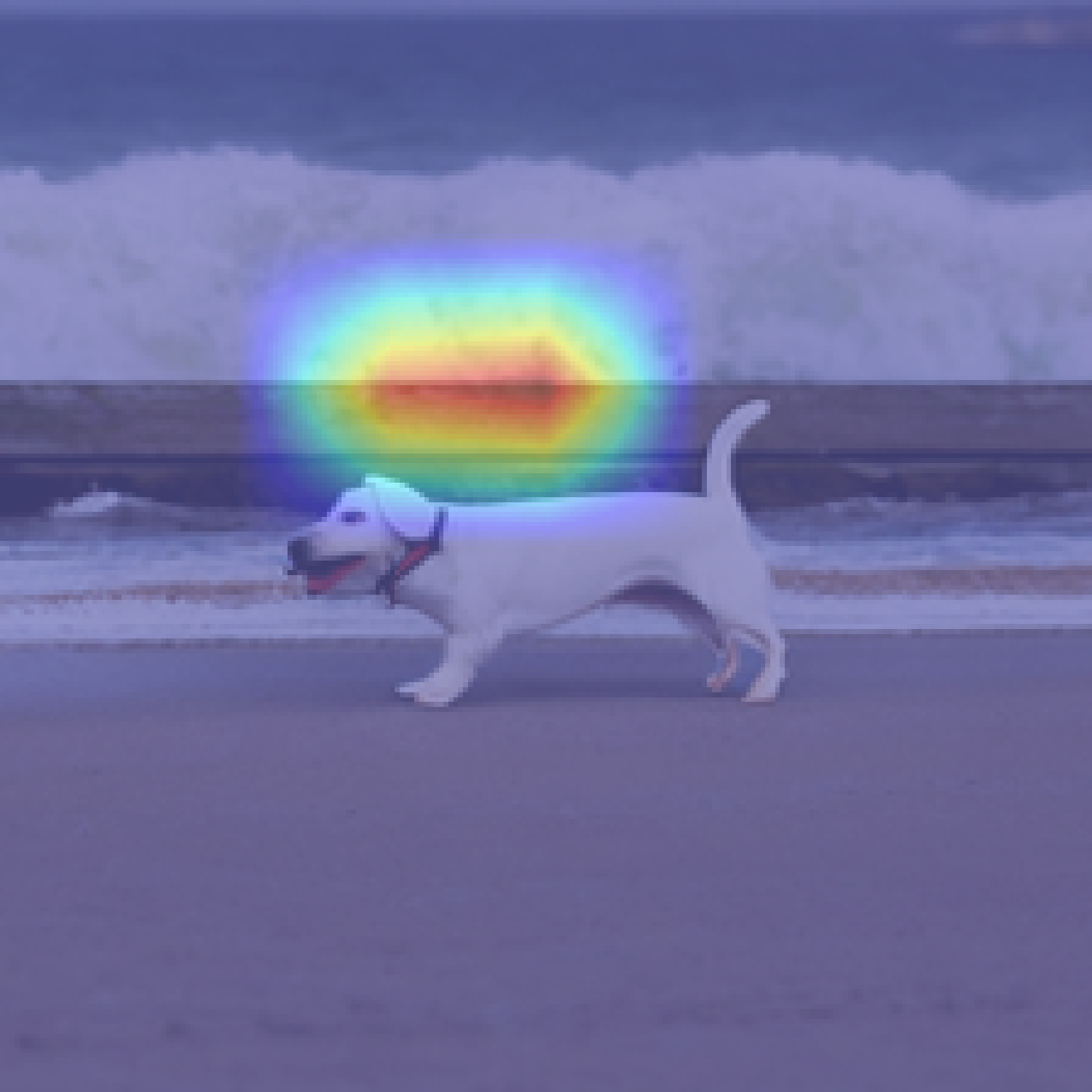} &
        \includegraphics[width=0.065\textwidth]{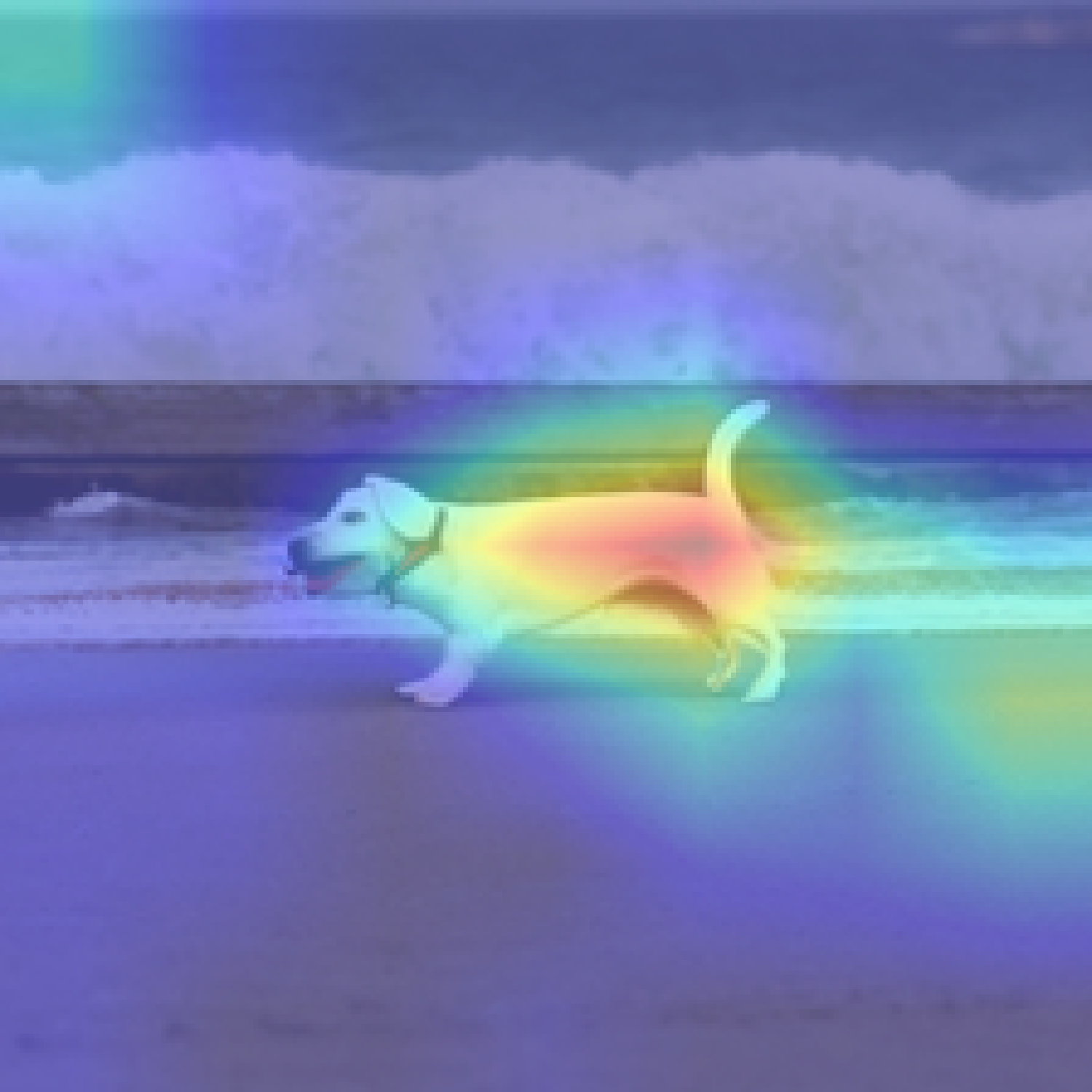} &
        \includegraphics[width=0.065\textwidth]{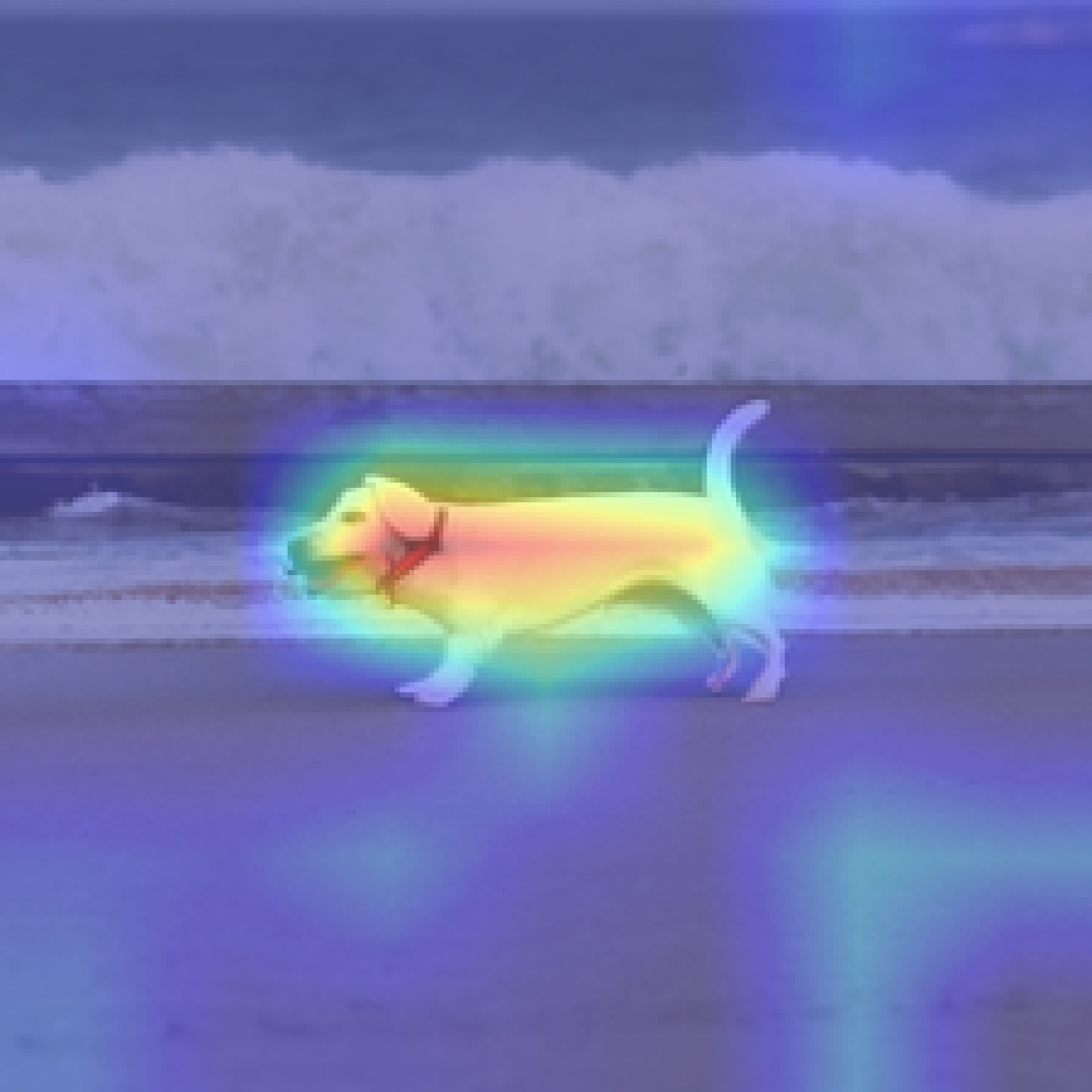} &
        \includegraphics[width=0.065\textwidth]{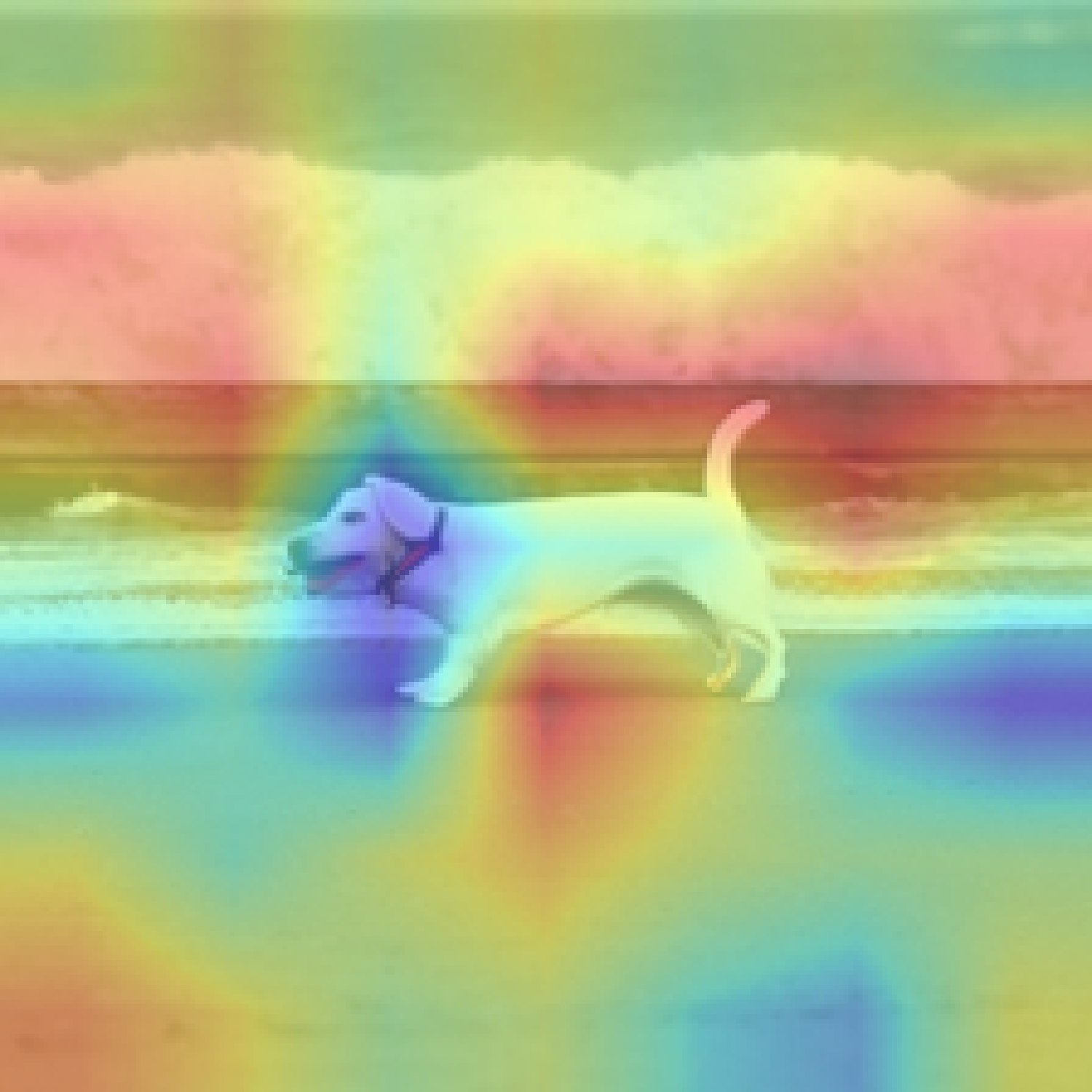} &
        \includegraphics[width=0.065\textwidth]{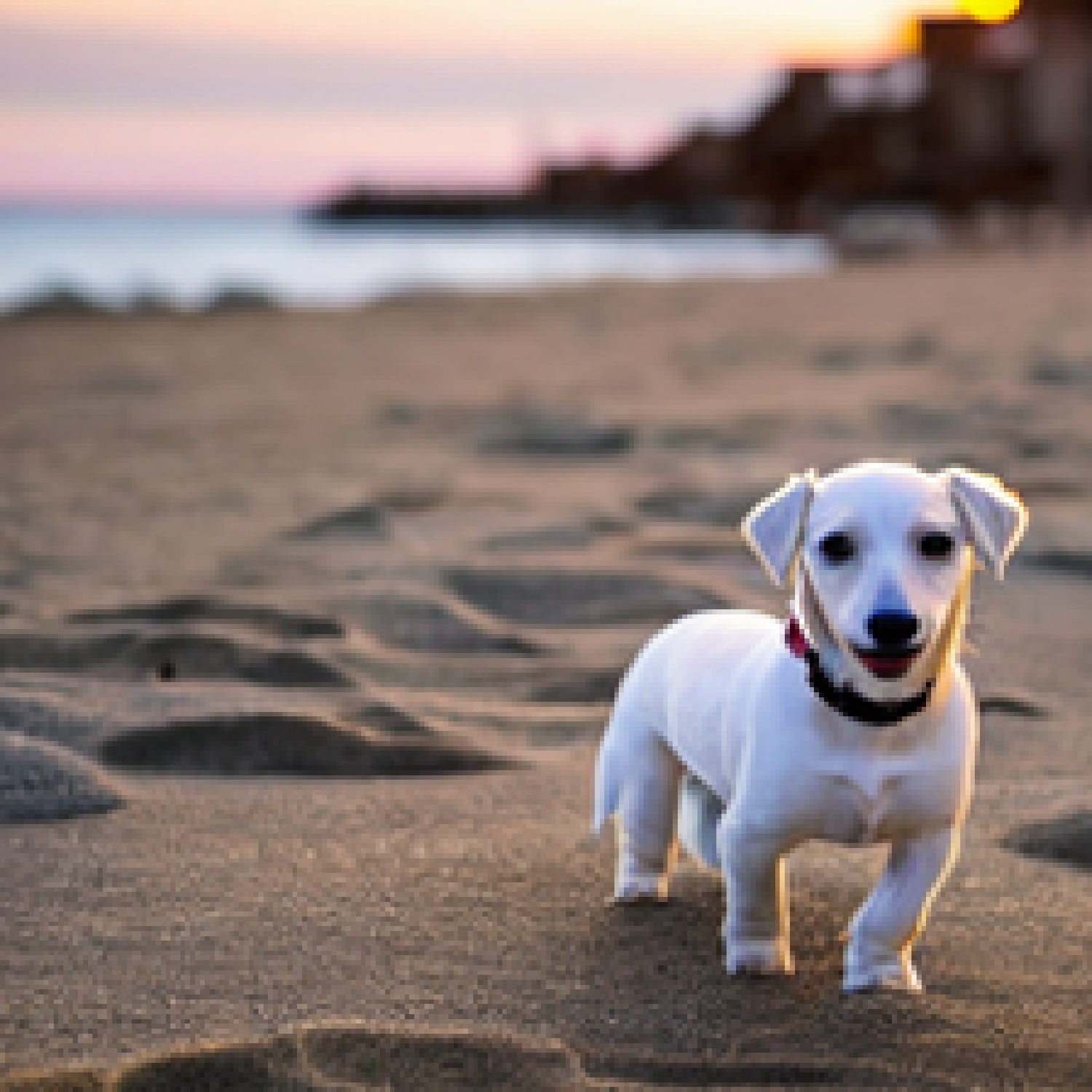} &
        \includegraphics[width=0.065\textwidth]{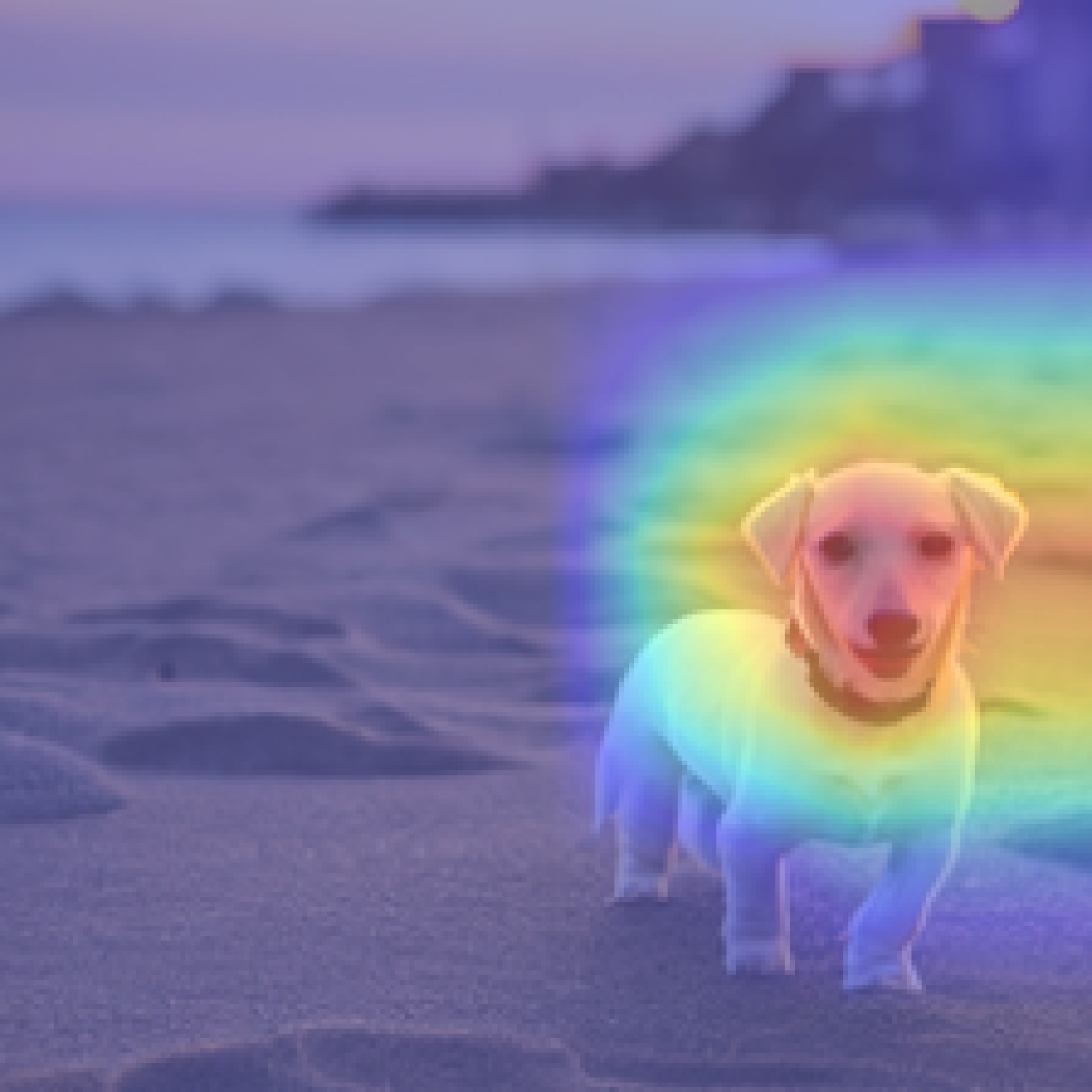} &
        \includegraphics[width=0.065\textwidth]{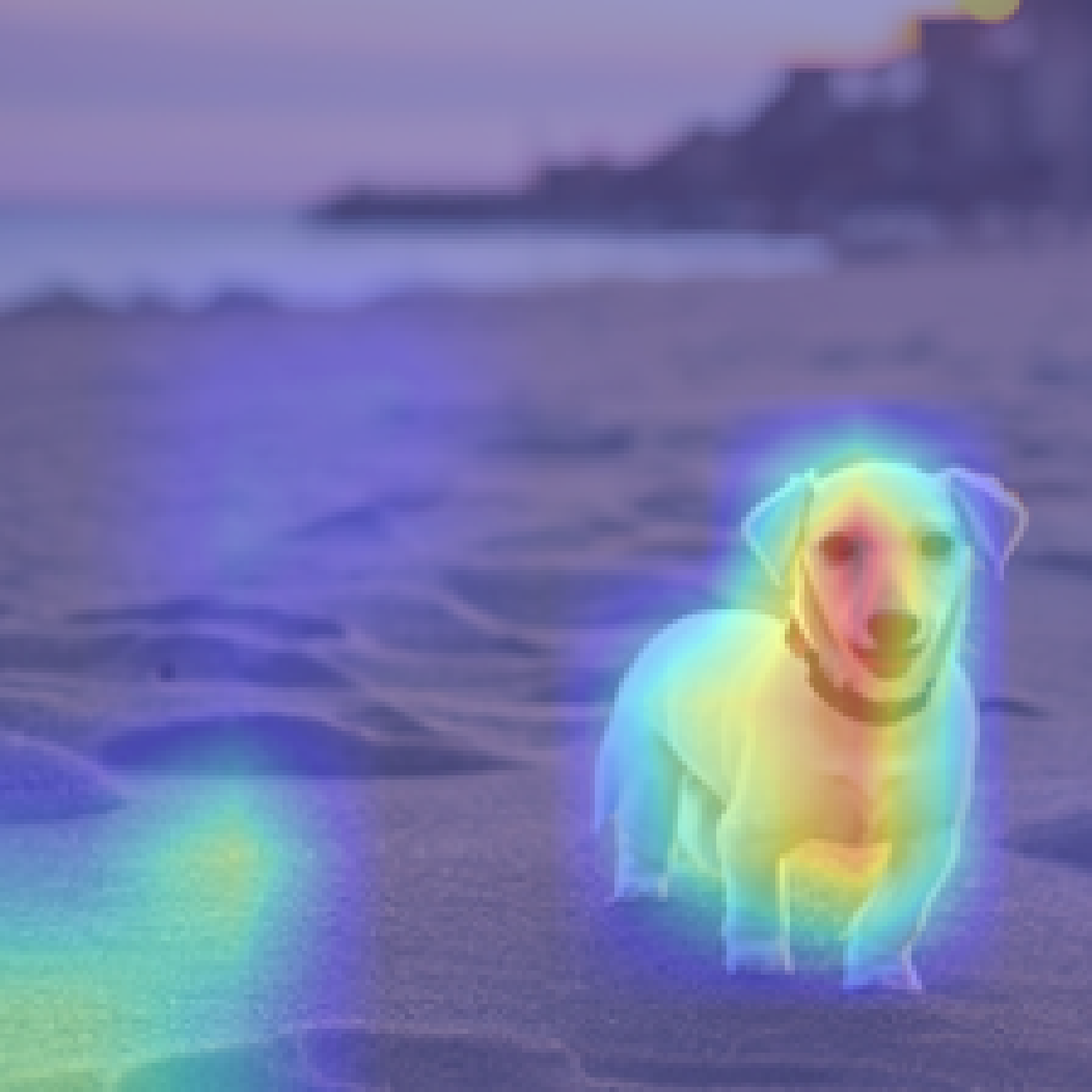} &
        \includegraphics[width=0.065\textwidth]{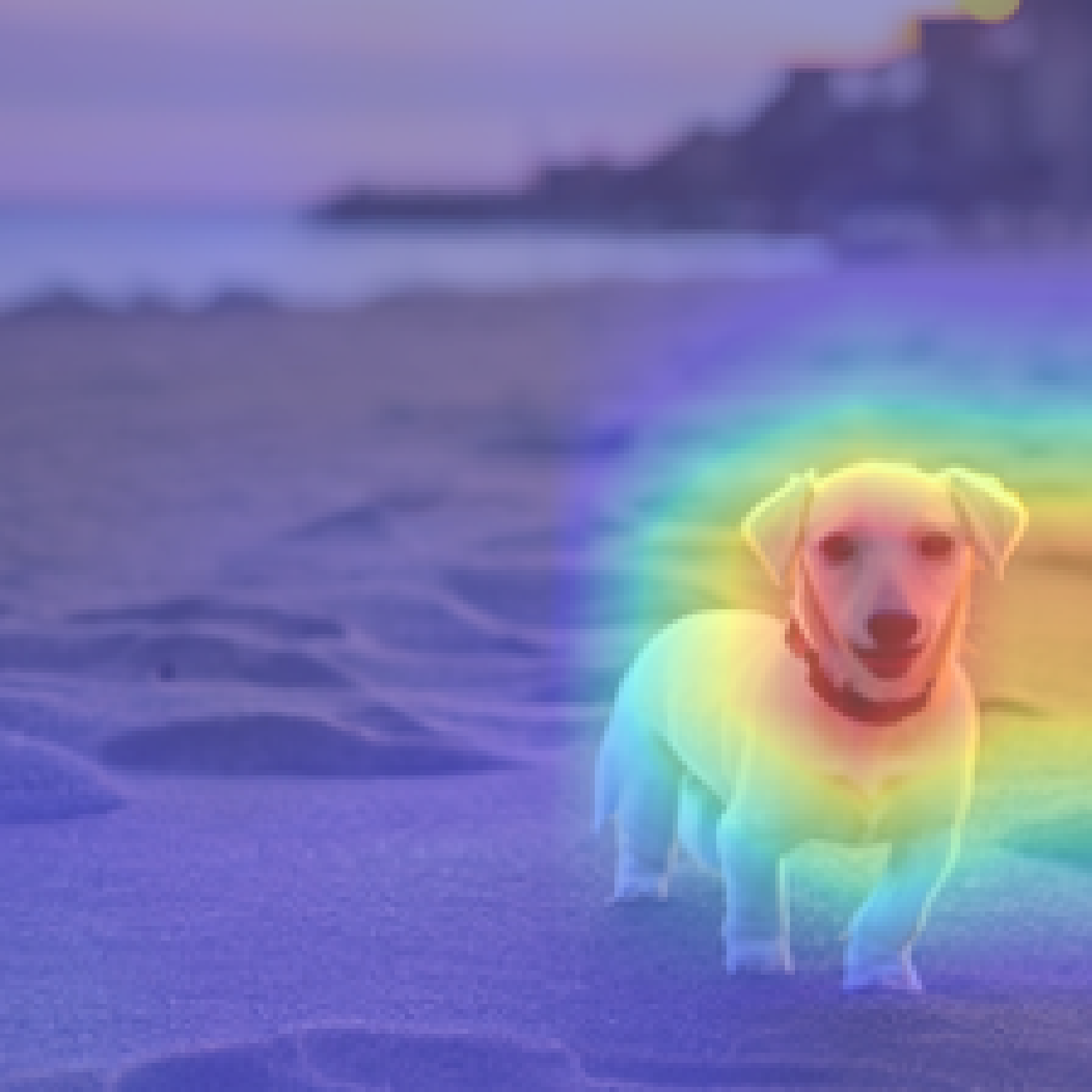} &
        \includegraphics[width=0.065\textwidth]{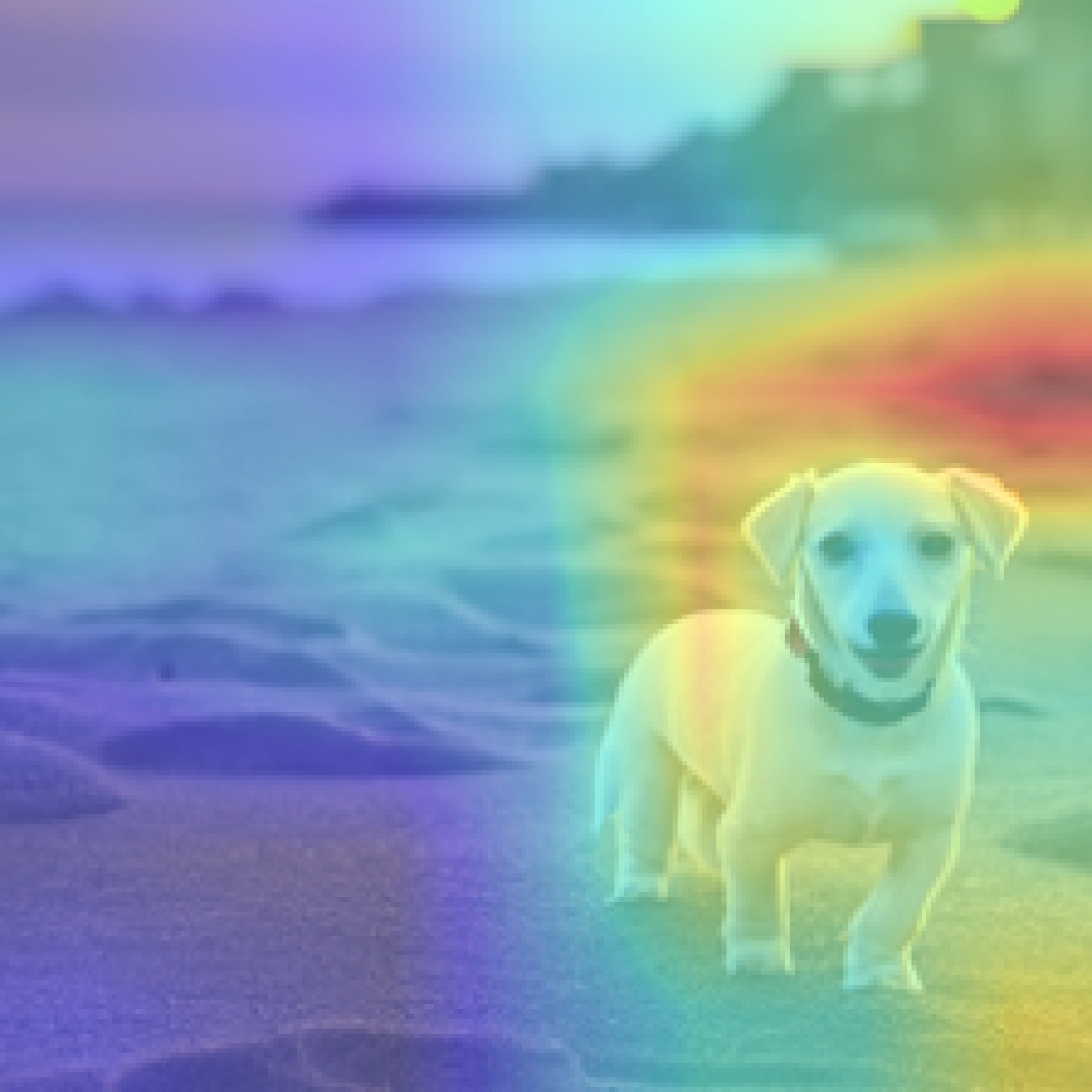} \\[0.3em]

        % Column labels
        \fontsize{8pt}{8pt}\selectfont
        (a) Original &\fontsize{8pt}{8pt}\selectfont (b) ERM &\fontsize{8pt}{8pt}\selectfont (c) CLIP &\fontsize{8pt}{8pt}\selectfont (d) $\omega_{\mathrm{rel}}$ \fontsize{8pt}{8pt}\selectfont &(e) $\omega_{\mathrm{irr}}$ &\fontsize{8pt}{8pt}\selectfont
        (a) Original &\fontsize{8pt}{8pt}\selectfont (b) ERM &\fontsize{8pt}{8pt}\selectfont (c) CLIP &\fontsize{8pt}{8pt}\selectfont (d) $\omega_{\mathrm{rel}}$ &\fontsize{8pt}{8pt}\selectfont (e) $\omega_{\mathrm{irr}}$ \\
    \end{tabular}
    \caption{Visualization of GradCAM maps across different models and datasets.  
Rows: (1) Waterbirds-95\%, (2) Waterbirds-100\%, (3) MetaShift, (4) Spawrious.  
Each group of five columns ((a)–(e)) shows: original image, GradCAM maps of ERM, CLIP, $\omega_{\mathrm{rel}}$, and $\omega_{\mathrm{irr}}$.  
In the left group, guided by superclass information from CLIP, SupER's classifier $\omega_{\mathrm{rel}}$ successfully learns to focus on superclass-relevant features.  
The right group shows occasional cases where CLIP exhibits internal bias; nonetheless, $\omega_{\mathrm{rel}}$ is able to focus on the correct and more complete superclass-relevant features through effective feature disentanglement.}
    \label{fig:gradcam_vis}
    \vspace{-0.2cm}
\end{figure*}

\subsubsection{Visualization Analysis of Feature Attention}

In this section, we analyze the visualized gradient-based attribution maps from different test samples across ERM, CLIP, and SupER to better understand each model's focus areas and feature disentanglement quality. More visualizations are provided in Appendix~\ref{app:AVR}.

\textbf{SupER achieves effective disentanglement of superclass-relevant and irrelevant features.}
As shown in the left five columns of Figure~\ref{fig:gradcam_vis}, while ERM tends to rely on spurious features for prediction, the attribution maps derived from CLIP can be considered as suitable guidance for superclass semantic information. Furthermore, in SupER, $\omega_{\mathrm{rel}}$ and $\omega_{\mathrm{irr}}$ exhibit clear attention to superclass-relevant and superclass-irrelevant features respectively, which validates our approach. 

\textbf{SupER can adjust internal biases in CLIP and significantly outperform CLIP.}
While CLIP’s attention in the left-hand (c) column of Figure~\ref{fig:gradcam_vis} can provide general guidance for superclass information, occasional cases from the right-hand (c) column reveal that internal biases in CLIP may lead it to focus on incorrect or incomplete features of the superclass. However, as shown in the right-hand (d) column, SupER is able to reduce these biases and redirect attention toward more accurate and comprehensive superclass-relevant features. 
Furthermore, Table~\ref{tab:clip_vs_super} shows that SupER achieves substantially higher accuracy than the CLIP teacher. This robustness and improvement stems from the superclass-level guidance, the disentanglement ability of the $\beta$-VAE, and our principled feature-usage mechanism (see the final paragraph of Section~\ref{sec:superclass-disentangle} for a detailed discussion).

\begin{table*}[t]
  \centering
  \begin{minipage}[t]{0.48\textwidth}
    \centering
    {\fontsize{8pt}{8.5pt}\selectfont
    \caption{Variance of accuracy across groups (\%) . \textbf{Bold} indicates the smallest across all baselines.}
    \label{tab:variance_wb_sc}
    \vspace{4pt}   
    \begin{tabular*}{\textwidth}{@{\extracolsep{\fill}}l c c}
      \toprule
      Method      & Waterbirds-95\% & Waterbirds-100\%\\
      \midrule
      ERM         & 245.9   & 778.1 \\
      CVaR\,DRO   & 154.6   & 528.8 \\
      LfF         & 89.2   & 442.1 \\
      GALS        & 126.7   & 516.5 \\
      JTT         & 6.1   & 405.0  \\
      CnC         & 16.3   & 347.6\\
      SupER (Ours) & \textbf{6.0} & \textbf{16.0}\\
      \midrule
      UW          & 12.7   & 536.2   \\
      IRM         & 127.2   & 479.8  \\
      GroupDRO    & 28.0   & 495.1 \\
      DFR         & 14.2   & 573.0 \\
      \bottomrule
    \end{tabular*}
    }
  \end{minipage}%
  \hfill
  \begin{minipage}[t]{0.49\textwidth}
    \centering
    {\fontsize{8pt}{8pt}\selectfont
    \captionof{figure}{Ablation of feature disentanglement (\(\beta\)) and superclass guidance (\(\lambda_2\)) strength.}
    \label{fig:ablation_combo}
    \includegraphics[width=0.71\textwidth]{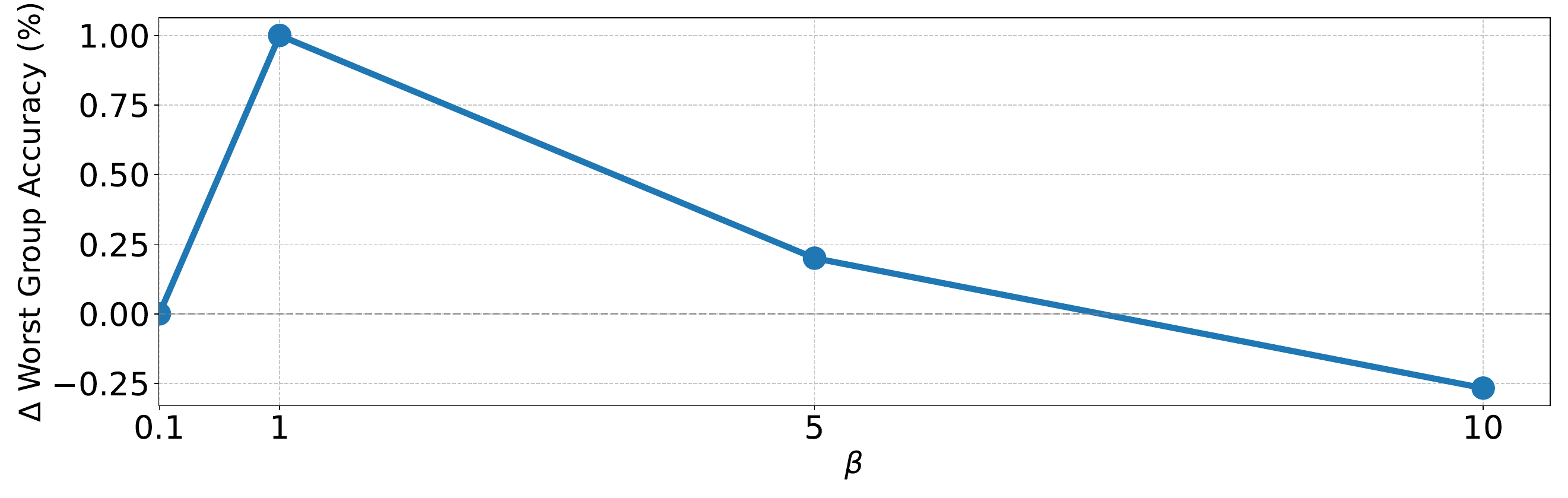}\\[-0pt] 
    \fontsize{8pt}{8pt}\selectfont (a) Effect of $\beta$ on SpuCo Dogs relative to the $\beta=0.1$ setting.\\[2pt]
    \includegraphics[width=0.71\textwidth]{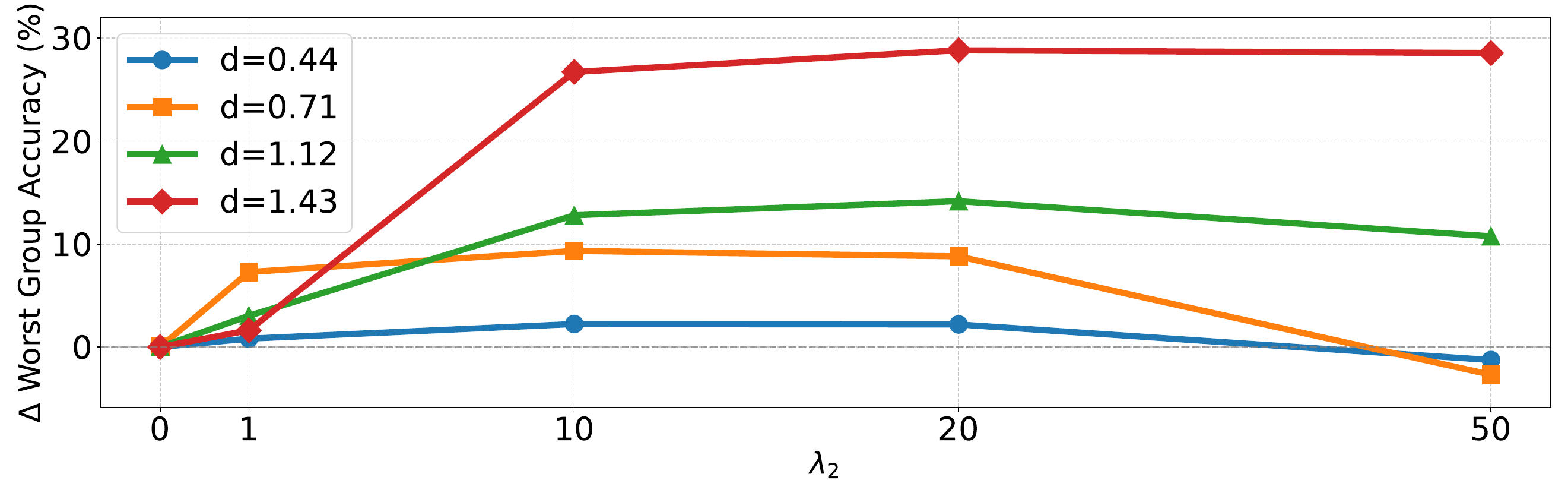}\\ [-0pt]
    \fontsize{8pt}{8pt}\selectfont (b) Effect of $\lambda_2$ on MetaShift relative to the $\lambda_2=0$ setting.
    }
  \end{minipage}
  \vspace{-0.2cm}
\end{table*}

\begin{table*}[t]
  \centering
  {\fontsize{8pt}{8.5pt}\selectfont
  \caption{Ablation results on Spawrious under different prompt configurations. All values indicate the change in worst group accuracy (\%) relative to the setting $m=1$, $\text{superclass}=\texttt{dog}$.}
  \label{tab:prompt_ablation}
  \setlength{\tabcolsep}{3pt}
  \begin{tabular*}{\textwidth}{@{\extracolsep{\fill}}%
      >{\centering\arraybackslash}p{0.9cm}%  #Prompts
      >{\centering\arraybackslash}p{1.1cm}%  Superclass
      >{\centering\arraybackslash}p{1.2cm}%  O2O-Easy
      >{\centering\arraybackslash}p{1.6cm}%  O2O-Medium
      >{\centering\arraybackslash}p{1.3cm}%  O2O-Hard
      >{\centering\arraybackslash}p{1.3cm}%  M2M-Easy
      >{\centering\arraybackslash}p{1.7cm}%  M2M-Medium
      >{\centering\arraybackslash}p{1.3cm}%  M2M-Hard
      >{\centering\arraybackslash}p{1.0cm}}%   Average
    \toprule
    \#Prompts & Superclass & O2O-Easy & O2O-Medium & O2O-Hard & M2M-Easy & M2M-Medium & M2M-Hard & Average \\
    \midrule
    1   & \texttt{dog}     & 0.0  & 0.0  & 0.0  & 0.0  & 0.0  & 0.0  & 0.0  \\
    2   & \texttt{dog}     & +0.7  & -2.2   & -2.3   & +0.8   & -9.6   & -2.4   & -2.5   \\
    5   & \texttt{dog}     & +0.3   & -2.5   & -2.9   & -1.1   & -9.4   & +1.1   & -2.4   \\
    1   & \texttt{animal}  & -4.0   & -1.1   & -5.6   & -4.9   & -7.5   & -5.6   & -4.8   \\
    \bottomrule
  \end{tabular*}
  }
  \vspace{-0.4cm}
\end{table*}

\subsection{Ablation Study}
\label{sec:ablation}
\textbf{Text prompt.}
To examine the impact of superclass guidance, we vary the text prompts provided to CLIP in two aspects. First, although our main experiments are based on a single text prompt, our general framework in Section~\ref{sec:method} allows for multiple prompts. Second, we are interested in the effect of prompt specificity, particularly in terms of superclass hierarchy. In Table~\ref{tab:prompt_ablation}, we report SupER's performance on the Spawrious datasets under different prompt configurations: (1) increasing the number of prompts, and (2) changing the superclass label $y^{\text{super}}$ from ``dog'' to the more general ``animal''. Results show that using multiple prompts generally hurts performance. This may occur because attention maps from different prompts could highlight distinct non-superclass regions due to imperfect guidance, and averaging them mixes biases from each prompt. Moreover, replacing the prompt ``dog'' with ``animal'' leads to a drop in accuracy, likely due to the coarser semantic alignment between the generalized superclass and the visual features. (Additional results and more ablation experiments are provided in Appendix~\ref{app:ablation}.)

\textbf{Feature disentanglement strength.}
We study how the strength of feature disentanglement, controlled by the $\beta$ coefficient in the $\beta$-VAE objective, affects model performance. We vary $\beta$ to observe its impact on SupER's worst group accuracy. Figure~\ref{fig:ablation_combo}(a) shows results on the SpuCo Dogs dataset. As $\beta$ increases, the worst group accuracy initially rises and then declines. This trend indicates that moderate feature disentanglement benefits semantic feature extraction and superclass-relevant feature utilization, whereas overly strong disentanglement can distort task-relevant information.

\textbf{Degree of superclass guidance.} We study the effect of varying the weight $\lambda_2$ of the alignment loss $\mathcal{L}^{\text{ATT}}_{\phi, \omega_{\mathrm{rel}}, \omega_{\mathrm{irr}}}(\mathbf{x}, y)$ in Algorithm~\ref{alg:training_dafd}, which governs the strength of superclass guidance from CLIP. Results on the four MetaShift subsets in Figure~\ref{fig:ablation_combo}(b) indicate that as $\lambda_2$ increases, worst group accuracy initially rises and then declines. The effectiveness of superclass guidance becomes more pronounced as distribution shift intensifies (larger $d$), with larger optimal values of $\lambda_2$.
These results clearly reveal a trade-off between external guidance and model autonomy: excessive reliance on superclass guidance may prevent the model from learning discriminative features, while ignoring guidance altogether increases the risk of learning spurious correlations between background and labels.

\section{Conclusion}
\label{sec:conclusion}
 In this work, we propose SupER, a group-label-free framework that leverages superclass-level semantics as a more intrinsic and reliable signal for mitigating spurious correlations. SupER integrates feature disentanglement with superclass guidance, together with a principled feature-usage mechanism supported by theory. Across multiple domain generalization benchmarks, SupER consistently delivers strong performance, and is particularly effective when the training and test groups are not identical or when spurious correlations are highly complex. Beyond spurious correlation mitigation, our framework also illustrates a broader paradigm in which a teacher (e.g., CLIP) provides weak and coarse supervision that enables a student (e.g., SupER) not only to reduce the teacher’s biases but also significantly surpass the teacher's performance.

\section*{Acknowledgments}
QL acknowledges support of NSF DMS-2523382. YS was supported in part by the National Institutes of Health (grant no. 1R01EB036530-01A1).

\bibliographystyle{plainnat}
\bibliography{references}

@article{lynch2023spawrious,
  title={Spawrious: A benchmark for fine control of spurious correlation biases},
  author={Lynch, Aengus and Dovonon, Gb{\`e}tondji JS and Kaddour, Jean and Silva, Ricardo},
  journal={arXiv preprint arXiv:2303.05470},
  year={2023}
}

@article{levy2020large,
  title={Large-scale methods for distributionally robust optimization},
  author={Levy, Daniel and Carmon, Yair and Duchi, John C and Sidford, Aaron},
  journal={Advances in Neural Information Processing Systems},
  volume={33},
  pages={8847--8860},
  year={2020}
}

@inproceedings{krueger2021out,
  title={Out-of-distribution generalization via risk extrapolation (rex)},
  author={Krueger, David and Caballero, Ethan and Jacobsen, Joern-Henrik and Zhang, Amy and Binas, Jonathan and Zhang, Dinghuai and Le Priol, Remi and Courville, Aaron},
  booktitle={International conference on machine learning},
  pages={5815--5826},
  year={2021},
  organization={PMLR}
}

@article{ye2024spurious,
  title={Spurious correlations in machine learning: A survey},
  author={Ye, Wenqian and Zheng, Guangtao and Cao, Xu and Ma, Yunsheng and Zhang, Aidong},
  journal={arXiv preprint arXiv:2402.12715},
  year={2024}
}

@article{peters2016causal,
  title={Causal inference by using invariant prediction: identification and confidence intervals},
  author={Peters, Jonas and B{\"u}hlmann, Peter and Meinshausen, Nicolai},
  journal={Journal of the Royal Statistical Society Series B: Statistical Methodology},
  volume={78},
  number={5},
  pages={947--1012},
  year={2016},
  publisher={Oxford University Press}
}

@article{arjovsky2019invariant,
  title={Invariant risk minimization},
  author={Arjovsky, Martin and Bottou, L{\'e}on and Gulrajani, Ishaan and Lopez-Paz, David},
  journal={arXiv preprint arXiv:1907.02893},
  year={2019}
}

@article{lee2021learning,
  title={Learning debiased representation via disentangled feature augmentation},
  author={Lee, Jungsoo and Kim, Eungyeup and Lee, Juyoung and Lee, Jihyeon and Choo, Jaegul},
  journal={Advances in Neural Information Processing Systems},
  volume={34},
  pages={25123--25133},
  year={2021}
}

@article{yang2022chroma,
  title={Chroma-vae: Mitigating shortcut learning with generative classifiers},
  author={Yang, Wanqian and Kirichenko, Polina and Goldblum, Micah and Wilson, Andrew G},
  journal={Advances in Neural Information Processing Systems},
  volume={35},
  pages={20351--20365},
  year={2022}
}

@inproceedings{zhang2022towards,
  title={Towards principled disentanglement for domain generalization},
  author={Zhang, Hanlin and Zhang, Yi-Fan and Liu, Weiyang and Weller, Adrian and Sch{\"o}lkopf, Bernhard and Xing, Eric P},
  booktitle={Proceedings of the IEEE/CVF conference on computer vision and pattern recognition},
  pages={8024--8034},
  year={2022}
}

@inproceedings{wang2024effect,
  title={On the effect of key factors in spurious correlation: A theoretical perspective},
  author={Wang, Yipei and Wang, Xiaoqian},
  booktitle={International Conference on Artificial Intelligence and Statistics},
  pages={3745--3753},
  year={2024},
  organization={PMLR}
}

@article{puli2021out,
  title={Out-of-distribution generalization in the presence of nuisance-induced spurious correlations},
  author={Puli, Aahlad and Zhang, Lily H and Oermann, Eric K and Ranganath, Rajesh},
  journal={arXiv preprint arXiv:2107.00520},
  year={2021}
}

@article{puli2022nuisances,
  title={Nuisances via negativa: Adjusting for spurious correlations via data augmentation},
  author={Puli, Aahlad and Joshi, Nitish and Wald, Yoav and He, He and Ranganath, Rajesh},
  journal={arXiv preprint arXiv:2210.01302},
  year={2022}
}

@inproceedings{makar2022causally,
  title={Causally motivated shortcut removal using auxiliary labels},
  author={Makar, Maggie and Packer, Ben and Moldovan, Dan and Blalock, Davis and Halpern, Yoni and D’Amour, Alexander},
  booktitle={International Conference on Artificial Intelligence and Statistics},
  pages={739--766},
  year={2022},
  organization={PMLR}
}

@inproceedings{creager2021environment,
  title={Environment inference for invariant learning},
  author={Creager, Elliot and Jacobsen, J{\"o}rn-Henrik and Zemel, Richard},
  booktitle={International Conference on Machine Learning},
  pages={2189--2200},
  year={2021},
  organization={PMLR}
}

@inproceedings{liu2021just,
  title={Just train twice: Improving group robustness without training group information},
  author={Liu, Evan Z and Haghgoo, Behzad and Chen, Annie S and Raghunathan, Aditi and Koh, Pang Wei and Sagawa, Shiori and Liang, Percy and Finn, Chelsea},
  booktitle={International Conference on Machine Learning},
  pages={6781--6792},
  year={2021},
  organization={PMLR}
}

@article{nam2020learning,
  title={Learning from failure: De-biasing classifier from biased classifier},
  author={Nam, Junhyun and Cha, Hyuntak and Ahn, Sungsoo and Lee, Jaeho and Shin, Jinwoo},
  journal={Advances in Neural Information Processing Systems},
  volume={33},
  pages={20673--20684},
  year={2020}
}

@article{sagawa2019distributionally,
  title={Distributionally robust neural networks for group shifts: On the importance of regularization for worst-case generalization},
  author={Sagawa, Shiori and Koh, Pang Wei and Hashimoto, Tatsunori B and Liang, Percy},
  journal={arXiv preprint arXiv:1911.08731},
  year={2019}
}

@inproceedings{teney2022evading,
  title={Evading the simplicity bias: Training a diverse set of models discovers solutions with superior ood generalization},
  author={Teney, Damien and Abbasnejad, Ehsan and Lucey, Simon and Van den Hengel, Anton},
  booktitle={Proceedings of the IEEE/CVF conference on computer vision and pattern recognition},
  pages={16761--16772},
  year={2022}
}

@article{pagliardini2022agree,
  title={Agree to disagree: Diversity through disagreement for better transferability},
  author={Pagliardini, Matteo and Jaggi, Martin and Fleuret, Fran{\c{c}}ois and Karimireddy, Sai Praneeth},
  journal={arXiv preprint arXiv:2202.04414},
  year={2022}
}

@article{higgins2017beta,
  title={beta-vae: Learning basic visual concepts with a constrained variational framework.},
  author={Higgins, Irina and Matthey, Loic and Pal, Arka and Burgess, Christopher P and Glorot, Xavier and Botvinick, Matthew M and Mohamed, Shakir and Lerchner, Alexander},
  journal={ICLR (Poster)},
  volume={3},
  year={2017}
}

@inproceedings{radford2021learning,
  title={Learning transferable visual models from natural language supervision},
  author={Radford, Alec and Kim, Jong Wook and Hallacy, Chris and Ramesh, Aditya and Goh, Gabriel and Agarwal, Sandhini and Sastry, Girish and Askell, Amanda and Mishkin, Pamela and Clark, Jack and others},
  booktitle={International conference on machine learning},
  pages={8748--8763},
  year={2021},
  organization={PMLR}
}

@inproceedings{selvaraju2017grad,
  title={Grad-cam: Visual explanations from deep networks via gradient-based localization},
  author={Selvaraju, Ramprasaath R and Cogswell, Michael and Das, Abhishek and Vedantam, Ramakrishna and Parikh, Devi and Batra, Dhruv},
  booktitle={Proceedings of the IEEE international conference on computer vision},
  pages={618--626},
  year={2017}
}

@article{sun2021recovering,
  title={Recovering latent causal factor for generalization to distributional shifts},
  author={Sun, Xinwei and Wu, Botong and Zheng, Xiangyu and Liu, Chang and Chen, Wei and Qin, Tao and Liu, Tie-Yan},
  journal={Advances in Neural Information Processing Systems},
  volume={34},
  pages={16846--16859},
  year={2021}
}

@article{kirichenko2022last,
  title={Last layer re-training is sufficient for robustness to spurious correlations},
  author={Kirichenko, Polina and Izmailov, Pavel and Wilson, Andrew Gordon},
  journal={arXiv preprint arXiv:2204.02937},
  year={2022}
}

@article{labonte2024towards,
  title={Towards last-layer retraining for group robustness with fewer annotations},
  author={LaBonte, Tyler and Muthukumar, Vidya and Kumar, Abhishek},
  journal={Advances in Neural Information Processing Systems},
  volume={36},
  year={2024}
}

@article{zhang2022correct,
  title={Correct-n-contrast: A contrastive approach for improving robustness to spurious correlations},
  author={Zhang, Michael and Sohoni, Nimit S and Zhang, Hongyang R and Finn, Chelsea and R{\'e}, Christopher},
  journal={arXiv preprint arXiv:2203.01517},
  year={2022}
}

@article{wang2024disentangled,
  title={Disentangled representation learning},
  author={Wang, Xin and Chen, Hong and Wu, Zihao and Zhu, Wenwu and others},
  journal={IEEE Transactions on Pattern Analysis and Machine Intelligence},
  year={2024},
  publisher={IEEE}
}

@article{bengio2013representation,
  title={Representation learning: A review and new perspectives},
  author={Bengio, Yoshua and Courville, Aaron and Vincent, Pascal},
  journal={IEEE transactions on pattern analysis and machine intelligence},
  volume={35},
  number={8},
  pages={1798--1828},
  year={2013},
  publisher={IEEE}
}

@inproceedings{lachapelle2023synergies,
  title={Synergies between disentanglement and sparsity: Generalization and identifiability in multi-task learning},
  author={Lachapelle, S{\'e}bastien and Deleu, Tristan and Mahajan, Divyat and Mitliagkas, Ioannis and Bengio, Yoshua and Lacoste-Julien, Simon and Bertrand, Quentin},
  booktitle={International Conference on Machine Learning},
  pages={18171--18206},
  year={2023},
  organization={PMLR}
}

@article{fumero2023leveraging,
  title={Leveraging sparse and shared feature activations for disentangled representation learning},
  author={Fumero, Marco and Wenzel, Florian and Zancato, Luca and Achille, Alessandro and Rodol{\`a}, Emanuele and Soatto, Stefano and Sch{\"o}lkopf, Bernhard and Locatello, Francesco},
  journal={Advances in Neural Information Processing Systems},
  volume={36},
  pages={27682--27698},
  year={2023}
}

@inproceedings{petryk2022guiding,
  title={On guiding visual attention with language specification},
  author={Petryk, Suzanne and Dunlap, Lisa and Nasseri, Keyan and Gonzalez, Joseph and Darrell, Trevor and Rohrbach, Anna},
  booktitle={Proceedings of the IEEE/CVF Conference on Computer Vision and Pattern Recognition},
  pages={18092--18102},
  year={2022}
}

@article{zhang2023diagnosing,
  title={Diagnosing and rectifying vision models using language},
  author={Zhang, Yuhui and HaoChen, Jeff Z and Huang, Shih-Cheng and Wang, Kuan-Chieh and Zou, James and Yeung, Serena},
  journal={arXiv preprint arXiv:2302.04269},
  year={2023}
}

@inproceedings{yenamandra2023facts,
  title={Facts: First amplify correlations and then slice to discover bias},
  author={Yenamandra, Sriram and Ramesh, Pratik and Prabhu, Viraj and Hoffman, Judy},
  booktitle={Proceedings of the IEEE/CVF International Conference on Computer Vision},
  pages={4794--4804},
  year={2023}
}

@inproceedings{Chattopadhay_2018,
   title={Grad-CAM++: Generalized Gradient-Based Visual Explanations for Deep Convolutional Networks},
   booktitle={2018 IEEE Winter Conference on Applications of Computer Vision (WACV)},
   publisher={IEEE},
   author={Chattopadhay, Aditya and Sarkar, Anirban and Howlader, Prantik and Balasubramanian, Vineeth N},
   year={2018},
   month=mar }

@inproceedings{yang2023mitigating,
  title={Mitigating spurious correlations in multi-modal models during fine-tuning},
  author={Yang, Yu and Nushi, Besmira and Palangi, Hamid and Mirzasoleiman, Baharan},
  booktitle={International Conference on Machine Learning},
  pages={39365--39379},
  year={2023},
  organization={PMLR}
}

@article{agarwal2021evaluating,
  title={Evaluating clip: towards characterization of broader capabilities and downstream implications},
  author={Agarwal, Sandhini and Krueger, Gretchen and Clark, Jack and Radford, Alec and Kim, Jong Wook and Brundage, Miles},
  journal={arXiv preprint arXiv:2108.02818},
  year={2021}
}

@inproceedings{nie2024out,
  title={Out-of-Distribution Detection with Negative Prompts},
  author={Nie, Jun and Zhang, Yonggang and Fang, Zhen and Liu, Tongliang and Han, Bo and Tian, Xinmei},
  booktitle={The Twelfth International Conference on Learning Representations},
  year={2024}
}

@inproceedings{liu2015deep,
  title={Deep learning face attributes in the wild},
  author={Liu, Ziwei and Luo, Ping and Wang, Xiaogang and Tang, Xiaoou},
  booktitle={Proceedings of the IEEE international conference on computer vision},
  pages={3730--3738},
  year={2015}
}

@article{joshi2023towards,
  title={Towards Mitigating more Challenging Spurious Correlations: A Benchmark \& New Datasets},
  author={Joshi, Siddharth and Yang, Yu and Xue, Yihao and Yang, Wenhan and Mirzasoleiman, Baharan},
  journal={arXiv preprint arXiv:2306.11957},
  year={2023}
}

@article{liang2022metashift,
  title={Metashift: A dataset of datasets for evaluating contextual distribution shifts and training conflicts},
  author={Liang, Weixin and Zou, James},
  journal={arXiv preprint arXiv:2202.06523},
  year={2022}
}

@article{phan2024controllable,
  title={Controllable prompt tuning for balancing group distributional robustness},
  author={Phan, Hoang and Wilson, Andrew Gordon and Lei, Qi},
  journal={arXiv preprint arXiv:2403.02695},
  year={2024}
}

@inproceedings{he2016deep,
  title={Deep residual learning for image recognition},
  author={He, Kaiming and Zhang, Xiangyu and Ren, Shaoqing and Sun, Jian},
  booktitle={Proceedings of the IEEE conference on computer vision and pattern recognition},
  pages={770--778},
  year={2016}
}

@article{deng2023robust,
  title={Robust learning with progressive data expansion against spurious correlation},
  author={Deng, Yihe and Yang, Yu and Mirzasoleiman, Baharan and Gu, Quanquan},
  journal={Advances in neural information processing systems},
  volume={36},
  pages={1390--1402},
  year={2023}
}

@article{han2024improving,
  title={Improving group robustness on spurious correlation requires preciser group inference},
  author={Han, Yujin and Zou, Difan},
  journal={arXiv preprint arXiv:2404.13815},
  year={2024}
}

@article{rudelson2013hanson,
  title={Hanson-Wright inequality and sub-Gaussian concentration},
  author={Rudelson, Mark and Vershynin, Roman},
  year={2013}
}

@article{vershynin2010introduction,
  title={Introduction to the non-asymptotic analysis of random matrices},
  author={Vershynin, Roman},
  journal={arXiv preprint arXiv:1011.3027},
  year={2010}
}

@article{wang2022causal,
  title={Causal balancing for domain generalization},
  author={Wang, Xinyi and Saxon, Michael and Li, Jiachen and Zhang, Hongyang and Zhang, Kun and Wang, William Yang},
  journal={arXiv preprint arXiv:2206.05263},
  year={2022}
}

@inproceedings{zhou2021examining,
  title={Examining and combating spurious features under distribution shift},
  author={Zhou, Chunting and Ma, Xuezhe and Michel, Paul and Neubig, Graham},
  booktitle={International Conference on Machine Learning},
  pages={12857--12867},
  year={2021},
  organization={PMLR}
}

@inproceedings{lim2023biasadv,
  title={Biasadv: Bias-adversarial augmentation for model debiasing},
  author={Lim, Jongin and Kim, Youngdong and Kim, Byungjai and Ahn, Chanho and Shin, Jinwoo and Yang, Eunho and Han, Seungju},
  booktitle={Proceedings of the IEEE/CVF conference on computer vision and pattern recognition},
  pages={3832--3841},
  year={2023}
}

@article{liu2025bridging,
  title={Bridging Distribution Shift and AI Safety: Conceptual and Methodological Synergies},
  author={Liu, Chenruo and Tang, Kenan and Qin, Yao and Lei, Qi},
  journal={arXiv preprint arXiv:2505.22829},
  year={2025}
}

@inproceedings{ni2021superclass,
  title={Superclass-conditional gaussian mixture model for learning fine-grained embeddings},
  author={Ni, Jingchao and Cheng, Wei and Chen, Zhengzhang and Asakura, Takayoshi and Soma, Tomoya and Kato, Sho and Chen, Haifeng},
  booktitle={International Conference on Learning Representations},
  year={2021}
}

@misc{hastie2009elements,
  title={The elements of statistical learning},
  author={Hastie, Trevor and Tibshirani, Robert and Friedman, Jerome and others},
  year={2009},
  publisher={Springer series in statistics New-York}
}

@misc{qwen2025qwen25technicalreport,
      title={Qwen2.5 Technical Report}, 
      author={Qwen and : and An Yang and Baosong Yang and Beichen Zhang and Binyuan Hui and Bo Zheng and Bowen Yu and Chengyuan Li and Dayiheng Liu and Fei Huang and Haoran Wei and Huan Lin and Jian Yang and Jianhong Tu and Jianwei Zhang and Jianxin Yang and Jiaxi Yang and Jingren Zhou and Junyang Lin and Kai Dang and Keming Lu and Keqin Bao and Kexin Yang and Le Yu and Mei Li and Mingfeng Xue and Pei Zhang and Qin Zhu and Rui Men and Runji Lin and Tianhao Li and Tianyi Tang and Tingyu Xia and Xingzhang Ren and Xuancheng Ren and Yang Fan and Yang Su and Yichang Zhang and Yu Wan and Yuqiong Liu and Zeyu Cui and Zhenru Zhang and Zihan Qiu},
      year={2025},
      eprint={2412.15115},
      archivePrefix={arXiv}
}

\newpage
%% Appendix starts here
\clearpage
\appendix 
\onecolumn% switch to appendix numbering
\cleardoublepage
\pagenumbering{arabic}  % reset numbering style
\setcounter{page}{0}    % start page from 1

% Appendix part title (unnumbered)
\part*{Appendix}
\thispagestyle{empty}
\addcontentsline{toc}{part}{Appendix}

% Local TOC: include up to subsections
\etocsetnexttocdepth{subsection}
\localtableofcontents

\newpage

\section{Algorithms for Gradient-based Attribution Maps}
\label{appendix:b}
\subsection{Gradient-based Attribution Map for CLIP}
\label{subsec:A.1}
\begin{algorithm}[H]
{\fontsize{9pt}{9pt}\selectfont
   \caption{Gradient-based Attribution Map for CLIP}
   \label{alg:CLIP}
\begin{algorithmic}
   \STATE {\bfseries Input:} Image $\mathbf{x}$, text $\mathbf{T}$, pre-trained ResNet50-based CLIP
   \STATE {\bfseries Output:} Normalized attribution map $L_{\text{CLIP}}^{\mathbf{T}}(\mathbf{x})$
   
   \STATE Pass $\mathbf{x}$ through CLIP's vision encoder to get the feature vector $\mathbf{z}$ and $K$ feature maps $\mathbf{A}_k \in \mathbb{R}^{h \times w}$ for $k = 1,2,\dots,K$, from the last convolutional layer of ResNet50
   \STATE Pass $\mathbf{T}$ through CLIP's text encoder to get text embedding $\mathbf{t}$
   \STATE Compute similarity score:
   \[
   s(\mathbf{x}, \mathbf{T}) = \frac{\mathbf{z} \cdot \mathbf{t}}{\Vert \mathbf{z}\Vert \Vert \mathbf{t}\Vert}
   \]
   
   \FOR{$k = 1$ to $K$}
        \FOR{each $(i,j)\in\{1..h\}\times\{1..w\}$}
            \STATE Calculate gradient $\displaystyle \frac{\partial s(\mathbf{x}, \mathbf{T})}{\partial \mathbf{A}_k^{ij}}$ for spatial location $(i,j)$
       \ENDFOR
       \STATE Compute importance weight $\alpha_k^{\mathbf{T}}$ through global average pooling:
       \(
       \alpha_k^{\mathbf{T}} = \frac{1}{hw} \sum_{i=1}^{h} \sum_{j=1}^{w} \frac{\partial s(\mathbf{x}, \mathbf{T})}{\partial \mathbf{A}_k^{ij}}
       \)
   \ENDFOR
   
   \STATE Combine feature maps weighted by importance: \( 
   L_{\text{CLIP}}^{\mathbf{T}}(\mathbf{x}) = \text{ReLU}\left(\sum_{k=1}^{K} \alpha_k^{\mathbf{T}} \mathbf{A}_k\right)
   \)

   \STATE Normalize $L_{\text{CLIP}}^{\mathbf{T}}(\mathbf{x})$ to the range $[0, 1]$ using min-max normalization
\end{algorithmic}
}
\end{algorithm}

\subsection{Gradient-based Attribution Map for SupER's $\omega_{\mathrm{rel}}$ and $\omega_{\mathrm{irr}}$}
\label{subsec:A.2}

\begin{algorithm}[H]
{\fontsize{9pt}{9pt}\selectfont
   \caption{Gradient-based Attribution Map for SupER's $\omega_{\mathrm{rel}}$ and $\omega_{\mathrm{irr}}$}
   \label{alg:VAE}
\begin{algorithmic}
   \STATE {\bfseries Input:} Image $\mathbf{x}$, true label $y$, ResNet50-based encoder $\phi$, classifiers $\omega_{\mathrm{rel}}$ (for $\mathbf{z}_{\mathrm{rel}}$) and $\omega_{\mathrm{irr}}$ (for $\mathbf{z}_{\mathrm{irr}}$)
   \STATE {\bfseries Output:} Normalized attribution maps $L_{\phi,\omega_{\mathrm{rel}}}(\mathbf{x},y)$ and $L_{\phi,\omega_{\mathrm{irr}}}(\mathbf{x},y)$
   
   \STATE Pass $\mathbf{x}$ through encoder to obtain latent feature $\mathbf{z}=[\mathbf{z}_{\mathrm{rel}};\mathbf{z}_{\mathrm{irr}}]$ with mean \(\boldsymbol{\mu} = [\boldsymbol{\mu}_{\mathrm{rel}};\boldsymbol{\mu}_{\mathrm{irr}}]\), and $K$ feature maps $\mathbf{A}_k \in \mathbb{R}^{h \times w}$ for $k = 1,2,\dots,K$, from the last convolutional layer of ResNet50
   \STATE Compute logits $g_1 = \omega_{\mathrm{rel}}(\boldsymbol{\mu}_{\mathrm{rel}})$ and $g_2 = \omega_{\mathrm{irr}}(\boldsymbol{\mu}_{\mathrm{irr}})$
   
   \FOR{$l = 1$ to $2$}   
       \STATE Let $s_l(\mathbf{x},y) = g_l[y]$  
       \FOR{$k = 1$ to $K$}
           \FOR{each $(i,j)\in\{1..h\}\times\{1..w\}$}
                \STATE Calculate gradient $\displaystyle \frac{\partial s_l(\mathbf{x},y)}{\partial \mathbf{A}_k^{ij}}$ for spatial location $(i,j)$
           \ENDFOR
           \STATE Compute importance weight $\alpha_k^l$ through global average pooling:
           \(isplaystyle \alpha^l_k = \frac{1}{hw} \sum_{i=1}^{h} \sum_{j=1}^{w} \frac{\partial s_l(\mathbf{x},y)}{\partial \mathbf{A}_k^{ij}}\)
       \ENDFOR
       \STATE Combine feature maps weighted by importance:
       \[
       L_{\phi,\omega_l}(\mathbf{x},y) = \mathrm{ReLU}\!\left(\sum_{k=1}^{K} \alpha^l_k \mathbf{A}_k\right)
       \]
       \STATE Normalize $L_{\phi,\omega_l}(\mathbf{x},y)$ to the range $[0, 1]$ using min-max normalization
   \ENDFOR
\end{algorithmic}
}
\end{algorithm}

\section{Theoretical Results}
\label{appendix:TR}

\begin{theorem}[Fixed-design setting: formal restatement of Theorem~\ref{thm:informal}]
\label{thm:formal-restatement}
Assume Assumptions~\ref{ass:lin}--\ref{ass:commute} hold. For any generating index \(c\in\{1,2\}\) with coefficient vector \(\boldsymbol{\beta}_c\in\mathbb{R}^p\), and any strategy \(S\in\{S_1,S_2,S_{1,2}\}\), let \(\hat{f}_{S,c,\boldsymbol{\beta}_c}\) be the ordinary least-squares (OLS) predictor obtained by fitting on the training data using the feature(s) specified by \(S\). Define the excess risk under the test distribution by
\[
\mathcal{E}(S;c,\boldsymbol{\beta}_c)
\;:=\;
\mathbb{E}_{Y^s}\mathbb{E}_{(Z_1,Z_2)\sim P_t}
\!\left[\,\bigl(\hat{f}_{S,c,\boldsymbol{\beta}_c}(Z_1,Z_2)-Z_c^\top\boldsymbol{\beta}_c\bigr)^2\,\right].
\]
Then:
\begin{enumerate}[label=\arabic*., leftmargin=*]
\item (Superclass-irrelevant feature should not be used.)
When the generating index is known to be \(c=1\), restricting to strategies \(\{S_1,S_{1,2}\}\), for all \(\boldsymbol{\beta}_1\in\mathbb{R}^p\),
\[
\min_{S\in\{S_1,S_{1,2}\}}\;\mathcal{E}(S;1,\boldsymbol{\beta}_1)
\;=\;
\frac{\sigma^2}{n}\,\sum_{i=1}^p \frac{d^{t}_{1,i}}{d^{s}_{1,i}},
\qquad\text{with equality achieved by } S=S_1.
\]
\item (All superclass-relevant features should be used.)
When the generating index \(c\) is uncertain and may be either \(1\) or \(2\), restricting to strategies \(\{S_1,S_2,S_{1,2}\}\), fix any \(r>0\) and define the set
\[
\mathcal{B}_r^{(c)} \;:=\; \bigl\{\,\boldsymbol{\beta}\in\mathbb{R}^p:\;\boldsymbol{\beta}^\top \Sigma^{t}_c\,\boldsymbol{\beta} \;=\; r\,\bigr\}.
\]
There exists a constant $C>0$, independent of both $n$ and $r$, such that whenever $n r > C$,
\[
\min_{S\in\{S_1,S_2,S_{1,2}\}}\;
\max_{c\in\{1,2\}}\;
\max_{\boldsymbol{\beta}_c\in \mathcal{B}_r^{(c)}}
\;\mathcal{E}(S;c,\boldsymbol{\beta}_c)
\;=\;
\frac{\sigma^2}{\,n\,(1-\rho^2)\,}\,
\sum_{i=1}^p\!\left(
\frac{d^{t}_{1,i}}{d^{s}_{1,i}}
\;+\;
\frac{d^{t}_{2,i}}{d^{s}_{2,i}}
\right),
\]
with equality achieved by \(S=S_{1,2}\).
\end{enumerate}
\end{theorem}

\begin{proof}[Proof of Part (1)]
By Assumption~\ref{ass:commute}, there exists an orthogonal matrix $U\in\mathbb{R}^{p\times p}$ such that
\[
\hat{\Sigma}^{s}_1
= U\,\mathrm{diag} \bigl(d^{s}_1\bigr)\,U^\top,\quad
\hat{\Sigma}^{s}_2
= U\,\mathrm{diag} \bigl(d^{s}_2\bigr)\,U^\top,\quad
\Sigma^{t}_1
= U\,\mathrm{diag} \bigl(d^{t}_1\bigr)\,U^\top,\quad
\Sigma^{t}_2
= U\,\mathrm{diag} \bigl(d^{t}_2\bigr)\,U^\top,
\]
for vectors $d^{s}_1,d^{s}_2,d^{t}_1,d^{t}_2\in\mathbb{R}_+^p$. We work under the generating model
\(
Y = Z_1^\top \boldsymbol{\beta}_1 + \varepsilon
\)
with $\varepsilon\sim\mathcal{N}(0,\sigma^2)$ independent of $(Z_1,Z_2)$, and fixed
$Z_1^{s},Z_2^{s}$.
The OLS estimator using $Z_1$ is
\[
\hat{\boldsymbol{\beta}}_1
=\bigl((Z_1^{s})^\top Z_1^{s}\bigr)^{-1}(Z_1^{s})^\top Y^{s}
=\boldsymbol{\beta}_1+\bigl((Z_1^{s})^\top Z_1^{s}\bigr)^{-1}(Z_1^{s})^\top \boldsymbol{\varepsilon}.
\]
Then,
\[
\operatorname{Cov}(\hat{\boldsymbol{\beta}}_1-\boldsymbol{\beta}_1)
=\sigma^2\bigl((Z_1^{s})^\top Z_1^{s}\bigr)^{-1}
=\frac{\sigma^2}{n}\,(\hat{\Sigma}_1^{s})^{-1}.
\]
Therefore the excess risk is
\[
\mathcal{E}(S_1;1,\boldsymbol{\beta}_1)
=\mathbb{E}_{Y^s}\mathbb{E}_{Z_1\sim P_{t}}\bigl[(Z_1^\top(\hat{\boldsymbol{\beta}}_1-\boldsymbol{\beta}_1))^2\bigr]
=\operatorname{tr} \bigl(\Sigma_1^{t}\operatorname{Cov}(\hat{\boldsymbol{\beta}}_1-\boldsymbol{\beta}_1)\bigr)
=\frac{\sigma^2}{n}\,\operatorname{tr} \bigl(\Sigma_1^{t}(\hat{\Sigma}_1^{s})^{-1}\bigr).
\]
Using the simultaneous diagonalization by $U$,
\[
\mathcal{E}(S_1;1,\boldsymbol{\beta}_1)
=\frac{\sigma^2}{n}\,\operatorname{tr} \bigl(\mathrm{diag}(d_1^{t})\,\mathrm{diag}(d_1^{s})^{-1}\bigr)
=\frac{\sigma^2}{n}\sum_{i=1}^p \frac{d^{t}_{1,i}}{d^{s}_{1,i}}.
\]

Furthermore, let $\boldsymbol{\beta}=\begin{bmatrix}\boldsymbol{\beta}_1\\ 0\end{bmatrix}\in\mathbb{R}^{2p}$ be the true parameter in the joint model and
$Z^{s}=[\,Z_1^{s}\;Z_2^{s}\,]$.
The OLS estimator satisfies
\[
\hat{\boldsymbol{\beta}} = \bigl((Z^{s})^\top Z^{s}\bigr)^{-1}(Z^{s})^\top Y^{s}
= \boldsymbol{\beta} + \bigl((Z^{s})^\top Z^{s}\bigr)^{-1}(Z^{s})^\top \boldsymbol{\varepsilon}.
\]
Hence,
\[
\operatorname{Cov}(\hat{\boldsymbol{\beta}}-\boldsymbol{\beta})
= \frac{\sigma^2}{n}\,\bigl(\hat{\Sigma}^{s}\bigr)^{-1},
\quad\text{where}\
\hat{\Sigma}^{s}
=\frac{1}{n}(Z^{s})^\top Z^{s}
=
\begin{bmatrix}
\hat{\Sigma}_1^{s} & \hat{\Sigma}_{1,2}^{s}\\[2pt]
\hat{\Sigma}_{2,1}^{s} & \hat{\Sigma}_2^{s}
\end{bmatrix}.
\]
Using the training-time correlation $\hat{\Sigma}_{1,2}^{s}
=\rho\,(\hat{\Sigma}_1^{s})^{1/2}(\hat{\Sigma}_2^{s})^{1/2}$ and the shared eigen-basis
$U'=\begin{bmatrix}U&0\\[2pt]0&U\end{bmatrix}$, we can write
\[
\hat{\Sigma}^{s}
=
U'
\begin{bmatrix}
\mathrm{diag} \bigl(d_1^{s}\bigr) & \rho\,\mathrm{diag}\bigl(\sqrt{d_1^{s}d_2^{s}}\bigr)\\[2pt]
\rho\,\mathrm{diag} \bigl(\sqrt{d_1^{s}d_2^{s}}\bigr) & \mathrm{diag} \bigl(d_2^{s}\bigr)
\end{bmatrix}
U'^\top,
\quad
\Sigma^{t}
=
U'
\begin{bmatrix}
\mathrm{diag} \bigl(d_1^{t}\bigr) & 0\\[2pt]
0 & \mathrm{diag} \bigl(d_2^{t}\bigr)
\end{bmatrix}
U'^\top,
\]
where we used $\Sigma_{1,2}^{t}=0$. The inverse of $\hat{\Sigma}^{s}$ can be computed as
\[
(\hat{\Sigma}^{s})^{-1}
=
\frac{1}{1-\rho^{2}} U'
\begin{bmatrix}
\mathrm{diag} \bigl(\frac{1}{d_{1}^{s}}\bigr) &
-\mathrm{diag} \Bigl(\frac{\rho}{\sqrt{d_{1}^{s}d_{2}^{s}}}\Bigr)\\[6pt]
-\mathrm{diag} \Bigl(\frac{\rho}{\sqrt{d_{1}^{s}d_{2}^{s}}}\Bigr) &
\mathrm{diag} \bigl(\frac{1}{d_{2}^{s}}\bigr)
\end{bmatrix}
U'^{\!\top}.
\]
Therefore,
\[
\mathcal{E}(S_{1,2};1,\boldsymbol{\beta}_1)
=
\mathbb{E}_{Y^s}\,\mathbb{E}_{(Z_1,Z_2)\sim P_{t}}
\bigl[(Z^\top(\hat{\boldsymbol{\beta}}-\boldsymbol{\beta}))^2\bigr]
=
\operatorname{tr} \bigl(\Sigma^{t}\operatorname{Cov}(\hat{\boldsymbol{\beta}}-\boldsymbol{\beta})\bigr)
=
\frac{\sigma^2}{n}\,\operatorname{tr}\bigl(\Sigma^{t}(\hat{\Sigma}^{s})^{-1}\bigr),
\]
which yields
\[
\mathcal{E}(S_{1,2};1,\boldsymbol{\beta}_1)
=
\frac{\sigma^2}{\,n\,(1-\rho^2)\,}
\sum_{i=1}^p\!\left(
\frac{d^{t}_{1,i}}{d^{s}_{1,i}}
+
\frac{d^{t}_{2,i}}{d^{s}_{2,i}}
\right).
\]

Subtracting the two expressions, we have
\(
\mathcal{E}(S_{1,2};1,\boldsymbol{\beta}_1)-\mathcal{E}(S_1;1,\boldsymbol{\beta}_1)
>0,
\)
for $\rho\in[-1,1]\setminus\{0\}$. Hence
\[
\min_{S\in\{S_1,S_{1,2}\}}\mathcal{E}(S;1,\boldsymbol{\beta}_1)
=\mathcal{E}(S_1;1,\boldsymbol{\beta}_1)
=\frac{\sigma^2}{n}\sum_{i=1}^p \frac{d^{t}_{1,i}}{d^{s}_{1,i}},
\]
with equality attained by $S=S_1$.
\end{proof}

\begin{proof}[Proof of Part (2)]
Case $S=S_{1,2}$.
By the same blockwise computation used in the proof of part (1), we have
\[
\mathcal{E}(S_{1,2};1,\boldsymbol{\beta}_1)
=
\frac{\sigma^2}{\,n\,(1-\rho^2)\,}
\sum_{i=1}^p\!\left(
\frac{d^{t}_{1,i}}{d^{s}_{1,i}}
+
\frac{d^{t}_{2,i}}{d^{s}_{2,i}}
\right).
\]
By symmetry (interchanging indices $1$ and $2$ in the same calculation),
\[
\mathcal{E}(S_{1,2};2,\boldsymbol{\beta}_2)
=
\frac{\sigma^2}{\,n\,(1-\rho^2)\,}
\sum_{i=1}^p\!\left(
\frac{d^{t}_{1,i}}{d^{s}_{1,i}}
+
\frac{d^{t}_{2,i}}{d^{s}_{2,i}}
\right).
\]
Therefore, for any $r>0$ and any $\boldsymbol{\beta}_c\in\mathcal{B}_r^{(c)}$,
\[
\max_{c\in\{1,2\}}\;\max_{\boldsymbol{\beta}_c\in \mathcal{B}_r^{(c)}}
\;\mathcal{E}(S_{1,2};c,\boldsymbol{\beta}_c)
\;=\;
\frac{\sigma^2}{\,n\,(1-\rho^2)\,}
\sum_{i=1}^p\!\left(
\frac{d^{t}_{1,i}}{d^{s}_{1,i}}
+
\frac{d^{t}_{2,i}}{d^{s}_{2,i}}
\right).
\]

Case $S=S_1$.
When $c=1$, by part (1),
\[
\mathcal{E}(S_1;1,\boldsymbol{\beta}_1)
=\frac{\sigma^2}{n}\sum_{i=1}^p \frac{d^{t}_{1,i}}{d^{s}_{1,i}}.
\]
When $c=2$, let
\[
\hat{\boldsymbol{\beta}}_1
=
\bigl((Z^{s}_1)^\top Z^{s}_1\bigr)^{-1}(Z^{s}_1)^\top Y^{s},
\qquad
Y^{s}=Z^{s}_2\boldsymbol{\beta}_2+\boldsymbol{\varepsilon}.
\]
Then
\[
\mathcal{E}(S_1;2,\boldsymbol{\beta}_2)
= \mathbb{E}_{Y^s}\,\mathbb{E}_{(Z_1,Z_2)\sim P_t}
[(Z_1^\top \hat{\boldsymbol{\beta}}_1- Z_2^\top \boldsymbol{\beta}_2)^2]
= r + \mathbb{E}_{Y^s}[\hat{\boldsymbol{\beta}}_1^\top \Sigma^{t}_1 \hat{\boldsymbol{\beta}}_1],
\]
because $\Sigma^{t}_{1,2}=0$ and $\boldsymbol{\beta}_2^\top\Sigma^{t}_2\boldsymbol{\beta}_2=r$ for $\boldsymbol{\beta}_2\in\mathcal{B}_r^{(2)}$.
Decompose
\[
\mathbb{E}_{Y^s}[\hat{\boldsymbol{\beta}}_1^\top \Sigma^{t}_1 \hat{\boldsymbol{\beta}}_1]
=
\bigl(\mathbb{E}_{Y^s}[\hat{\boldsymbol{\beta}}_1]\bigr)^\top \Sigma^{t}_1 \bigl(\mathbb{E}_{Y^s}[\hat{\boldsymbol{\beta}}_1]\bigr)
\;+\;
\operatorname{tr} \bigl(\Sigma^{t}_1\operatorname{Var}(\hat{\boldsymbol{\beta}}_1)\bigr).
\]
With zero-mean noise,
\[
\mathbb{E}_{Y^s}[\hat{\boldsymbol{\beta}}_1]
=
\bigl((Z^{s}_1)^\top Z^{s}_1\bigr)^{-1}(Z^{s}_1)^\top Z^{s}_2\,\boldsymbol{\beta}_2
=
(\hat{\Sigma}^{s}_1)^{-1}\hat{\Sigma}^{s}_{1,2}\,\boldsymbol{\beta}_2
=
\rho\,(\hat{\Sigma}^{s}_1)^{-1/2}(\hat{\Sigma}^{s}_2)^{1/2}\boldsymbol{\beta}_2.
\]
Hence
\[
\bigl(\mathbb{E}_{Y^s}[\hat{\boldsymbol{\beta}}_1]\bigr)^\top \Sigma^{t}_1 \bigl(\mathbb{E}_{Y^s}[\hat{\boldsymbol{\beta}}_1]\bigr)
=
\rho^2\,\boldsymbol{\beta}_2^\top
(\hat{\Sigma}^{s}_2)^{1/2}
(\hat{\Sigma}^{s}_1)^{-1/2}
\Sigma^{t}_1
(\hat{\Sigma}^{s}_1)^{-1/2}
(\hat{\Sigma}^{s}_2)^{1/2}
\boldsymbol{\beta}_2.
\]
Move to the common eigenbasis $U$ and write $\boldsymbol{\beta}_2=U\gamma$. Using
$\hat{\Sigma}^{s}_\ell=U\operatorname{diag}(d^{s}_\ell)U^\top$ and
$\Sigma^{t}_\ell=U\operatorname{diag}(d^{t}_\ell)U^\top$, we get
\[
\bigl(\mathbb{E}_{Y^s}[\hat{\boldsymbol{\beta}}_1]\bigr)^\top \Sigma^{t}_1 \bigl(\mathbb{E}_{Y^s}[\hat{\boldsymbol{\beta}}_1]\bigr)
=
\rho^2 \sum_{i=1}^p \gamma_i^2\,
\frac{d^{s}_{2,i}\,d^{t}_{1,i}}{d^{s}_{1,i}},
\qquad
\text{subject to}\quad
\sum_{i=1}^p \gamma_i^2\,d^{t}_{2,i} = r.
\]
Maximizing under this weighted $\ell_2$-constraint concentrates mass on
\[
i^\star \in \arg\max_{i}\;
\frac{d^{s}_{2,i}\,d^{t}_{1,i}}{d^{s}_{1,i}\,d^{t}_{2,i}},
\]
hence
\[
\max_{\boldsymbol{\beta}_2\in\mathcal{B}_r^{(2)}}
\bigl(\mathbb{E}_{Y^s}[\hat{\boldsymbol{\beta}}_1]\bigr)^\top \Sigma^{t}_1 \bigl(\mathbb{E}_{Y^s}[\hat{\boldsymbol{\beta}}_1]\bigr)
=
\rho^2\,r\;\max_{i}\frac{d^{s}_{2,i}\,d^{t}_{1,i}}{d^{s}_{1,i}\,d^{t}_{2,i}}.
\]
Furthermore, since
\(
\operatorname{Var}(\hat{\boldsymbol{\beta}}_1)=\sigma^2\bigl((Z^{s}_1)^\top Z^{s}_1\bigr)^{-1}
=\sigma^2(\hat{\Sigma}^{s}_1)^{-1}/n,
\)
we have
\[
\operatorname{tr} \bigl(\Sigma^{t}_1\,\operatorname{Var}(\hat{\boldsymbol{\beta}}_1)\bigr)
=
\frac{\sigma^2}{n}\operatorname{tr} \bigl(\Sigma^{t}_1(\hat{\Sigma}^{s}_1)^{-1}\bigr)
=
\frac{\sigma^2}{n}\sum_{i=1}^p \frac{d^{t}_{1,i}}{d^{s}_{1,i}}
\;=\;
\mathcal{E}(S_1;1,\boldsymbol{\beta}_1).
\]
Therefore,
\[
\max_{\boldsymbol{\beta}_2\in\mathcal{B}_r^{(2)}}\;\mathcal{E}(S_1;2,\boldsymbol{\beta}_2)
=
r\!\left(1+\rho^2\,\max_{i}\frac{d^{s}_{2,i}\,d^{t}_{1,i}}{d^{s}_{1,i}\,d^{t}_{2,i}}\right)
+
\frac{\sigma^2}{n}\sum_{i=1}^p \frac{d^{t}_{1,i}}{d^{s}_{1,i}}
\;\ge\;
\mathcal{E}(S_1;1,\boldsymbol{\beta}_1).
\]
Hence
\[
\max_{c\in\{1,2\}}\;\max_{\boldsymbol{\beta}_c\in \mathcal{B}_r^{(c)}}
\;\mathcal{E}(S_{1};c,\boldsymbol{\beta}_c)
=
r\!\left(1+\rho^2\,\max_{i}\frac{d^{s}_{2,i}\,d^{t}_{1,i}}{d^{s}_{1,i}\,d^{t}_{2,i}}\right)
+
\frac{\sigma^2}{n}\sum_{i=1}^p \frac{d^{t}_{1,i}}{d^{s}_{1,i}}.
\]

Case $S=S_2$.
By symmetry with the case $S=S_1$ after interchanging indices $1\leftrightarrow 2$, we obtain
\[
\max_{c\in\{1,2\}}\;\max_{\boldsymbol{\beta}_c\in \mathcal{B}_r^{(c)}}
\;\mathcal{E}(S_{2};c,\boldsymbol{\beta}_c)
=
r\!\left(1+\rho^2\,\max_{i}\frac{d^{s}_{1,i}\,d^{t}_{2,i}}{d^{s}_{2,i}\,d^{t}_{1,i}}\right)
+
\frac{\sigma^2}{n}\sum_{i=1}^p \frac{d^{t}_{2,i}}{d^{s}_{2,i}}.
\]

Let 
\[
C \;=\; \frac{\sigma^2}{(1-\rho^2)} \cdot \frac{1}{C'} \sum_{i=1}^p\!\left(\frac{d^{t}_{1,i}}{d^{s}_{1,i}}+\frac{d^{t}_{2,i}}{d^{s}_{2,i}}\right),
\]
where
\[
C' \;=\; 1+\rho^2\,\min\left\{\max_{i}\frac{d^{s}_{1,i}\,d^{t}_{2,i}}{d^{s}_{2,i}\,d^{t}_{1,i}}, \,\max_{i}\frac{d^{s}_{2,i}\,d^{t}_{1,i}}{d^{s}_{1,i}\,d^{t}_{2,i}}\right\}.
\]
Then whenever $n > C/r$, we have
\[
\mathcal{E}(S_{1,2};1,\boldsymbol{\beta}_1),\;\;\mathcal{E}(S_{1,2};2,\boldsymbol{\beta}_2)
\;<\;
\min_{S\in\{S_1,S_2\}}\;
\max_{c\in\{1,2\}}\;
\max_{\boldsymbol{\beta}_c\in \mathcal{B}_r^{(c)}}
\;\mathcal{E}(S;c,\boldsymbol{\beta}_c),
\]
and hence the claim holds.
\end{proof}

In addition, we provide further theoretical analysis under the random-design setting.

\begin{assumption}\label{ass:rand}
Under the training distribution $P_s$, $Z=(Z_1,Z_2)$ follows a Gaussian distribution
\[
Z \sim \mathcal{N}\Biggl(0,\,
\begin{bmatrix}
\Sigma^{s}_1 & \Sigma^{s}_{1,2}\\[2pt]
\Sigma^{s}_{2,1} & \Sigma^{s}_2
\end{bmatrix}
\Biggr),
\]
where
\(\Sigma_1^s=\mathbb{E}_{Z_1\sim P_s}[Z_1Z_1^\top]\), 
\(\Sigma_2^s=\mathbb{E}_{Z_2\sim P_s}[Z_2Z_2^\top]\), and 
\(\Sigma_{1,2}^s=\mathbb{E}_{(Z_1,Z_2)\sim P_s}[Z_1Z_2^\top]=(\Sigma_{2,1}^s)^\top\).
\end{assumption}

\begin{theorem}[Random-design setting]
\label{thm:random-formal}
Assume Assumptions~\ref{ass:lin} and~\ref{ass:rand} hold. In addition, suppose the population counterparts of Assumptions~\ref{ass:spur} and \ref{ass:commute} hold; namely, \(\Sigma_{1,2}^s \;=\; \rho\,(\Sigma_1^s)^{1/2}(\Sigma_2^s)^{1/2}\), and \(\Sigma_1^s,\Sigma_2^s,\Sigma_1^t,\Sigma_2^t\) commute with positive eigenvalues $\{d_{1,i}^s\}_{i=1}^p$, $\{d_{2,i}^s\}_{i=1}^p$, $\{d_{1,i}^t\}_{i=1}^p$, and $\{d_{2,i}^t\}_{i=1}^p$.
For any generating index \(c\in\{1,2\}\) with coefficient vector \(\boldsymbol{\beta}_c\in\mathbb{R}^p\), and any strategy \(S\in\{S_1,S_2,S_{1,2}\}\), let \(\hat{f}_{S,c,\boldsymbol{\beta}_c}\) be the OLS predictor trained on \(n\) i.i.d.\ samples drawn from \(P_s\) using the feature(s) specified by \(S\).
Define the random-design excess risk
\[
\mathcal{E}^{\mathrm{rd}}(S;c,\boldsymbol{\beta}_c)
\;:=\;
\mathbb{E}_{(Z_1,Z_2)\sim P_t}
\!\left[\,\bigl(\hat{f}_{S,c,\boldsymbol{\beta}_c}(Z_1,Z_2)-Z_c^\top\boldsymbol{\beta}_c\bigr)^2\,\right],
\]
Then:
\begin{enumerate}[label=\arabic*., leftmargin=*]
\item (Superclass-irrelevant feature should not be used.)
When the generating index is known to be \(c=1\), restricting to strategies \(\{S_1,S_{1,2}\}\), for all \(\boldsymbol{\beta}_1\in\mathbb{R}^p\), if \(n \gg p\) with \(p\) sufficiently large, we have, with high probability,
\[
\min_{S\in\{S_1,S_{1,2}\}}\;\mathcal{E}^{\mathrm{rd}}(S;1,\boldsymbol{\beta}_1)
\quad\text{is attained by}\quad S=S_1.
\]
\item (All superclass-relevant features should be used.)
When the generating index \(c\) is uncertain and may be either \(1\) or \(2\), restricting to strategies \(\{S_1,S_2,S_{1,2}\}\), fix any \(r>0\) and define
\[
\mathcal{B}_r^{(c)} \;:=\; \bigl\{\,\boldsymbol{\beta}\in\mathbb{R}^p:\;\boldsymbol{\beta}^\top \Sigma^{t}_c\,\boldsymbol{\beta} \;=\; r\,\bigr\}.
\]
There exists a constant $C>0$ independent of both $n$ and $r$, such that whenever $n r > C$, we have with high probability,
\[
\min_{S\in\{S_1,S_2,S_{1,2}\}}\;
\max_{c\in\{1,2\}}\;
\max_{\boldsymbol{\beta}_c\in \mathcal{B}_r^{(c)}}
\;\mathcal{E}^{\mathrm{rd}}(S;c,\boldsymbol{\beta}_c)
\quad\text{is attained by}\quad S=S_{1,2}.
\]
\end{enumerate}
\end{theorem}

\begin{proof}[Proof of Part (1)]
Fix any $\boldsymbol{\beta}_1\in\mathbb{R}^p$.  
By Lemma~\ref{lem:S1-hp}, we have a high-probability upper bound on $\mathcal{E}^{\mathrm{rd}}(S_1;1,\boldsymbol{\beta}_1)$.  
By Lemma~\ref{lem:S12-lb-c1}, we have a high-probability lower bound on $\mathcal{E}^{\mathrm{rd}}(S_{1,2};1,\boldsymbol{\beta}_1)$. 

When $n\gg p$, both bounds satisfy that $\epsilon(\delta_2)$ can be made arbitrarily small, and when \(p\) is sufficiently large, the logarithmic terms (multiplied by $c_1$) are $o(1)$ relative to the spectral sums. In this regime, the leading term of the lower bound on $\mathcal{E}^{\mathrm{rd}}(S_{1,2};1,\boldsymbol{\beta}_1)$ dominates the leading term of the upper bound on $\mathcal{E}^{\mathrm{rd}}(S_1;1,\boldsymbol{\beta}_1)$, therefore
\[
\mathcal{E}^{\mathrm{rd}}(S_{1,2};1,\boldsymbol{\beta}_1)\;\ge\;\mathcal{E}^{\mathrm{rd}}(S_1;1,\boldsymbol{\beta}_1)
\quad\text{with high probability.}
\]
Therefore, the minimum over $S\in\{S_1,S_{1,2}\}$ is attained by $S=S_1$ with high probability.
\end{proof}

\begin{proof}[Proof of Part (2)]
First, by Corollary~\ref{cor:S12-ub-c1} and by symmetry between the cases $c=1$ and $c=2$, with probability at least $(1-\delta_1)(1-\delta_2)$, the strategy $S_{1,2}$ satisfies the uniform bound (independent of $c$ and $\boldsymbol{\beta}_c$)
\[
\max_{c\in\{1,2\}}\;\max_{\boldsymbol{\beta}_c\in\mathcal{B}_r^{(c)}}
\mathcal{E}^{\mathrm{rd}}(S_{1,2};c,\boldsymbol{\beta}_c)
\;\le\;
\frac{\sigma^2}{\,n\,(1-\rho^2)\,(1-\varepsilon(\delta_2))\,}
\!\left(
\sum_{i=1}^p\!\left(\frac{d^{t}_{1,i}}{d^{s}_{1,i}}+\frac{d^{t}_{2,i}}{d^{s}_{2,i}}\right) + \frac{1}{2}c_1g_{\rho}\log\frac{1}{\delta_1}
\right).
\]
Next, consider $S=S_1$ under $c=2$. For any $\boldsymbol{\beta}_2\in\mathcal{B}_r^{(2)}$,
\[
\mathcal{E}^{\mathrm{rd}}(S_1;2,\boldsymbol{\beta}_2)
=\mathbb{E}_{(Z_1,Z_2)\sim P_t}\!\bigl[(Z_1^\top\hat{\boldsymbol{\beta}}_1 - Z_2^\top\boldsymbol{\beta}_2)^2\bigr]
= r + \hat{\boldsymbol{\beta}}_1^{\!\top}\Sigma_1^t\hat{\boldsymbol{\beta}}_1
\;\ge\; r,
\]
since $\Sigma_{1,2}^t=0$ and $\Sigma_1^t\succeq 0$. Hence
\[
\max_{c\in\{1,2\}}\;\max_{\boldsymbol{\beta}_c\in\mathcal{B}_r^{(c)}}\;
\mathcal{E}^{\mathrm{rd}}(S_1;c,\boldsymbol{\beta}_c)\;\ge\; r.
\]
By symmetry, the same lower bound holds for $S=S_2$. Choose
\[
C
\;=\;
\frac{\sigma^2\left(\sum_{i=1}^p(d^{t}_{1,i}/d^{s}_{1,i}+d^{t}_{2,i}/d^{s}_{2,i}) + 1/2\cdot c_1g_{\rho}\log(1/\delta_1)\right)}{(1-\rho^2)\,(1-\varepsilon(\delta_2))}.
\]
Then whenever $n>C/r$, with high probability (at least $(1-\delta_1)(1-\delta_2)$),
\[
\max_{c\in\{1,2\}}\;\max_{\boldsymbol{\beta}_c\in\mathcal{B}_r^{(c)}}\;
\mathcal{E}^{\mathrm{rd}}(S_{1,2};c,\boldsymbol{\beta}_c)
\;\le\; r
\;\le\;
\min\!\left\{
\max_{c\in\{1,2\}}\;\max_{\boldsymbol{\beta}_c\in\mathcal{B}_r^{(c)}}\mathcal{E}^{\mathrm{rd}}(S_1;c,\boldsymbol{\beta}_c),\;
\max_{c\in\{1,2\}}\;\max_{\boldsymbol{\beta}_c\in\mathcal{B}_r^{(c)}}\mathcal{E}^{\mathrm{rd}}(S_2;c,\boldsymbol{\beta}_c)
\right\}.
\]
Therefore,
\[
\min_{S\in\{S_1,S_2,S_{1,2}\}}
\max_{c\in\{1,2\}}\max_{\boldsymbol{\beta}_c\in\mathcal{B}_r^{(c)}}
\mathcal{E}^{\mathrm{rd}}(S;c,\boldsymbol{\beta}_c)
\quad\text{is attained by}\quad S=S_{1,2}.
\]
\end{proof}

\begin{lemma}
\label{lem:S1-hp}
Under the conditions of Theorem~\ref{thm:random-formal} with generating index $c=1$, there exist universal constants $c_1,c_2>0$ such that for all $0<\delta_1,\delta_2<1$ chosen suitably small, with probability at least $(1-\delta_1)(1-\delta_2)$ over the draw of the $n$ training samples,
\[
\mathcal{E}^{\mathrm{rd}}(S_1;1,\boldsymbol{\beta}_1)
\;\le\;
\frac{\sigma^2}{\,n\,(1-\epsilon(\delta_2))\,}
\Biggl(
\sum_{i=1}^p \frac{d^{t}_{1,i}}{d^{s}_{1,i}}
\;+\;
c_1\,g\left(\{d_{1,i}^s\}_{i=1}^p, \{d_{1,i}^t\}_{i=1}^p \right)\,\log\frac{1}{\delta_1}
\Biggr),
\]
where
\[
\epsilon(\delta_2)
\;=\;
c_2
\max\Biggl\{
\sqrt{\frac{\,p+\log\bigl(\tfrac{2}{\delta_2}\bigr)\,}{n}}\;,\;
\frac{\,p+\log\bigl(\tfrac{2}{\delta_2}\bigr)\,}{n}
\Biggr\},
\qquad
g\left(\{d_{1,i}^s\}_{i=1}^p, \{d_{1,i}^t\}_{i=1}^p \right) \;=\; \max_{i}\frac{d^{t}_{1,i}}{d^{s}_{1,i}}.
\]
\end{lemma}

\begin{proof}[Proof]
First, there exists an orthogonal matrix $U\in\mathbb{R}^{p\times p}$ such that
\[
\Sigma^{s}_1
= U\,\mathrm{diag} \bigl(d^{s}_1\bigr)\,U^\top,\quad
\Sigma^{s}_2
= U\,\mathrm{diag} \bigl(d^{s}_2\bigr)\,U^\top,\quad
\Sigma^{t}_1
= U\,\mathrm{diag} \bigl(d^{t}_1\bigr)\,U^\top,\quad
\Sigma^{t}_2
= U\,\mathrm{diag} \bigl(d^{t}_2\bigr)\,U^\top,
\]
for vectors $d^{s}_1,d^{s}_2,d^{t}_1,d^{t}_2\in\mathbb{R}_+^p$. For the strategy $S_1$ (using $Z_1$ only), the proof proceeds in three steps.

\noindent\textbf{Step 1.}
We show that there exists a function $B(Z_1^{s},\delta_1)$ depending on the training data $Z_1^{s}$ and $\delta_1$ such that
\[
\mathbb{P}\Bigl(
\mathcal{E}^{\mathrm{rd}}(S_1;1,\boldsymbol{\beta}_1)
\;\le\;
B(Z_1^{s},\delta_1)
\;\Big|\;
Z_1^{s}
\Bigr)
\;\ge\; 1-\delta_1.
\]

\noindent\textbf{Step 2.}
We show the existence of $\epsilon(\delta_2)$ such that
\[
\mathbb{P}\Bigl(
(1-\epsilon(\delta_2))\,\Sigma^{s}_1 \;\preceq\; \hat\Sigma^{s}_1 \;\preceq\; (1+\epsilon(\delta_2))\,\Sigma^{s}_1
\Bigr)
\;\ge\; 1-\delta_2.
\]

\noindent\textbf{Step 3.}
Let
\[
E \;=\; \bigl\{\,Z^{s}_1 : (1-\epsilon(\delta_2))\,\Sigma^{s}_1 \preceq \hat\Sigma^{s}_1 \preceq (1+\epsilon(\delta_2))\,\Sigma^{s}_1 \bigr\}.
\]
On the event $E$, we choose a constant $B(\delta_1,\delta_2)$, independent of $Z^{s}_1$, such that
$B(Z^{s}_1,\delta_1)\le B(\delta_1,\delta_2)$.
Hence,
\[
\begin{aligned}
\mathbb{P}\Bigl(\mathcal{E}^{\mathrm{rd}}(S_1;1,\boldsymbol{\beta}_1)\le B(\delta_1,\delta_2)\Bigr)
&= \mathbb{E}_{Z_1^{s}}\Bigl[
    \mathbb{P}\bigl(\mathcal{E}^{\mathrm{rd}}(S_1;1,\boldsymbol{\beta}_1)\le B(\delta_1,\delta_2)\,\big|\,Z_1^{s}\bigr)
  \Bigr] \\[4pt]
&\ge \mathbb{E}_{Z_1^{s}}\Bigl[
    \mathbf{1}_{E}(Z^{s}_1)\,
    \mathbb{P}\bigl(\mathcal{E}^{\mathrm{rd}}(S_1;1,\boldsymbol{\beta}_1)\le B(Z^{s}_1,\delta_1)\,\big|\,Z_1^{s}\bigr)
  \Bigr] \\[4pt]
&\ge (1-\delta_1)\,\mathbb{E}_{Z_1^{s}}[\mathbf{1}_{E}(Z^{s}_1)]
= (1-\delta_1)\,\mathbb{P}(Z^{s}_1\in E)
\;\ge\; (1-\delta_1)(1-\delta_2).
\end{aligned}
\]

\medskip
\noindent\textbf{Details for Step 1.}
Condition on $Z_1^{s}\in\mathbb{R}^{n\times p}$. The OLS error satisfies
\[
\hat{\boldsymbol{\beta}}_1-\boldsymbol{\beta}_1
=\bigl((Z_1^{s})^\top Z_1^{s}\bigr)^{-1}(Z_1^{s})^\top\boldsymbol{\varepsilon},
\qquad
\boldsymbol{\varepsilon}\sim\mathcal{N}(0,\sigma^2 I_n),
\]
so
\(
\hat{\boldsymbol{\beta}}_1-\boldsymbol{\beta}_1
\sim \mathcal{N}\!\bigl(0,\sigma^2(\hat{\Sigma}_1^{s})^{-1}/n\bigr),
\)
with \(\hat{\Sigma}_1^{s}=(Z_1^{s})^\top Z_1^{s}/n\).
Thus
\[
\mathcal{E}^{\mathrm{rd}}(S_1;1,\boldsymbol{\beta}_1)
=\mathbb{E}_{Z_1\sim P_t}\!\bigl[(Z_1^\top(\hat{\boldsymbol{\beta}}_1-\boldsymbol{\beta}_1))^2\bigr]
=(\hat{\boldsymbol{\beta}}_1-\boldsymbol{\beta}_1)^\top \Sigma_1^{t}(\hat{\boldsymbol{\beta}}_1-\boldsymbol{\beta}_1)
\;\stackrel{d}{=}\;
\frac{\sigma^2}{n}\,v^\top A v,
\]
where \(v\sim\mathcal{N}(0,I_p)\) and
\(
A := (\hat{\Sigma}_1^{s})^{-1/2}\,\Sigma_1^{t}\,(\hat{\Sigma}_1^{s})^{-1/2}.
\)
By the Hanson–Wright inequality~\citep{rudelson2013hanson},
there exist absolute constants \(c>0\) such that for any \(t>0\),
\[
\mathbb{P}\!\left(v^\top A v - \operatorname{tr}(A) \ge t \,\middle|\, Z_1^{s}\right)
\le \exp\!\left(-c\,\min\!\left\{\frac{t^2}{\|A\|_F^2},\,\frac{t}{\|A\|_2}\right\}\right).
\]
Taking \(c_1=1/c\) and \(t=c_1\|A\|_2\log(1/\delta_1)\), when \(\delta_1\) is small such that the second regime applies, we have with probability at least $1-\delta_1$,
\[
\mathcal{E}^{\mathrm{rd}}(S_1;1,\boldsymbol{\beta}_1)
\;\le\;
\frac{\sigma^2}{n}\Bigl(\operatorname{tr}(A) + c_1\|A\|_2\,\log\frac{1}{\delta_1}\Bigr)
=: B(Z_1^{s},\delta_1).
\]

\noindent\textbf{Details for Step 2.}
By concentration inequalities for covariance matrices~\citep{vershynin2010introduction},
there exists a universal constant \(c_2>0\) such that, for any \(\delta_2\in(0,1)\), with probability at least \(1-\delta_2\),
\[
(1-\epsilon(\delta_2))\,\Sigma_1^{s}\ \preceq\ \hat{\Sigma}_1^{s}\ \preceq\ (1+\epsilon(\delta_2))\,\Sigma_1^{s},
\]
where
\[
\epsilon(\delta_2)=
c_2\max\Biggl\{
\sqrt{\frac{\,p+\log\bigl(\tfrac{2}{\delta_2}\bigr)\,}{n}}\;,\;
\frac{\,p+\log\bigl(\tfrac{2}{\delta_2}\bigr)\,}{n}
\Biggr\},
\]
and we assume \(n\) is large enough that \(\epsilon(\delta_2) < 1\).

\noindent\textbf{Details for Step 3.}
On \(E\) we have \((\hat{\Sigma}_1^{s})^{-1}\preceq \tfrac{1}{\,1-\epsilon(\delta_2)\,}(\Sigma_1^{s})^{-1}\).
Hence,
\[
\operatorname{tr}(A)
=\operatorname{tr}\bigl((\hat{\Sigma}_1^{s})^{-1}\Sigma_1^{t}\bigr)
\le \frac{1}{\,1-\epsilon(\delta_2)\,}\,\operatorname{tr}\bigl((\Sigma_1^{s})^{-1}\Sigma_1^{t}\bigr)
= \frac{1}{\,1-\epsilon(\delta_2)\,}\sum_{i=1}^p \frac{d^{t}_{1,i}}{d^{s}_{1,i}},
\]
Moreover,
\[
\|A\|_2
=\bigl\|(\Sigma_1^{t})^{1/2}(\hat{\Sigma}_1^{s})^{-1}(\Sigma_1^{t})^{1/2}\bigr\|_2
\le \frac{1}{\,1-\epsilon(\delta_2)\,}\,
\bigl\|(\Sigma_1^{t})^{1/2}(\Sigma_1^{s})^{-1}(\Sigma_1^{t})^{1/2}\bigr\|_2
= \frac{1}{\,1-\epsilon(\delta_2)\,}\,\max_{i}\frac{d^{t}_{1,i}}{d^{s}_{1,i}}.
\]
Consequently, on \(E\),
\[
B(Z_1^{s},\delta_1)
\le
\frac{\sigma^2}{\,n\,(1-\epsilon(\delta_2))\,}
\Biggl(
\sum_{i=1}^p \frac{d^{t}_{1,i}}{d^{s}_{1,i}}
\;+\;
c_1 \max_{i}\frac{d^{t}_{1,i}}{d^{s}_{1,i}}\,
\log\frac{1}{\delta_1}
\Biggr):= B(\delta_1,\delta_2).
\]
Combining the three steps yields the stated high-probability bound with
\(
g\big(\{d_{1,i}^s\}_{i=1}^p,\{d_{1,i}^t\}_{i=1}^p\big)
=\max_{i} d_{1,i}^t/d_{1,i}^s.
\)
\end{proof}

\begin{lemma}
\label{lem:S12-lb-c1}
Under the conditions of Theorem~\ref{thm:random-formal} with generating index $c=1$, there exist universal constants $c_1,c_2>0$ such that for all $0<\delta_1,\delta_2<1$ chosen suitably small, with probability at least $(1-\delta_1)(1-\delta_2)$ over the draw of the $n$ training samples,
\[
\begin{aligned}
\mathcal{E}^{\mathrm{rd}}(S_{1,2};1,\boldsymbol{\beta}_1)
\;\ge\;&\;
\frac{\sigma^2}{\,n\,(1-\rho^2)\,}
\Biggl(
\frac{1}{1+\epsilon(\delta_2)}\,
\sum_{i=1}^p\!\left(\frac{d^{t}_{1,i}}{d^{s}_{1,i}}+\frac{d^{t}_{2,i}}{d^{s}_{2,i}}\right)\\
&\;-\;
\frac{c_1}{\,2(1-\epsilon(\delta_2))\,}\,
g_{\rho}\big(\{d_{1,i}^s\}_{i=1}^p,\{d_{2,i}^s\}_{i=1}^p,\{d_{1,i}^t\}_{i=1}^p,\{d_{2,i}^t\}_{i=1}^p\big)
\log\frac{1}{\delta_1}
\Biggr).
\end{aligned}
\]
where
\[
\epsilon(\delta_2)
\;=\;
c_2
\max\Biggl\{
\sqrt{\frac{\,2p+\log \bigl(\tfrac{2}{\delta_2}\bigr)\,}{n}}\;,\;
\frac{\,2p+\log \bigl(\tfrac{2}{\delta_2}\bigr)\,}{n}
\Biggr\},
\]
and
\[
g_{\rho}\big(\{d_{1,i}^s\}_{i=1}^p,\{d_{2,i}^s\}_{i=1}^p,\{d_{1,i}^t\}_{i=1}^p,\{d_{2,i}^t\}_{i=1}^p\big)
\;=\;
\max_{i}
\left(
\frac{d^{t}_{1,i}}{d^{s}_{1,i}}
+
\frac{d^{t}_{2,i}}{d^{s}_{2,i}}
+
\sqrt{\left(\frac{d^{t}_{1,i}}{d^{s}_{1,i}}-\frac{d^{t}_{2,i}}{d^{s}_{2,i}}\right)^2
+4\rho^2\,\frac{d^{t}_{1,i}}{d^{s}_{1,i}}\frac{d^{t}_{2,i}}{d^{s}_{2,i}}}
\right).
\]
\end{lemma}

\begin{proof}[Proof]
First, there exists an orthogonal matrix $U\in\mathbb{R}^{p\times p}$ such that
\[
\Sigma^{s}_1
= U\,\mathrm{diag} \bigl(d^{s}_1\bigr)\,U^\top,\quad
\Sigma^{s}_2
= U\,\mathrm{diag} \bigl(d^{s}_2\bigr)\,U^\top,\quad
\Sigma^{t}_1
= U\,\mathrm{diag} \bigl(d^{t}_1\bigr)\,U^\top,\quad
\Sigma^{t}_2
= U\,\mathrm{diag} \bigl(d^{t}_2\bigr)\,U^\top,
\]
for vectors $d^{s}_1,d^{s}_2,d^{t}_1,d^{t}_2\in\mathbb{R}_+^p$.
For the strategy $S_{1,2}$ (using $Z=(Z_1,Z_2)$), the proof proceeds in three steps.

\noindent\textbf{Step 1.}
We show there exists a function $C(Z^{s},\delta_1)$ depending on the full training data $Z^{s}=[\,Z_1^{s}\;Z_2^{s}\,]\in\mathbb{R}^{n\times 2p}$ and $\delta_1$ such that
\[
\mathbb{P}\Bigl(
\mathcal{E}^{\mathrm{rd}}(S_{1,2};1,\boldsymbol{\beta}_1)
\;\ge\;
C(Z^{s},\delta_1)
\;\Big|\;
Z^{s}
\Bigr)
\;\ge\; 1-\delta_1.
\]

\noindent\textbf{Step 2.}
We show the existence of $\epsilon(\delta_2)$ such that
\[
\mathbb{P}\Bigl(
(1-\epsilon(\delta_2))\,\Sigma^{s}
\;\preceq\;
\hat\Sigma^{s}
\;\preceq\;
(1+\epsilon(\delta_2))\,\Sigma^{s}
\Bigr)
\;\ge\; 1-\delta_2,
\]
where $\Sigma^{s}=\mathbb{E}_{(Z_1,Z_2)\sim P_s}[ZZ^\top]$ and $\hat\Sigma^{s}=(Z^{s})^\top Z^{s}/n$ are the population and empirical covariances of $Z=(Z_1,Z_2)$.

\noindent\textbf{Step 3.}
Let
\[
E \;=\; \bigl\{\,Z^{s}:\ (1-\epsilon(\delta_2))\,\Sigma^{s}\preceq \hat\Sigma^{s}\preceq (1+\epsilon(\delta_2))\,\Sigma^{s}\bigr\}.
\]
On the event $E$, we choose a constant $C(\delta_1,\delta_2)$, independent of $Z^{s}$, such that
$C(Z^{s},\delta_1)\;\ge\; C(\delta_1,\delta_2)$.
Hence
\[
\begin{aligned}
\mathbb{P} \Bigl(\mathcal{E}^{\mathrm{rd}}(S_{1,2};1,\boldsymbol{\beta}_1)\ge C(\delta_1,\delta_2)\Bigr)
&= \mathbb{E}_{Z^{s}}\Bigl[
\mathbb{P}\bigl(\mathcal{E}^{\mathrm{rd}}(S_{1,2};1,\boldsymbol{\beta}_1)\ge C(\delta_1,\delta_2)\,\big|\,Z^{s}\bigr)
\Bigr]\\[3pt]
&\ge \mathbb{E}_{Z^{s}}\Bigl[\mathbf{1}_{E}(Z^{s})\,
\mathbb{P}\bigl(\mathcal{E}^{\mathrm{rd}}(S_{1,2};1,\boldsymbol{\beta}_1)\ge C(Z^{s},\delta_1)\,\big|\,Z^{s}\bigr)
\Bigr]\\[3pt]
&\ge (1-\delta_1)\,\mathbb{E}_{Z^{s}}\bigl[\mathbf{1}_{E}(Z^{s})\bigr]
= (1-\delta_1)\,\mathbb{P}(Z^{s}\in E)\\[3pt]
&\ge (1-\delta_1)(1-\delta_2).
\end{aligned}
\]

\medskip
\noindent\textbf{Details for Step 1.}
Let $\hat{\boldsymbol{\beta}}$ be the OLS estimator in the joint model trained on $Z^{s}$.
The standard OLS formula yields
\[
\hat{\boldsymbol{\beta}}-\boldsymbol{\beta}
=\bigl((Z^{s})^\top Z^{s}\bigr)^{-1}(Z^{s})^\top\boldsymbol{\varepsilon},
\qquad
\boldsymbol{\beta}=\begin{bmatrix}\boldsymbol{\beta}_1\\ 0\end{bmatrix},
\]
hence
\(
\hat{\boldsymbol{\beta}}-\boldsymbol{\beta}
\sim \mathcal{N}\bigl(0,\sigma^2(\hat{\Sigma}^{s})^{-1}/n\bigr)
\)
.
So
\[
\mathcal{E}^{\mathrm{rd}}(S_{1,2};1,\boldsymbol{\beta}_1)
=\mathbb{E}_{(Z_1,Z_2)\sim P_t}\!\bigl[(Z^\top(\hat{\boldsymbol{\beta}}-\boldsymbol{\beta}))^2\bigr]
=(\hat{\boldsymbol{\beta}}-\boldsymbol{\beta})^\top \Sigma^{t}(\hat{\boldsymbol{\beta}}-\boldsymbol{\beta})
\;\stackrel{d}{=}\;
\frac{\sigma^2}{n}\,v^\top A v,
\]
where $v\sim\mathcal{N}(0,I_{2p})$ and
\(
A := (\hat{\Sigma}^{s})^{-1/2}\,\Sigma^{t}\,(\hat{\Sigma}^{s})^{-1/2}.
\)
By the Hanson–Wright inequality \citep{rudelson2013hanson},
there exists an absolute constant $c>0$ such that, for any $t>0$,
\[
\mathbb{P}\!\left(v^\top A v - \operatorname{tr}(A) \le -t \,\middle|\, Z^{s}\right)
\le \exp\!\left(-c\,\min\!\left\{\frac{t^2}{\|A\|_F^2},\,\frac{t}{\|A\|_2}\right\}\right).
\]
Similar to the proof in Lemma \ref{lem:S1-hp}, when \(\delta_1\) is small, there exists \(c_1 >0\) such that with probability at least $1-\delta_1$,
\[
\mathcal{E}^{\mathrm{rd}}(S_{1,2};1,\boldsymbol{\beta}_1)
\;\ge\;
\frac{\sigma^2}{n}\Bigl(\operatorname{tr}(A) - c_1\|A\|_2\,\log\frac{1}{\delta_1}\Bigr)
=: C(Z^{s},\delta_1).
\]

\noindent\textbf{Details for Step 2.}
By concentration inequalities for covariance matrices \citep{vershynin2010introduction},
there exists a universal constant $c_2>0$ such that, for any $\delta_2\in(0,1)$, with probability at least $1-\delta_2$,
\[
(1-\epsilon(\delta_2))\,\Sigma^{s}
\ \preceq\
\hat{\Sigma}^{s}
\ \preceq\
(1+\epsilon(\delta_2))\,\Sigma^{s},
\]
where
\[
\epsilon(\delta_2)
=
c_2 \max \Biggl\{
\sqrt{\frac{\,2p+\log\bigl(\frac{2}{\delta_2}\bigr)\,}{n}}\;,\;
\frac{\,2p+\log\bigl(\frac{2}{\delta_2}\bigr)\,}{n}
\Biggr\},
\]
and we assume \(n\) is large enough that \(\epsilon(\delta_2) < 1\).

\noindent\textbf{Details for Step 3.}
On $E$,
\(
(\hat{\Sigma}^{s})^{-1}\succeq \tfrac{1}{\,1+\epsilon(\delta_2)\,}(\Sigma^{s})^{-1}
\)
and
\(
(\hat{\Sigma}^{s})^{-1}\preceq \tfrac{1}{\,1-\epsilon(\delta_2)\,}(\Sigma^{s})^{-1}.
\)
Thus, by Loewner monotonicity of trace and spectral norm,
\[
\operatorname{tr}(A)
=\operatorname{tr} \bigl((\hat{\Sigma}^{s})^{-1}\Sigma^{t}\bigr)
\ \ge\
\frac{1}{\,1+\epsilon(\delta_2)\,}\,
\operatorname{tr}\bigl((\Sigma^{s})^{-1}\Sigma^{t}\bigr),
\qquad
\|A\|_2
\ \le\
\frac{1}{\,1-\epsilon(\delta_2)\,}\,
\bigl\|(\Sigma^{t})^{1/2}(\Sigma^{s})^{-1}(\Sigma^{t})^{1/2}\bigr\|_2.
\]

We further simultaneously diagonalize with
$U'=\begin{bmatrix}U&0\\[2pt]0&U\end{bmatrix}$ and write
\[
(\Sigma^{s})^{-1}
=\frac{1}{1-\rho^2}\,
U'\!
\begin{bmatrix}
\mathrm{diag}\bigl(\frac{1}{d_{1}^{s}}\bigr) & -\,\mathrm{diag}\bigl(\frac{\rho}{\sqrt{d_{1}^{s}d_{2}^{s}}}\bigr)\\[4pt]
-\,\mathrm{diag}\bigl(\frac{\rho}{\sqrt{d_{1}^{s}d_{2}^{s}}}\bigr) & \mathrm{diag}\bigl(\frac{1}{d_{2}^{s}}\bigr)
\end{bmatrix}
U'^{\!\top},
\qquad
\Sigma^{t}
=U'\!
\begin{bmatrix}
\mathrm{diag}(d_{1}^{t}) & 0\\[2pt]
0 & \mathrm{diag}(d_{2}^{t})
\end{bmatrix}
U'^{\!\top}.
\]
Therefore,
\[
\operatorname{tr} \bigl((\Sigma^{s})^{-1}\Sigma^{t}\bigr)
=\frac{1}{1-\rho^2}\sum_{i=1}^p\!\left(\frac{d^{t}_{1,i}}{d^{s}_{1,i}}+\frac{d^{t}_{2,i}}{d^{s}_{2,i}}\right),
\]
and
\[
(\Sigma^{t})^{1/2}(\Sigma^{s})^{-1}(\Sigma^{t})^{1/2}
=\frac{1}{1-\rho^2}\,
U'\!
\begin{bmatrix}
\mathrm{diag}\bigl(\frac{d^{t}_{1}}{d^{s}_{1}}\bigr)
& -\,\rho\,\mathrm{diag}\Bigl(\frac{\sqrt{d^{t}_{1}d^{t}_{2}}}{\sqrt{d^{s}_{1}d^{s}_{2}}}\Bigr)\\[6pt]
-\,\rho\,\mathrm{diag}\Bigl(\frac{\sqrt{d^{t}_{1}d^{t}_{2}}}{\sqrt{d^{s}_{1}d^{s}_{2}}}\Bigr)
& \mathrm{diag}\bigl(\frac{d^{t}_{2}}{d^{s}_{2}}\bigr)
\end{bmatrix}
U'^{\!\top}:= C.
\]
By a suitable simultaneous permutation of
rows and columns, \(C\) is similar to $\bigoplus_{i=1}^p C_i$ with
\[
C_i=
\begin{bmatrix}
\frac{d^{t}_{1,i}}{d^{s}_{1,i}} & -\,\rho\,\sqrt{\frac{d^{t}_{1,i}d^{t}_{2,i}}{d^{s}_{1,i}d^{s}_{2,i}}}\\[4pt]
-\,\rho\,\sqrt{\frac{d^{t}_{1,i}d^{t}_{2,i}}{d^{s}_{1,i}d^{s}_{2,i}}} & \frac{d^{t}_{2,i}}{d^{s}_{2,i}}
\end{bmatrix}.
\]
Hence
\[
\bigl\|(\Sigma^{t})^{1/2}(\Sigma^{s})^{-1}(\Sigma^{t})^{1/2}\bigr\|_2
=\frac{1}{1-\rho^2}\,\max_{i}\lambda_{\max}(C_i)
\ \le\ \frac{1}{2(1-\rho^2)}\,
g_{\rho}\big(\{d_{1,i}^s\}_{i=1}^p,\{d_{2,i}^s\}_{i=1}^p,\{d_{1,i}^t\}_{i=1}^p,\{d_{2,i}^t\}_{i=1}^p\big),
\]
where $g_\rho(\cdot)$ is as in the lemma statement.
Combining the bounds on $\operatorname{tr}(A)$ and $\|A\|_2$ on $E$ yields
\[
\begin{aligned}
 C(Z^{s},\delta_1)
\;\ge\;&\;
\frac{\sigma^2}{\,n\,(1-\rho^2)\,}
\Biggl(
\frac{1}{1+\epsilon(\delta_2)}\,
\sum_{i=1}^p\!\left(\frac{d^{t}_{1,i}}{d^{s}_{1,i}}+\frac{d^{t}_{2,i}}{d^{s}_{2,i}}\right)\\
&\;-\;
\frac{c_1}{\,2(1-\epsilon(\delta_2))\,}\,
g_{\rho}\big(\{d_{1,i}^s\}_{i=1}^p,\{d_{2,i}^s\}_{i=1}^p,\{d_{1,i}^t\}_{i=1}^p,\{d_{2,i}^t\}_{i=1}^p\big)
\log\frac{1}{\delta_1}
\Biggr):= C(\delta_1, \delta_2).
\end{aligned}
\]
which completes the proof.
\end{proof}

\begin{corollary}
\label{cor:S12-ub-c1}
Under the conditions of Theorem~\ref{thm:random-formal} with generating index $c=1$, there exist universal constants $c_1,c_2>0$ such that for all $0<\delta_1,\delta_2<1$ chosen suitably small, with probability at least $(1-\delta_1)(1-\delta_2)$ over the draw of the $n$ training samples,
\[
\begin{aligned}
\mathcal{E}^{\mathrm{rd}}(S_{1,2};1,\boldsymbol{\beta}_1)
\;\le\;&\;
\frac{\sigma^2}{\,n\,(1-\rho^2)\,(1-\varepsilon(\delta_2))}
\Biggl(
\sum_{i=1}^p\!\left(\frac{d^{t}_{1,i}}{d^{s}_{1,i}}+\frac{d^{t}_{2,i}}{d^{s}_{2,i}}\right)\\
&\;+\;
\frac{1}{2}c_1
g_{\rho}\big(\{d_{1,i}^s\}_{i=1}^p,\{d_{2,i}^s\}_{i=1}^p,\{d_{1,i}^t\}_{i=1}^p,\{d_{2,i}^t\}_{i=1}^p\big)\,
\log\frac{1}{\delta_1}
\Biggr),
\end{aligned}
\]
where
\[
\varepsilon(\delta_2)
\;=\;
c_2
\max\Biggl\{
\sqrt{\frac{\,2p+\log \bigl(\frac{2}{\delta_2}\bigr)\,}{n}}\;,\;
\frac{\,2p+\log \bigl(\frac{2}{\delta_2}\bigr)\,}{n}
\Biggr\},
\]
and $g_{\rho}(\cdot)$ is as in Lemma~\ref{lem:S12-lb-c1}.
\end{corollary}

\section{Additional Experimental Details}
\label{appendix:exp}
\subsection{Dataset Statistics}
\label{subapp:ds}

\textbf{Waterbirds-95\% statistics}:
Label set $\mathcal{Y} = \{\text{waterbird}, \text{landbird}\}$. Attribute set $\mathcal{Z} = \{\text{water}, \text{land}\}$.

\begin{table}[htbp]
\centering
{\fontsize{8pt}{9pt}\selectfont
\caption{Dataset statistics for Waterbirds-95\%.}
\label{tab:waterbirds95_stats}
\begin{tabular}{lcccc}
\toprule
Split & (waterbird, water) & (waterbird, land) & (landbird, water) & (landbird, land) \\
\midrule
Train      &1,057     &56     &184     &3,498     \\
Validation &133     &133     &466     &467     \\
Test       & 642    &642     &2,255     &2,255     \\
\bottomrule
\end{tabular}
}
\end{table}

\textbf{Waterbirds-100\% statistics}:  
Label set $\mathcal{Y} = \{\text{waterbird}, \text{landbird}\}$.  
Attribute set $\mathcal{Z} = \{\text{water}, \text{land}\}$.

\begin{table}[htbp]
\centering
{\fontsize{8pt}{9pt}\selectfont
\caption{Dataset statistics for Waterbirds-100\%.}
\label{tab:waterbirds100_stats}
\begin{tabular}{lcccc}
\toprule
Split & (waterbird, water) & (waterbird, land) & (landbird, water) & (landbird, land) \\
\midrule
Train      &1,101       &0       &0       &3,694       \\
Validation &133       &133       &466       &467       \\
Test       &642       &642       &2,255       &2,255       \\
\bottomrule
\end{tabular}
}
\end{table}

\textbf{SpuCo Dogs statistics}:  
Label set $\mathcal{Y} = \{\text{small dog}, \text{big dog}\}$.  
Attribute set $\mathcal{Z} = \{\text{indoor}, \text{outdoor}\}$.

\begin{table}[H]
\centering
{\fontsize{8pt}{9pt}\selectfont
\caption{Dataset statistics for SpuCo Dogs.}
\label{tab:spucodogs_stats}
\begin{tabular}{lcccc}
\toprule
Split &(big dog, indoor) & (big dog, outdoor) & (small dog, indoor) & (small dog, outdoor) \\
\midrule
Train      & 500   & 10,000 & 10,000 & 500   \\
Validation & 25    & 500   & 500   & 25    \\
Test       & 500   & 500   & 500   & 500   \\
\bottomrule
\end{tabular}
}
\end{table}

\textbf{MetaShift statistics}:  
Label set $\mathcal{Y} = \{\text{cat}, \text{dog}\}$.  
Attribute set  
\[
\mathcal{Z} = \{\text{sofa}, \text{bed}, \text{shelf}, \text{cabinet}, \text{bag}, \text{box}, \text{bench}, \text{bike}, \text{boat}, \text{surfboard}\}.
\]
We consider four subsets in~\citep{liang2022metashift}, each differing only in the two attributes paired with dog in the training data. According to the distances to (dog, shelf) reported in~\citep{liang2022metashift}, these subsets are:
\[
\begin{array}{llc}
\text{(a)} & \{\text{cabinet},\,\text{bed}\} & d = 0.44,\\
\text{(b)} & \{\text{bag},\,\text{box}\}     & d = 0.71,\\
\text{(c)} & \{\text{bench},\,\text{bike}\} & d = 1.12,\\
\text{(d)} & \{\text{boat},\,\text{surfboard}\} & d = 1.43.
\end{array}
\]
Larger $d$ indicates a more challenging spurious correlation. We partition a portion of the test set into a validation set following a \(15:85\) ratio, as in~\citep{phan2024controllable}.

\begin{table}[htbp]
\centering
{\fontsize{8pt}{9pt}\selectfont
\caption{Data statistics for MetaShift subset (a): cabinet \& bed ($d=0.44$).}
\label{tab:metashift_a}
\begin{tabular}{lcccccc}
\toprule
Split 
  & (cat, sofa) & (cat, bed) & (dog, cabinet) & (dog, bed) & (cat, shelf) & (dog, shelf) \\
\midrule
Train      &231       &380       &314       &244       &0       &0       \\
Validation &0       &0       &0       &0       &34       &47       \\
Test       &0       &0       &0       &0       &201       &259       \\
\bottomrule
\end{tabular}
}
\end{table}

\begin{table}[H]
\centering
{\fontsize{8pt}{9pt}\selectfont
\caption{Data statistics for MetaShift subset (b): bag \& box ($d=0.71$).}
\label{tab:metashift_b}
\begin{tabular}{lcccccc}
\toprule
Split 
  & (cat, sofa) & (cat, bed) & (dog, bag) & (dog, box) & (cat, shelf) & (dog, shelf) \\
\midrule
Train      &231       &380       &202       &193       &0       &0       \\
Validation &0       &0       &0       &0       &34       &47       \\
Test       &0       &0       &0       &0       &201       &259       \\
\bottomrule
\end{tabular}
}
\end{table}

\begin{table}[H]
\centering
{\fontsize{8pt}{9pt}\selectfont
\caption{Data statistics for MetaShift subset (c): bench \& bike ($d=1.12$).}
\label{tab:metashift_c}
\begin{tabular}{lcccccc}
\toprule
Split 
  & (cat, sofa) & (cat, bed) & (dog, bench) & (dog, bike) & (cat, shelf) & (dog, shelf) \\
\midrule
Train      &231       &380       &145       &367       &0       &0      \\
Validation &0       &0       &0       &0       &34       &47       \\
Test       &0       &0       &0       &0       &201       &259       \\
\bottomrule
\end{tabular}
}
\end{table}

\begin{table}[H]
\centering
{\fontsize{8pt}{9pt}\selectfont
\caption{Data statistics for MetaShift subset (d): boat \& surfboard ($d=1.43$).}
\label{tab:metashift_d}
\begin{tabular}{lcccccc}
\toprule
Split 
  & (cat, sofa) & (cat, bed) & (dog, boat) & (dog, surfboard) & (cat, shelf) & (dog, shelf) \\
\midrule
Train      &231       &380       &459       &318       &0       &0       \\
Validation &0       &0       &0       &0       &34       &47       \\
Test       &0       &0       &0       &0       &201       &259       \\
\bottomrule
\end{tabular}
}
\end{table}

\textbf{Spawrious statistics}:  
Label set \(\mathcal{Y} = \{\text{Bulldog}, \text{Dachshund}, \text{Corgi}, \text{Labrador}\}\). Attribute set
\[
\mathcal{Z} = \{\text{Beach}, \text{Desert}, \text{Dirt}, \text{Jungle}, \text{Mountain}, \text{Snow}\}.
\]
The Spawrious dataset includes two modes of spurious correlation:  
(1) One-to-one (O2O): each class is associated with exactly one attribute during training. At test time, the model encounters novel label–attribute combinations.  
(2) Many-to-many (M2M): a subset of classes is correlated with a subset of attributes during training, and this correlation is permuted in the test environment.

Each mode is divided into three subsets labeled as ``easy,” ``medium,” and ``hard” following the original paper’s naming convention, resulting in six subsets in total.   
For each subset, the original Spawrious dataset provides two training domains and one test domain.  
To align with the setup of other datasets, we merge the two training domains into a single training set, and for each group in the test domain, we split 10\% of the test samples into a validation set.

\begin{table}[H]
\centering
{\fontsize{8pt}{9pt}\selectfont
\caption{Data statistics for Spawrious subset: O2O–Easy}
\label{tab:spawrious_o2o_easy}
\begin{tabularx}{\textwidth}{
    >{\raggedright\arraybackslash}X  % first column left‐aligned
    *{2}{>{\centering\arraybackslash}X} % Train Env 1: Class, Attr
    *{2}{>{\centering\arraybackslash}X} % Train Env 2: Class, Attr
    >{\centering\arraybackslash}X       % Test Env
}
\toprule
          & \multicolumn{2}{c}{Train I} 
          & \multicolumn{2}{c}{Train II} 
          & Test \\
\midrule
Bulldog   &3,072 Desert            &96 Beach           
          &2,756 Desert            &412 Beach           
          &3,168 Dirt  \\
Dachshund &3,072 Jungle            &96 Beach           
          &2,756 Jungle            &412 Beach           
          &3,168 Snow  \\
Corgi     &3,072 Snow            &96 Beach           
          &2,756 Snow            &412 Beach           
          &3,168 Jungle  \\
Labrador  &3,072 Dirt            &96 Beach           
          &2,756 Dirt            &412 Beach           
          &3,168 Desert  \\
\bottomrule
\end{tabularx}
}
\end{table}

\begin{table}[H]
\centering
{\fontsize{8pt}{9pt}\selectfont
\caption{Data statistics for Spawrious subset: O2O–Medium}
\label{tab:spawrious_o2o_medium}
\begin{tabularx}{\textwidth}{
  >{\raggedright\arraybackslash}X  
  Y M                            % Train Env 1: Class, Attr
  Y M                            % Train Env 2: Class, Attr
  M                              % Test Env
}
\toprule
          & \multicolumn{2}{c}{Train I} 
          & \multicolumn{2}{c}{Train II} 
          & Test \\
\midrule
Bulldog   &3,072 Mountain            &96 Desert           
          &2,756 Mountain            &412 Desert           
          &3,168 Jungle  \\
Dachshund &3,072 Beach            &96 Desert           
          &2,756 Beach            &412 Desert           
          &3,168 Dirt  \\
Corgi     &3,072 Jungle            &96 Desert           
          &2,756 Jungle           &412 Desert           
          &3,168 Snow  \\
Labrador  &3,072 Dirt            &96 Desert           
          &2,756 Dirt            &412 Desert           
          &3,168 Beach  \\
\bottomrule
\end{tabularx}
}
\end{table}

\begin{table}[H]
\centering
{\fontsize{8pt}{9pt}\selectfont
\caption{Data statistics for Spawrious subset: O2O–Hard}
\label{tab:spawrious_o2o_hard}
\begin{tabularx}{\textwidth}{
  >{\raggedright\arraybackslash}X  
  Y Z                            % Train Env 1: Class, Attr
  Y Z                            % Train Env 2: Class, Attr
  Y                              % Test Env
}
\toprule
          & \multicolumn{2}{c}{Train I} 
          & \multicolumn{2}{c}{Train II} 
          & Test \\
\midrule
Bulldog   &3,072 Jungle            &96 Beach           
          &2,756 Jungle            &412 Beach           
          &3,168 Mountain  \\
Dachshund &3,072 Mountain            &96 Beach           
          &2,756 Mountain            &412 Beach           
          &3,168 Snow  \\
Corgi     &3,072 Desert            &96 Beach           
          &2,756 Desert           &412 Beach           
          &3,168 Jungle  \\
Labrador  &3,072 Snow            &96 Beach           
          &2,756 Snow            &412 Beach           
          &3,168 Desert  \\
\bottomrule
\end{tabularx}
}
\end{table}

\begin{table}[H]
\centering
{\fontsize{8pt}{9pt}\selectfont
\caption{Data statistics for Spawrious subset: M2M-Easy}
\label{tab:spawrious_m2m_easy}
\begin{tabularx}{\textwidth}{
    >{\raggedright\arraybackslash}X
    *{4}{>{\centering\arraybackslash}X}
}
\toprule
            & \multicolumn{1}{c}{Train I} 
            & \multicolumn{1}{c}{Train II} 
            & \multicolumn{2}{c}{Test} \\
\midrule
Bulldog     &3,168 Desert                   &3,168 Mountain                   &3,168 Dirt                   &3,168 Jungle                    \\
Dachshund   &3,168 Mountain                   &3,168 Desert                   &3,168 Dirt                   &3,168 Jungle                    \\
Corgi       &3,168 Jungle                   &3,168 Dirt                   &3,168 Desert                   &3,168 Mountain                    \\
Labrador    &3,168 Dirt                   & 3,168 Jungle                   &3,168 Desert                   &3,168 Mountain                    \\
\bottomrule
\end{tabularx}
}
\end{table}

\begin{table}[H]
\centering
{\fontsize{8pt}{9pt}\selectfont
\caption{Data statistics for Spawrious subset: M2M-Medium}
\label{tab:spawrious_m2m_medium}
\begin{tabularx}{\textwidth}{
    >{\raggedright\arraybackslash}X
    *{4}{>{\centering\arraybackslash}X}
}
\toprule
            & \multicolumn{1}{c}{Train I} 
            & \multicolumn{1}{c}{Train II} 
            & \multicolumn{2}{c}{Test} \\
\midrule
Bulldog     &3,168 Beach                   &3,168 Snow                   &3,168 Desert                   &3,168 Mountain                    \\
Dachshund   &3,168 Snow                   &3,168 Beach                   &3,168 Desert                   &3,168 Mountain                    \\
Corgi       &3,168 Desert                   &3,168 Mountain                   &3,168 Beach                   &3,168 Snow                    \\
Labrador    &3,168 Mountain                   & 3,168 Desert                   &3,168 Beach                   &3,168 Snow                    \\
\bottomrule
\end{tabularx}
}
\end{table}

\begin{table}[H]
\centering
{\fontsize{8pt}{9pt}\selectfont
\caption{Data statistics for Spawrious subset: M2M-Hard}
\label{tab:spawrious_m2m_hard}
\begin{tabularx}{\textwidth}{
    >{\raggedright\arraybackslash}X
    *{4}{>{\centering\arraybackslash}X}
}
\toprule
            & \multicolumn{1}{c}{Train I} 
            & \multicolumn{1}{c}{Train II} 
            & \multicolumn{2}{c}{Test} \\
\midrule
Bulldog     &3,168 Dirt                   &3,168 Jungle                   &3,168 Snow                   &3,168 Beach                    \\
Dachshund   &3,168 Jungle                   &3,168 Dirt                   &3,168 Snow                   &3,168 Beach                    \\
Corgi       &3,168 Beach                   &3,168 Snow                   &3,168 Dirt                  &3,168 Jungle                    \\
Labrador    &3,168 Snow                   & 3,168 Beach                   &3,168 Dirt                   &3,168 Jungle                    \\
\bottomrule
\end{tabularx}
}
\end{table}

\subsection{Hyperparameter Selection}
\label{subapp:hs}

\textbf{SupER.} Our SupER model employs a $\beta$-VAE encoder built upon the ResNet50 backbone architecture. For consistency, the CLIP model also uses ResNet50. We perform a grid search to assess the performance of SupER under different hyperparameter configurations and select the optimal values for each dataset as summarized in Table \ref{tab:hyperparams}.  Specifically, the hyperparameters are as follows: $\beta$ denotes the weighting factor of $\beta$-VAE; $\lambda_1$ is the weight for the loss $\mathcal{L}^{\text{Beta}}_{\theta, \phi}(\mathbf{x})$; $\lambda_2$ is the weight for the loss $\mathcal{L}^{\text{ATT}}_{\phi, \omega_{\mathrm{rel}}, \omega_{\mathrm{irr}}}(\mathbf{x}, y)$; $\lambda_3$ controls the $L_2$ regularization term $||\omega_{\mathrm{rel}}||_2^2$, where $n_1$ denotes the number of parameters in $\omega_{\mathrm{rel}}$; $\eta$ is the learning rate; $B$ is the batch size; $T$ is the number of epochs; $\gamma$ denotes the weight decay coefficient used in the Adam optimizer; and $d$ specifies the dimensionality of features $\mathbf{z}_{\mathrm{rel}}$ and $\mathbf{z}_{\mathrm{irr}}$. Early stopping is adopted when applicable, and training is terminated once the worst-group accuracy on the validation set reaches its maximum. Note that our method does not require any group information; this criterion is used only for fair comparison with previous work.
For the number of superclass-specific text prompts $m$, unless stated otherwise, we set $m=1$. The text prompts used for each dataset are detailed in Table~\ref{tab:prompts}. 

\begin{table*}[htbp]
\centering
{\fontsize{8pt}{9pt}\selectfont
\caption{SupER hyperparameter settings across datasets.}
\label{tab:hyperparams}
\begin{tabular*}{\textwidth}{@{\extracolsep{\fill}}lccccccccc}
\toprule
Dataset 
  & $\beta$ 
  & $\lambda_{1}$ 
  & $\lambda_{2}$ 
  & $\lambda_{3}$ 
  & $\eta$ 
  & $B$ 
  & $T$ 
  & $\gamma$ 
  & $d$ \\
\midrule
Waterbirds-95\%       & 1 & 1 & 40 & \(1000/n_1\) & \(10^{-5}\) & 32 & 50 & \(10^{-4}\) & 256 \\
Waterbirds-100\%      & 1 & 1 & 40 & \(1000/n_1\) & \(10^{-5}\) & 32 & 50 & \(10^{-4}\) & 256 \\
SpuCo Dogs            & 1 & 1 & 40 & \(100/n_1\) & \(10^{-6}\) & 32 & 30 & \(10^{-2}\) & 256 \\
MetaShift (a)         & 5 & 1 & 1 & \(100/n_1\) & \(10^{-5}\) & 32 & 100 & \(10^{-2}\) & 256 \\
MetaShift (b)         & 5 & 1 & 1 & \(100/n_1\) & \(10^{-5}\) & 32 & 100 & \(10^{-2}\) & 256 \\
MetaShift (c)         & 5 & 1 & 20 & \(100/n_1\) & \(10^{-5}\) & 32 & 100 & \(10^{-2}\) & 256 \\
MetaShift (d)         & 10 & 1 & 20 & \(100/n_1\) & \(10^{-5}\) & 32 & 100 & \(10^{-2}\) & 256 \\
Spawrious O2O–Easy    & 10 & 1 & 10 & \(100/n_1\) & \(10^{-6}\) & 32 & 30 & \(10^{-4}\) & 256 \\
Spawrious O2O–Medium  & 1 & 1 & 80 & \(100/n_1\) & \(10^{-6}\) & 32 & 30 & \(10^{-4}\) & 256 \\
Spawrious O2O–Hard    & 1 & 1 & 80 & \(100/n_1\) & \(10^{-6}\) & 32 & 30 & \(10^{-4}\) & 256 \\
Spawrious M2M–Easy    & 10 & 1 & 50 & \(100/n_1\) & \(10^{-6}\) & 32 & 30 & \(10^{-4}\) & 256 \\
Spawrious M2M–Medium  & 1 & 1 & 50 & \(100/n_1\) & \(10^{-6}\) & 32 & 30 & \(10^{-4}\) & 256 \\
Spawrious M2M–Hard    & 1 & 1 & 50 & \(100/n_1\) & \(10^{-6}\) & 32 & 30 & \(10^{-4}\) & 256 \\
\bottomrule
\end{tabular*}
}
\end{table*}

\begin{table*}[htbp]
\centering
{\fontsize{8pt}{9pt}\selectfont
\caption{Superclass text prompts for each dataset}
\label{tab:prompts}
\begin{tabular*}{0.5\textwidth}{@{\extracolsep{\fill}}l l}
\toprule
Dataset & Prompt \\
\midrule
Waterbirds-95\%      & $\mathtt{a\ bird}$ \\
Waterbirds-100\%     & $\mathtt{a\ bird}$ \\
SpuCo Dogs           & $\mathtt{a\ dog}$ \\
MetaShift      & $\mathtt{a\ cat\ or\ a\ dog}$ \\
Spawrious  & $\mathtt{a\ dog}$ \\
\bottomrule
\end{tabular*}
}
\end{table*}

\textbf{Baselines.} For baseline methods considered in our experiments, we similarly employ ResNet50 backbone architectures and determine their optimal hyperparameters via grid search. We specifically evaluate learning rates $\eta \in \left\{10^{-6}, 10^{-5}, 10^{-4}\right\}$ and weight decay $\gamma \in \left\{10^{-4}, 10^{-2}\right\}$, with the batch size and number of training epochs for each dataset as specified in Table~\ref{tab:hyperparams}. Note that for all the above configurations, as well as additional model-specific hyperparameters, we directly use the values provided or recommended in the original papers whenever available.

\subsection{Full Worst Group Accuracy, Average Accuracy, and Group Accuracy Variance for All Datasets}
\label{app:WGAAAV}

\textbf{Worst group and average accuracy.} Tables \ref{tab:wb_results_app_0}, \ref{tab:sc_results_app_0}, \ref{tab:sc_results_app}, \ref{tab:ms_results_app0}, \ref{tab:ms_results_app}, and  \ref{tab:wb_results_app} summarize the worst group accuracy and average accuracy for all datasets and selected baseline methods. \textbf{Bold} indicates the best across all selected baselines; \underline{Underlined} indicates the best among methods without group information;  ``--'' indicates omitted result due to consistently subpar or unstable performance, even after comprehensive hyperparameter tuning using the original codebase.

\begin{table}[H]
  \captionsetup{skip=4pt}
  \centering
  {\fontsize{8pt}{9pt}\selectfont
  \caption{Worst and average group accuracy (\%) for Waterbirds-95\% and Waterbirds-100\%.}
  \label{tab:wb_results_app_0}
  \begin{tabular*}{\textwidth}{@{\extracolsep{\fill}} l  c  c  c  c  c  c}
    \toprule
    \multirow{2}{*}{Method}
      & \multirow{2}{*}{\shortstack{Group\\Info}}
      & \multirow{2}{*}{\shortstack{Train\\Twice}}
      & \multicolumn{2}{c}{Waterbirds-95\%}
      & \multicolumn{2}{c}{Waterbirds-100\%} \\
    \cmidrule(lr){4-5} \cmidrule(lr){6-7}
      &  & 
      & Worst & Avg 
      & Worst & Avg \\
    \midrule
    ERM          & $\times$     & $\times$     
                 & 64.9$_{\pm1.5}$  & 90.7$_{\pm1.0}$
                 & 46.4$_{\pm6.9}$  & 74.8$_{\pm3.0}$ \\
    CVaR\,DRO    & $\times$     & $\times$     
                 & 73.1$_{\pm7.1}$ & 90.7$_{\pm0.7}$
                 & 58.0$_{\pm2.2}$ & 79.0$_{\pm1.2}$ \\
    LfF          & $\times$     & $\times$     
                 & 79.1$_{\pm2.5}$  & \underline{91.9}$_{\pm0.7}$ 
                 & 61.5$_{\pm2.8}$ & 80.6$_{\pm1.2}$ \\
    GALS         & $\times$     & $\times$     
                 & 75.4$_{\pm2.2}$  & 89.0$_{\pm0.5}$
                 & 55.0$_{\pm5.5}$ & 79.7$_{\pm0.4}$ \\
    JTT          & $\times$     & $\checkmark$ 
                 & 86.4$_{\pm1.0}$ & 89.5$_{\pm0.5}$ 
                 & 61.3$_{\pm 5.5}$ & 79.7$_{\pm 3.0}$ \\
    CnC          & $\times$     & $\checkmark$ 
                 & \underline{86.5}$_{\pm 5.9}$  & 91.0$_{\pm 0.5}$ 
                 & 62.1$_{\pm 0.9}$ & 81.9$_{\pm 1.5}$ \\
    SupER (Ours)        & $\times$     & $\times$     
                 & 84.4$_{\pm2.3}$  & 87.3$_{\pm0.6}$ 
                 & \textbf{\underline{79.7}}$_{\pm1.7}$ & \textbf{\underline{85.0}}$_{\pm1.4}$ \\
    \midrule
    UW           & $\checkmark$ & $\times$     
                 & 89.3$_{\pm1.5}$  & \textbf{94.5}$_{\pm0.9}$ 
                 & 56.4$_{\pm2.3}$ & 78.6$_{\pm0.8}$ \\
    IRM          & $\checkmark$ & $\times$     
                 & 76.2$_{\pm6.3}$  & 89.4$_{\pm0.9}$ 
                 & 57.0$_{\pm5.4}$ & 80.5$_{\pm5.0}$ \\
    GroupDRO     & $\checkmark$ & $\times$     
                 & 87.2$_{\pm1.3}$ & 93.2$_{\pm0.4}$
                 & 56.5$_{\pm 1.4}$ & 79.4$_{\pm 0.3}$ \\
    DFR          & $\checkmark$ & $\checkmark$ 
                 & \textbf{89.7}$_{\pm2.4}$ & 93.6$_{\pm0.6}$ 
                 & 48.2$_{\pm 0.4}$ & 76.4$_{\pm 0.2}$ \\
    \bottomrule
  \end{tabular*}
  }
\end{table}

\begin{table}[H]
  \captionsetup{skip=4pt}
  \setlength{\tabcolsep}{3pt}
  \centering
  {\fontsize{8pt}{9pt}\selectfont
  \caption{Worst group accuracy (\%) for the six Spawrious subsets.}
  \label{tab:sc_results_app_0}
  \begin{tabular*}{\textwidth}{@{\extracolsep{\fill}} l  c  c  c  c  c  c  c  c  c}
    \toprule
    \multirow{2}{*}{Method}
      & \multirow{2}{*}{\shortstack{Group\\ Info?}}
      & \multirow{2}{*}{\shortstack{Train\\Twice?}}
      & \multicolumn{3}{c}{One--To--One}
      & \multicolumn{3}{c}{Many--To--Many}
      & \multirow{2}{*}{Average} \\
    \cmidrule(lr){4-6} \cmidrule(lr){7-9}
      &  & 
      & Easy & Medium & Hard
      & Easy & Medium & Hard
      &  \\
    \midrule
    ERM       & $\times$     & $\times$ 
              & 78.4$_{\pm 1.8}$ & 63.4$_{\pm 2.3}$ & 71.1$_{\pm 3.7}$
              & 72.9$_{\pm 1.3}$ & 52.7$_{\pm 2.9}$ & 50.7$_{\pm 1.0}$ 
              & 64.9$_{\pm 11.3}$ \\
    CVaR DRO  & $\times$     & $\times$ 
              & 81.7$_{\pm 0.5}$ & 66.4$_{\pm 1.4}$ & 61.2$_{\pm 1.6}$
              & 69.7$_{\pm 0.8}$ & 50.3$_{\pm 3.9}$ & 45.9$_{\pm 0.2}$ 
              & 62.5$_{\pm 13.1}$ \\
    LfF       & $\times$     & $\times$ 
              & 74.6$_{\pm 7.7}$ & -- & 62.9$_{\pm 3.6}$
              & 72.7$_{\pm 3.5}$ & 50.0$_{\pm 4.0}$ & 48.6$_{\pm 3.7}$ 
              & -- \\
    GALS      & $\times$     & $\times$ 
              & 89.1$_{\pm 1.9}$ & 60.0$_{\pm 5.4}$ & 81.0$_{\pm 3.0}$
              & 74.0$_{\pm 4.8}$ & 44.9$_{\pm 0.3}$ & 46.9$_{\pm 2.4}$ 
              & 66.0$_{\pm 18.3}$ \\
    JTT       & $\times$     & $\checkmark$      
              & 80.9$_{\pm 2.1}$ & -- & 59.7$_{\pm 4.9}$
              & 71.2$_{\pm 2.0}$ & 49.7$_{\pm 3.5}$ & 45.2$_{\pm 1.8}$ 
              & -- \\
    CnC       & $\times$ & $\checkmark$      
              & \underline{\textbf{90.0}}$_{\pm 1.4}$ & 73.5$_{\pm 4.6}$ & 81.3$_{\pm 3.1}$
              & 82.8$_{\pm 2.1}$ & 62.5$_{\pm 5.2}$ & 78.7$_{\pm 4.9}$ 
              & 78.1$_{\pm 9.4}$ \\
    SupER (Ours)     & $\times$     & $\times$ 
              & 82.7$_{\pm 2.0}$ & \underline{\textbf{80.3}}$_{\pm 4.6}$ & \underline{\textbf{83.8}}$_{\pm 3.4}$
              & \underline{\textbf{87.4}}$_{\pm 1.3}$ & \underline{\textbf{83.4}}$_{\pm 2.3}$ & \underline{\textbf{79.9}}$_{\pm 4.7}$ 
              & \underline{\textbf{82.9}}$_{\pm 2.7}$ \\
    \midrule
    UW        & $\checkmark$  & $\times$
              & 87.4$_{\pm 1.1}$ & 67.9$_{\pm 2.1}$ & 75.9$_{\pm 2.9}$
              & 72.9$_{\pm 1.3}$ & 52.7$_{\pm 2.9}$ & 50.7$_{\pm 1.0}$ 
              & 67.9$_{\pm 14.1}$ \\
    IRM       & $\checkmark$  & $\times$
              & 78.4$_{\pm 1.0}$ & 64.5$_{\pm 3.2}$ & 64.9$_{\pm 2.2}$
              & 77.9$_{\pm 3.7}$ & 57.1$_{\pm 2.9}$ & 50.7$_{\pm 1.1}$ 
              & 65.6$_{\pm 11.1}$ \\
    GroupDRO  & $\checkmark$  & $\times$
              & 86.7$_{\pm 1.2}$ & 67.2$_{\pm 0.7}$ & 76.4$_{\pm 2.2}$
              & 74.3$_{\pm 0.9}$ & 55.7$_{\pm 1.4}$ & 49.9$_{\pm 0.8}$ 
              & 68.3$_{\pm 13.7}$ \\
    DFR       & $\checkmark$  & $\checkmark$    
              & 79.1$_{\pm 5.2}$ & 64.3$_{\pm 1.9}$ & 70.0$_{\pm 1.9}$
              & 76.4$_{\pm 1.9}$ & 58.7$_{\pm 2.2}$ & 54.1$_{\pm 2.2}$ 
              & 67.1$_{\pm 9.9}$ \\
    \bottomrule
  \end{tabular*}
  }
\end{table}

\begin{table}[H]
\captionsetup{skip=4pt}
  \setlength{\tabcolsep}{3pt}
  \centering
  {\fontsize{8pt}{9pt}\selectfont
  \caption{Average accuracy (\%) for the six Spawrious subsets.}
  \label{tab:sc_results_app}
  \begin{tabular*}{\textwidth}{@{\extracolsep{\fill}} l  c  c  c  c  c  c  c  c  c}
    \toprule
    \multirow{2}{*}{Method}
      & \multirow{2}{*}{\shortstack{Group\\ Info?}}
      & \multirow{2}{*}{\shortstack{Train\\Twice?}}
      & \multicolumn{3}{c}{One--To--One}
      & \multicolumn{3}{c}{Many--To--Many}
      & \multirow{2}{*}{Average} \\
    \cmidrule(lr){4-6} \cmidrule(lr){7-9}
      &  & 
      & Easy & Medium & Hard
      & Easy & Medium & Hard
      &  \\
    \midrule
    ERM       & $\times$     & $\times$ 
              & 85.5$_{\pm 2.6}$ & 76.7$_{\pm 1.3}$ & 82.0$_{\pm 1.0}$
              & 89.5$_{\pm 0.6}$ & 74.5$_{\pm 1.4}$ & 70.7$_{\pm 2.1}$ 
              & 79.8$_{\pm 7.1}$ \\
    CVaR DRO  & $\times$     & $\times$ 
              & 89.4$_{\pm 0.1}$ & 86.0$_{\pm 3.7}$ & 80.7$_{\pm 0.6}$
              & 88.5$_{\pm 0.6}$ & 74.0$_{\pm 1.0}$ & 67.7$_{\pm 0.6}$ 
              & 81.0$_{\pm 8.7}$ \\
    LfF       & $\times$     & $\times$ 
              & 84.1$_{\pm 1.5}$ & -- & 76.9$_{\pm 1.2}$
              & 89.6$_{\pm 0.8}$ & 73.8$_{\pm 2.6}$ & 69.1$_{\pm 1.1}$ 
              & -- \\
    GALS      & $\times$     & $\times$ 
              & 93.5$_{\pm 0.9}$ & 86.6$_{\pm 0.9}$ & 90.0$_{\pm 0.4}$
              & 87.8$_{\pm 0.2}$ & 74.0$_{\pm 0.3}$ & 69.8$_{\pm 1.9}$ 
              & 83.6$_{\pm 9.5}$ \\
    JTT       & $\times$     & $\checkmark$      
              & 86.1$_{\pm 1.3}$ & -- & 77.5$_{\pm 1.7}$
              & 89.2$_{\pm 0.5}$ & 72.8$_{\pm 0.9}$ & 66.6$_{\pm 0.8}$ 
              & -- \\
    CnC       & $\times$ & $\checkmark$      
              & \underline{\textbf{94.4}}$_{\pm 1.1}$ & 87.8$_{\pm 2.5}$ & 89.6$_{\pm 0.9}$
              & 92.6$_{\pm 1.0}$ & 80.8$_{\pm 4.0}$ & 88.8$_{\pm 1.2}$ 
              & 89.0$_{\pm 4.7}$ \\
    SupER (Ours)     & $\times$     & $\times$ 
              & 90.9$_{\pm 0.5}$ & \underline{\textbf{90.1}}$_{\pm 3.2}$ & \underline{\textbf{90.5}}$_{\pm 2.0}$
              & \underline{\textbf{94.9}}$_{\pm 0.9}$ & \underline{\textbf{91.6}}$_{\pm 1.8}$ & \underline{\textbf{91.4}}$_{\pm 1.5}$ 
              & \underline{\textbf{91.6}}$_{\pm 1.7}$ \\
    \midrule
    UW        & $\checkmark$  & $\times$
              & 93.5$_{\pm 0.3}$ & 82.6$_{\pm 0.7}$ & 86.5$_{\pm 0.6}$
              & 89.5$_{\pm 0.6}$ & 74.5$_{\pm 1.4}$ & 70.7$_{\pm 2.1}$ 
              & 82.9$_{\pm 8.8}$ \\
    IRM       & $\checkmark$  & $\times$
              & 87.3$_{\pm 0.3}$ & 76.9$_{\pm 0.4}$ & 82.7$_{\pm 0.4}$
              & 90.9$_{\pm 0.9}$ & 76.7$_{\pm 2.6}$ & 71.2$_{\pm 1.0}$ 
              & 80.9$_{\pm 7.4}$ \\
    GroupDRO  & $\checkmark$  & $\times$
              & 92.7$_{\pm 0.3}$ & 89.5$_{\pm 0.3}$ & 86.5$_{\pm 1.5}$
              & 89.4$_{\pm 0.6}$ & 77.3$_{\pm 0.5}$ & 68.4$_{\pm 1.7}$ 
              & 84.0$_{\pm 9.3}$ \\
    DFR       & $\checkmark$  & $\checkmark$    
              & 87.5$_{\pm 3.3}$ & 80.9$_{\pm 1.1}$ & 79.4$_{\pm 1.3}$
              & 89.4$_{\pm 0.4}$ & 75.1$_{\pm 0.1}$ & 72.4$_{\pm 1.9}$ 
              & 80.8$_{\pm 6.7}$ \\
    \bottomrule
  \end{tabular*}
  }
\end{table}

\begin{table}[H]
\captionsetup{skip=4pt}
  \setlength{\tabcolsep}{3pt}
  \centering
  {\fontsize{8pt}{9pt}\selectfont
  \caption{Worst group accuracy (\%) for the four MetaShift subsets.}
  \label{tab:ms_results_app0}
  \begin{tabular*}{\textwidth}{@{\extracolsep{\fill}} l  c  c  c  c  c  c  c}
    \toprule
    \multirow{2}{*}{Method}
      & \multirow{2}{*}{\shortstack{Group\\Info?}}
      & \multirow{2}{*}{\shortstack{Train\\Twice?}}
      & \multicolumn{4}{c}{MetaShift Subsets}
      & \multirow{2}{*}{Average} \\
    \cmidrule(lr){4-7}
      &  & 
      & (a) $d=0.44$ & (b) $d=0.71$ & (c) $d=1.12$ & (d) $d=1.43$ 
      &  \\
    \midrule
    ERM       & $\times$     & $\times$ 
              & 78.8$_{\pm 1.0}$ & 75.8$_{\pm 0.8}$ & 61.9$_{\pm 5.9}$ & 52.6$_{\pm 2.6}$ 
              & 67.3$_{\pm 12.2}$ \\
    CVaR DRO  & $\times$     & $\times$ 
              & 77.8$_{\pm 2.5}$ & 72.5$_{\pm 2.8}$ & 65.1$_{\pm 0.2}$ & 54.7$_{\pm 3.2}$ 
              & 67.5$_{\pm 10.0}$ \\
    LfF       & $\times$     & $\times$ 
              & 77.2$_{\pm 1.7}$ & 73.9$_{\pm 0.6}$ & 69.5$_{\pm 1.0}$ & 59.5$_{\pm 3.1}$ 
              & 70.0$_{\pm 7.7}$ \\
    GALS      & $\times$     & $\times$ 
              & 74.8$_{\pm 3.9}$ & 68.8$_{\pm 2.0}$ & 70.6$_{\pm 2.2}$ & 50.0$_{\pm 0.9}$ 
              & 66.0$_{\pm 11.0}$ \\
    JTT       & $\times$     & $\checkmark$      
              & 76.7$_{\pm 2.3}$ & 73.2$_{\pm 0.8}$ & 67.1$_{\pm 4.6}$ & 53.0$_{\pm 1.6}$ 
              & 67.5$_{\pm 10.4}$ \\
    CnC       & $\times$ & $\checkmark$      
              & \underline{\textbf{81.1}}$_{\pm 1.4}$ & 71.4$_{\pm 2.4}$ & 65.4$_{\pm 6.8}$
              & 49.6$_{\pm 1.6}$ & 66.9$_{\pm 13.2}$ \\
    SupER (Ours)     & $\times$     & $\times$ 
              & 79.8$_{\pm 3.6}$ & \underline{\textbf{78.4}}$_{\pm 1.9}$ & \underline{\textbf{77.6}}$_{\pm 2.1}$ & \underline{\textbf{71.4}}$_{\pm 2.1}$ 
              & \underline{\textbf{76.8}}$_{\pm 3.7}$ \\
    \bottomrule
  \end{tabular*}
  }
\end{table}

\vspace{-10pt}

\begin{table}[H]
\captionsetup{skip=4pt}
  \setlength{\tabcolsep}{3pt}
  \centering
  {\fontsize{8pt}{9pt}\selectfont
  \caption{Average accuracy (\%) for the four MetaShift subsets.}
  \label{tab:ms_results_app}
  \begin{tabular*}{\textwidth}{@{\extracolsep{\fill}} l  c  c  c  c  c  c  c}
    \toprule
    \multirow{2}{*}{Method}
      & \multirow{2}{*}{\shortstack{Group\\Info?}}
      & \multirow{2}{*}{\shortstack{Train\\Twice?}}
      & \multicolumn{4}{c}{MetaShift Subsets}
      & \multirow{2}{*}{Average} \\
    \cmidrule(lr){4-7}
      &  & 
      & (a) $d=0.44$ & (b) $d=0.71$ & (c) $d=1.12$ & (d) $d=1.43$ 
      &  \\
    \midrule
    ERM       & $\times$     & $\times$ 
              & 80.5$_{\pm 0.8}$ & 78.0$_{\pm 0.2}$ & 73.6$_{\pm 0.5}$ & 69.2$_{\pm 1.3}$ 
              & 75.3$_{\pm 5.0}$ \\
    CVaR DRO  & $\times$     & $\times$ 
              & 80.7$_{\pm 1.2}$ & 78.2$_{\pm 0.3}$ & 74.6$_{\pm 0.5}$ & 69.8$_{\pm 2.1}$ 
              & 75.8$_{\pm 4.7}$ \\
    LfF       & $\times$     & $\times$ 
              & 79.2$_{\pm 0.9}$ & 77.2$_{\pm 1.4}$ & 74.8$_{\pm 1.1}$ & 69.1$_{\pm 0.7}$ 
              & 75.1$_{\pm 4.4}$ \\
    GALS      & $\times$     & $\times$ 
              & 80.5$_{\pm 1.8}$ & 77.4$_{\pm 1.2}$ & 78.3$_{\pm 0.7}$ & 69.1$_{\pm 1.3}$ 
              & 76.3$_{\pm 5.0}$ \\
    JTT       & $\times$     & $\checkmark$      
              & 80.8$_{\pm 1.3}$ & 76.4$_{\pm 1.0}$ & 73.2$_{\pm 0.5}$ & 69.3$_{\pm 0.6}$ 
              & 74.9$_{\pm 4.9}$ \\
    CnC       & $\times$ & $\checkmark$      
              & \underline{\textbf{82.1}}$_{\pm 1.4}$ & 77.0$_{\pm 2.2}$ & 74.4$_{\pm 1.6}$
              & 66.7$_{\pm 1.4}$ & 75.1$_{\pm 6.4}$ \\
    SupER (Ours)     & $\times$     & $\times$ 
              & 81.7$_{\pm 1.9}$ & \underline{\textbf{80.5}}$_{\pm 1.4}$ & \underline{\textbf{79.2}}$_{\pm 1.9}$ & \underline{\textbf{76.6}}$_{\pm 1.4}$ 
              & \underline{\textbf{79.5}}$_{\pm 2.2}$ \\
    \bottomrule
  \end{tabular*}
  }
\end{table}

\vspace{-10pt}

\begin{table}[H]
\captionsetup{skip=4pt}
  \setlength{\tabcolsep}{3pt}
  \centering
  {\fontsize{8pt}{9pt}\selectfont
  \caption{Worst and average group accuracy (\%) for Spuco Dogs.}
  \label{tab:wb_results_app}
  \begin{tabular*}{0.6\textwidth}{@{\extracolsep{\fill}} l  c  c  c  c}
    \toprule
    \multirow{2}{*}{Method}
      & \multirow{2}{*}{\shortstack{Group\\Info?}}
      & \multirow{2}{*}{\shortstack{Train\\Twice?}}
      & \multicolumn{2}{c}{Spuco Dogs} \\
    \cmidrule(lr){4-5}
      &  &  & Worst & Avg \\
    \midrule
    ERM          & $\times$     & $\times$     
                 & 54.5$_{\pm 1.3}$ & 77.4$_{\pm 1.6}$ \\
    CVaR\,DRO    & $\times$     & $\times$     
                 & 56.3$_{\pm 3.1}$ & 78.5$_{\pm 2.2}$ \\
    LfF          & $\times$     & $\times$     
                 & 52.6$_{\pm 2.5}$ & 77.1$_{\pm 1.9}$ \\
    GALS         & $\times$     & $\times$     
                 & -- & -- \\
    JTT          & $\times$     & $\checkmark$ 
                 & 50.4$_{\pm 0.2}$ & 77.9$_{\pm 0.1}$ \\
    CnC          & $\times$ & $\checkmark$ 
                 & 65.6$_{\pm 0.7}$ & \underline{82.0}$_{\pm 0.5}$ \\
    SupER (Ours)        & $\times$     & $\times$     
                 & \underline{69.7}$_{\pm 4.4}$ & 76.0$_{\pm 2.3}$ \\
    \midrule
    UW           & $\checkmark$ & $\times$     
                 & \textbf{84.7}$_{\pm 2.0}$ & 87.4$_{\pm 0.5}$ \\
    IRM          & $\checkmark$ & $\times$     
                 & 50.0$_{\pm 5.5}$ & 75.2$_{\pm 5.7}$ \\
    GroupDRO     & $\checkmark$ & $\times$     
                 & 83.8$_{\pm 0.4}$ & \textbf{87.6}$_{\pm 0.5}$ \\
    DFR          & $\checkmark$ & $\checkmark$ 
                 & 71.3$_{\pm 4.4}$ & 83.3$_{\pm 2.8}$ \\
    \bottomrule
  \end{tabular*}
  }
\end{table}

\textbf{Variance of accuracy across groups.} Tables \ref{tab:variance_wb_sc_app}, \ref{tab:variance_metashift}, and \ref{tab:variance_spawrious} summarize the variance of accuracy across groups for all datasets and selected baseline methods. \textbf{Bold} indicates the smallest across all selected baselines; \underline{Underlined} indicates the smallest among methods without group information. ;  ``--'' indicates omitted result due to consistently subpar or unstable performance, even after comprehensive hyperparameter tuning using the original codebase.

\begin{table}[H]
\captionsetup{skip=4pt}
  \setlength{\tabcolsep}{3pt}
  \centering
  {\fontsize{8pt}{9pt}\selectfont
  \caption{Variance of accuracy across groups (\%) for Waterbirds-95\%, Waterbirds-100\%, and SpuCo Dogs.}
  \label{tab:variance_wb_sc_app}
  \begin{tabular*}{0.6\textwidth}{@{\extracolsep{\fill}} l  c  c  c}
    \toprule
    Method      & Waterbirds-95\% & Waterbirds-100\% & SpuCo Dogs \\
    \midrule
      ERM         & 245.9   & 778.1 & 603.9\\
      CVaR\,DRO   & 154.6   & 528.8 & 558.5\\
      LfF         & 89.2   & 442.1 & 582.9\\
      GALS        & 126.7   & 516.5 & --\\
      JTT         & 6.1   & 405.0  & 621.4\\
      CnC         & 16.3   & 347.6 & 261.7\\
      SupER (Ours) & \underline{\textbf{6.0}} & \underline{\textbf{16.0}} & \underline{28.4}\\
      \midrule
      UW          & 12.7   & 536.2   & \textbf{6.5}\\
      IRM         & 127.2   & 479.8  & 776.7\\
      GroupDRO    & 28.0   & 495.1 & 10.0\\
      DFR         & 14.2   & 573.0 & 282.9\\
    \bottomrule
  \end{tabular*}
  }
\end{table}

% Table 6b: MetaShift
\begin{table}[H]
\captionsetup{skip=4pt}
  \setlength{\tabcolsep}{3pt}
  \centering
  {\fontsize{8pt}{9pt}\selectfont
  \caption{Variance of accuracy across groups (\%) for the four MetaShift subsets.}
  \label{tab:variance_metashift}
  \begin{tabular*}{0.7\textwidth}{@{\extracolsep{\fill}} l  c  c  c  c}
    \toprule
    Method       & (a) $d=0.44$ & (b) $d=0.71$ & (c) $d=1.12$ & (d) $d=1.43$ \\
    \midrule
    ERM          & 10.3        & 14.2        & 411.8        & 722.1        \\
    CVaR\,DRO    & 21.3        & 97.5        & 237.7        & 599.7        \\
    LfF          & 15.3        & 30.3        & 73.3        & 258.8        \\
    GALS         & 82.1        & 197.7        & 157.7        & 955.6        \\
    JTT          & 31.2        & 24.9        & 128.6        & 699.8        \\
    CnC          & \underline{\textbf{2.3}}        & 71.5        & 262.5        & 769.6        \\
SupER &9.3&\underline{\textbf{13.1}}&\underline{\textbf{7.7}}&\underline{\textbf{49.4}}\\
    \bottomrule
  \end{tabular*}
  }
\end{table}

% Table 6c: Spawrious
\begin{table}[H]
\captionsetup{skip=4pt}
  \setlength{\tabcolsep}{3pt}
  \centering
  {\fontsize{8pt}{9pt}\selectfont
  \caption{Variance of accuracy across groups (\%) for the six Spawrious subsets.}
  \label{tab:variance_spawrious}
  \begin{tabular*}{\textwidth}{@{\extracolsep{\fill}} l  c  c  c  c  c  c}
    \toprule
    Method       & O2O-Easy & O2O-Medium & O2O-Hard & M2M-Easy & M2M-Medium & M2M-Hard \\
    \midrule
    ERM          & 50.9 & 109.8 & 101.1 & 92.6 & 246.3 & 373.8 \\
    CVaR\,DRO    & 63.3 & 261.0 & 241.4 & 109.4 & 323.0 & 469.0 \\
    LfF          & 88.1 & -- & 254.2 & 80.9 & 290.2 & 413.8 \\
    GALS         & \underline{\textbf{12.6}} & 490.8 & 61.3 & 161.6 & 563.8 & 584.9 \\
    JTT          & 31.4 & -- & 310.2 & 108.3 & 293.2 & 473.2 \\
    CnC          & 19.2 & 169.2 & 62.9 & 48.1 & 129.1 & \underline{\textbf{47.1}} \\
    SupER (Ours) &64.7&\underline{\textbf{92.1}}&\underline{\textbf{41.2}}&\underline{\textbf{22.8}}&\underline{\textbf{33.8}}&58.0\\
    \midrule
    UW           & 29.3 & 109.8 & 99.6 & 92.6 & 246.3 & 373.8 \\
    IRM          & 76.5 & 107.8 & 183.8 & 70.0 & 291.2 & 383.4 \\
    GroupDRO     & 33.5 & 221.7 & 85.5 & 79.0 & 190.7 & 397.1 \\
    DFR          & 70.3 & 359.1 & 149.3 & 68.0 & 293.1 & 327.5 \\
    \bottomrule
  \end{tabular*}
  }
\end{table}

\textbf{CLIP guidance.} 
The goal of SupER is fundamentally different from extracting or replicating CLIP’s features.
Instead, CLIP only provides superclass guidance and does not contribute any information useful for distinguishing class labels, since the superclass is shared across different class labels. 
Moreover, in Table~\ref{tab:clip_vs_super} we report the performance of directly using CLIP for prediction compared to SupER. 
Directly applying CLIP leads to a noticeable drop in accuracy, which suggests that CLIP itself may also rely on spurious correlations. 
Therefore, using CLIP as superclass guidance can both give SupER enough autonomy to learn features on its own, and avoid the spurious correlations that CLIP might exploit for fine-grained class prediction.

\begin{table}[H]
\centering
{\fontsize{8pt}{9pt}\selectfont
\caption{Comparison of worst group accuracy (\%) between CLIP and SupER on Waterbirds. 
CLIP (zero-shot) means directly using the pretrained CLIP model for classification. 
CLIP (fine-tuned) denotes standard fine-tuning of CLIP on the downstream dataset.}
\label{tab:clip_vs_super}
\begin{tabular}{lcc}
\toprule
Method & Waterbirds-95\% & Waterbirds-100\% \\
\midrule
CLIP (zero-shot) & 41.6 & 47.9 \\
CLIP (fine-tuned) & 70.2 & 48.8 \\
SupER             & 84.4 & 79.7 \\
\bottomrule
\end{tabular}
}
\end{table}

\subsection{Visualization Results}
\label{app:AVR}

\begin{figure}[H]
\vskip 0.2in
\centering
\includegraphics[width=0.5\linewidth]{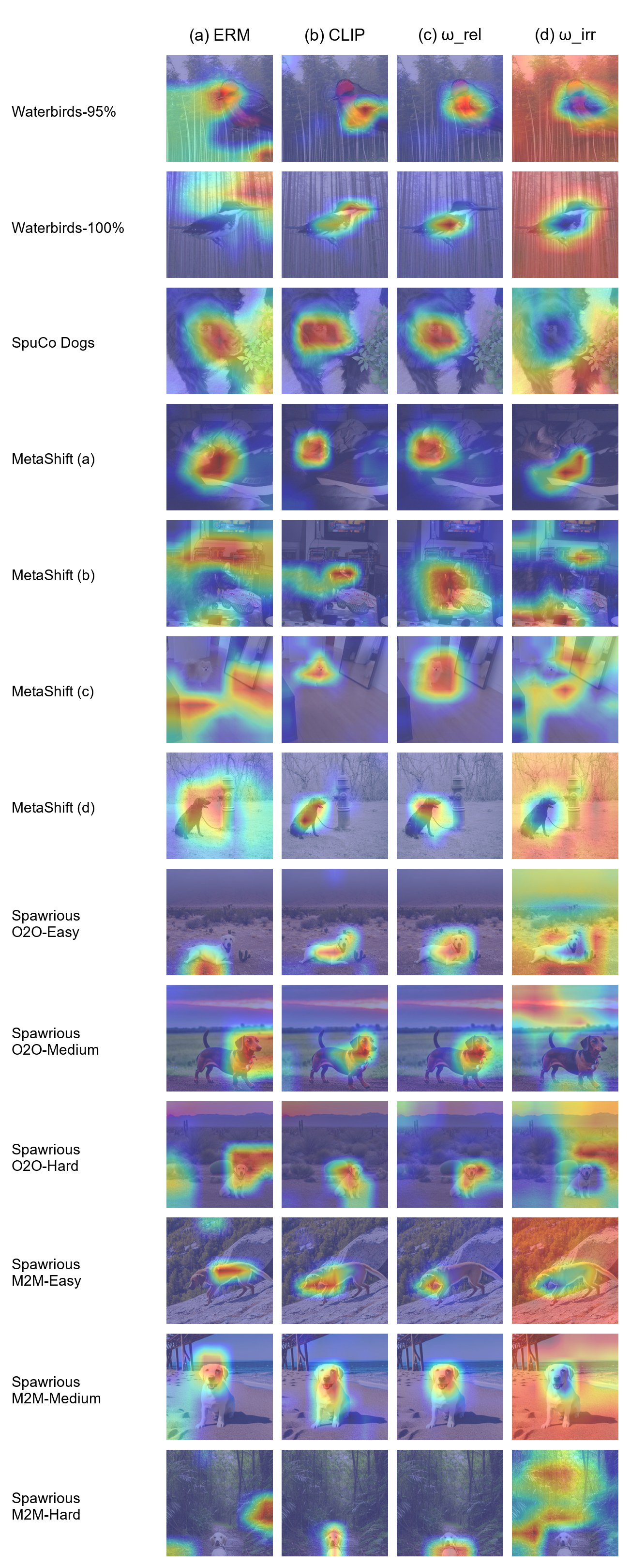}
\caption{Visualization of GradCAM maps across all datasets to assess feature disentanglement.
Each row corresponds to one representative sample per dataset subset.
Columns (a)–(d) show: GradCAM maps from ERM, CLIP, SupER's classifier $\omega_{\mathrm{rel}}$ (superclass-relevant), and $\omega_{\mathrm{irr}}$ (superclass-irrelevant).}
\label{fig:gradcam_vis_1}
\vskip -0.2in
\end{figure}

\textbf{SupER achieves effective disentanglement of superclass-relevant and irrelevant features.}
Figures~\ref{fig:gradcam_vis_1} illustrates gradient-based attention visualizations from one representative samples per subset across all datasets. For each sample, we present the GradCAM attribution maps from the ERM baseline, CLIP, SupER's $\omega_{\mathrm{rel}}$ and $\omega_{\mathrm{irr}}$. The results show that SupER consistently succeeds in separating superclass-relevant and superclass-irrelevant features by leveraging guidance from CLIP across diverse datasets.

\begin{figure}[H]
\centering
\includegraphics[width=0.5\linewidth]{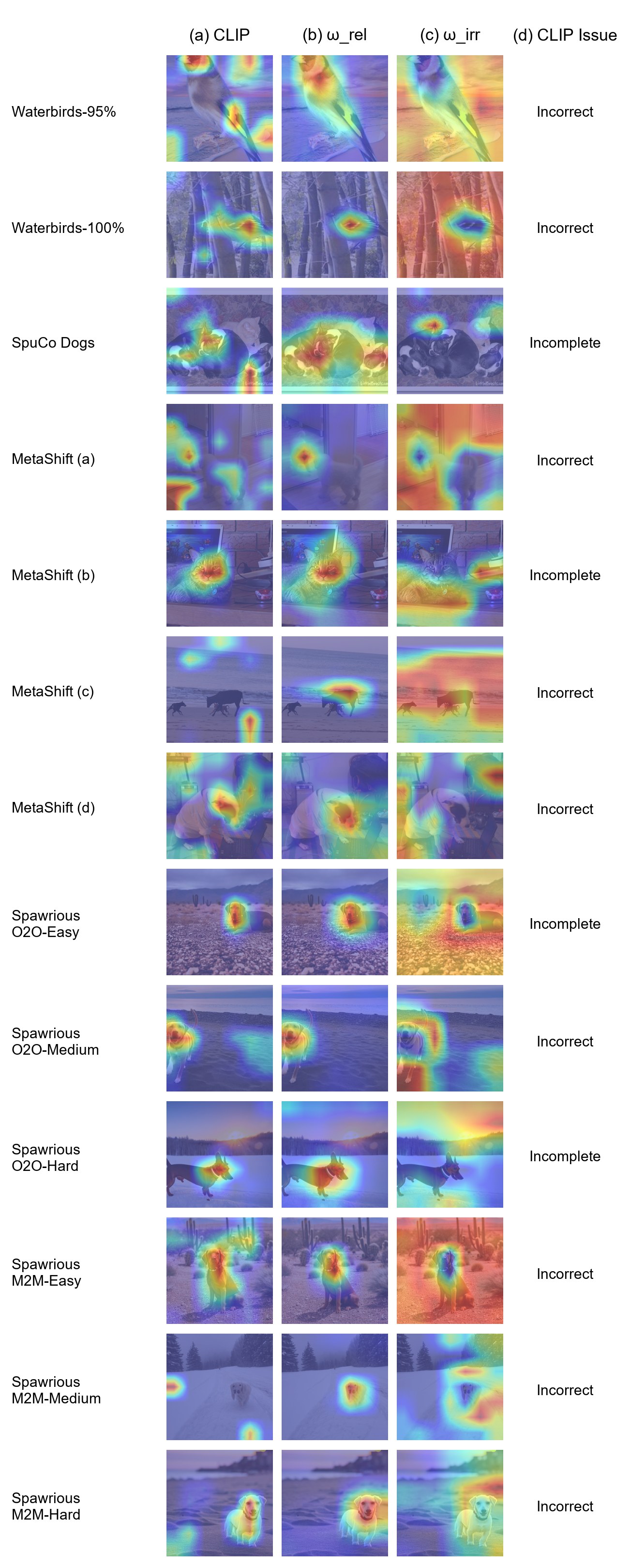}
\caption{Visualization of GradCAM maps highlighting CLIP's internal bias and SupER's correction.
Each row presents one representative sample per dataset subset.
Columns (a)–(d) show: GradCAM maps from CLIP, SupER's classifier $\omega_{\mathrm{rel}}$ (superclass-relevant), $\omega_{\mathrm{irr}}$ (superclass-irrelevant), and an illustration of the primary CLIP bias.}
\label{fig:gradcam_vis_clipbias}
\vskip -0.2in
\end{figure}

\textbf{SupER can adjust internal biases in CLIP.}
Figure~\ref{fig:gradcam_vis_clipbias} illustrates gradient-based attention visualizations from one representative sample per subset across all datasets. Each sample includes the GradCAM attribution maps from CLIP, SupER's classifiers ($\omega_{\mathrm{rel}}$, $\omega_{\mathrm{irr}}$), and an illustration of the primary issue observed in CLIP's attention (e.g., focusing on incomplete or incorrect features). The results demonstrate that SupER, by emphasizing feature disentanglement, can effectively mitigate internal biases in CLIP's attention.

\subsection{Ablation Results}
\label{app:ablation}

In this section, we examine the contributions of the core components of SupER. We focus on (i) text prompts, (ii) the strength of feature disentanglement, and (iii) the degree of superclass guidance, as these constitute the primary design elements of the method. We also study the impact of $L_2$ regularization and the relative strength of the two classifiers.  
To better isolate the effect of each factor, we keep all other hyperparameters fixed during each ablation study.  
This includes adopting a consistent protocol for random-seed selection across repeated trials, while in Appendix~\ref{app:WGAAAV}, we do not enforce fixed random seeds across runs.

\textbf{Text prompt.}
We evaluate the impact of text prompt configurations across all datasets. Tables~\ref{tab:prompt_ablation_spawrious}, \ref{tab:prompt_ablation_metashift}, \ref{tab:prompt_ablation_waterbirds}, and \ref{tab:prompt_ablation_spucodogs} present the change in worst group accuracy relative to the reference setting for the Spawrious, MetaShift, Waterbirds, and SpuCo Dogs datasets, respectively. The exact text prompts used are listed in Table~\ref{tab:prompt_variants}. Overall, performance tends to degrade as the number of prompts increases and as the superclass becomes more abstract.

Beyond the above prompt ablations, we further study how SupER behaves when the superclass prompt is intentionally mismatched with our definition in Section~\ref{subsec:problem setup}. In Table~\ref{tab:prompt_mismatch}, on Waterbirds we intentionally replace the reference superclass prompt ``bird'' with ``waterfowl'' and ``songbird'' such that they only cover a strict subset of the intended superclass. We observe a consistent but relatively small drop in worst-group accuracy compared to the reference setting, likely because the CLIP attribution maps induced by ``waterfowl'' and ``songbird'' remain visually close to those induced by ``bird''.

We also explore whether superclasses can be extracted automatically from fine-grained class names using a large language model, which would further reduce manual prompt engineering.
Specifically, we use \emph{Qwen2.5-7B-Instruct}~\citep{qwen2025qwen25technicalreport} and query it with the following prompt:
\begin{quote}
{\small\ttfamily
We have fine-grained class names: \{fine-grained class label\}. Give ONE English word that is their shared superclass.
Return ONLY the single word, lowercase, no punctuation.
}
\end{quote}
The extracted superclasses are highly reasonable across datasets (e.g., ``bird'' for Waterbirds, ``dog'' for Spawrious, and ``animal'' for MetaShift), suggesting that LLM-based extraction is a practical way to define superclasses and may help avoid inadvertent human specification errors.

\begin{figure}[H]
  \centering

  \begin{minipage}[t]{0.48\linewidth}
  \centering
    \subfloat[\fontsize{8pt}{8pt}\selectfont Effect of $\beta$ on SpuCo Dogs relative to the $\beta=0.1$ setting.\label{fig:beta_spucodogs}]{\includegraphics[width=0.7\textwidth]{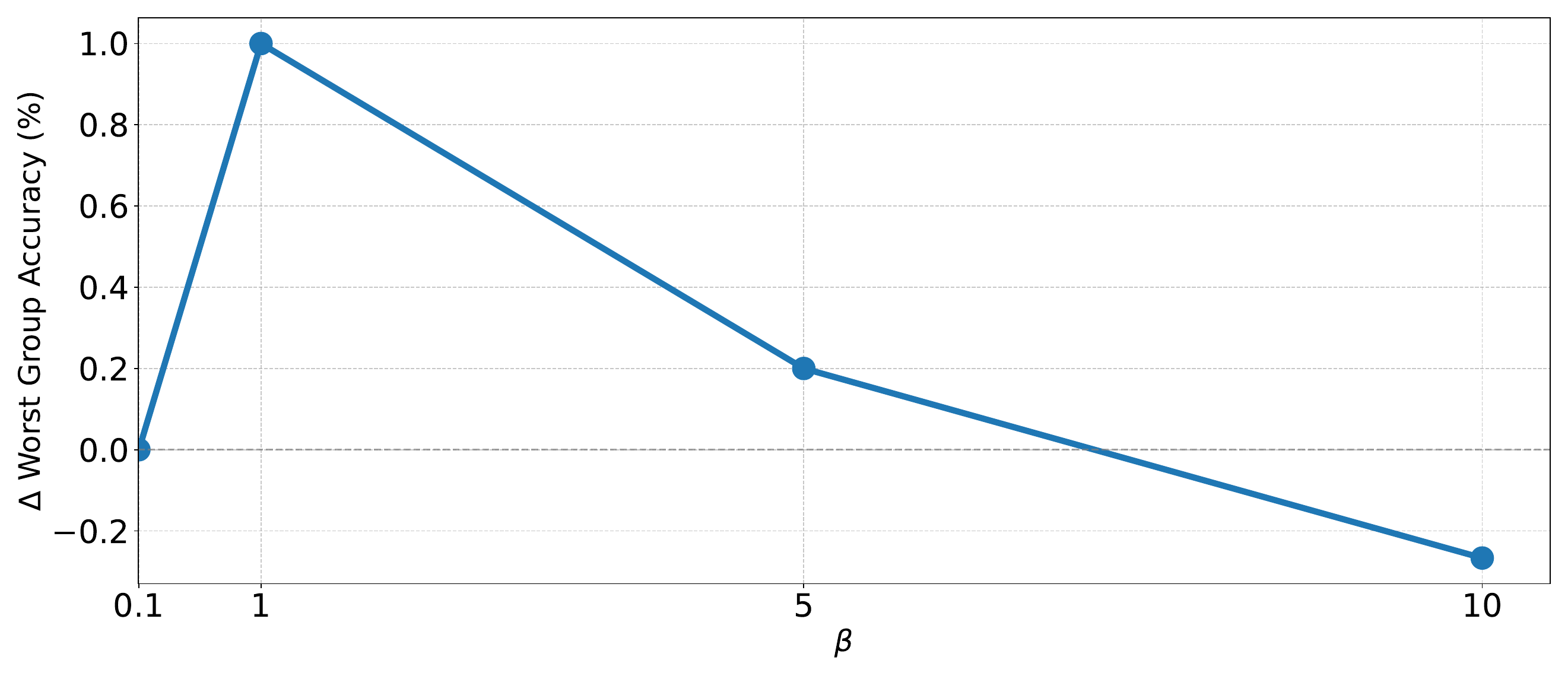}
    }
  \end{minipage}\hfill
  \begin{minipage}[t]{0.48\linewidth}
  \centering
    \subfloat[\fontsize{8pt}{8pt}\selectfont Effect of $\beta$ on Waterbirds relative to the $\beta=0.1$ setting.\label{fig:beta_waterbirds}]{\includegraphics[width=0.7\textwidth]{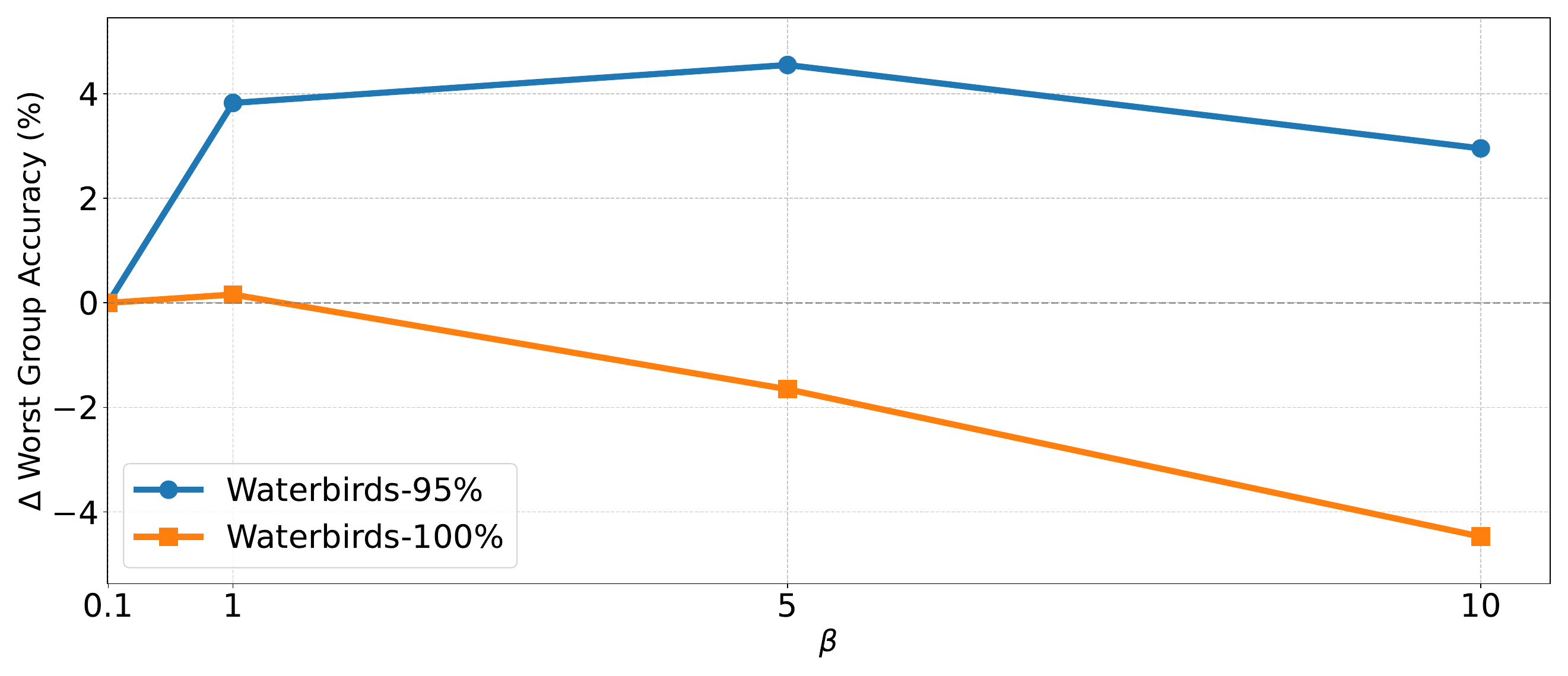}
    }
  \end{minipage}

  \vspace{1em} 

  \begin{minipage}[t]{0.48\linewidth}
  \centering
    \subfloat[\fontsize{8pt}{8pt}\selectfont Effect of $\beta$ on Spawrious  relative to the $\beta=0.1$ setting.\label{fig:beta_spawrious}]{\includegraphics[width=0.7\textwidth]{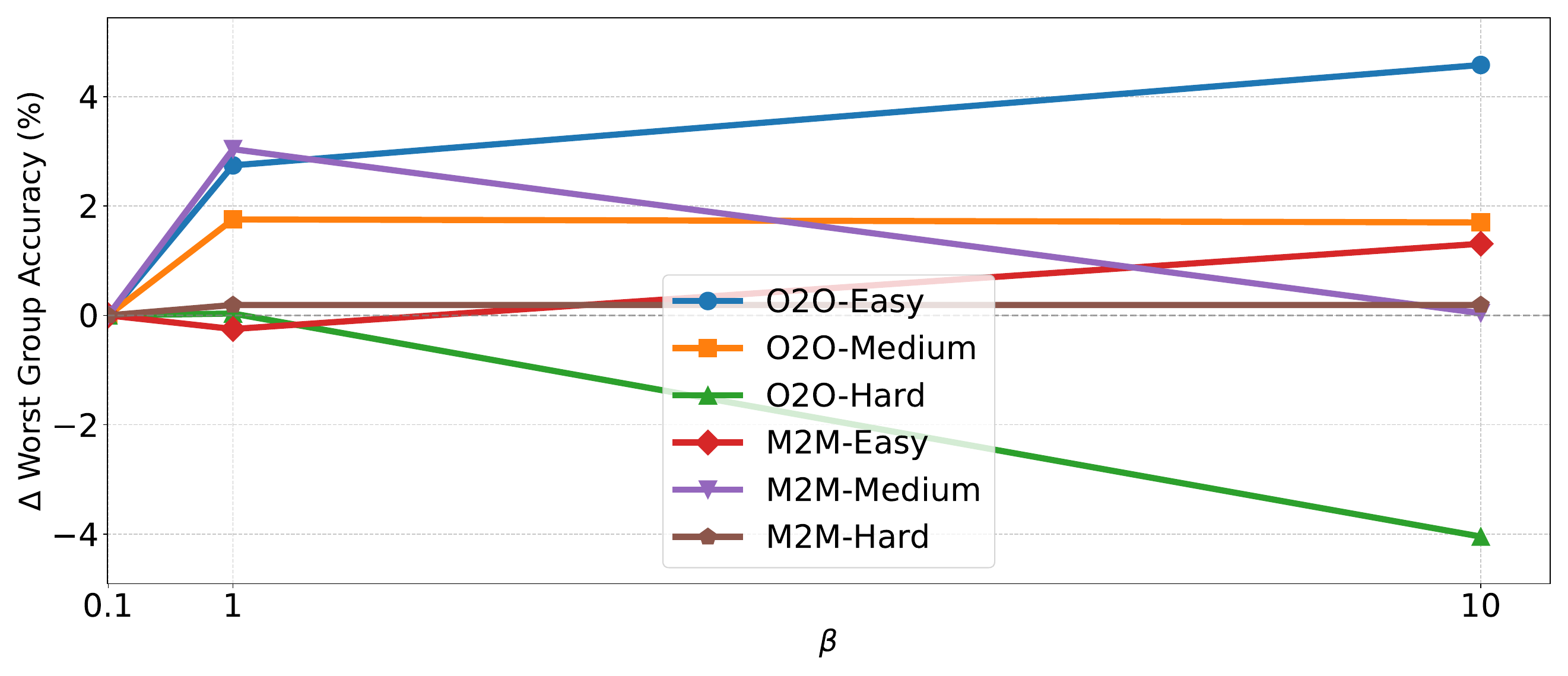}
    }
  \end{minipage}\hfill
  \begin{minipage}[t]{0.48\linewidth}
  \centering
    \subfloat[\fontsize{8pt}{8pt}\selectfont Effect of $\beta$ on MetaShift relative to the $\beta=0.1$ setting.\label{fig:beta_metashift}]{\includegraphics[width=0.7\textwidth]{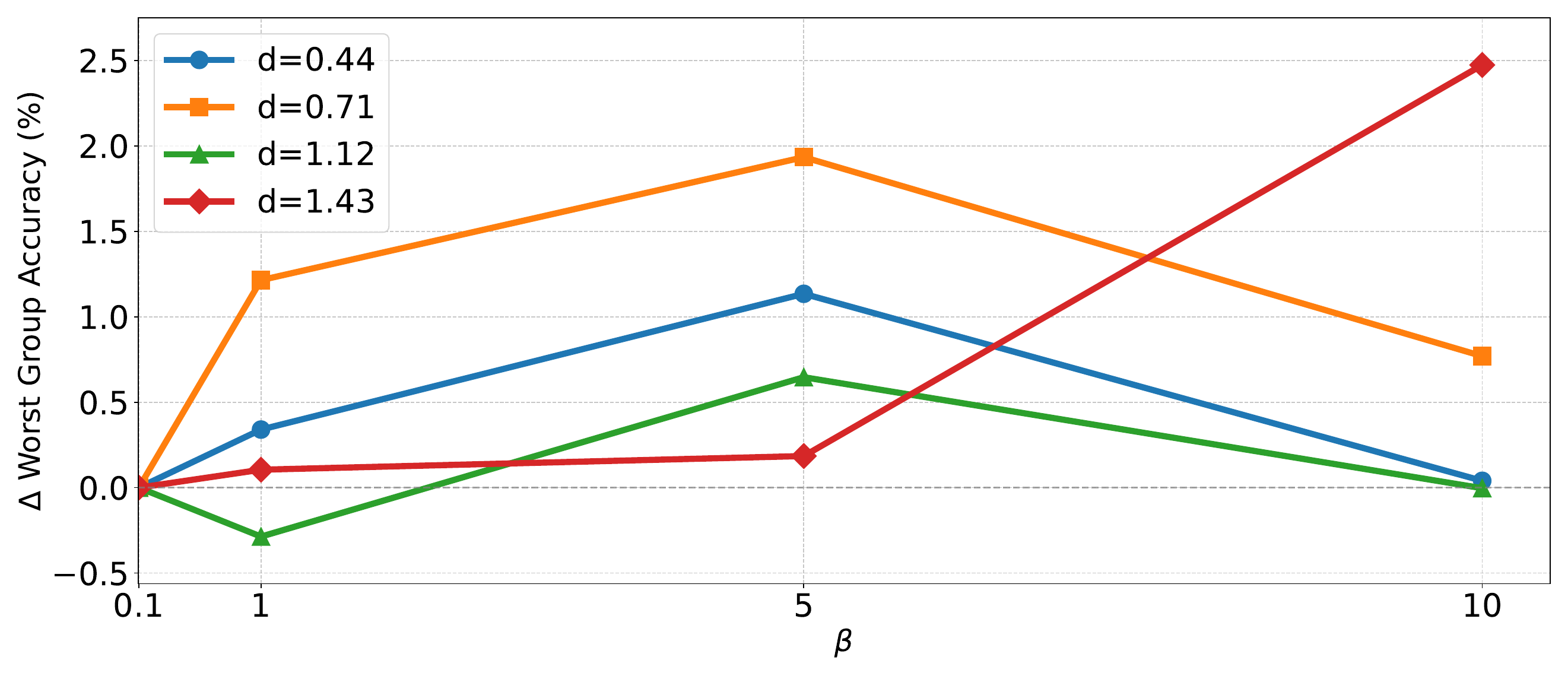}
    }
  \end{minipage}

  \caption{\fontsize{8pt}{8pt}\selectfont Ablation of feature disentanglement strength $\beta$ across all datasets.}
  \label{fig:ablation_beta_appendix}
\end{figure}

\textbf{Feature disentanglement strength.}
We evaluate the effect of varying the feature disentanglement coefficient $\beta$ across all datasets. Figure~\ref{fig:ablation_beta_appendix} shows the worst group accuracy as $\beta$ changes on selected datasets. Overall, both insufficient feature disentanglement (i.e., low $\beta$) and excessive disentanglement (i.e., overly large $\beta$) can lead to degraded model performance.

\begin{figure}[H]
\vskip 0.2in
\centering
\begin{minipage}[t]{0.48\linewidth}
\centering
  \subfloat[\fontsize{8pt}{8pt}\selectfont Effect of $\lambda_2$ on Waterbirds relative to the $\lambda_2=0$ setting.\label{fig:lambda_waterbirds}]{%
    \includegraphics[width=0.7\textwidth]{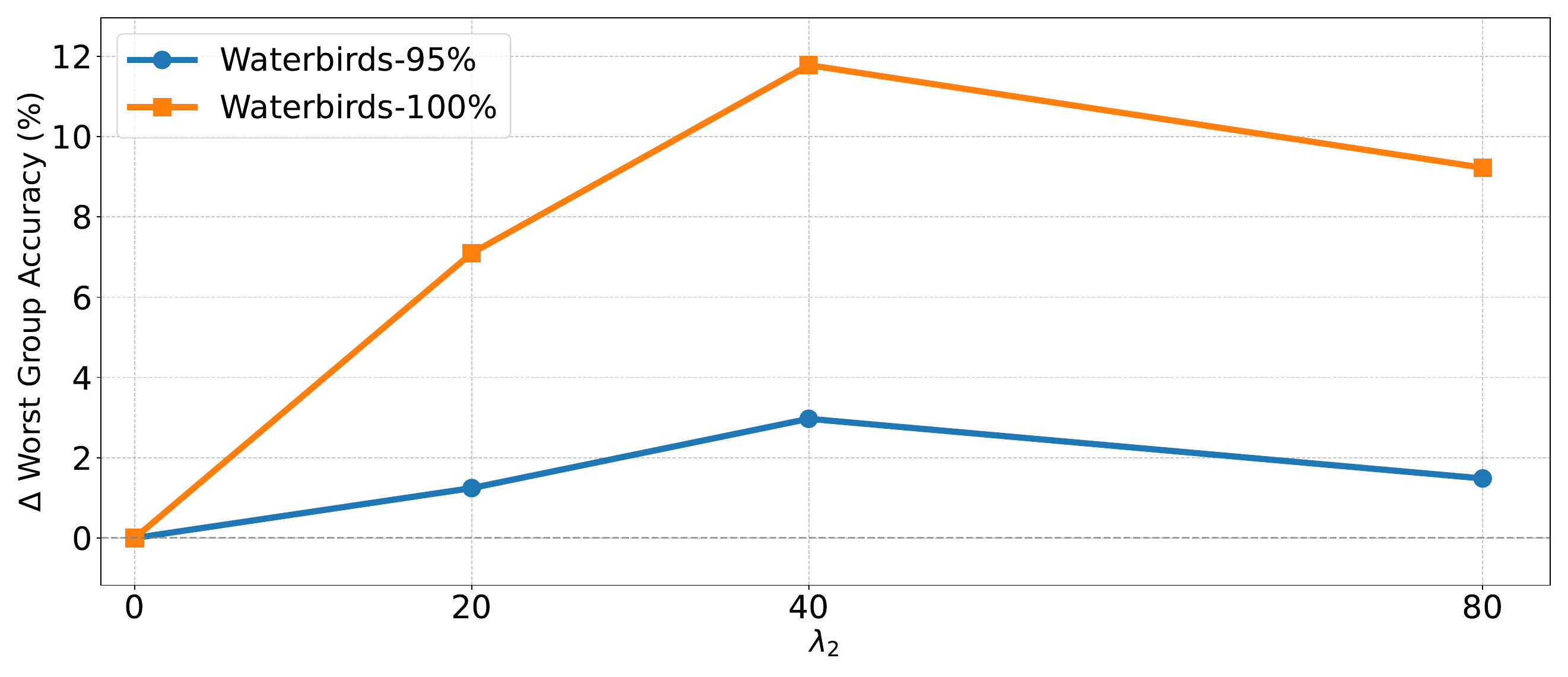}%
  }
\end{minipage}\hfill
\begin{minipage}[t]{0.48\linewidth}
\centering
  \subfloat[\fontsize{8pt}{8pt}\selectfont Effect of $\lambda_2$ on MetaShift relative to the $\lambda_2=0$ setting.\label{fig:lambda_metashift}]{%
    \includegraphics[width=0.7\textwidth]{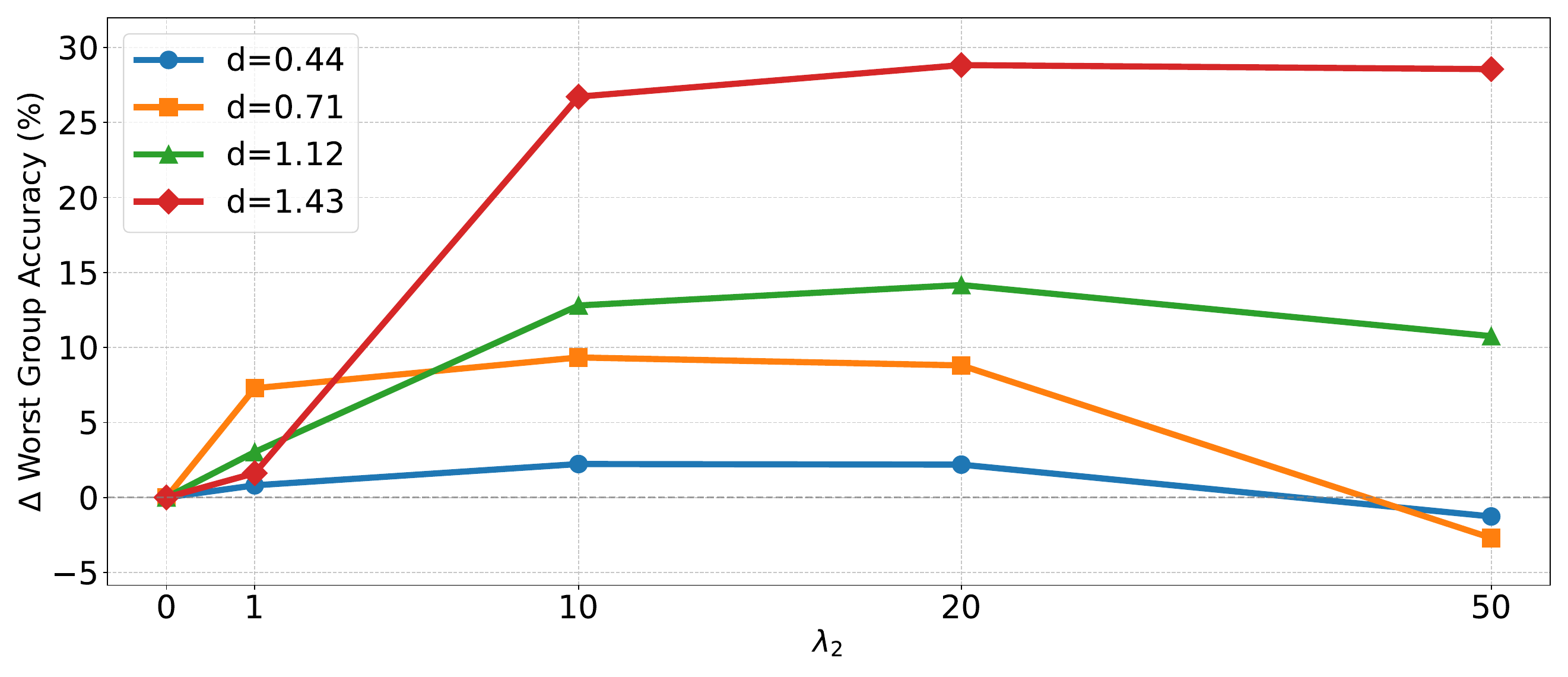}%
  }
\end{minipage}

\caption{\fontsize{8pt}{8pt}\selectfont Ablation of the degree of superclass guidance $\lambda_2$ on Waterbirds and MetaShift.}
\label{fig:ablation_lambda_appendix}
\end{figure}

\textbf{Degree of superclass guidance.}
We evaluate the effect of varying the superclass guidance weight $\lambda_2$ across all datasets. Figure~\ref{fig:ablation_lambda_appendix} reports the worst group accuracy under different values of $\lambda_2$ on selected datasets.  Overall, both insufficient guidance (i.e., low $\lambda_2$) and overly strong guidance (i.e., excessively large $\lambda_2$) can lead to degraded model performance.

\textbf{$L_2$ regularization.}
We ablate the $L_2$ term $\|\omega_{\mathrm{rel}}\|_2^2$ in Algorithm~\ref{alg:training_dafd}. Table~\ref{tab:ablation_l2_appendix} reports the change in worst group accuracy when removing $L_2$ regularization (i.e., setting $\lambda_3=0$) on selected datasets. Overall, removing $L_2$ degrades performance, which indicates that encouraging the use of all superclass-relevant features improves domain generalization.

\textbf{Relative strength of the two classifiers.}
In Algorithm~\ref{alg:training_dafd}, the losses
$\mathcal{L}^{\mathrm{CE}}_{\phi,\omega_{\mathrm{rel}}}(\mathbf{x},y)$ and
$\mathcal{L}^{\mathrm{CE}}_{\phi,\omega_{\mathrm{irr}}}(\mathbf{x},y)$ are, by default, weighted equally. Here we fix the coefficient of
$\mathcal{L}^{\mathrm{CE}}_{\phi,\omega_{\mathrm{rel}}}(\mathbf{x},y)$ to $1$ and vary the weight on
$\mathcal{L}^{\mathrm{CE}}_{\phi,\omega_{\mathrm{irr}}}(\mathbf{x},y)$ to assess its effect on performance. Results in Table~\ref{tab:ablation_cls_weight_appendix} show that either ignoring or overemphasizing $\omega_{\mathrm{irr}}$ degrades performance, and a balanced strength between $\omega_{\mathrm{rel}}$ and $\omega_{\mathrm{irr}}$ better support guidance for both superclass-relevant and irrelevant features.

\begin{table}[H]
\captionsetup{skip=10pt}
  \setlength{\tabcolsep}{3pt}
\centering
{\fontsize{8pt}{9pt}\selectfont
\caption{Prompt variants used for different values of $m$. Each prompt includes the superclass placeholder, formatted as $\mathtt{a/an\ [superclass]}$.}
\label{tab:prompt_variants}
\begin{tabular*}{0.6\textwidth}{@{\extracolsep{\fill}}c l}
\toprule
\#Prompts ($m$) & Prompt Variant \\
\midrule
1 & $\mathtt{a/an\ [superclass]}$ \\
\midrule
\multirow{2}{*}{2} 
  & $\mathtt{a/an\ [superclass]}$ \\
  & $\mathtt{a\ photo\ of\ a/an\ [superclass]}$ \\
\midrule
\multirow{5}{*}{5} 
  & $\mathtt{a/an\ [superclass]}$ \\
  & $\mathtt{a\ photo\ of\ a/an\ [superclass]}$ \\
  & $\mathtt{a\ picture\ of\ a/an\ [superclass]}$ \\
  & $\mathtt{an\ image\ of\ a/an\ [superclass]}$ \\
  & $\mathtt{a/an\ [superclass]\ photograph}$ \\
\bottomrule
\end{tabular*}
}
\end{table}

\begin{table}[H]
\captionsetup{skip=10pt}
  \setlength{\tabcolsep}{3pt}
  \centering
  {\fontsize{8pt}{8pt}\selectfont
  \caption{Ablation results on Spawrious under different prompt configurations. All values indicate the change in worst group accuracy (\%) relative to the setting $m=1$, $\text{superclass}=\texttt{dog}$.}
  \label{tab:prompt_ablation_spawrious}
  \setlength{\tabcolsep}{3pt}
  \begin{tabular*}{\textwidth}{@{\extracolsep{\fill}}%
      >{\centering\arraybackslash}p{0.9cm}%  #Prompts
      >{\centering\arraybackslash}p{1.1cm}%  Superclass
      >{\centering\arraybackslash}p{1.2cm}%  O2O-Easy
      >{\centering\arraybackslash}p{1.6cm}%  O2O-Medium
      >{\centering\arraybackslash}p{1.3cm}%  O2O-Hard
      >{\centering\arraybackslash}p{1.3cm}%  M2M-Easy
      >{\centering\arraybackslash}p{1.7cm}%  M2M-Medium
      >{\centering\arraybackslash}p{1.3cm}%  M2M-Hard
      }%   Average
    \toprule
    \#Prompts & Superclass & O2O-Easy & O2O-Medium & O2O-Hard & M2M-Easy & M2M-Medium & M2M-Hard\\
    \midrule
    1   & \texttt{dog}     & 0.0  & 0.0  & 0.0  & 0.0  & 0.0  & 0.0  \\
    2   & \texttt{dog}     & +0.7  & -2.2   & -2.3   & +0.8   & -9.6   & -2.4   \\
    5   & \texttt{dog}     & +0.3   & -2.5   & -2.9   & -1.1   & -9.4   & +1.1   \\
    1   & \texttt{animal}  & -4.0   & -1.1   & -5.6   & -4.9   & -7.5   & -5.6  \\
    \bottomrule
  \end{tabular*}
  }
\end{table}

\begin{table}[H]
\captionsetup{skip=10pt}
  \setlength{\tabcolsep}{3pt}
  \centering
  {\fontsize{8pt}{8pt}\selectfont
  \caption{Ablation results on MetaShift under different prompt configurations. All values indicate the change in worst group accuracy (\%) relative to the setting $m=1$, $\text{superclass}=\texttt{dog or cat}$.}
  \label{tab:prompt_ablation_metashift}
  \begin{tabular*}{\textwidth}{@{\extracolsep{\fill}} c c c c c c}
    \toprule
    \#Prompts & Superclass & (a) $d=0.44$ & (b) $d=0.71$ & (c) $d=1.12$ & (d) $d=1.43$\\
    \midrule
    1   & \texttt{dog or cat}     & 0.0  & 0.0  & 0.0  & 0.0  \\
    2   & \texttt{dog or cat}     & -1.5   & +0.4   & -0.9   & -2.4   \\
    5  & \texttt{dog or cat}     & -0.8   & -0.8   & -2.5   & -0.1   \\
    1   & \texttt{animal}  & -1.1   & -0.3   & -8.3   & -6.3   \\
    \bottomrule
  \end{tabular*}
  }
\end{table}

\begin{table}[H]
\captionsetup{skip=10pt}
  \setlength{\tabcolsep}{3pt}
  \centering
  {\fontsize{8pt}{8pt}\selectfont
  \caption{Ablation results on Waterbirds-95\% and Waterbirds-100\% under different prompt configurations. All values indicate the change in worst group accuracy (\%) relative to the setting $m=1$, $\text{superclass}=\texttt{bird}$.}
  \label{tab:prompt_ablation_waterbirds}
  \begin{tabular*}{0.7\textwidth}{@{\extracolsep{\fill}} c c c c}
    \toprule
    \#Prompts & Superclass & Waterbirds-95\% & Waterbirds-100\%\\
    \midrule
    1   & \texttt{bird}     & 0.0  & 0.0  \\
    2   & \texttt{bird}     & -2.6   & +3.2   \\
    5  & \texttt{bird}     & -1.1   & -2.2   \\
    1   & \texttt{animal}  & -29.2   & -44.8   \\
    \bottomrule
  \end{tabular*}
  }
\end{table}

\begin{table}[H]
\captionsetup{skip=10pt}
  \setlength{\tabcolsep}{3pt}
  \centering
  {\fontsize{8pt}{8pt}\selectfont
  \caption{Ablation results on SpuCo Dogs under different prompt configurations. All values indicate the change in worst group accuracy (\%) relative to the setting $m=1$, $\text{superclass}=\texttt{dog}$.}
  \label{tab:prompt_ablation_spucodogs}
  \begin{tabular*}{0.5\textwidth}{@{\extracolsep{\fill}} c c c}
    \toprule
    \#Prompts & Superclass & SpuCo Dogs\\
    \midrule
    1   & \texttt{dog}     & 0.0  \\
    2   & \texttt{dog}     & -0.5   \\
    5  & \texttt{dog}     & +0.9   \\
    1   & \texttt{animal}  & -14.1   \\
    \bottomrule
  \end{tabular*}
  }
\end{table}

\begin{table}[H]
\captionsetup{skip=10pt}
  \setlength{\tabcolsep}{3pt}
  \centering
  {\fontsize{8pt}{8pt}\selectfont
  \caption{Superclass mismatch results on Waterbirds-95\% and Waterbirds-100\%. All values indicate the change in worst group accuracy (\%) relative to the reference setting $\text{superclass}=\texttt{bird}$.}
  \label{tab:prompt_mismatch}
  \begin{tabular*}{0.5\textwidth}{@{\extracolsep{\fill}} c c c}
    \toprule
    Superclass & Waterbirds-95\% & Waterbirds-100\%\\
    \midrule
    \texttt{bird}      & 0.0   & 0.0   \\
    \texttt{waterfowl} & -0.6  & -0.2  \\
    \texttt{songbird}  & -2.8  & -1.2  \\
    \bottomrule
  \end{tabular*}
  }
\end{table}

\vspace{-20pt}

\begin{table}[H]
\captionsetup{skip=10pt}
\setlength{\tabcolsep}{3pt}
\centering
{\fontsize{8pt}{8pt}\selectfont
\caption{Ablation results when removing $L_2$ regularization. 
All values indicate the drop in worst group accuracy (\%) relative to the full model. 
For MetaShift, we report the mean over the four subsets (a)–(d).}
\label{tab:ablation_l2_appendix}
\begin{tabular*}{0.8\textwidth}{@{\extracolsep{\fill}} c c c c c}
\toprule
 & Waterbirds-95\% & Waterbirds-100\% & MetaShift & SpuCo Dogs \\
\midrule
$\lambda_3=0$ & -1.2 & -0.1 & -1.1 & -3.5 \\
\bottomrule
\end{tabular*}
}
\end{table}

\begin{table}[H]
\captionsetup{skip=10pt}
\setlength{\tabcolsep}{3pt}
\centering
{\fontsize{8pt}{8pt}\selectfont
\caption{Results on Waterbirds-95\%, Waterbirds-100\%, MetaShift (mean of 4 subsets), and SpuCo Dogs under different cross-entropy loss weights on $\omega_{\mathrm{irr}}$. 
The cross-entropy loss weight on $\omega_{\mathrm{rel}}$ is fixed at 1. 
All values indicate the change in worst group accuracy (\%) relative to the setting with weight $=1$.}
\label{tab:ablation_cls_weight_appendix}
\begin{tabular*}{0.75\textwidth}{@{\extracolsep{\fill}} c c c c c}
\toprule
Weight & Waterbirds-95\% & Waterbirds-100\% & MetaShift & SpuCo Dogs \\
\midrule
0   & -3.5 & -4.7 & +1.1 & -7.3 \\
1   &  0.0 &  0.0 & 0.0 & 0.0 \\
10  & -1.4 & -1.1 & -1.3 & -5.2 \\
\bottomrule
\end{tabular*}
}
\end{table}

\subsection{Compute Resources}
\label{appendix:compute-resources}

We used a single NVIDIA A100-SXM4 GPU (40 GB VRAM), an Intel Xeon CPU @ 2.20 GHz with 12 cores, and 83 GB of system RAM. Table \ref{tab:time-per-epoch} shows the average time per epoch (in seconds) for each dataset. For epoch counts and specific hyperparameters, see Appendix \ref{subapp:hs}.

\begin{table}[H]
\captionsetup{skip=10pt}
  \setlength{\tabcolsep}{3pt}
  \centering
  {\fontsize{8pt}{9pt}\selectfont
  \caption{Average time per epoch (s) for each dataset}
  \label{tab:time-per-epoch}
  \begin{tabular}{lc}
    \toprule
    Dataset                      & Time per epoch (s) \\
    \midrule
    Waterbirds-95\% \& 100\%     & 41               \\
    SpuCo Dogs                   & 216               \\
    MetaShift                    & 12               \\
    Spawrious                    & 195               \\
    \bottomrule
  \end{tabular}
  }
\end{table}

\section{SupER under Internal Spurious Correlation}
\label{app:internal_spurious}

In Section~\ref{sec:exp}, we have already demonstrated that SupER achieves significant generalization improvements under various types and degrees of spurious correlations, in particular when new groups appear at test time and when spurious features in the training data are perfectly correlated with the labels. In this section, we further consider a special case where prior knowledge indicates that spurious correlations arise entirely within the superclass. We examine this scenario because the superclass guidance from CLIP is now less dominant compared to the contributions of the $\beta$-VAE and $L_2$ regularization. Importantly, this does not present a contradiction: SupER is designed to address the more general setting in Section~\ref{subsec:problem setup}, where both the source of spurious features and the group distribution of training and test data are unknown, and Theorems~\ref{thm:informal}-\ref{thm:random-formal} already establishes SupER as the optimal choice under this setting. On the other hand, when prior knowledge suggests that spurious correlations arise entirely within the superclass, we show in this section that SupER still exhibits relatively effective performance and can be further enhanced by integrating it with existing approaches that do not require group annotations.

\textbf{Datasets.} 
We evaluate SupER on BFFHQ~\citep{lee2021learning}, CelebA~\citep{liu2015deep}, and Color MNIST~\citep{zhang2022correct, arjovsky2019invariant}. 
Specifically, BFFHQ contains spurious correlations between age (a superclass-relevant feature) and gender labels, with the minority group ratio being only 0.5\%. 
CelebA is a large-scale dataset exhibiting spurious correlations between hair color (a superclass-relevant feature) and gender. 
The data statistics of BFFHQ and CelebA are shown in Tables~\ref{tab:bffhq_stats} and~\ref{tab:celeba_stats}, respectively. 
In addition, Color MNIST introduces a spurious correlation between color (a superclass-relevant feature) and the label $y$. 
In this setting, the target label is
\(
  y \in \mathcal{Y} = \{(0,1), (2,3), (4,5), (6,7), (8,9)\},
\)
the spurious attribute $z$ takes one of five colors, and the spurious correlation ratio is $99.5\%$ during training. 
For evaluation, Color MNIST adopt a regime where colors are assigned uniformly at random to each sample. Note that we use human as the superclass for both BFFHQ and CelebA, and digit as the superclass for Color MNIST.

\begin{table}[H]
\centering
{\fontsize{8pt}{9pt}\selectfont
\caption{Dataset statistics for BFFHQ.}
\label{tab:bffhq_stats}
\begin{tabular}{lcccc}
\toprule
Split & (young, female) & (young, male) & (old, female) & (old, male) \\
\midrule
Train & 9,552 & 48 & 48 & 9,552 \\
Test  & 250   & 250 & 250 & 250 \\
\bottomrule
\end{tabular}
}
\end{table}

\begin{table}[H]
\centering
{\fontsize{8pt}{9pt}\selectfont
\caption{Dataset statistics for CelebA.}
\label{tab:celeba_stats}
\begin{tabular}{lcccc}
\toprule
Split & (not blond, female) & (not blond, male) & (blond, female) & (blond, male) \\
\midrule
Train & 71,629 & 66,874 & 22,880 & 1,387 \\
Test  & 9,767  & 7,535  & 2,480  & 180   \\
\bottomrule
\end{tabular}
}
\end{table}

\textbf{Results.} 
First, we present the performance of SupER on BFFHQ. 
Following~\citep{lee2021learning}, we report the accuracy on the bias-conflicting groups, i.e., the accuracy of two minority groups (young, male) and (old, female). 
As shown in Table~\ref{tab:bffhq_results}, in the dataset with a strong spurious correlation (99.5\%) and highly complex spurious feature (age), SupER still achieves competitive performance compared to other baselines that do not require group labels. 

\begin{table}[H]
\captionsetup{skip=10pt}
\setlength{\tabcolsep}{3pt}
\centering
{\fontsize{8pt}{9pt}\selectfont
\caption{Bias-conflicting group accuracy (\%) on BFFHQ for SupER and baselines that do not require group labels.  
For baselines, we use reported results from prior work~\citep{lee2021learning,lim2023biasadv} whenever they are stronger than our own implementations.}
\label{tab:bffhq_results}
\begin{tabular*}{0.35\textwidth}{@{\extracolsep{\fill}} l c}
\toprule
Method & Bias-conflicting \\
\midrule
ERM       & 57.0$_{\pm 0.9}$ \\
% CVaR DRO  & xx.x$_{\pm x.x}$ \\
LfF       & 62.2$_{\pm 1.0}$ \\
JTT       & 62.2$_{\pm 1.3}$ \\
CnC       & 63.1$_{\pm 1.0}$ \\
SupER (Ours) & 62.8$_{\pm 0.9}$ \\
\bottomrule
\end{tabular*}
}
\end{table}

Second, we show that SupER can be further improved by easily combining it with other approaches that do not require group annotations. 
Specifically, we combine SupER with JTT by upweighting the loss $\mathcal{L}^{\text{CE}}_{\phi, \omega_{\mathrm{rel}}}(\mathbf{x}, y)$ for samples identified in the first step of the original JTT procedure~\citep{liu2021just}, where a standard ERM model is first trained to identify potential samples with spurious correlations based on misclassification. 
As shown in Table~\ref{tab:celebA_internal}, we evaluate both JTT and our combined SupER+JTT method on the CelebA the Color MNIST datasets. 
Results show that our combined method achieves higher worst group accuracy compared to JTT alone. 
This suggests that the identification of spurious samples by JTT complements SupER’s feature disentanglement and its emphasis on leveraging all relevant superclass features for prediction. We leave further investigations on different integrations as an important future direction.

\begin{table}[H]
\captionsetup{skip=10pt}
  \setlength{\tabcolsep}{3pt}
\centering
{\fontsize{8pt}{9pt}\selectfont
\caption{
Comparison of JTT and SupER+JTT on CelebA and Color MNIST. 
SupER+JTT achieves improved worst group accuracy (\%) across both datasets. 
When applicable, shared hyperparameters are set to the same values across both methods.
}
\label{tab:celebA_internal}
\begin{tabular*}{0.55\textwidth}{@{\extracolsep{\fill}}lcc}
\toprule
Method & CelebA & Color MNIST \\
\midrule
JTT           & 80.7$_{\pm 1.2}$ & 83.3$_{\pm 2.7}$ \\
SupER+JTT     & 83.8$_{\pm 2.1}$ & 84.4$_{\pm 2.0}$ \\
\bottomrule
\end{tabular*}
}
\end{table}

\section{Licenses for External Assets}
\label{appendix:licenses}

We use the following publicly available datasets and pretrained models in our work:
\begin{itemize}
  \item \textbf{Pretrained models:}
    \begin{itemize}
      \item CLIP, MIT, available at \url{https://github.com/openai/CLIP}.
      \item ResNet50, BSD-3-Clause, available at \url{https://pytorch.org/vision/stable/models/generated/torchvision.models.resnet50.html}.
    \end{itemize}
  \item \textbf{Datasets:}
    \begin{itemize}
      \item \textbf{Waterbirds-95\%} and \textbf{Waterbirds-100\%}, MIT, available at \url{https://github.com/kohpangwei/group_DRO} and \url{https://github.com/spetryk/GALS}.
      \item \textbf{SpuCo Dogs}, MIT, available at \url{https://github.com/BigML-CS-UCLA/SpuCo}.
      \item \textbf{MetaShift}, MIT, available at \url{https://github.com/Weixin-Liang/MetaShift}.
      \item \textbf{Spawrious}, CC BY 4.0, available at \url{https://github.com/aengusl/spawrious}.
      \item \textbf{BFFHQ} (derived from FFHQ), CC BY-NC-SA 4.0 (inherited from FFHQ), FFHQ available at \url{https://github.com/NVlabs/ffhq-dataset}. 
      \item \textbf{CelebA}, Non-commercial, available at \url{https://mmlab.ie.cuhk.edu.hk/projects/CelebA.html}.
      \item \textbf{Color MNIST} (synthetic variant of MNIST), CC BY-SA 3.0 (inherited from MNIST), MNIST available via \url{https://keras.io/api/datasets/mnist/}.
    \end{itemize}
\end{itemize}

\end{document}